\documentclass[12pt,letterpaper]{report}
\newcommand{\thesistitle}{Facilitating Emergency Vehicle Passage in Congested Urban Areas Using Multi-agent Deep Reinforcement Learning}
\newcommand{\thesisauthor}{Haoran Su}
\newcommand{\thesisadvisor}{Joseph Y.J. Chow}
\newcommand{\secondthesisadvisor}{Li Jin}
\newcommand{\graddate}{January 2025} % like January XX, May 20XX, September 20XX

\usepackage{setspace}

\usepackage{xspace}
\usepackage{microtype}
\usepackage{color}
\usepackage{url}
\usepackage{lipsum}
\usepackage{fancyhdr}
\fancypagestyle{plain}{%
\fancyhf{}
\rhead{\thepage}
}

\usepackage{times}
\usepackage{helvet}
\usepackage{courier}

\usepackage{amsmath, amssymb, amsfonts, amsthm, bm}
\usepackage{cases}

\usepackage[final]{graphicx} % Use 'final' to include graphics in the document
\usepackage{subcaption} 
\usepackage{placeins}
\usepackage{booktabs}
\usepackage{multirow}
\usepackage[dvipsnames]{xcolor}

\usepackage[ruled,noend,linesnumbered]{algorithm2e}
\usepackage{appendix}
\usepackage{indentfirst}
\usepackage{etoolbox}
\BeforeBeginEnvironment{appendix}{\clearpage}

\usepackage{microtype} % Improves spacing and reduces overfull lines
\usepackage{soul}      % Text highlighting
\usepackage{xspace}    % Adds spacing for macros
\usepackage{listings}  % Code listings

\usepackage{hyperref}
\hypersetup{
    colorlinks=true,
    linkcolor=blue,
    filecolor=blue,
    urlcolor=blue,
    citecolor=blue
}
\usepackage{url}
\usepackage{todonotes}
\usepackage[all,cmtip]{xy}

\newcommand{\hs}[1]{{\color{black}{#1}}}

\newcommand{\modi}[1]{{\color{black}{#1}}}

\newtheorem{definition}{Definition}
\newtheorem{remark}{Remark}
\begin{document}
%% Produces a test "layout" page, for "debugging" purposes only.
%% Comment out for final version.
%\layout % requires package layout (see above, on this same file)
%% Sets page numbering to "roman style" i, ii, iii, iv, etc:

%%%%%% Cover page %%%%%%%%%%%
%% Sets page numbering to "roman style" i, ii, iii, iv, etc:
\pagenumbering{roman}
\thispagestyle{empty}
\begin{center}
{\bfseries 
  {\large\thesistitle}
  \vspace{1in}
  
 {\large {\bf DISSERTATION}}\\
  \vspace{.5in}
  
  \begin{doublespace}
  {\large  
  Submitted in Partial Fulfillment of\\
  % \vspace{.1in}
  the Requirements for\\
  % \vspace{.1in}
  the Degree of\\}
  \end{doublespace}
  \vspace{.5in}
  
  {\large DOCTOR OF PHILOSOPHY (Transportation Systems)}\\
  \vspace{.5in}
  
  at the \\
  \vspace{.2in}
  
  {\large
  NEW YORK UNIVERSITY\\
  \vspace{-0.05in}
  TANDON SCHOOL OF ENGINEERING\\
  }
  \vspace{.2in}
  
  by
  \vspace{.5in}

  {\large\thesisauthor}
  \vspace{.5in}
  % \vfill

  % {\large\graddate}
}

\end{center}

\newpage

%%%%%% Title page %%%%%%%%%%%
%
\setcounter{page}{1}
%% No numbering in the title page:
\thispagestyle{empty}
\begin{center}
{\bfseries 
  {\large\thesistitle}
  \vspace{.25in}
  
  DISSERTATION\\
  \vspace{.25in}
  
  \begin{doublespace}
  Submitted in Partial Fulfillment of\\
  % \vspace{.1in}
  the Requirements for\\
  % \vspace{.1in}
  the Degree of\\
  \end{doublespace}
  \vspace{.25in}
  
  DOCTOR OF PHILOSOPHY (Transportation Systems)\\
  \vspace{.25in}
  
  at the \\
  \vspace{.1in}
  
  {\large
  NEW YORK UNIVERSITY\\
  \vspace{-0.05in}
  TANDON SCHOOL OF ENGINEERING\\
  }
  \vspace{.2in}
  
  by
  \vspace{.3in}

  \thesisauthor
  \vspace{.3in}
  % \vfill
}

\end{center}
% \vfill

\vspace{0.2in}

% \noindent
% \makebox[\textwidth]{\hfill\makebox[2.5in]{Approved: \hfill}}
% \vspace{0.1in}

% \noindent
% \makebox[\textwidth]{\hfill\makebox[2.5in]{\hrulefill}}\\
% \makebox[\textwidth]{\hfill\makebox[2.5in]{\hfill Department Chair Signature\hfill}}
% \vspace{0.05in}

% \noindent
% \makebox[\textwidth]{\hfill\makebox[2.5in]{\hrulefill}}\\
% \makebox[\textwidth]{\hfill\makebox[2.5in]{\hfill Date \hfill}}

% \noindent

% University ID: {N11216022}\\ % Add your University ID
% Net ID: \hspace{.415in} {hs1854}\\ % Add your Net ID

\newpage

\doublespacing

%%%%%%%%%%%%%% Microfilm / Publishing Page %%%%%%%%%%%%%%%%%
\begin{center}
Microfilm or other copies of this dissertation are obtainable from
\vspace{4in}

UMI Dissertation Publishing\\
ProQuest CSA\\
789 E. Eisenhower Parkway\\
P.O. Box 1346\\
Ann Arbor, MI 48106-1346

\end{center}
\newpage

%%%%%%%%%%%%%% Vita %%%%%%%%%%%%%%%%%
\section*{Vita}
\addcontentsline{toc}{section}{Vita}
\textbf{Haoran Su} was born in 1995 in Wuxi, China. He earned dual Bachelor’s degrees in Civil Engineering and Computer Science in 2017, followed by a Master of Science in Systems Engineering in 2018, all from University of California, Berkeley. In Fall 2018, he started his doctoral studies in the Transportation Systems program at the Tandon School of Engineering, New York University. He is affiliated with the C2SMARTER Center, a Tier-1 University Transportation Research Center designated by the United States Department of Transportation. His research focuses on enhancing emergency vehicle passage in congested urban areas through multi-agent reinforcement learning methods.
\newpage

%%%%%%%%%%%%%% Ackknowlegment %%%%%%%%%%%%%%%%%
%% Comment out the following lines if you do not want to acknowledge
%% anyone's help...
\section*{Acknowledgements}
\addcontentsline{toc}{section}{Acknowledgements}
%!TEX root = thesis.tex

%% Write your acknowledgements in this file. If you do not want to acknowledge anyone,
%% you can delete this file and comment out the corresponding part in the "thesis.tex"
%% file.
I would like to express my deepest gratitude to my advisors, Professor Joseph Y.J. Chow and Professor Li Jin, for their patience, guidance, and inspiration, without which this dissertation would not have been possible. Their guidance and encouragement have shaped not only this dissertation but also my growth as a researcher and as an individual.

To my colleagues at C2SMARTER—Yueshuai He, Saeid Rasulkhani, Gyugeun Yoon, Ted Pantelidis, Qi Liu, Srushti Rath, Xi Xiong, Yu Tang, Qian Xie, and others—you have made this journey both meaningful and memorable.

Special thanks to my co-author and my mentor at Siemens, Yaofeng Desmond Zhong, for an inspiring summer internship that culminated in the honor of presenting our work at NeurIPS and AAAI. 

I would also like to acknowledge the financial support by Dwight David Eisenhower Transportation Fellowship, C2SMARTER, NSFC Project 62103260, SJTU UM Joint Institute, and J.Wu\&J.Sun Endowment Fund.

My parents are my role models and my best friends. Though I have been frustrated and struggling, they never doubt what I could achieve. Every storm runs out of rain and I am sincerely proud to be your son.

I still remember the reviews received for my very first manuscript: "\textit{this work makes very trivial contributions to academics or the society.}" I hope this dissertation can be of a little help when it comes to saving patients and fighting fires or crimes in the future.

\noindent
\makebox[\textwidth]{\hfill\makebox[3in]{\hfill Haoran Su\hfill}}
% \makebox[\textwidth]{\hfill\makebox[3in]{\hfill\graddate\hfill}}

\newpage

%%%%%%%%%%%%%% Dedication Page %%%%%%%%%%%%%%%%%
%% Comment out the following lines if you do not want to dedicate
%% this to anyone...
\vspace*{\fill}
\begin{center}
  To my parents\\
  for your constant love and support. 
\end{center}
\vfill
\newpage

%%%%%%%%%%%%%% Abstract %%%%%%%%%%%%%%%%%
\section*{}
\begin{center}
{\bfseries 
  %{\large\thesistitle}
  \vspace{.25in}  
  {\bf ABSTRACT}\\
  \vspace{.25in}
  {\bf \thesistitle}\\  
  \vspace{.25in}
  {\bf by}\\  
  \vspace{.5in}
  {\bf \thesisauthor}\\
  \vspace{.5in}
  {\bf Advisors: Prof. \thesisadvisor, Ph.D.,P.E. and Prof. \secondthesisadvisor, Ph.D. }\\
  \vspace{.25in}
  {\bf Submitted in Partial Fulfillment of the Requirements for}\\
  {\bf the Degree of Doctor of Philosophy (Transportation Systems)}\\
  \vspace{.25in}
  {\bf \graddate}  
  \vspace{.25in}
}
\end{center}
\addcontentsline{toc}{section}{Abstract}
Emergency Response Time (ERT) serves as a pivotal metric of urban resiliency and public safety, encapsulating a city’s capacity to respond promptly and effectively to medical, fire, and crime emergencies. Over the last decade, increasing urbanization and traffic congestion have severely exacerbated ERT, undermining emergency management systems and public trust. In New York City (NYC), average ERT for medical emergencies has surged by 72\% from 7.89 minutes in 2014 to 14.27 minutes in 2024, with over 50\% of delays attributable to prolonged Emergency Vehicle (EMV) travel times. These escalating delays have dire consequences: every minute of delay during a stroke can lead to the loss of 2 million brain cells, and survival rates for cardiac arrest diminish by 7--10\% per minute. This underscores the urgency of addressing EMV travel inefficiencies through advanced routing and traffic management strategies.

This dissertation responds to these critical challenges by advancing the state of EMV facilitation through three substantive contributions. First, it introduces \textit{EMVLight}, a decentralized multi-agent reinforcement learning (MARL) framework that dynamically integrates EMV routing and adaptive traffic signal pre-emption. \textit{EMVLight} leverages real-time traffic data and decentralized coordination between agents to optimize EMV travel paths while reducing disruptions to non-EMV traffic. Experimental results demonstrate a 42.6\% reduction in EMV travel times and a 23.5\% decrease in the average travel time for non-EMVs across synthetic and real-world traffic networks.

Second, the dissertation develops the \textit{Dynamic Queue-Jump Lane (DQJL)} system, which employs Multi-Agent Proximal Policy Optimization (MAPPO) to enable coordinated lane-clearing maneuvers in mixed-traffic environments comprising autonomous and human-driven vehicles. By dynamically forming queue-jump lanes in response to real-time traffic conditions, DQJL minimizes lane-change maneuvers and reduces EMV travel times by up to 40\%, with benefits amplified under higher autonomous vehicle penetration rates.

Third, the dissertation conducts an equity-focused evaluation of Emergency Medical Services (EMS) accessibility in NYC, integrating demographic, infrastructural, and traffic data to uncover disparities in response times across boroughs. Findings reveal systemic inequities, with Staten Island experiencing the longest delays due to sparse signalized intersections, while Manhattan, despite its dense network, faces severe congestion. The study proposes actionable interventions, including optimized EMS station placements, improved intersection configurations, and equitable resource allocation, to address these disparities.

Collectively, these contributions provide a robust foundation for enhancing EMV mobility, reducing response times, and ensuring equitable access to emergency services. The methodologies and insights presented herein offer significant implications for policymakers, transportation engineers, and urban planners, advancing the development of safer, more efficient, and resilient urban transportation systems.

\textbf{Keywords:} Emergency Response Time, Multi-Agent Reinforcement Learning, Traffic Signal Pre-emption, Dynamic Queue-Jump Lane, Emergency Medical Services, Urban Resiliency, Transportation Equity.
\newpage

%%%% Table of Contents %%%%%%%%%%%%
\tableofcontents
% \clearpage
% \pagestyle{headings}

%%%%% List of Figures %%%%%%%%%%%%%
%% Comment out the following two lines if your thesis does not
%% contain any figures. The list of figures contains only
%% those figures included withing the "figure" environment.
\listoffigures\addcontentsline{toc}{section}{\listfigurename}
\newpage

%%%%% List of Tables %%%%%%%%%%%%%
%% Comment out the following two lines if your thesis does not
%% contain any tables. The list of tables contains only
%% those tables included withing the "table" environment.
\listoftables\addcontentsline{toc}{section}{\listtablename}
\newpage

%%%%% Body of thesis starts %%%%%%%%%%%%
\pagenumbering{arabic} % switches page numbering to arabic: 1, 2, 3, etc.

%% Introduction. If your thesis has no introduction, or chapter 1 is
%% meant to be the introduction, then comment out the lines below.
%% \section*{Introduction}\addcontentsline{toc}{section}{Introduction}
%\input{intro}

%%If your thesis has different "Parts", use commands such as the following:
\chapter{Introduction}

In this chapter, we present the background and motivations underpinning this dissertation in Section~\ref{sec:background}, providing a comprehensive overview of the problem domain and its significance. Section~\ref{sec:challenges} identifies the key research challenges that this work aims to address, highlighting gaps in the existing literature and practical constraints. Finally, Section~\ref{sec:contributions} outlines the major contributions and structure of this dissertation.

\section{Background and Motivation}\label{sec:background}

Emergency Response Time (ERT) is a critical metric for assessing urban resiliency and public safety, reflecting a city's capacity to respond promptly and effectively to a diverse range of emergencies, including medical incidents, fire outbreaks, and criminal activities. Shorter ERTs are indicative of well-coordinated emergency response systems, efficient infrastructure, and sufficient resource allocation. Conversely, delays in ERT exacerbate the severity of incidents, increasing risks to human life and property.

Over the past decade, urban centers have faced mounting challenges in maintaining optimal ERT levels, primarily due to increasing traffic congestion. As urban populations grow and vehicular density escalates, the mobility of Emergency Vehicles (EMVs) is increasingly compromised. This trend has led to significant delays in emergency response, with profound implications for public health and safety. Prolonged ERTs correlate with higher mortality rates in critical medical cases, more extensive damage during fire incidents, and reduced effectiveness in deterring or containing criminal activities. Collectively, these factors undermine public trust in emergency management systems.

\begin{figure}[h]
    \centering
    \includegraphics[width=.75\linewidth]{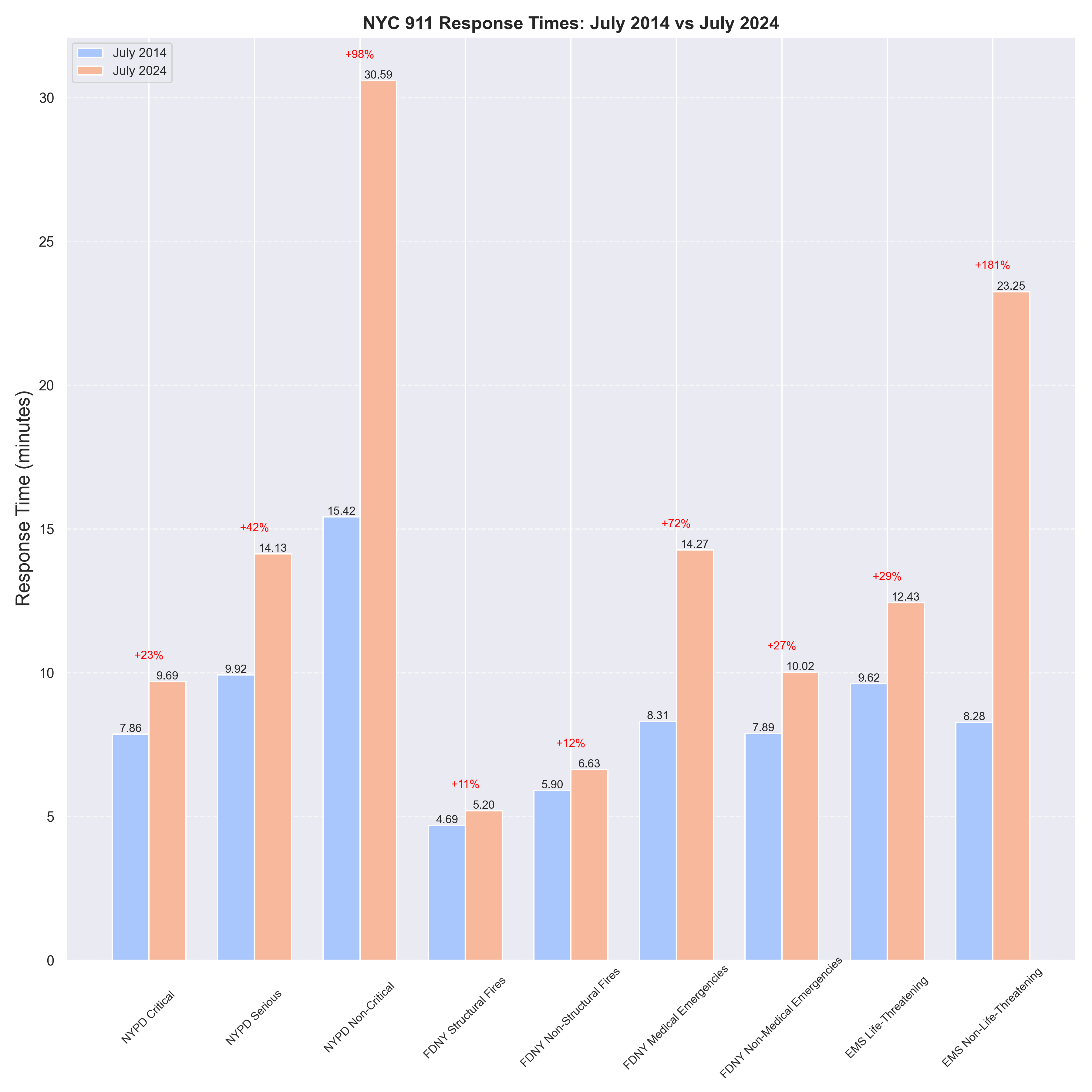}
    \caption{NYC 911 End-to-End ERTs for July 2014 and July 2024, across all incident categories.}
    \label{fig:ERT_NYC}
\end{figure}

Taking New York City (NYC) as an illustrative case, Fig.~\ref{fig:ERT_NYC} presents the end-to-end ERT for all emergency incident types, comparing data from July 2014 to July 2024, as reported in the NYC 911 End-to-End Response Time dataset~\cite{NYC911Data}. The values annotated above each bar represent the average ERT (minutes), with the percentage increase over the ten-year period highlighted in red for each incident type. Notably, ERTs have increased significantly across all categories, with non-critical and non-life-threatening cases experiencing the most severe delays.

\begin{figure}[ht]
    \centering
    \includegraphics[width=.75\linewidth]{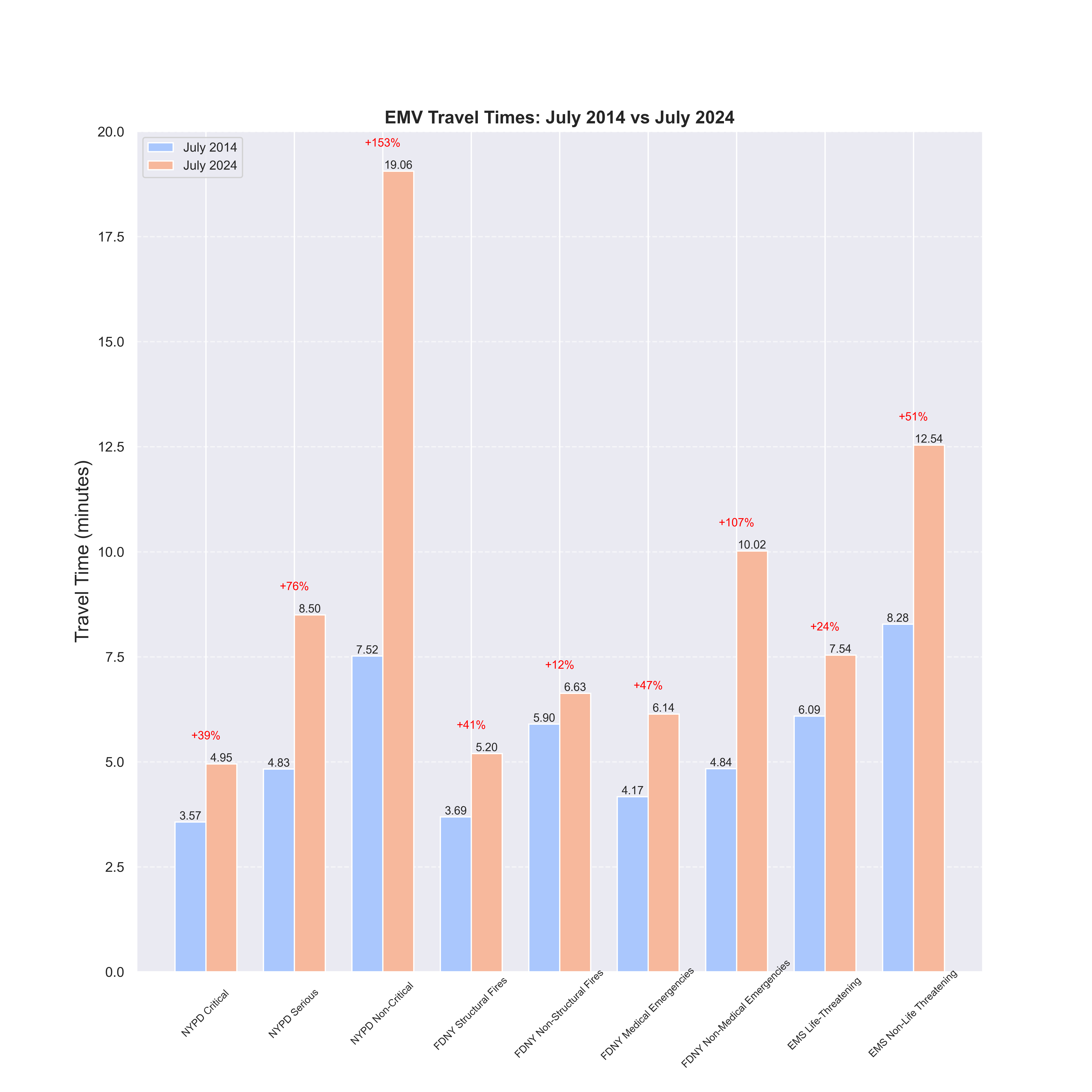}
    \caption{NYC 911 EMV travel times for July 2014 and July 2024, across all incident categories.}
    \label{fig:EMV_travel_NYC}
\end{figure}

Among the various stages of emergency response, spanning from incident reporting to on-scene arrival, the increased travel time of EMVs accounts for over 50\% of the observed growth in ERT within NYC, as illustrated in Fig.~\ref{fig:EMV_travel_NYC}. This escalation is primarily attributed to worsening traffic congestion in NYC. In 2024, INRIX, a global traffic analytics service, identified NYC as the most congested city globally~\cite{INRIXScorecard}, with drivers enduring an average of 101 hours annually in traffic delays---a significant increase compared to the 91 hours recorded in 2014, marking a ten-hour rise over the decade.

The implications of delayed emergency responses are severe. For instance, a stroke victim loses approximately 2 million brain cells for every minute of delayed medical intervention~\cite{Time2006Jeffrey}, while survival rates during cardiac arrest decline by 7--10\% with each minute of delay~\cite{Heart2013}. Despite having designated rights-of-way on urban roads, EMVs face mounting challenges in navigating through increasingly congested streets to reach emergency scenes in a timely manner. 

Escalating traffic congestion not only impedes the mobility of EMVs but also undermines their capacity to meet critical response time benchmarks, thereby compounding risks to public health and safety. This alarming trend underscores the pressing need for innovative strategies to facilitate the mobility of EMVs, ensuring that life-saving interventions can be delivered without avoidable hindrance.

\section{Research Challenges}\label{sec:challenges}

\subsection{Coupling of EMV Navigation and Traffic Signal Pre-emption}

Efficient EMV passage in urban traffic networks presents two interrelated challenges: dynamically routing EMVs under time-dependent traffic conditions and coordinating traffic signal pre-emption to minimize delays for both EMVs and non-EMVs.

The first challenge arises from the inherently dynamic nature of urban traffic, where congestion levels fluctuate across road segments in real time. As an EMV progresses through the network, its routing decisions must be continuously updated to account for evolving traffic conditions. Traditional approaches, such as recalculating shortest paths at every intersection, are computationally expensive and fail to meet the real-time requirements of EMV passage. Therefore, an efficient and adaptive routing algorithm is essential to ensure timely and computationally feasible updates to the EMV's route.

The second challenge concerns the coordination of traffic signals to achieve dual objectives: prioritizing EMV passage while minimizing network-wide disruptions for non-EMVs. Conventional methods often focus exclusively on reducing EMV travel time, disregarding the broader impact on non-EMVs. Vehicles along an EMV's path are often required to stop or pull over, yet the absence of clear and coordinated guidance from traffic signals frequently results in unnecessary delays. Moreover, non-EMVs at adjacent intersections may experience additional disruptions due to the lack of coordination in addressing the cascading delays caused by EMV pre-emption. Addressing this challenge requires traffic signals to operate collaboratively across the network, balancing the urgent need for EMV prioritization with minimizing delays for non-EMVs.

Compounding these challenges are the stochastic and unpredictable characteristics of urban traffic, which introduce significant uncertainties into both routing and signal control decisions. These uncertainties are further exacerbated by the cascading disruptions caused by EMV pre-emption, which ripple through the traffic network and amplify congestion. Addressing these issues necessitates a holistic approach capable of dynamically adapting to real-time traffic conditions, coordinating actions across the network, and accounting for the inherent uncertainties of urban traffic environments. Chapter~\ref{chap:emvlight} aims to address these challenges.

\subsection{Intra-link Movement for EMV}

From a microscopic perspective, the intra-link traversal of EMVs poses significant challenges. While traffic laws mandate yielding the right-of-way to EMVs, and most drivers instinctively comply, their uncoordinated and unpredictable maneuvers often result in suboptimal outcomes, especially on congested roadways. This lack of coordination leads to inefficiencies and delays that undermine the effectiveness of emergency responses.

Despite advancements in vehicle-to-everything (V2X) technologies designed to enhance EMV mobility, sirens remain the primary tool relied upon in practice. However, traditional siren systems frequently fail to provide sufficient warning time to other drivers, particularly in heavily congested traffic conditions. Furthermore, ambiguity regarding which route should be cleared often causes confusion, delaying the EMV's progress. These uncertainties exacerbate delays and contribute to accident rates that are 4 to 17 times higher~\cite{Buchenscheit2009AVE}, along with increased collision severity~\cite{Yasmin2012Effects}. These factors not only hinder EMV traversal but also pose additional risks to traffic safety, highlighting the urgent need for effective solutions.

Current lane-clearing strategies in emergency scenarios predominantly rely on mixed-integer linear programming (MILP) formulations. While these approaches offer theoretical insights, they often oversimplify real-world traffic dynamics and fail to provide actionable decisions in real time. The reliance on computationally intensive models makes them unsuitable for dynamic, time-sensitive scenarios. Additionally, these methods struggle to account for the stochastic nature of traffic flows, unexpected driver behaviors, and the cascading disruptions caused by lane-clearing maneuvers.

The core challenge, therefore, is how to quickly establish a cleared and easy-to-maneuver route for EMV passage on a road segment while minimizing additional disturbances for non-EMVs. Addressing this challenge requires innovative strategies capable of capturing real-time traffic information, managing uncertainties, and delivering computationally efficient solutions that can infer optimal actions within milliseconds. Meeting this need is critical for ensuring timely and safe EMV traversal through congested urban traffic. Chapter~\ref{chap:mappo-dqjl} focuses on providing a solution.

\section{Dissertation Contributions and Summary}\label{sec:contributions}

This dissertation addresses the aforementioned challenges associated with efficient EMV passage in congested urban areas, adopting a comprehensive approach that spans network-level EMV route optimization to intra-link EMV passage strategies. By systematically reducing EMV travel time and mitigating downstream traffic disturbances, the proposed methods aim to enhance emergency response efficiency. Furthermore, a qualitative case study on the emergency accessibility of NYC is conducted to evaluate the potential of the proposed solutions in a real-world setting. The research topics explored in this dissertation represent some of the earliest contributions in EMV passage, addressing critical gaps between theoretical concepts and their practical applications in real-world scenarios.

\subsection{EMVLight}

Chapter~\ref{chap:emvlight} introduces \textit{EMVLight}, a decentralized multi-agent reinforcement learning (MARL) framework that addresses two critical challenges mentioned above: dynamically routing EMVs under time-dependent traffic conditions and coordinating traffic signal pre-emption to minimize delays for both EMVs and non-EMVs. Dynamic EMV routing is particularly challenging due to the evolving nature of road congestion, which requires real-time updates to routing decisions without the computational burden of recalculating shortest paths at every intersection. Similarly, traffic signal pre-emption must balance the need to prioritize EMVs along their routes while ensuring network-wide efficiency, a task that demands coordinated adjustments across the entire traffic network. To address these challenges, \textit{EMVLight} models each intersection as an autonomous agent capable of optimizing local traffic flow and coordinating with neighboring agents to facilitate EMV passage. The framework incorporates three key contributions: a mathematical model for emergency lane formation, which evaluates road segment capacity and dynamically reallocates lanes to enable full-speed EMV travel; a decentralized path-finding algorithm that leverages real-time traffic information to adapt EMV routes efficiently; and an integrated MARL approach that jointly optimizes EMV routing and traffic signal control through specialized agents with tailored reward functions. Experimental results demonstrate that \textit{EMVLight} achieves up to a $42.6\%$ reduction in EMV travel time and a $23.5\%$ decrease in the average travel time for all vehicles, significantly outperforming existing methods across both synthetic and real-world traffic networks for this problem.

\subsection{MAPPO-DQJL}

Chapter~\ref{chap:mappo-dqjl} introduces dynamic queue-jump lanes (DQJLs) as a novel intra-link coordination strategy designed to expedite EMV passage through congested road segments. Extending the traditional concept of queue-jump lanes (QJLs) used in bus operations, DQJLs dynamically form lanes during an EMV's approach or traversal, leveraging V2X communication technologies to coordinate connected autonomous vehicles (CAVs) while accounting for non-connected human-driven vehicles (HDVs). To address the inherent complexity of mixed traffic environments, we model DQJL formation as a partially observable Markov decision process (POMDP), which captures both the controllable behavior of CAVs and the stochastic, unpredictable nature of HDVs. To optimize the DQJL formation process, we develop a multi-agent proximal policy optimization algorithm (MAPPO-DQJL) that employs centralized training with decentralized execution, enabling CAVs to efficiently coordinate lane-clearing maneuvers while minimizing disruptions to all non-EMVs. This framework effectively balances the dual objectives of reducing EMV passage time and reducing disturbance to overall traffic, demonstrating scalability across varying CAV penetration rates and traffic densities in multilane roadways. Through extensive SUMO-based simulations, the proposed framework is validated, achieving up to 39.8\% reduction in EMV passage times and up to 55.7\% reduction in lane-changing maneuvers compared to baseline methods, with particularly pronounced benefits under scenarios of increasing CAV adoption. These contributions address critical gaps in current EMV passage strategies, providing a robust and scalable solution for integrating DQJLs into EMV management systems.

\subsection{EMS Accessibility Study of NYC}

Chapter~\ref{chap:ems_accessibility} presents a comprehensive intersection-aware EMS accessibility model to address the critical challenge of EMS accessibility in congested urban environments, focusing on NYC as a case study. By integrating road network characteristics, intersection density, and population demographics, the proposed model provides a granular evaluation of EMS accessibility, identifying vulnerable regions where response times exceed critical benchmarks. The study introduces a novel metric that incorporates intersection-induced delays into travel time calculations, capturing the complexities of urban traffic networks more realistically. Additionally, it highlights the disparities in EMS coverage across NYC boroughs, with significant accessibility gaps in Staten Island, Queens, and parts of Manhattan. The implementation of \textit{EMVLight}, introduced in Chapter~\ref{chap:emvlight}, is further explored to demonstrate its potential in reducing intersection delays, improving hospital accessibility, and ensuring that over 95\% of NYC residents are served within the benchmark response time. These contributions lay a robust foundation for advancing emergency traffic management strategies and guiding urban planning decisions aimed at equitable and efficient EMS accessibility.

\subsection{Dissertation Outline}

The remainder of this dissertation is organized as follows. Chapter~\ref{chap:emvlight} introduces \textit{EMVLight}, a multi-agent reinforcement learning framework for EMV decentralized routing and traffic signal control. Chapter~\ref{chap:mappo-dqjl} proposes MAPPO-DQJL, an innovative approach that facilitates intra-link EMV movements through cooperative and efficient lane-clearing strategies. Chapter~\ref{chap:ems_accessibility} presents a case study on EMS accessibility in New York City, providing valuable insights into the potential impacts of deploying \textit{EMVLight} in real-world settings. Finally, Chapter~\ref{chap:conclusion} concludes with a summary of research contributions, a discussion of research limitations, and potential directions for future work.

%!TEX root = ../thesis.tex
\chapter[EMVLight: a Multi-agent Reinforcement Learning Framework for an Emergency Vehicle Decentralized Routing and Traffic Signal Control System]%
{EMVLight: a Multi-agent Reinforcement Learning Framework for an Emergency Vehicle Decentralized Routing and Traffic Signal Control System\footnote{This chapter has been published in Transportation Research Part C: Emerging Technologies, and can be referenced as: Su, H., Zhong, Y. D., Chow, J. Y., Dey, B., \& Jin, L. (2023). EMVLight: A multi-agent reinforcement learning framework for an emergency vehicle decentralized routing and traffic signal control system. \textit{Transportation Research Part C: Emerging Technologies}, \textit{146}, 103955.}}
\label{chap:emvlight}
\section{Introduction}
Emergency vehicles (EMVs) include ambulances, fire trucks, and police cars, which respond to critical events such as medical emergencies, fire disasters, and public security crisis. Emergency response time is the key indicator of a city's incidents management ability and resiliency. Reducing response time saves lives and prevents property losses. For instance, Berdowski et al.~\cite{berdowski2010global} indicated that the survivor rate from a sudden cardiac arrest without treatment drops 7\% - 10\% for every minute elapsed, and there is barely any chance to survive after 8 minutes. EMV travel time, the time interval for an EMV to travel from a rescue station to an incident site, accounts for a major portion of the emergency response time. However, overpopulation and urbanization have exacerbated road congestion, making it more challenging to reduce the average EMV travel time. Records \cite{end-to-end-response-times} have shown that even with a decline in average emergency response time, the average EMV travel time increased from 7.2 minutes in 2015 to 10.1 minutes in 2021 in New York City, an approximately 40\% increase over six years even accounting for post-Covid traffic conditions. Therefore, there is a severe urgency and significant benefit for shortening the average EMV travel time on increasingly crowded roads.

Existing work has studied strategies to reduce the travel time of EMVs by route optimization and traffic signal pre-emption \cite{Lu2019Literature, humagain2020systematic}. Route optimization refers to the search for a time-based shortest path. The traffic network (e.g., city road map) is modeled as a graph with intersections as nodes and road segments between intersections as edges. Based on the time a vehicle needs to travel through each edge (road segment), route optimization calculates an optimal route with the minimal EMV travel time \cite{humagain2020systematic}. In addition, as the EMV needs to be as fast as possible, the law in most places requires non-EMVs to yield to emergency vehicles sounding sirens, regardless of the traffic signals at intersections \cite{de1991lights}. Even though this practice gives the right-of-way to EMVs, it poses safety risks for vehicles and pedestrians at intersections \cite{grant2017human}. To address this safety concern, existing methods \cite{nelson2000impact, qin2012control, humagain2020systematic, huang2015design} have also studied traffic signal pre-emption which refers to the process of deliberately altering the signal phases at each intersection to prioritize EMV passage.
%
% introduce the current method
% The travel time of an EMV refers to the time interval for the EMV to travel from a source location (e.g. hospital, fire station, police office) to a emergency site (e.g. a residential community where a fire outbreak happens). \dz{As EMV travel time accounts for a majority of EMV response time, 
% The time that a vehicle needs to travel through each edge (road segment) is based on history data. 
% The most straightforward pre-emption strategy is WALABI \cite{bieker2019modelling}, where the signal phases are altered to let an EMV pass each intersection without stop. 
% cons of the current method
%

However, a major challenge for adaptive EMV operation is the coupling between
EMV route optimization and traffic light pre-emption \cite{humagain2020systematic}. As the traffic condition constantly changes, static route optimization can potentially become suboptimal as an EMV travels through the network; i.e. the traffic is highly time-dependent and exhibits transient properties during a dispatch \cite{Coogan2015Compartmental}. Moreover, traffic signal pre-emption has a significant impact on the traffic flow, which would change the fastest route as well. Thus, the optimal route should be updated with real-time traffic flow information, i.e., the route optimization should be solved in a dynamic (time-dependent) way. As an optimal route can change as an EMV travels through the traffic network, the traffic signal pre-emption would need to adapt accordingly. In other words, the subproblems of dynamic route optimization and traffic signal pre-emption are coupled and should be solved ideally simultaneously in real-time. Existing approaches have limited consideration to this coupling.

% discuss the reduce traffic impact problem. 
% As traffic signal pre-emption prioritize the route of the EMV, the travel time of non-EMVs will be largely influenced. 
% our approach
%
In addition, most of the existing models on emergency vehicle service have a single objective of reducing the EMV travel time \cite{Haghani2004Simulation, haghani2003optimization, panahi2008gis, shaaban2019strategy}. As a result, their traffic signal control strategies have an undesirable effect of increasing the travel time of non-EMVs. Non-EMVs on the path of approaching EMVs would pull over or stop, and they usually do not receive clear guidance from traffic signals on resuming their trips \cite{hsiao2018preventing}, causing unnecessary delay. Non-EMVs elsewhere are also likely indirectly and negatively impacted if adjacent intersections lack coordination to address the incurred delay \cite{humagain2020systematic}. Therefore, traffic signal control strategies accommodating both EMVs and non-EMVs need to be recommended.

In this chapter, we aim to perform route optimization and traffic signal pre-emption to not only reduce EMV travel time but also to reduce the average travel time of non-EMVs. In particular, we address the following two key challenges:
\begin{enumerate}
    \item \textbf{How to dynamically route an EMV to a destination under time-dependent traffic conditions in a computationally efficient way?} As the congestion level of each road segment changes over time, the routing algorithm should be able to update the remaining route as the EMV passes each intersection. Running the shortest-path algorithm each time the EMV passes through an intersection is not efficient. A computationally efficient decentralized routing algorithm is desired. \label{challenge_1}
    \item \textbf{How to coordinate traffic signals to not only reduce EMV travel time but reduce the average travel time of non-EMVs as well?} To reduce EMV travel time, only the traffic signals along the route of the EMV need to be altered. However, to further reduce average non-EMV travel time, traffic signals in the whole traffic network need to cooperate. \label{challenge_2} 
\end{enumerate}

Reinforcement learning (RL), which gained significant traction in assorted domains of traffic signal control recently, has been extensively studied and proven effective for learning stochastic traffic conditions and dealing with randomness.
Thus, to tackle the above challenges, we propose \textbf{EMVLight}, a decentralized multi-agent reinforcement learning framework with a dynamic routing algorithm to control traffic signal phases for efficient EMV passage. \modi{In the proposed RL design, each intersection is an agent and each agent is responsible for deciding local traffic signal phases. Multiple agents coordinate to facilitate EMV passage as well as alleviate congestion.} Our experimental results demonstrate that EMVLight outperforms traditional traffic engineering methods and existing RL methods under two metrics - EMV travel time and the average travel time of all vehicles - on different traffic configurations. We extend the preliminary work \cite{Su2021EMVLight} by taking into account extra capacity of each road segments and the possibility of forming ``emergency lanes" for full speed EMV passage. In addition, we demonstrate EMVLight's performance on synthetic and real-world maps with extra capacities. We also present the difference in EMV routing between EMVLight and benchmark methods with an emphasize on the number of successfully formed emergency lanes.

Our contributions are threefold. First, we capture the emergency capacity in road segments for emergencies and incidents. We also propose a mathematical model to decide whether an emergency lane can be formed for full speed EMV passage based on emergency capacity and number of vehicles of a road segment.
Second, we incorporate a decentralized path-finding scheme for EMVs based on real time traffic information. Third, we propose to solve EMV routing and traffic signal control problems simultaneously in a multi-agent reinforcement learning framework. In particular, we set up different types of reinforcement learning agents based on the location of the EMV and design different rewards for each type. This leads to up to a $42.6\%$ reduction in EMV travel time as well as an $23.5\%$ shorter average travel time of all trips completed in the network compared with existing benchmark methods.

The rest of the paper is organized as follows. Section \ref{sec_related_work} reviews relevant literature. Section \ref{sec_preliminaries} introduces our definition of pressure and emergency capacity. Section \ref{sec_methodology} presents our EMVLight methodology, i.e.,  dynamic routing and reinforcement learning. Benchmark methods and experimental setup are presented in Section \ref{sec_experimentation}. Section \ref{sec_result} discussed and compared the performance of EMVLight and benchmark methods in terms of EMV travel time, average travel time of all vehicles as well as EMV route choices. We conclude in Section \ref{sec_conclusion} and share inspirations on future directions.

\section{Literature Review}\label{sec_related_work}

\textbf{Conventional routing optimization and traffic signal pre-emption for EMVs.}
Although routing and pre-emption are coupled in reality, existing methods usually solve them separately. Many of the existing approaches leverage Dijkstra's shortest path algorithm to determine the optimal route~\cite{wang2013development, Mu2018Route, kwon2003route, JOTSHI20091}. Nordin et al.~\cite{nordin2012finding} proposed an A* algorithm for ambulance routing, assuming that routes and traffic conditions are fixed and static, which fails to address the dynamic nature of real-world traffic flows. Another line of work considers the temporal dynamics of traffic flows. For instance, Ziliaskopoulos et al.~\cite{ziliaskopoulos1993time} developed a shortest-path algorithm for time-dependent traffic networks, assuming that travel times associated with each edge are known in advance. Similarly, Musolino et al.~\cite{musolino2013travel} proposed routing strategies tailored to specific times of the day (e.g., peak/non-peak hours) based on historical traffic data. However, in the problem under consideration, routing and pre-emption strategies can significantly influence the travel time of each edge during the EMV passage, and the existing methods fail to handle such real-time changes effectively. Haghani et al.~\cite{haghani2003optimization} formulated the dynamic shortest path problem as a mixed-integer programming model, and Koh et al.~\cite{koh2020real} utilized reinforcement learning (RL) for real-time vehicle navigation and routing. Related studies~\cite{miller2000least, gao2006optimal, kim2005optimal, fan2005shortest, yang2014constraint, huang2012optimal, gao2012real, samaranayake2012tractable, nie2012optimal, thomas2007dynamic} explored adaptive routing problems in various stochastic and time-dependent traffic scenarios. However, none of these works considered the coupling of traffic signal pre-emption with EMV routing or the specific context of EMV passage when solving shortest path problems.

Once an optimal route for the EMV has been determined, traffic signal pre-emption is deployed to further reduce the EMV travel time. A common pre-emption strategy is to extend the green phases of traffic signals to allow the EMV to pass through intersections along a fixed optimal route~\cite{wang2013development, bieker2019modelling}. Pre-emption strategies for handling multiple EMV requests were introduced by Asaduzzaman et al.~\cite{Asaduzzaman2017APriority}. Wu et al.~\cite{wu2020emergency} approached the lane-clearing problem for emergency vehicles from a microscopic motion planning perspective, while Hosseinzadeh et al.~\cite{hosseinzadeh2022mpc} developed an EMV-centered traffic control scheme for multiple intersections to alleviate congestion. While these studies offer valuable insights into pre-emption strategies, they do not consider the dynamic routing of EMVs to determine the optimal path.

For a thorough survey of conventional routing optimization and traffic signal pre-emption methods, readers are referred to Lu et al.~\cite{Lu2019Literature} and Humagain et al.~\cite{humagain2020systematic}. It is worth noting that conventional methods prioritize EMV passage, often causing significant disturbances to traffic flow and increasing the average travel time for non-EMVs.

\textbf{RL-based traffic signal control.}
Traffic signal pre-emption only adjusts traffic phases at intersections where an EMV travels. However, reducing congestion often requires coordinated phase adjustments at nearby intersections. The problem of coordinating traffic signals to mitigate congestion has been addressed using deep reinforcement learning (RL) in an increasing body of research. Abdulhai et al.~\cite{abdulhai2003reinforcement} were among the first to apply Q-learning, a model-free RL method, to adaptive traffic signal control. Their work demonstrated the concept for both isolated traffic signal controllers and networks of controllers. However, as the state representation grows exponentially with the number of traffic signals, the learning process becomes computationally expensive, and their study was limited to isolated intersections. Prashanth et al.~\cite{prashanth2010reinforcement} introduced feature-based state representations to address the curse of dimensionality, reducing computational complexity while improving performance. El-Tantawy et al.~\cite{el2013multiagent} incorporated game-theoretic approaches into Q-learning, allowing agents to converge to best-response policies relative to their neighbors.

The advent of deep neural networks enabled more advanced applications of RL. Van der Pol et al.~\cite{van2016coordinated} integrated deep Q-networks (DQN) into traffic coordination tasks, which inspired subsequent research leveraging the Q-learning framework for traffic signal control. For instance, CoLight~\cite{wei2019colight} utilized graph attentional networks to enhance communication and cooperation between traffic signals, while FRAP~\cite{zheng2019frap} introduced a phase competition model to improve RL-based methods. PressLight~\cite{wei2019presslight} incorporated the max pressure strategy~\cite{varaiya2013max, LI2019Backpressure, LEVIN2020maxpressure, WANG2022Learning, Lazar2021Routing} into reward design, achieving better results than traditional designs. ThousandLights~\cite{ThousandLights} combined FRAP and PressLight to achieve city-level traffic signal control, and Zang et al.~\cite{Zang_Yao_Zheng_Xu_Xu_Li_2020} employed meta-learning algorithms to accelerate Q-learning.

Actor-critic methods represent another category of RL approaches that address the scalability challenge in multi-agent settings. Aslani et al.~\cite{aslani2017adaptive} demonstrated the effectiveness of actor-critic controllers for adaptive signal control under different disruption scenarios, while Xu et al.~\cite{xu2021hierarchically} introduced a hierarchical actor-critic method to foster cooperation between intersections. Chu et al.~\cite{chu2019multi} used a multi-agent independent advantage actor-critic algorithm, and Ma et al.~\cite{Ma2020Feudal} proposed a manager-worker hierarchy to manage traffic locally within actor-critic frameworks. While existing RL-based traffic control methods effectively reduce network congestion, they are not designed specifically for EMV pre-emption. In contrast, the proposed RL framework builds on state-of-the-art methods, such as max pressure, to minimize both EMV travel time and overall congestion. Moreover, the centralized training with decentralized execution paradigm enables compatibility with stochastic traffic settings while incurring minimal communication costs. Mo et al.~\cite{Mo2020CVLight} investigated traffic signal control strategies based on connected vehicle communications, illustrating the potential of RL in this domain.

For a detailed review of RL-based traffic signal control, see Noaeen et al.~\cite{RLSurvey2022Mohammad} and Wei et al.~\cite{wei2019survey}. While RL methods have demonstrated effectiveness in traffic signal control, no existing studies have explicitly addressed the unique challenges posed by EMV passages or their impact on non-EMV traffic flow.

\textbf{Intra-link EMV traversal strategies.}
Advances in vehicle-to-vehicle communication technology~\cite{LeBrun2005Knowledge} have sparked interest in optimizing EMV traversal strategies on congested road links. Agarwal et al.~\cite{Agarwal2016V2V} proposed Fixed Lane and Best Lane strategies for EMV traversal, which were further analyzed by Insaf et al.~\cite{Insaf2019Emergency}, showing that Fixed Lane outperforms Best Lane under conditions of low speed variance. Hannoun et al.~\cite{Hannoun2019Facilitating, hannoun2021sequential} developed a semi-automated warning system that instructs downstream traffic to yield using mixed-integer linear programming. This system adapts to varying connected vehicle penetration rates and is validated as computationally efficient for real-time deployment. Su et al.~\cite{su2021dynamic} proposed a dynamic queue-jump lane (DQJL) strategy leveraging multi-agent reinforcement learning.
\section{Preliminaries}\label{sec_preliminaries}
In this section, we introduce relevant preliminary terms and definitions which facilitate the problem formulation.

\subsection{Traffic map, movement and signal phase}
\begin{figure}[h]
\includegraphics[width=0.65\linewidth]{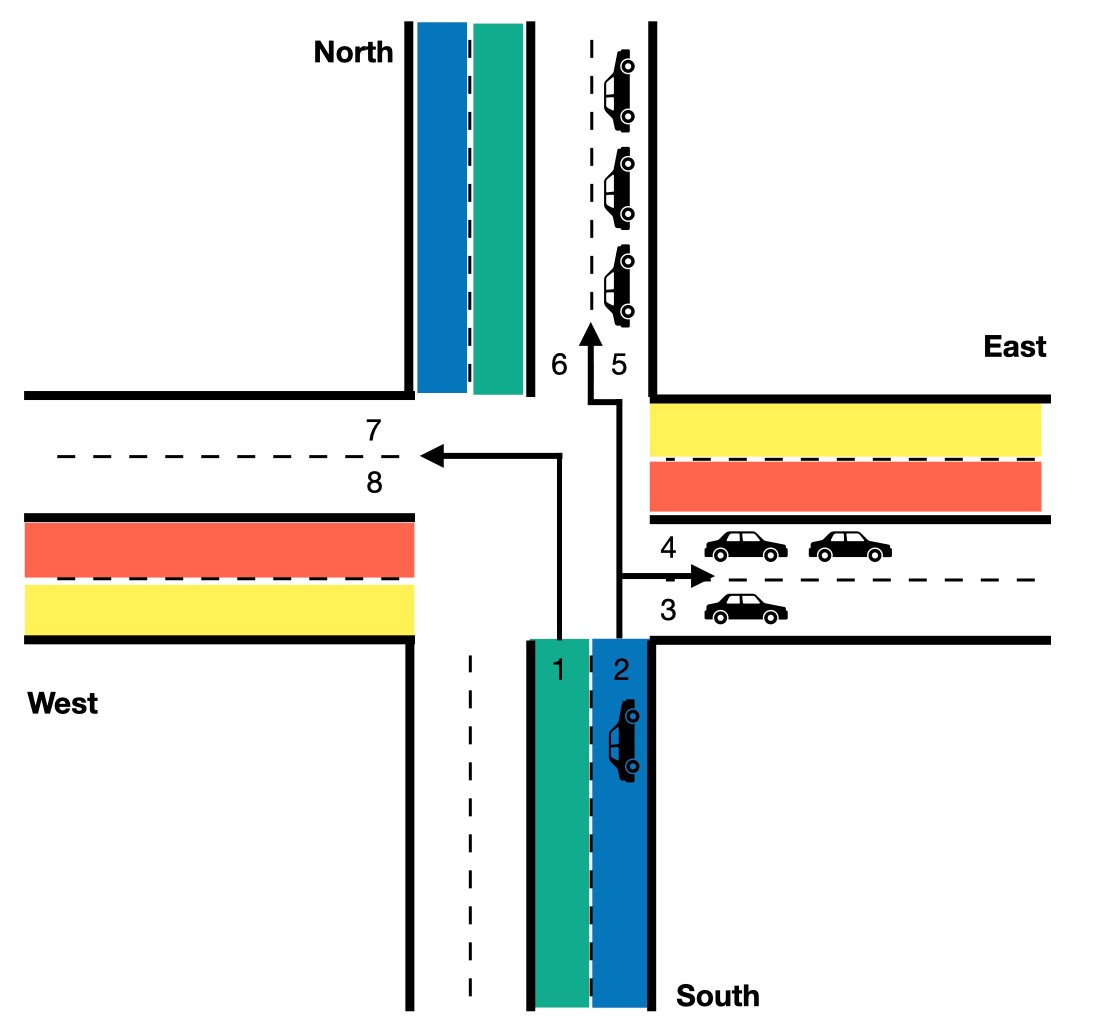}
\centering
\caption{Traffic movements of a four-way two-lane bidirectional intersection.}
\label{fig_movements}
\end{figure}
A traffic map can be represented by a graph $G(\mathcal{V}, \mathcal{E})$, with intersections as nodes and road segments between intersections as edges. We refer to a one-directional road segment between two intersections as a link. A link has a fixed number of lanes, denoted as $h(l)$ for lane $l$. Vehicles are allowed to switch lane between two intersections. Fig.~\ref{fig_movements} shows 8 links and each link has 2 lanes. 
    
% we make it clear it is per lane
% Traffic movements
A traffic movement $(l,m)$ is defined as the traffic traveling across an intersection from an incoming lane $l$ to an outgoing lane $m$. The intersection shown in Fig.~\ref{fig_movements} has 24 permissible traffic movements. As an example, vehicles on lane 1 are turning left, and vehicles on lane 2 may go straight or turn right. After turning into the link, vehicles will enter either lane. Thus, the incoming South link has the potential traffic movements of $\{(1, 7), (1, 8), (2, 5), (2, 6), (2, 3), (2, 4)\}$. The set of all permissible traffic movements of an intersection is denoted as $\mathcal{M}$.

\begin{figure}[t]
    \includegraphics[width=\linewidth]{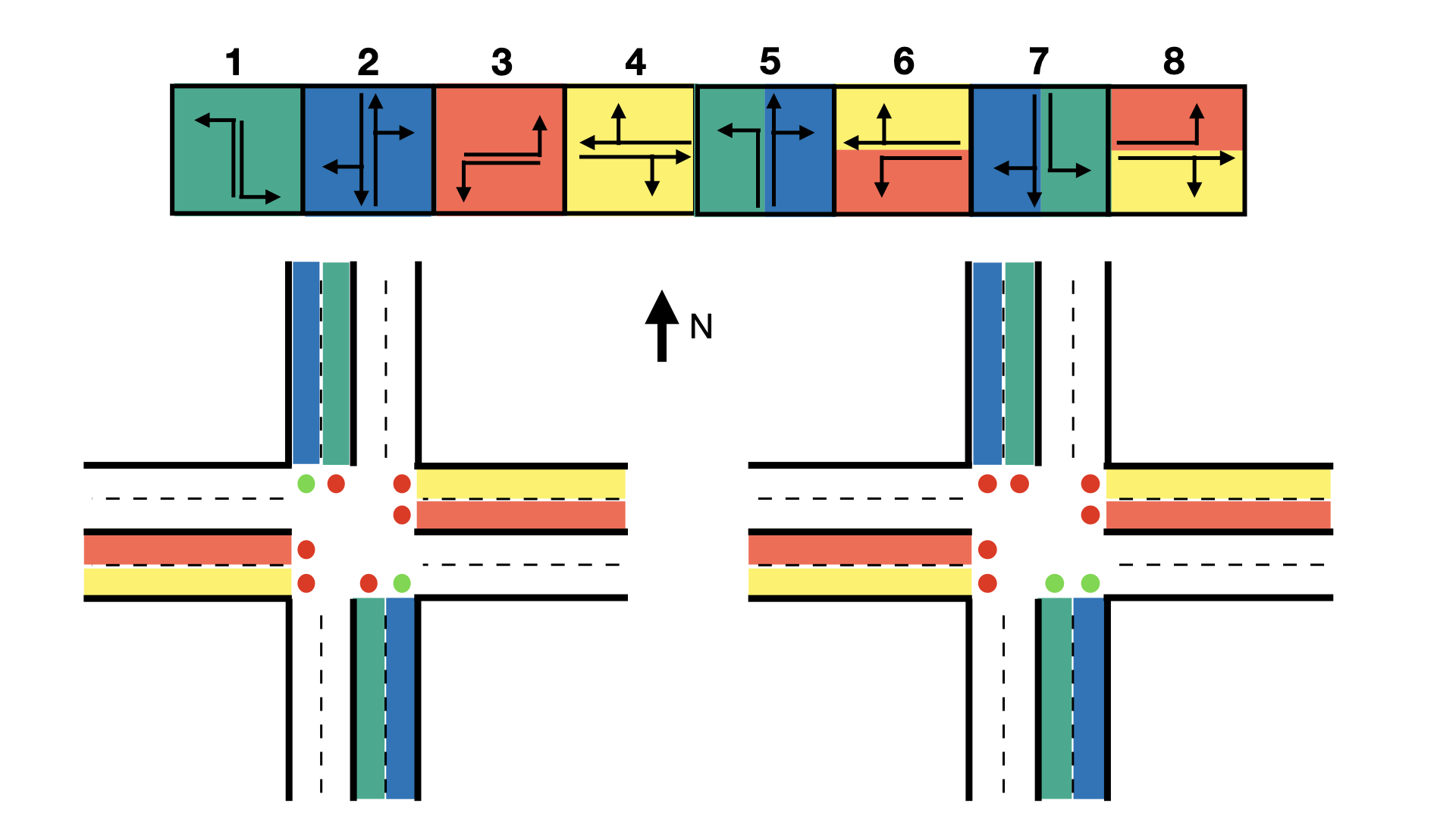}
    \centering
  \caption{\emph{Top}: 8 signal phases; \emph{Left}: phase \#2 illustration; \emph{Right}: phase \#5 illustration.
%   Example of a normal traffic phase on the left and example of a pre-emption phase on the right. All traffic phases are presented at the bottom.
  }
  \label{fig_phases}
  \end{figure}

%Traffic signal phase
A traffic signal phase is defined as the set of permissible traffic movements. 
As shown in Fig.~\ref{fig_phases}, an intersection with 4 links has 8 phases.
% During the normal traffic phases as shown in Fig.\ref{fig_movements}, lanes highlighted with the same color allow traffic movements in the opposite directions. 
% Only one pair of traffic movements is allowed anytime during the normal phases when there is no EMV, see Fig. \ref{fig_normal_phases}.
%   \begin{figure}[b]
%     \includegraphics[width=\linewidth]{LaTeX/images/fig_pressure_example.png}
%     \centering
%   \caption{Calculation of the lane pressure for Lane 2.}
%   \label{fig_pressure_example}
%   \end{figure}
%
% \subsection{Pre-emption phase} 
% When an EMV on duty approaches an intersection, however, both lanes along the direction of EMV traveling would have green lights to facilitate the intra-link passage of the EMV. We denote such phases as coordination phases.
%
\subsection{Pressure}
% Pressure of an incoming lane
The pressure of an incoming lane $l$ measures the unevenness of vehicle density between lane $l$ and corresponding out going lanes in permissible traffic movements. The vehicle density of a lane is $x(l)/x_{max}(l)$, where $x(l)$ is the number of vehicles on lane $l$ and $x_{max}(l)$ is the vehicle capacity on lane $l$, which is related to the length of a lane. Then the pressure of an incoming lane $l$ is 
\begin{equation} 
    w(l) = \left|\frac{x(l)}{x_{max}(l)} - \sum_{\{m|(l, m)\in \mathcal{M}\}}\frac{1}{h(m)}\frac{x(m)}{x_{max}(m)}\right|,
    \label{eqn:lane_pressure}
\end{equation}
where $h(m)$ is the number of lanes of the outgoing link which contains $m$. In Fig.~\ref{fig_movements}, $h(m)=2$ for all the outgoing lanes. An example for Eqn.~\eqref{eqn:lane_pressure} is shown in Fig.~\ref{fig_movements}.

Taking the intersection's traffic conditions shown in Fig.\ref{fig_movements} as an example, assuming the maximum capacity for each lane is 5 vehicles, we can calculate the lane pressure for lane 2 to be
\begin{equation*}
    w(2) = \left| \underbrace{\frac{1}{5}}_{\textrm{lane 2}} - \frac{1}{2}(\underbrace{\frac{1}{5} + \frac{2}{5}}_{\textrm{lane 3 and 4}}) - \frac{1}{2}(\underbrace{\frac{3}{5} + \frac{0}{5}}_{\textrm{lane 5 and 6}})\right| = \frac{2}{5}
\end{equation*}

% The pressure $P$ of an intersection is the average of the pressure of all incoming lanes, i.e. $P = \sum w(l)$. In the example shown in Fig.\ref{fig_movements}, the intersection pressure is $\frac{25}{10}$.

The pressure of an intersection indicates the unevenness of vehicle density between incoming and outgoing lanes in an intersection. Intuitively, reducing the pressure leads to more evenly distributed traffic, which indirectly reduce congestion and average travel time of vehicles. EMVLight defines pressure of an intersection in EMVLight as the average of the pressure of all incoming lanes,
\begin{equation*}\label{eq:EMVLight_pressure}
    P_{i} = \frac{1}{|\mathcal{I}_i|}\sum _{l\in \mathcal{I}_i} w(l),
\end{equation*}
where $\mathcal{I}_i$ represents the set of all incoming lanes of intersection $i$. According to such definition, the intersection pressure shown in Fig.\ref{fig_movements} is computed to be $\frac{25}{80}$.

% Here we present the key difference in pressure definition between our work and PressLight 
% \paragraph{Pressure in PressLight}
PressLight \cite{wei2019presslight} assumes that traffic movements are lane-to-lane, i.e., vehicles in one lane can only move into a particular lane in a link. Because of the lane-to-lane assumption, in PressLight, the pressure is defined per movement. PressLight defines the pressure of a movement as the difference of the vehicle density between an incoming lane $l$ and the outgoing lane $m$, i.e., 
\begin{equation*}
    w^{*}(l, m) = \frac{x(l)}{x_{max}(l)} - \frac{x(m)}{x_{max}(m)},
\end{equation*}
For instance, lane 2, shown in Fig.\ref{fig_movements}, carries $w^{*}(2, 3)$, $w^{*}(2, 4)$, $w^{*}(2, 5)$, $w^{*}(2, 6)$, $w^{*}(2, 7)$ and $w^{*}(2, 8)$. Taking the permissible traffic movement from lane 2 to lane 4 as an example, we can get the pressure for this movement as
\begin{equation*}
    w^{*}(2, 4) = \frac{1}{5} - \frac{2}{5} = -\frac{1}{5}
\end{equation*}

PressLight then defines the pressure of an intersection $i$ as the absolute value of the sum of pressure of movements of intersection $i$, i.e., 
When calculating the pressure intersection, PressLight has
\begin{equation*}\label{eq:PressLight_pressure}
    P^{*}_{i} = \left|\sum _{(l, m)\in \mathcal{M}_i} w^{*}(l, m)\right|,
\end{equation*}
where $\mathcal{M}_i$ is the set of permissible traffic movements of intersection $i$. According to PressLight's definition of intersection pressure, Fig.\ref{fig_movements} shows an intersection with pressure of 6.

\paragraph{Pressure in our work}
EMVLight assumes a lane-to-link style traffic movement as vehicles can enter either lane on the target link, see Fig.\ref{fig_movements}. 
Following lane pressure defined in \eqref{eqn:lane_pressure},

\paragraph{Comparison}
The first difference between the two definitions is that $w^{*}(l, m)$ can be both positive or negative, but $w(l)$ can only take positive values that measures the unevenness of the vehicle density in the incoming lane and that of the corresponding outgoing lanes. We take the absolute value since the direction of pressure is irrelevant here, and the goal of each agent is to minimize this unevenness. The second difference is that at the intersection level, $P^{*}_{i}$ takes a sum but $P_{i}$ takes an average. The average is more suitable for our purpose since it scales the pressure down and the unit penalty for normal agents would be relatively large as compared to rewards for pre-emption agents (Eqn. (3)).  This design puts the efficient passage of EMV vehicles at the top priority. Our experimentation results, presented in Sec.\ref{sec_result} indicate the proposed pressure design produces a more robust reward signal during training and outperforms PressLight in congestion reduction.

\subsection{Emergency capacity}

\begin{figure}[ht]
\includegraphics[width=0.85\linewidth]{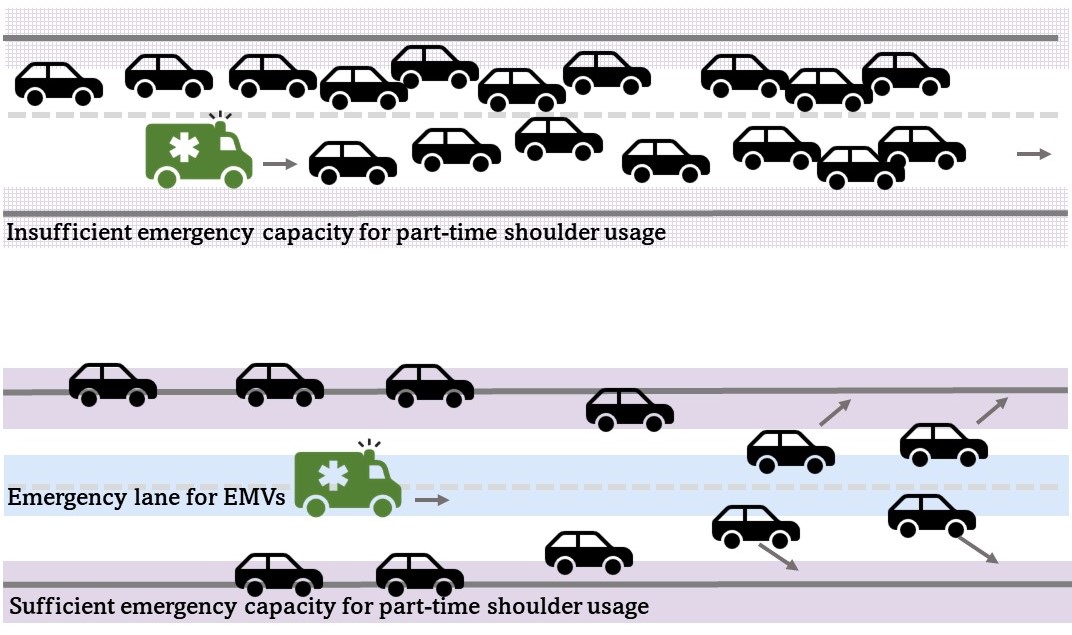}
\centering
\caption{A demonstration of an emergency lane for EMV passage. \emph{Top}: the EMV has to follow the congested queue due to insufficient emergency capacity;
\emph{Bottom}: the EMV is traversing on the yielded emergency lane due to sufficient emergency capacity.}
\label{fig_yielding}
\end{figure}

A roadway segment may have additional capacity, e.g. shoulder, parking lanes, bike lanes, dedicated to providing extra space that can be used under emergencies and incidents.
In the presence of EMVs, existing vehicles are allowed to pull-over or park on the shoulders temporarily, forming an \emph{emergency lane} for the emergency vehicle to pass. An emergency lane is an lane formed between original lanes dedicated for EMV passage, see Fig.\ref{fig_yielding} bottom. EMVs are assumed to travel freely on the emergency lane through the dense traffic. This is referred as an \emph{emergency yielding} and non-EMVs are experiencing an \emph{part-time shoulder use} \cite{part-time-shoulder-use}.

Adaptive traffic management strategies based on part-time shoulder use, such as dynamic hard shoulder running (D-HSR) \cite{Jiaqi2016Dynamic}, have proven beneficial and cost-effective for such scenarios. 

% Whether or not such an emergency lane for EMV passage can be established depends on several factors such as the segment's shoulder width, lane width, geometric clearance and other factors. 
We use an intuitive mathematical model to determine whether such an emergency lane for EMV passage can be established. 
First, we define \emph{emergency capacity} $C_{i}^{\textrm{EC}}$ of a link $i$ to be the additional capacity in the link for emergencies and incidents. The emergency capacity depends on the segment's shoulder width, lane width, geometric clearance and other factors. We say that a link is \emph{emergency-capacitated} if it has a nonzero emergency capacity.
% A larger emergency capacity enables an emergency lane for EMVs even when the road is more crowded. 
In order to express the maximum number of vehicles allowed in a link, for forming an emergency lane, we assume there are $n_i$ non-EMVs in the link $i$ with an average speed of $s_i$. We also assume that the normal capacity $k_i$ is evenly distributed among $l_i$ lanes so that each lane has a capacity of $k_i/l_i$. 
The overall capacity of the link is then $k_i + C_{i}^{\textrm{EC}}$. 
To form an emergency lane for EMV passage with maximum speed, all the non-EMVs need to move out of the emergency lane. Thus the maximum number of vehicles allowed is $k_i + C_{i}^{\textrm{EC}}- k_i/l_i$. As a result, the travel speed of the EMV is 
% Mathematically, the speed of an EMV passing through a particular segment $i$ can be determined. Without making oversimplified assumptions, we observe $n_{i}$ non-EMVs downstream with an average speed of $s_i$. 
% As an attribute of the segment, the normal capacity $k_i$ is uniformly distributed over $l_i$ lanes, resulting an average lane capacity of $\frac{k_i}{l_i}$ for this segment. Denoting $C_{i}^{\textrm{EC}}$ as the emergency capacity for this segment, we are able to approximate the travel speed of the EMV as
\begin{equation}\label{eqn:EMV_speed}
s_{i}^{\textrm{EMV}} = \begin{cases}
s_{f} & n_{i} \leq k_{i} + C_{i}^{\textrm{EC}} - \frac{k_i}{l_i},\\
s_{i} & \textrm{else,}
\end{cases}
\end{equation}
where $s_{f}$ represents the maximum speed allowed for EMVs.
% 10 <= 10 + 5 - 10 / 2
% 10 <= 10 + 4 - 10 / 2
% 

\section{Methodology}\label{sec_methodology}
In this section, we elaborate the methodology of EMVLight. We begin with implementing a decentralized shortest path onto EMV navigation in traffic networks, and then incorporate it into the proposed multi-class RL agent design. Subsequently, we introduce the multi-agent advantage actor-critic framework as well as the RL training workflow in details.
\subsection{Decentralized Routing for EMVLight}
Dijkstra's algorithm is an algorithm that finds the shortest path between a given node and every other nodes in a graph. The time-based Dijkstra's algorithm finds the fastest path and has been widely used for EMV routing. In order to find such a path, the EMV travel time along each link need to be estimated first and we refer to it as the \emph{intra-link travel time.}
Dijkstra's algorithm takes as input the traffic graph, the intra-link travel time and a destination, and can return the time-based shortest path as well as estimated travel time from each intersection to the destination. The latter is usually referred to as the \emph{estimated time of arrival} (ETA) of each intersection.

% Pre-determined route-based signal pre-emption techniques often yield sub-optimal paths due to stochastic traffic dynamics, and overall network-level traffic conditions are significantly exacerbated.
% Navigating emergency vehicles in congested traffic networks relies on dynamically finding the time-based shortest path, \dz{meaning routing decisions are updated every time they pass an intersection. Todo: inaccurate, decision distance.}

% \subsection{Estimation of Intra-link EMV Travel Time}
% After reviewing very limited literature on the investigation of traffic impacts on intra-link segment, we propose a meso-scopic approach based on the fundamental diagram to estimate the intra-link travel time for EMVs. See Figure. 
% \subsection{Dynamic Dijkstra's algorithm for EMV routing}

In a traffic network, the intra-link travel time usually depends on the link's emergency capacity and number of vehicles in that link. In our model, this dependency is captured by EMV speed Eqn.~\eqref{eqn:EMV_speed}. The intra-link travel time is then calculated as the link length divided by the EMV speed. 
However, traffic conditions are constantly changing and so does EMV travel time along each link. Moreover, EMV pre-emption techniques alter traffic signal phases, which will significantly change the traffic condition as the EMV travels. The pre-determined shortest path might become congested due to stochasticity and pre-emption. Thus, updating the optimal route dynamically can facilitate EMV passage. One option is to run Dijkstra's algorithm repeatedly as the EMV travels through the network in order to take into account the updated EMV intra-link travel time.
\hs{However, this requires global traffic information of the entire traffic network throughout EMVs' trips. Even when a centralized controller is established to navigate the EMV, the synchronization and communication cost grows exponentially when the network size increases and the nonscability disallows the centralized scheme to be real-time for the navigation. 

Decentralized routing approaches \cite{ADACHER2014routeguidance, Chen2006Riskaverse,He2015Kshortest} were introduced to find the shortest path with partial observability of the system. However, these models require network decomposition or partitioning in advance, and solve optimal paths with polynomial-bounded time complexity at best \cite{JOHNSON2016partitioning}. Some other approach heavily relies on V2I communication \cite{Mostafizi2021decentralized}. Considering the massive amount of iterations of trial-and-error in reinforcement learning, none of these decentralized routing methods provide a suitable design for a learning-based framework.
}

To achieve efficient decentralized dynamic routing, we extend Dijkstra's algorithm to update the optimal route based on the updated intra-link travel times. As shown in Algorithm~\ref{alg:ETA_prepopulation}, first a pre-populating process is carried out where a standard Dijkstra's algorithm is run to get the $\mathsf{ETA}^0$ from each intersection to the destination. For each intersection, the next intersection $\mathsf{Next}^0$ along the shortest path is also calculated. For an intersection $i$, the result $\mathsf{ETA}_i^0$ and $\mathsf{Next}_i^0$ are stored locally in the intersection. We assume this process can be done before the EMV leaves the dispatching hub. This is reasonable since a sequence of processes, including call-taker processing, are performed before the EMVs are dispatched. Once the pre-populating process is finished, we can update $\mathsf{ETA}$ and $\mathsf{Next}$ for each intersection efficiently in parallel in a decentralized way, since the update only depends on information of neighboring intersections. 
\begin{figure}[h]
\includegraphics[width=\linewidth]{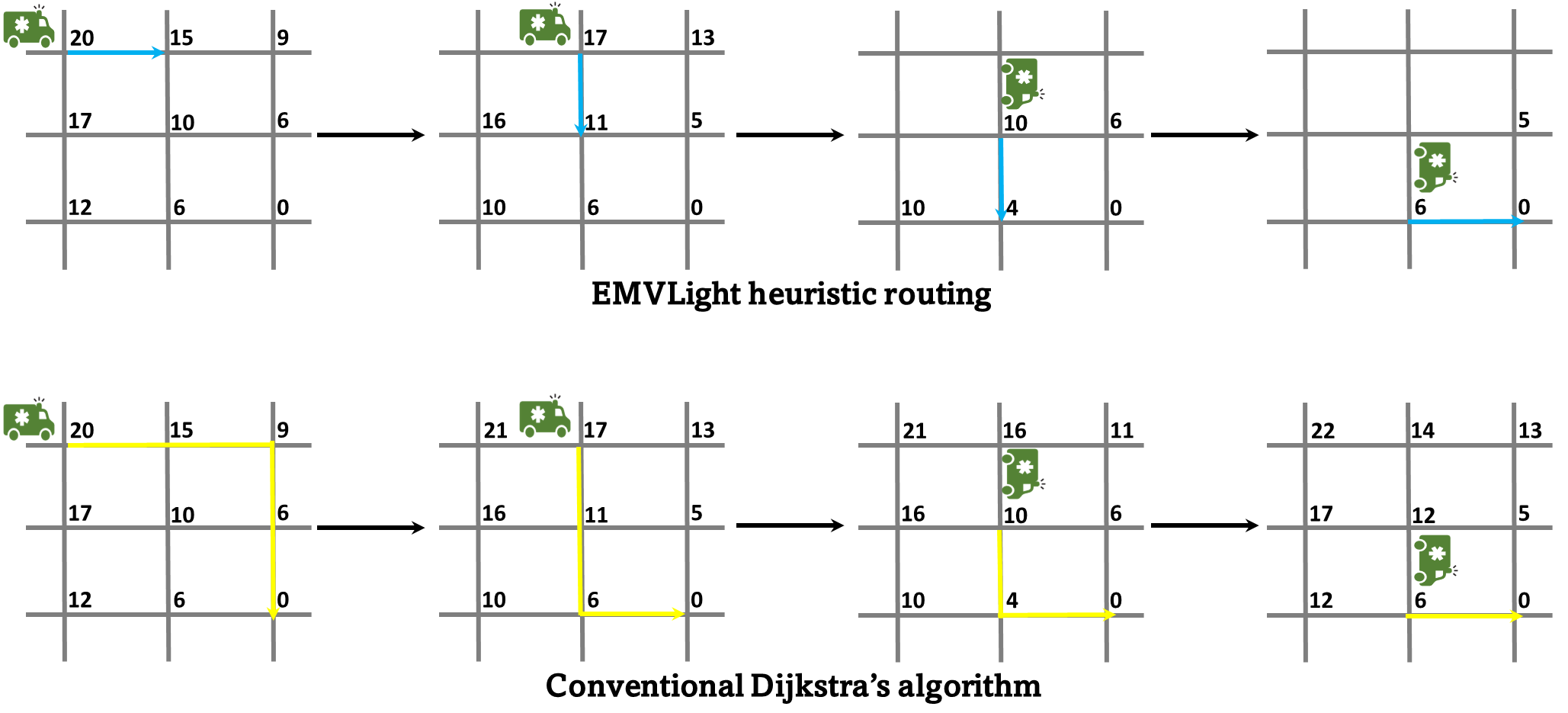}
\centering
\caption{EMVLight routing (top) vs conventional Dijkstra's routing (bottom).}
\label{fig_EMVLight_routing}
\end{figure}
\hs{
Fig. \ref{fig_EMVLight_routing} provides an example comparing between Algorithm.\ref{alg:ETA_prepopulation} and conventional Dijkstra's algorithm on an 3-by-3 traffic network. The numerical value represents the $\mathsf{ETA}$ of each intersection, and it gets updated as described above. EMVLight, rather than solving the full shortest path like the conventional Dijkstra's algorithm, only determines the next link to travel each iteration. A comparison of the worst-case time complexity between the EMVLight routing heuristic and a Dijkstra-based dynamic shortest path method is provided in Table\ref{tab_routing_comparison}. Notice that the Dijkstra' algorithm can be implemented with a Fibonacci heap min-priority queue and solve the shortest path with a time complexity of $\mathcal{O}(|\mathcal{V}|\log{}|\mathcal{V}| + |\mathcal{E}|)$ \cite{Fredman1984Fibonacci}.
\begin{table}[h]
\centering
\fontsize{10.0pt}{10.0pt} \selectfont
\begin{tabular}{@{}ccc@{}}
\toprule
                     & EMVLight heuristic & Dynamic shortest-path based on Dijkstra's \\ \midrule
Initialization & $\mathcal{O}(|\mathcal{V}|\log{}|\mathcal{V}| + |\mathcal{E}|)$    &   -                          \\
updating &   $\mathcal{O}(|\mathcal{V}|)$        &   $\mathcal{O}(|\mathcal{V}|\log{}|\mathcal{V}| + |\mathcal{E}|)$      \\
updating frequency &   $|\mathcal{V}|$  & $M$  \\ \bottomrule
\end{tabular}
\caption{Time complexities of the proposed routing heuristic and the dynamic shortest-path approach. $M$ can be arbitrarily set to determine the updating frequency. The larger $M$ is, the shorter selected route can be.}
\label{tab_routing_comparison}
\end{table}

By adopting the proposed heuristic routing algorithm, we facilitate the RL agent design, which is introduced in Sec. \ref{sec:agent_design}.
}

\begin{algorithm}[t]
    \caption{Decentralized routing for EMVs}
    \label{alg:ETA_prepopulation}
    \SetEndCharOfAlgoLine{}
    \SetKwInOut{Input}{Input}
    \SetKwInOut{Output}{Output}
    \SetKwData{ETA}{ETA}
    \SetKwData{Next}{Next}
    \SetKwFor{ParrallelForEach}{foreach}{do (in parallel)}{endfor}
    \Input{\\\hspace{-3.7em}
        \begin{tabular}[t]{l @{\hspace{3.3em}} l}
        $G=(\mathcal{V}, \mathcal{E})$ & traffic map as a graph \\
        $T^t = [T_{ij}^t]$ & intra-link travel time at time $t$ \\
        $i_d$  & index of the destination
        \end{tabular}
    }
    \Output{\\\hspace{-3.7em}
        \begin{tabular}[t]{l @{\hspace{1.5em}} l}
        $\mathsf{ETA}^t = [\mathsf{ETA}^t_i]$ & ETA of each intersection \\
        $\mathsf{Next}^t = [\mathsf{Next}^t_i]$ & next intersection to go \\
        & from each intersection
        \end{tabular}
    }
    \tcc{pre-populating}
    % \ForEach{$i \in V\setminus \{d\}$}{
    %     \ETA$_i^0 \gets \infty$\;
    %     \Next$_i^0 \gets \text{Undefined}$\;
    % }
    $\mathsf{ETA}^0, \mathsf{Next}^0$ $=$ \texttt{Dijkstra}$(G, T^0, i_d)$\;
    \tcc{dynamic routing}
    \For{$t = 0 \to T$}{
        \ParrallelForEach{$i \in \mathcal{V}$}{
            $\mathsf{ETA}_i^{t+1} \gets \min_{(i, j)\in \mathcal{E}} (\mathsf{ETA}_j^t + T_{ji}^t)$\;
            $\mathsf{Next}_i^{t+1} \gets \arg\min_{\{j|(i, j)\in \mathcal{E}\}}(\mathsf{ETA}_j^{t} + T_{ji}^t$)\;}}
\end{algorithm}
\begin{remark}
    In static Dijkstra's algorithm, the shortest path is obtained by repeatedly querying the $\mathsf{Next}$ attribute of each node from the origin until we reach the destination. In our dynamic Dijkstra's algorithm, since the shortest path changes, at a intersection $i$, we only care about the immediate next intersection to go to, which is exactly $\mathsf{Next}_i$.
\end{remark}

\subsection{Reinforcement Learning Agent Design}\label{sec:agent_design}
While dynamic routing directs the EMV to the destination, it does not take into account the possible waiting times for red lights at the intersections. Thus, traffic signal pre-emption is also required for the EMV to arrive at the destination in the least amount of time. However, since traditional pre-emption only focuses on reducing the EMV travel time, the average travel time of non-EMVs can increase significantly. Thus, we set up traffic signal control for efficient EMV passage as a decentralized RL problem. \modi{In our problem, each intersection is an RL agent. Each agent makes decisions on the traffic signal phases of this intersection based on local information. Aside from communicating with the approaching EMV, agents are also communicating with each other.} Multiple agents coordinate the control signal phases of intersections cooperatively to \textbf{(1)} reduce EMV travel time and \textbf{(2)} reduce the average travel time of non-EMVs. First we design 3 agent types. Then we present agent design and multi-agent interactions.

\subsubsection{Types of agents for EMV passage}
\begin{figure}[h]
    \centering
    \includegraphics[width=\linewidth]{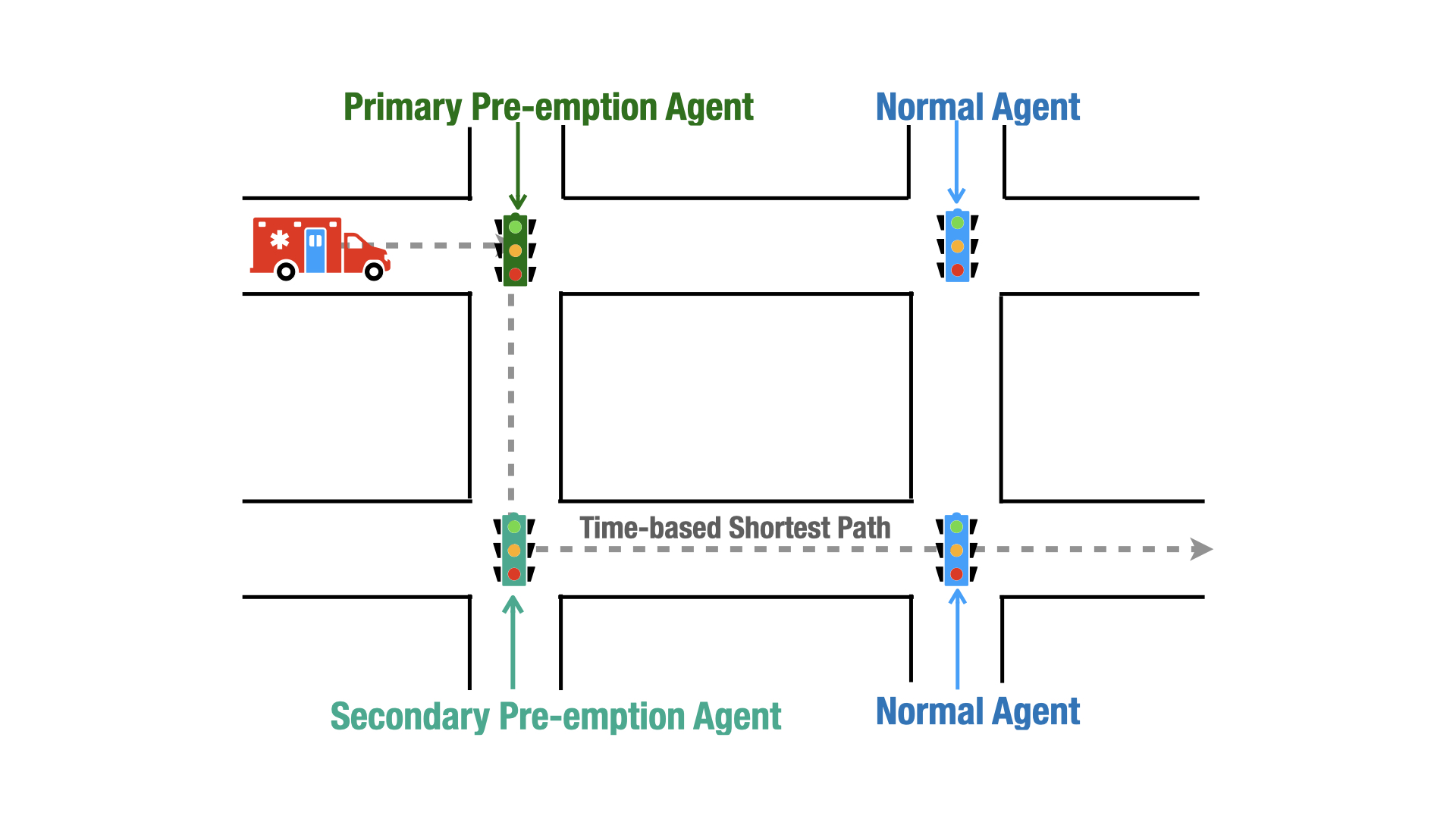}
  \caption{Types of agents through the EMV's passage to the destination.}
  \label{fig_secondary}
\end{figure}

% pre-emption phases can help it pass through an intersection more quickly than normal phases. This motivates us to
When an EMV is on duty, we distinguish 3 types of traffic control agents based on EMV location and routing (Fig.~\ref{fig_secondary}). An agent is a \emph{primary pre-emption agent} $i_p$ if an EMV is on one of its incoming links. The agent of the next intersection $i_s = \mathsf{Next}_{i_p}$ is refered to as a \emph{secondary pre-emption agent}.
The rest of the agents are \emph{normal agents}. We design these types since different agents have different local goals, which is reflected in their reward designs. 
% We assume normal agents can only execute normal phases, while primary and secondary pre-emption agents can execute both normal phases and pre-emption phases. 

\subsubsection{Agent design}
%\item \textbf{State}: The local state of an agent $i$ includes the number of vehicles on each outgoing lanes and incoming lanes, the distance of the EMV to the intersection, the estimated time of arrival and which link the EMV will be routed to:
\textbf{State}: The state of an agent $i$ at time $t$ is denoted as $s^t_i$ and it includes the number of vehicles on each outgoing lanes and incoming lanes, the distance of the EMV to the intersection, the estimated time of arrival ($\mathsf{ETA}$), and which link the EMV will be routed to ($\mathsf{Next}$), i.e.,
\begin{equation}
    s^t_i = \{x^t(l), x^t(m),  d^t_{\text{EMV}}[L_{ji}], \mathsf{ETA}^{t'}_i, \mathsf{Next}^{t'}_i \},%_{ji, im\in E, l \in L_{ji}},
\end{equation}
where $L_{ji}$ represents the links incoming to intersection  $i$ from its adjacent intersections $j\in\mathcal{N}_{i}$. With a slight abuse of notation, $l$ and $m$ denote the set of incoming and outgoing lanes, respectively. The vector $d^t_{\text{EMV}}$ contains the information about the distance of an EMV to the intersection is an EMV is present. For the intersection shown in Fig.~\ref{fig_movements}, $d^t_{\text{EMV}}$ is a vector of four elements. In particular, for primary pre-emption agents, one of the elements represents the distance of EMV to the intersection in the corresponding link and the rest of the elements are set to -1. For all other agents, $d^t_{\text{EMV}}$ are padded with -1. 
%where $L_{ji}$ represents the links incoming to intersection  $i$, $l$ includes all incoming lanes, i.e., $ji\in \mathcal{E}, l\in L_{ji}$, and $m$ includes all outgoing lanes, i.e., $ik\in \mathcal{E}, m \in L_{ik}$. For grid networks, $d^t$ contains four elements. For primary pre-emption agents, one of the elements represents the distance of EMV to the intersection in the corresponding link. The rest of the elements are set to -1. For all other agents, $d^t$ are padded with -1. 

\textbf{Action}: Prior work has focused on using phase switch, phase duration and phase itself as actions. In this work, we define the action of an agent as one of the 8 phases in Fig.~\ref{fig_phases}; this enables more flexible signal patterns as compared to the traditional cyclical patterns. 
Due to safety concerns, once a phase has been initiated, it should remain unchanged for a minimum amount of time, e.g. 5 seconds. Because of this, we set our MDP time step length to be 5 seconds to avoid rapid switch of phases. 
% We also assume that once a phase is executed, it must remain unchanged for at least a fixed number of time steps to avoid rapid phase changing.

\textbf{Reward}: PressLight has shown that minimizing the pressure is an effective way to encourage efficient vehicle passage. For normal agents, we adopt a similar idea, as shown in Eqn.~\ref{eqn:reward1}. For secondary pre-emption agents, we additionally encourage less vehicles on the link where the EMV is about to enter in order to encourage efficient EMV passage. Thus, the reward is a weighted sum of the pressure and this additional term, with a weight $\beta$, as shown in Eqn.~\ref{eqn:reward2}. For a default setting of balancing EMV navigation and traffic congestion alleviation, we choose $\beta=0.5$. For primary pre-emption agents, we simply assign a unit penalty at each time step to encourage fast EMV passage, as shown in Eqn.~\ref{eqn:reward3}. Thus, depending on the agent type, the local reward for agent $i$ at time $t$ is as follows
% Each intersection is controlled by an agent and each agent has partial observation of the entire traffic network. By proactively determining the traffic signal phases of each agent, we are trying to \textbf{(1)} dynamically navigate the EMV so it can arrive at the incident scene as soon as possible, and to\textbf{(2)} disrupt the original traffic flow as little as possible.
\begin{subnumcases}{r_{i}^{t} = \label{eqn:reward}}
  -P_{i}^{t} & $i \notin \{i_p, i_s\}$, \label{eqn:reward1}\\
  - \beta P_{i_s}^{t} - \frac{1-\beta}{|L_{i_pi_s}|}\sum\limits_{l\in L_{i_pi_s}} \frac{x(l)}{x_{max}(l)}  & $i=i_s,$ \label{eqn:reward2}\\
  -1 & $i=i_p$. \label{eqn:reward3} 
\end{subnumcases} 
\textbf{Justification of agent design.} The quantities in local agent state can be obtained at each intersection using various technologies. Numbers of vehicles on each lane $(x^t(l), x^t(m))$ can be obtained by vehicle detection technologies, such as inductive loop \cite{gajda2001vehicle} based on the hardware installed underground. The distance of the EMV to the intersection $d^t_{EMV}[L_{ji}]$ can be obtained by \emph{vehicle-to-infrastructure} technologies such as VANET\cite{buchenscheit2009vanet}, which broadcasts the real-time position of a vehicle to an intersection. Prior work by \cite{wang2013design} and  \cite{noori2016connected} have explored these technologies for traffic signal pre-emption. 

The dynamic routing algorithm (Algorithm~\ref{alg:ETA_prepopulation}) can provide $(\mathsf{ETA}, \mathsf{Next})$ for each agent at every time step. However, due to the stochastic nature of traffic flows, updating the route too frequently might confuse the EMV driver, since the driver might be instructed a new route, say, every 5 seconds. 
% Moreover, different route might lead to frequent change of agent types. For example, an agent might be a secondary pre-emption agent after one update and be a normal agent a few seconds later. This is undesirable since the local reward for different agents are different. 
There are many ways to ensure reasonable frequency. One option is to inform the driver only once while the EMV travels on a single link. We implement it by updating the state of a RL agent $(\mathsf{ETA}^{t'}_i, \mathsf{Next}^{t'}_i)$ at the time step when the EMV travels through half of a link. For example, if the EMV travels through a link to agent $i$ from time step 11 to 20 in constant speed, then dynamic routing information in $s_i^{16}$ to $s_i^{20}$ are the same, which is $(\mathsf{ETA}_i^{15}, \mathsf{Next}_i^{15})$, i.e., $t'=15$.
% We use the notation $t'$ to indicate that $s_i^t$ might contain these information from different time steps.

As for the reward design, one might wonder how an agent can know its type. As we assume an agent can observe the state of its neighbors, agent type can be inferred from the observation. This will become clearer in Section~\ref{sec:MA2C}.

\subsection{Multi-agent Advantage Actor-critic}
\label{sec:MA2C}
We adopt a multi-agent advantage actor-critic (MA2C) framework similar to \cite{chu2019multi} to address the coupling of EMV navigating and traffic signal control simultaneously in a decentralized manner. The difference is that our local state includes dynamic routing information and our local reward encourages efficient passage of EMV. Here we briefly introduce the MA2C framework.

In a multi-agent network $G(\mathcal{V}, \mathcal{E})$, the neighborhood of agent $i$ is denoted as $\mathcal{N}_i = \{ j | ji\in \mathcal{E} \textrm{ or } ij\in \mathcal{E}\}$. The local region of agent $i$ is $\mathcal{V}_i = \mathcal{N}_i \cup i$. We define the distance between two agents $d(i, j)$ as the minimum number of edges that connect them. For example, $d(i, i) = 0$ and $d(i, j)=1, \forall j \in \mathcal{N}_i$. In MA2C, each agent learns a policy $\pi_{\theta_i}$ (actor) and the corresponding value function $V_{\phi_i}$ (critic), where ${\theta_i}$ and ${\phi_i}$ are learnable neural network parameters of agent $i$.

\textbf{Local Observation.} In an ideal setting, agents can observe the states of every other agent and leverage this global information to make a decision. However, this is not practical in our problem due to communication latency and will cause scalability issues. We assume an agent can observe its own state and the states of its neighbors, i.e., $s^t_{\mathcal{V}_i} = \{s^t_j|j\in \mathcal{V}_i\}$. The agents feed this observation to its policy network $\pi_{\theta_i}$ and value network $V_{\phi_i}$.

% $\pi_{\theta_i}$ to decide signal phases. As for the value network, we leverage the spatial discount factor $\alpha$ to let the value function focus less on neighboring agents. Thus, the adjusted observation that are fed into the value network $V_{\phi_i}$ is $s^t_{\mathcal{V}_i} = s^t_i \cup \{ s^t_j|j\in \mathcal{N}_i \}$.

\textbf{Fingerprint.} In multi-agent training, each agent treats other agents as part of the environment, but the policy of other agents are changing over time. \cite{foerster2017stabilising} introduce \emph{fingerprints} to inform agents about the changing policies of neighboring agents in multi-agent Q-learning. \cite{chu2019multi} bring fingerprints into MA2C. Here we use the probability simplex of neighboring policies $\pi^{t-1}_{\mathcal{N}_i} = \{\pi^{t-1}_j|j\in \mathcal{N}_i\}$ as fingerprints, and include it into the input of policy network and value network. Thus, our policy network can be written as $\pi_{\theta_i}(a_i^t|s^t_{\mathcal{V}_i}, \pi^{t-1}_{\mathcal{N}_i})$ and value network as $V_{\phi_i}(s^t_{\mathcal{V}_i}, \pi^{t-1}_{\mathcal{N}_i})$, where $s^t_{\mathcal{V}_i}$ is the local observation with spatial discount factor introduced below.

\textbf{Spatial Discount Factor and Adjusted Reward.} MA2C agents cooperatively optimize a global cumulative reward. We assume the global reward is decomposable as $r_t = \sum_{i\in \mathcal{V}} r^t_i$, where $r^t_i$ is defined in Eqn.~\eqref{eqn:reward}. Instead of optimizing the same global reward for every agent, here we employ the spatial discount factor $\alpha$, introduced by \cite{chu2019multi}, to let each agent pay less attention to rewards of agents farther away. The adjusted reward for agent $i$ is 
\begin{equation}
    \Tilde{r}_i^t = \sum_{d=0}^{D_i}\Big( \sum_{j\in\mathcal{V}|d(i, j)=d} (\alpha)^d r^t_j\Big),
\end{equation}
where $D_i$ is the maximum distance of agents in the graph from agent $i$. When $\alpha > 0$, the adjusted reward include global information, it seems this is in contradiction to the local communication assumption. However, since reward is only used for offline training, global reward information is allowed. Once trained, the RL agents can control a traffic signal without relying on global information. 

\textbf{Temporal Discount Factor and Return.} 
The local return $\Tilde{R}^t_i$ is defined as the cumulative adjusted reward $\Tilde{R}^t_i := \sum_{\tau=t}^T \gamma^{\tau-t} \Tilde{r}^\tau_i$, where $\gamma$ is the temporal discount factor and $T$ is the length of an episode. We can estimate the local return using value function,
\begin{equation}
    \Tilde{R}^t_i = \Tilde{r}^t_i + \gamma V_{\phi_i^-}(s^{t+1}_{\mathcal{V}_i}, \pi^{t}_{\mathcal{N}_i}|\pi_{\theta_{-i}^-}),
\end{equation}
where $\phi_i^-$ means parameters $\phi_i$ are frozen and $\theta_{-i}^-$ means the parameters of policy networks of all other agents are frozen. 

\begin{figure}[h]
    \centering
    \includegraphics[width=0.9\linewidth]{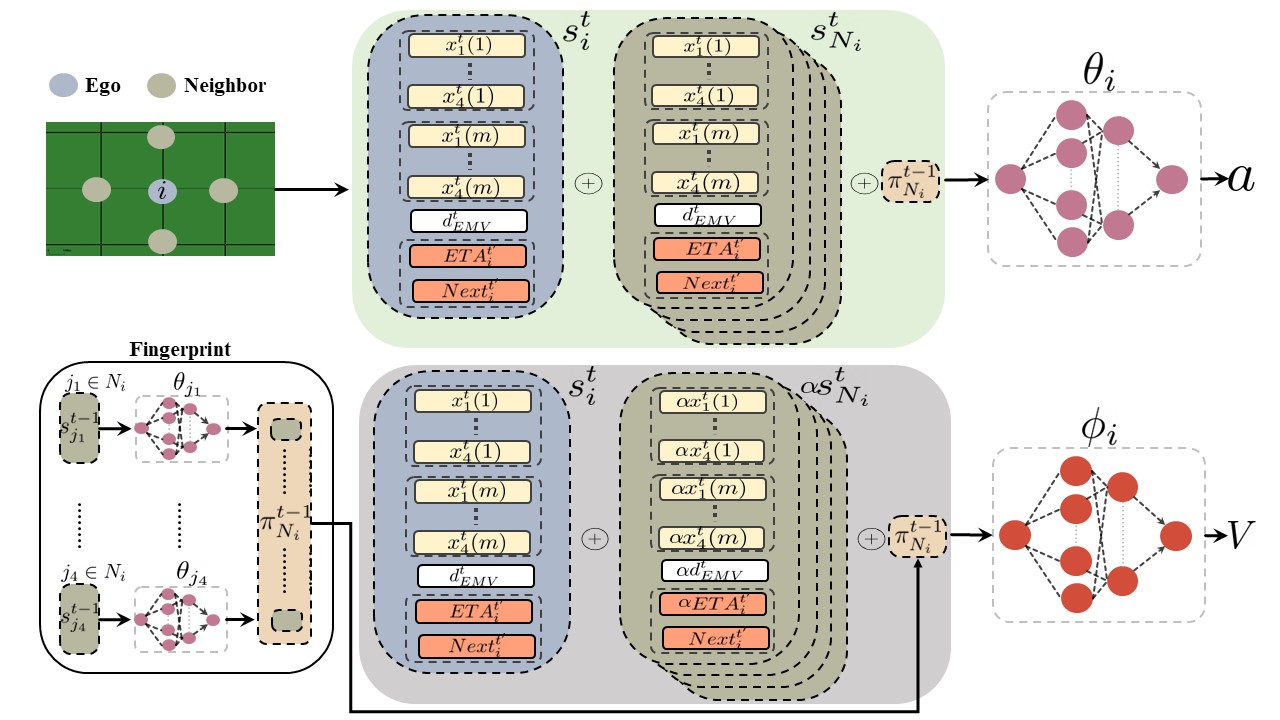}  
    \caption{Overview of MA2C framework for EMVLight's navigation and traffic signal control.}
    \label{fig_model}
\end{figure}

\textbf{Network architecture and training.} 
As traffic flow data are spatial temporal, we leverage a long-short term memory (LSTM) layer along with fully connected (FC) layers for policy network (actor) and value network (critic). Fig. \ref{fig_model} provides an overview for the MA2C frameworks for EMVLight.
% We provide neural architecture details, policy loss expression,  value loss expression as well as a training pseudocode in the Appendix.

% \section{Network Training}
% \begin{table}[]
% \centering
% \begin{tabular}{@{}cc@{}}
% \toprule
% Hyper-parameters                & Value    \\ \midrule
% Optimizer                       & Adam     \\
% temporal discount factor $\gamma$ & 0.99     \\
% spatial discounted factor  $\gamma_{spatial}$    & 0.90     \\
% batch size                      & 1000     \\
% Initial learning rate           & 1e-3     \\
% learning rate decay             & constant \\
% MDP step length  $\Delta t$              & 5s       \\
% $\beta$ for secondary pre-emption & 0.5      \\
% entropy coefficient              & 0.01     \\ \bottomrule
% \end{tabular}
% \caption{Hyper-parameters used for training.}
% \label{tab_hyperparameters}
% \end{table}
\textbf{Value loss function}
With a batch of data $B = \{(s_i^t, \pi_i^t, a_i^t, s_i^{t+1}, r_i^t)_{i\in \mathcal{V}}^{t\in \mathcal{T}}\}$, each agent's value network is trained by minimizing the difference between bootstrapped estimated value and neural network approximated value
\begin{equation}
    \label{eqn:L_v}
    \mathcal{L}_v(\phi_i) = \frac{1}{2|B|} \sum_{B}\Big( \Tilde{R}^t_i - V_{\phi_i}(s^t_{\mathcal{V}_i}, \pi^{t-1}_{\mathcal{N}_i}) \Big)^2.
\end{equation}

\textbf{Policy loss function}
Each agent's policy network is trained by minimizing its policy loss
\begin{align}
    \label{eqn:L_p}
    \mathcal{L}_p(\theta_i) = -& \frac{1}{|B|}\sum_{B} \bigg(\ln \pi_{\theta_i}(a_i^t|s^t_{\mathcal{V}_i}, \pi^{t-1}_{\mathcal{N}_i}) \Tilde{A}^t_i \\
    &- \lambda \sum_{a_i \in \mathcal{A}_i} \pi_{\theta_i} \ln \pi_{\theta_i} (a_i | s^t_{\mathcal{V}_i}, \pi^{t-1}_{\mathcal{N}_i}) \bigg),
\end{align}
where $\Tilde{A}^t_i = \Tilde{R}^t_i - V_{\phi_i^-}(s^t_{\mathcal{V}_i}, \pi^{t-1}_{\mathcal{N}_i})$ is the estimated advantage which measures how much better the action $a^t_i$ is as compared to the average performance of the policy $\pi_{\theta_i}$ in the state $s_i^t$. The second term is a regularization term that encourage initial exploration, where $\mathcal{A}_i$ is the action set of agent $i$. For an intersection as shown in Fig. 1, $\mathcal{A}_i$ contains 8 traffic signal phases.

\textbf{Training algorithm}
Algorithm \ref{alg:training} shows the multi-agent A2C training process. %
\setcounter{algocf}{0}
\renewcommand{\thealgocf}{S\arabic{algocf}}
\begin{algorithm}[ht]
    \caption{Multi-agent A2C Training}
    \label{alg:training}
    \SetEndCharOfAlgoLine{}
    \SetKwInOut{Input}{Input}
    \SetKwInOut{Output}{Output}
    \SetKwData{ETA}{ETA}
    \SetKwData{Next}{Next}
    \SetKwFor{ParrallelForEach}{foreach}{do (in parallel)}{endfor}
    \Input{\\\hspace{-3.7em}
        \begin{tabular}[t]{l @{\hspace{3.3em}} l}
        $T$ & maximum time step of an episode \\
        $N_{\mathrm{bs}}$ & batch size \\
        $\eta_\theta$  & learning rate for policy networks \\
        $\eta_\phi$    & learning rate for value networks \\
        $\alpha$       & spatial discount factor \\
        $\gamma$       & (temporal) discount factor\\ $\lambda$      & regularizer coefficient
        \end{tabular}
    }
    \Output{\\\hspace{-3.7em}
        \begin{tabular}[t]{l @{\hspace{1.4em}} l}
        $\{\phi_i\}_{i\in\mathcal{V}}$ & learned parameters in value networks \\
        $\{\theta_i\}_{i\in\mathcal{V}}$ & learned parameters in policy networks \\
        \end{tabular}
    }
    \textbf{initialize} $\{\phi_i\}_{i\in\mathcal{V}}$, $\{\theta_i\}_{i\in\mathcal{V}}$, $k \gets 0$, $B \gets \varnothing$;
    \textbf{initialize} SUMO, $t \gets 0$, \textbf{get} $\{s^0_i\}_{i\in\mathcal{V}}$\;
    \Repeat{Convergence}{
        \tcc{generate trajectories}
        \ParrallelForEach{$i \in \mathcal{V}$}{
            \textbf{sample} $a^t_i$ from $\pi^t_i$\;
            \textbf{receive} $\Tilde{r}^t_i$ and $s^{t+1}_i$\;
        }
        $B \gets B \cup \{(s_i^t, \pi_i^t, a_i^t, s_i^{t+1}, r_i^t)_{i\in \mathcal{V}}\}$\;
        $t \gets t+1$, $k \gets k+1$\;
        \If{$t == T$}{
            \textbf{initialize} SUMO, $t \gets 0$, \textbf{get} $\{s^0_i\}_{i\in\mathcal{V}}$\;
        }
        \tcc{update actors and critics}
        \If{$k == N_{\mathrm{bs}}$}{
            \ParrallelForEach{$i \in \mathcal{V}$}{
                \textbf{calculate} $\Tilde{r}^t_i$ (Eqn. (4)), $\Tilde{R}^t_i$ (Eqn. (5))\;
                $\phi_i \gets \phi_i - \eta_\phi \nabla \mathcal{L}_v(\phi_i)$\;
                $\theta_i \gets \theta_i - \eta_\theta \nabla \mathcal{L}_p(\theta_i)$\;
            }
            $k \gets 0, B \gets \varnothing$\;
        }
    }
\end{algorithm}
\FloatBarrier
\section{Experimentation}\label{sec_experimentation}
In this section, we demonstrate our RL framework using Simulation of Urban MObility (SUMO) \cite{lopez2018microscopic}
% to simulate the dynamic routing of EMVs in congested traffic network.
SUMO is an open-source traffic simulator capable of simulating both microscopic and macroscopic traffic dynamics, suitable for capturing the EMV's impact on the regional traffic as well as monitoring the overall traffic flow. An RL-simulator training pipeline is established between the proposed RL framework and SUMO, i.e., the agents collect observations from SUMO and  preferred signal phases are fed back into SUMO. \modi{Notice that in order to reflect the proposed emergency lane in Fig.3, we enable the built-in \emph{Sublane model} \cite{sumo-sublane} and \emph{Blue light device} \cite{bluelight-device} to establish the emergency lane for EMV passage. Under this scenarios, non-EMVs pull over to the sides when the EMV is driving between, and they resume normal driving when the EMV leaves the segment, see \ref{appendix_c} for more details.
}

\subsection{Datasets and Maps Descriptions}
\begin{figure}[h]
    \centering
    \includegraphics[width=0.9\linewidth]{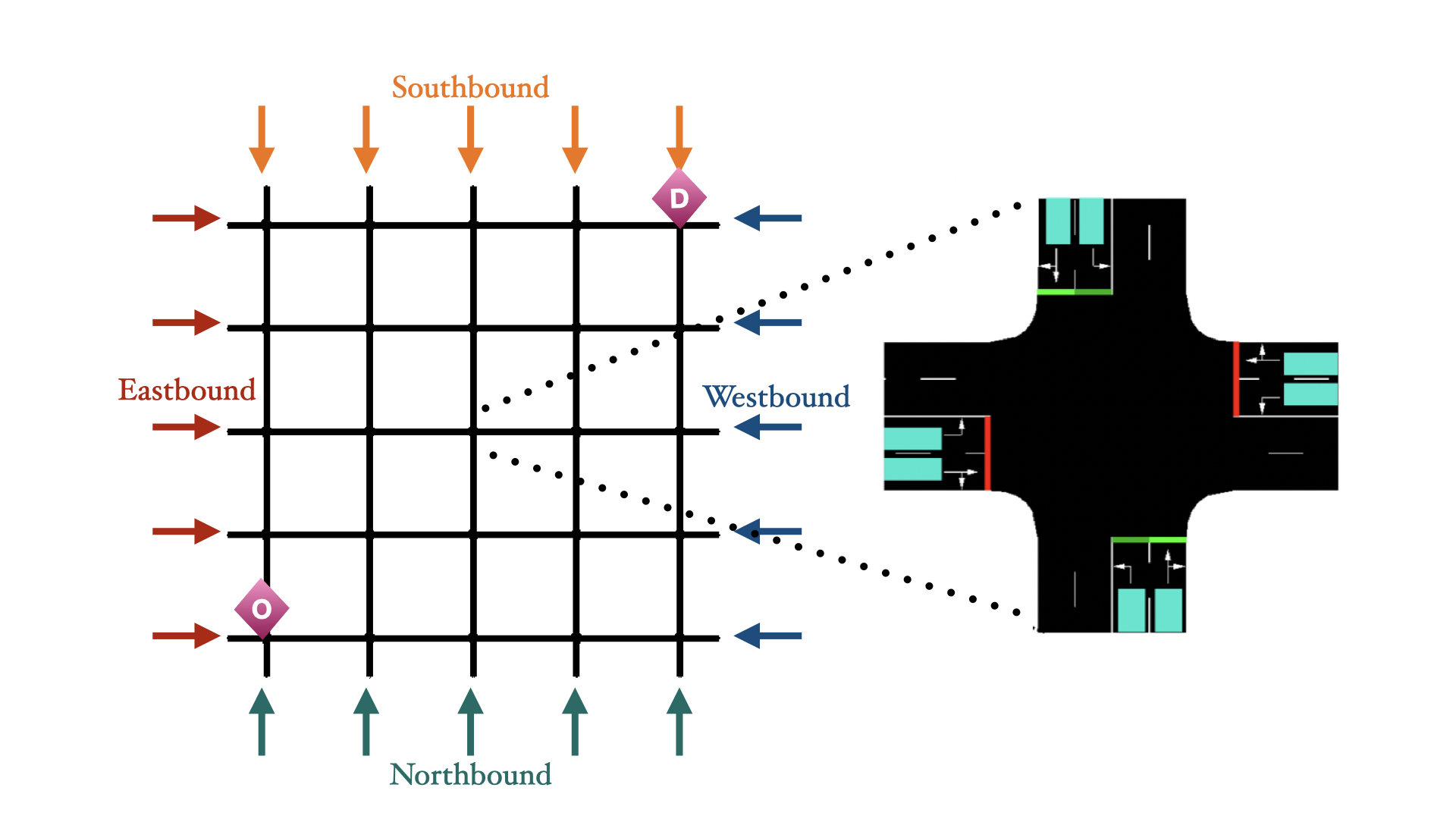}  
    \caption{\emph{Left}: the synthetic $\text{grid}_{5\times 5}$. Origin and destination for EMV are labeled. \emph{Right}: an intersection illustration in SUMO, the teal area are inductive loop detected area.}
  \label{fig_synthetic_map}
\end{figure}

\begin{figure}[h]
    \centering
    \includegraphics[width=0.9\linewidth]{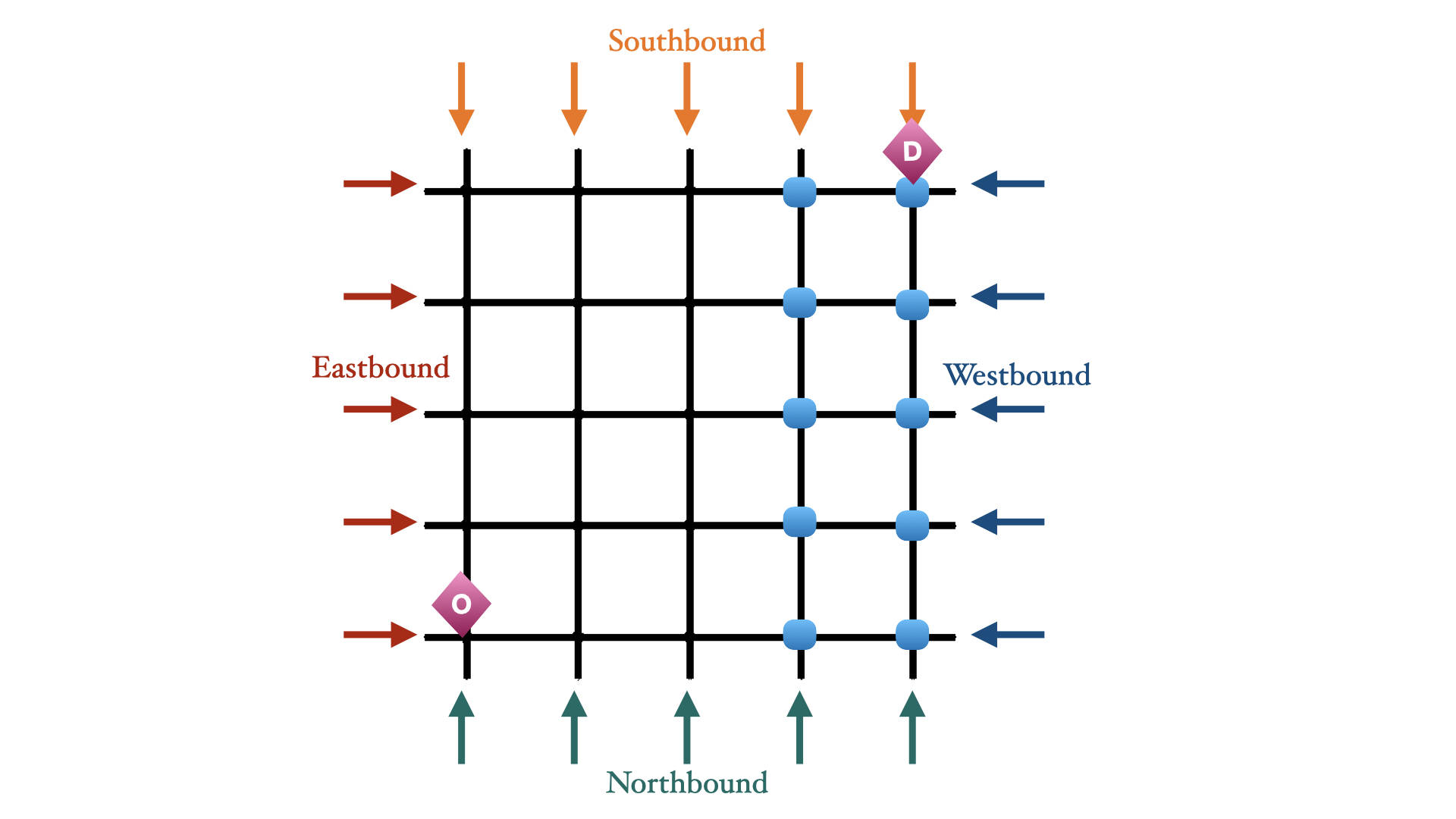}
  \caption{Emergency-capacitated Synthetic $\text{Grid}_{5\times 5}$. Segments towards intersections highlighted by blue have emergency capacities.}
  \label{fig_synthetic_map_reserved}
\end{figure}

We conduct the following experiments based on both synthetic and real-world map. 
\paragraph{Synthetic $\text{Grid}_{5\times 5}$} We synthesize a $5 \times 5$ traffic grid, where intersections are connected with bi-directional links. Each link contains two lanes. We assume all the links have zero emergency capacity. We design 4 configurations of time-varying traffic flows, listed in Table \ref{tab_synthetic_configuration}. 
% As for the traffic flow, trip information including the origin, destination and departure time are generated upon the configurations. 
The origin (O) and destination (D) of the EMV are labelled in Fig.~\ref{fig_synthetic_map}.
The traffic for this map has a time span of 1200s. We dispatch the EMV at $t =600s$ to ensure the roads are compacted when it starts travel.
\begin{table}[h]
\centering
\fontsize{10.0pt}{10.0pt} \selectfont
\begin{tabular}{@{}ccccc@{}}
\toprule[1pt]
\multicolumn{1}{c}{\multirow{2}{*}{Config}} & \multicolumn{2}{l}{Traffic Flow (veh/lane/hr)} & \multirow{2}{*}{Origin}                                                            & \multirow{2}{*}{Destination}                                                     \\ \cmidrule(lr){2-3}
\multicolumn{1}{c}{}                               & Non-peak                 & Peak                &                                                                                    &                                                                                  \\ 
\cmidrule{1-5}
1                                                  & 200                      & 240                 & \multirow{2}{*}{N,S} & \multirow{2}{*}{E,W} \\
\cmidrule{1-3}
2                                                  & 160                      & 320                 &                                                                                    &                                                                                  \\
\cmidrule{1-5}
3                                                  & 200                      & 240                 & \multicolumn{2}{c}{Randomly}                                                                                                               \\
\cmidrule{1-3}
4                                                  & 160                      & 320                 & \multicolumn{2}{c}{generated}     \\ \bottomrule[1pt]                                                                                                                                   
% \multirow{2}{*}{Randomly \\generated}
% \multicolumn{2}{c}{}
% \begin{tabular}[c]{@{}l@{}}N\\ S\end{tabular}
% \begin{tabular}[c]{@{}l@{}}E\\ W\end{tabular}
\end{tabular}
\caption{Configuration for Synthetic $\text{Grid}_{5\times 5}$. Peak flow is assigned from 400s to 800s and non-peak flow is assigned out of this period. For Config. 1 and 2, the vehicles enter the grid from North and South, and exit toward East and West.}
\label{tab_synthetic_configuration}
\end{table}
\paragraph{Emergency-capacitated (EC) Synthetic $\text{Grid}_{5\times 5}$}
This map adopts the same network layout as the Synthetic $\text{Grid}_{5\times 5}$ but with emergency-capacitated segments. As shown in Fig. \ref{fig_synthetic_map_reserved}, segments
% different emergency capacities on segments, see Fig. \ref{fig_synthetic_map_reserved}. Segments
approaching intersections highlighted by blue are emergency-capacitated with $C^{EC} = 0.2k$, with $k$ represents the normal vehicle capacity of this segment. All other segments are not emergency-capacitated.

\begin{figure}[h]
    \centering
    \includegraphics[width=\linewidth]{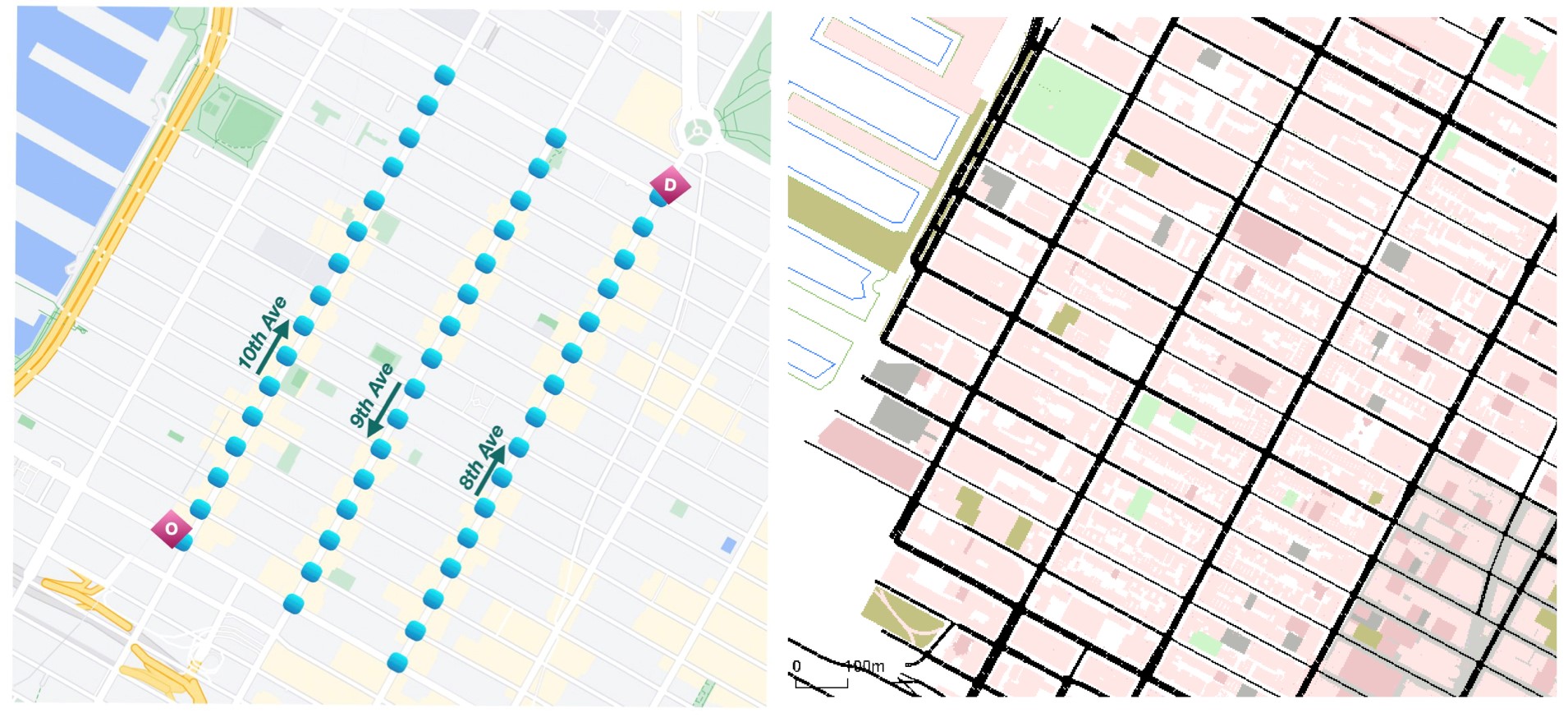}
  \caption{$\textrm{Manhattan}_{16 \times 3}$: a 16-by-3 traffic network in Hell's Kitchen area. Origin and destination for the EMV dispatching are labeled. \emph{Left}: on Google Map; \emph{Right}: in SUMO simulator.}
  \label{fig_manhattan}
\end{figure}

\paragraph{$\text{Manhattan}_{16\times 3}$}
This is a $16 \times 3$ traffic network extracted from Manhattan Hell's Kitchen area (Fig.~\ref{fig_manhattan}) and customized for demonstrating EMV passage. In this traffic network, intersections are connected
by 16 one-directional streets and 3 one-directional avenues. We assume each avenue contains four lanes and each street contains two lanes so that the right-of-way of EMVs and pre-emption can be demonstrated. 
We assume the emergency capacity for avenues and streets are $C^{EC}_{\textrm{avenue}} = 0.2k_{\textrm{avenue}}$ and $C^{EC}_{\textrm{street}} = 0.15k_{\textrm{street}}$, respectively.
% We also assume different emergency capacities for avenues and streets correspondingly.
The traffic flow for this map is generated from open-source NYC taxi data. Both the map and traffic flow data are publicly available.\footnote{https://traffic-signal-control.github.io/}
The origin and destination of EMV are set to be far away as shown in Fig.~\ref{fig_manhattan}. 
% Again, for a better demonstration purpose, we set the origin and destination as far as possible so the difference between methods are more obvious to see, see Figure. \ref{fig_manhattan}.

% \begin{figure}
% \begin{subfigure}{.5\textwidth}
%   \centering
%   \includegraphics[width=\linewidth]{images/fig_Manhattan_raw.jpeg}
%   \caption{Hell's Kitchen on Google Map}
% \end{subfigure}%
% \begin{subfigure}{.5\textwidth}
%   \centering
%   \includegraphics[width=\linewidth]{images/fig_manhattan_sumo.png}
%   \caption{Hell's Kitchen in SUMO simulator}
% \end{subfigure}
% \caption{$\textrm{Manhattan}_{16 \times 3}$: a 16-by-3 traffic network in Hell's Kitchen area. Origin and destination for the EMV dispatching are labeled.}
% \label{fig_manhattan}
% \end{figure}

\begin{figure}[h]
    \centering
    \includegraphics[width=\linewidth]{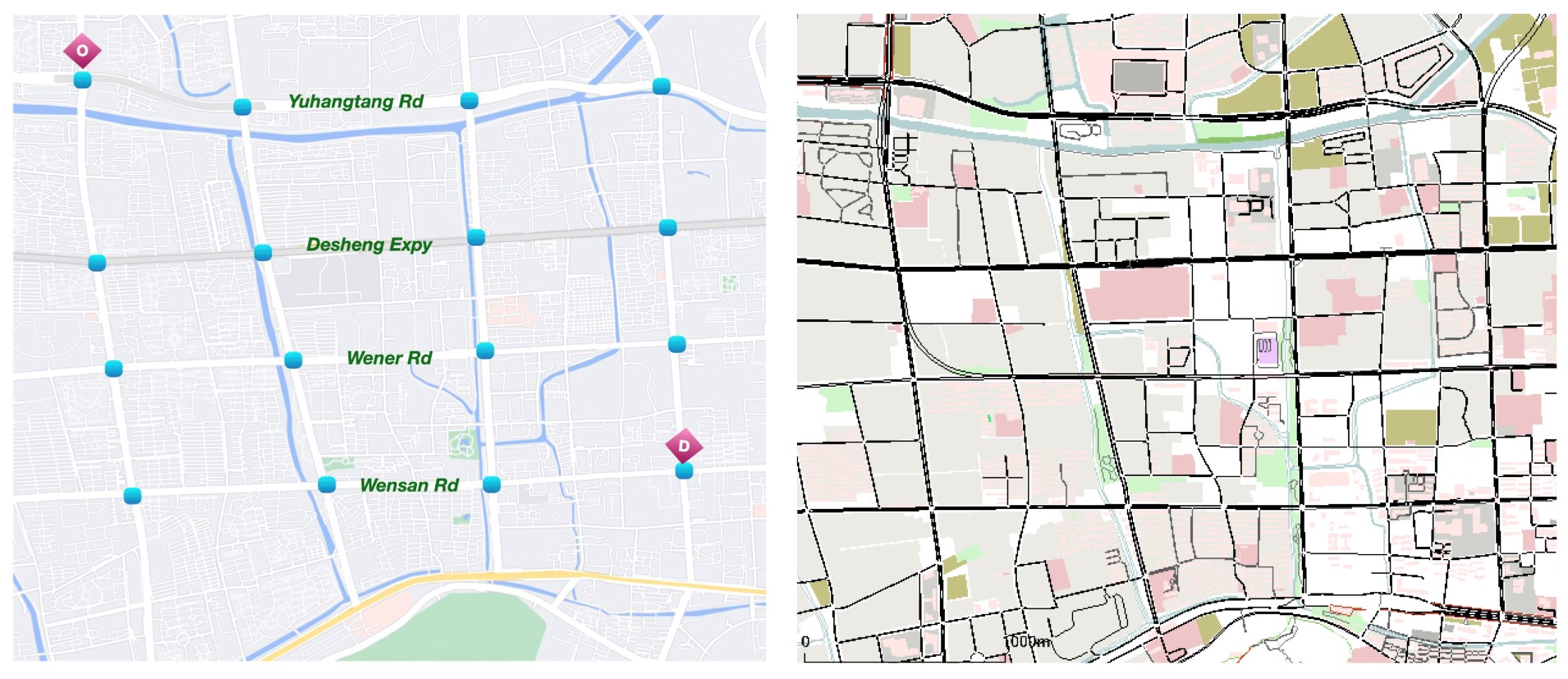}
  \caption{$\textrm{Hangzhou}_{4 \times 4}$: a 4-by-4 irregular and asymmetric network. Origin and destination for the EMV dispatching are labeled. \emph{Left}: on Google Map; \emph{Right}: in SUMO simulator.}
  \label{fig_gudang}
\end{figure}

\paragraph{$\textrm{Hangzhou}_{4\times 4}$}
An irregular $4 \times 4$ road network represents major avenues in Gudang sub-district in Hangzhou, China. All the road segments are bi-directional with two lanes in each direction. Both the map and traffic flow data are publicly available.
We set the origin and destination for EMV routing as shown in Fig.~\ref{fig_gudang}. The emergency capacity for each segment is set as $C^{EC} = 0.2k$.
\modi{Notice that this map only includes primary arterials and intersections because secondary arterials and intersections usually lack ITS infrastructure to assist EMVs' passage. They are also hard to monitor and control, considering Hangzhou Gudang district is a densely populated area. It is common to see double parking or merchants occupying road space on these secondary roads. The associated traffic flow dataset \cite{dataset} also excludes vehicle trajectories information on these secondary arterials for simplification.
}

\subsection{Benchmark Methods}
Due to the lack of existing RL methods for efficient EMV passage, we select traditional methods and RL methods for each subproblem and combine them to set up benchmarks. 
% we integrate techniques from pre-emption, routing and traffic signal control together to serve as the benchmarks.

For traffic signal pre-emption, the most intuitive and widely-used approach is the idea of extending green light period for EMV passage at each intersection which results in a \emph{Green Wave} \cite{corman2009evaluation}. 
\textbf{Walabi (W)} \cite{bieker2019modelling} is 
an effective rule-based method that implemented Green Wave for EMVs in SUMO environment. We integrate Walabi with combinations of routing and traffic signal control strategies introduced below as benchmarks.
% As \textbf{Walabi} \hs{saves EMV travel time in the most obvious manner,} we incorporate it in each of the benchmark methods introduced below.
We first present two routing benchmarks. 
% \emph{Routing benchmarks:}
% as briefed in Section \ref{sec_related_work}. To be specific: 
\begin{itemize}
    \item \textbf{Static routing}: static routing is performed only when EMV starts to travel and the route remains fixed. We adopt A* search as the implementation of static routing since it is a powerful extension to the Dijkstra's shortest path algorithm and is used in many real-time applications. \footnote{Our implementation of A* search employs a Manhattan distance as the heuristic function.}
    % shortest path is predetermined based on the shortest distance. 
    % In highly symmetric maps such as a traffic grid, more than one distance-based shortest path may exist and one of these paths is randomly selected. 
    % The static path is found using Dijkstra's algorithm and remains fixed as the EMV travels.
    \item \textbf{Dynamic routing}: dynamic routing updates the route by taking into account real-time information of traffic conditions. The route is then updated by repeatedly running static routing algorithms. To set up the dynamic routing benchmark, we run A* every 50s as EMV travels. The update interval is set to 50s since running the full A* to update the route is not as efficient as our proposed dynamic Dijkstra's algorithm. 
    % To serve as a benchmark method, we enable the \textbf{A*} search to be performed based on the global traffic network information, namely all agents' states.
    % Considering the time complexity of \textbf{A*} search algorithm, we allow the EMV to perform an \textbf{A*} right before dispatching and every 50s passed by.
\end{itemize}

% For traffic signal control strategies on reducing congestion, we select both conventional transportation engineering benchmarks as well as RL benchmarks:
\emph{Traffic signal control benchmarks:}
\begin{itemize}
    \item \textbf{Fixed Time (FT)}: Cyclical fixed time traffic phases with random offset \cite{roess2004traffic} is a policy that split all phases with an predefined green ratio. The coordination between traffic signals are predefined so it is not updated based on real-time traffic. Because of its simplicity, it is the default strategy in real traffic signal control for steady traffic flow.
    \item \textbf{Max Pressure (MP)}: \cite{varaiya2013max} studies max pressure control and use it as the main criterion for selecting traffic signal phases. It defines pressure for each signal phases and aggressively selects the traffic signal phase with maximum pressure to smooth congestion. Hence the name Max Pressure. It is the state-of-the-art network-level signal control strategy that is not based on learning.
    \item \textbf{Coordinated Deep Reinforcement Learners (CDRL)}: CDRL \cite{van2016coordinated} is a Q-learning based coordinator which directly learns joint local value functions for adjacent intersections. It extends Q-learning from single-agent scenarios to multi-agent scenarios.
    It also employs transfer planning and max-plus coordination strategies for joint intersection coordination. 
    \item \textbf{PressLight (PL)}: PL \cite{wei2019presslight} is also a Q-learning based method for traffic signal coordination. It aims at optimizing the pressure at each intersection. However, it defines pressure for each intersection, which is slightly different from the definition in Max Pressure. Our definition of pressure Eqn.~\eqref{eqn:reward}. is also different from that in PL.
    \item \textbf{CoLight (CL)}: CoLight \cite{wei2019colight} uses a graph-attentional-network-based reinforcement learning method for large scale traffic signal control. It adjusts queue length with information from neighbor intersections.
\end{itemize}

% Adopting \textbf{Walabi} as the pre-emption method, we can comprise one method each from path finding and traffic signal control to assemble as an end-to-end benchmark.
% \subsection{Evaluation Metrics}
% \textbf{EMV travel time} reflects the shortest path finding ability of the methods. To measure the delay experienced by non-EMVs, we choose \textbf{average travel time}, according to existing related work to compare the congestion reducing capabilities. 
% Proven as the most effective metric in transportation engineering practices, it calculates the average time vehicles spent in the system.

\subsection{Metrics}
We evaluate performance of all strategies under two metrics: \emph{EMV travel time}, which reflects the routing and pre-emption ability, and \emph{average travel time}, which indicates the ability of traffic signal control for efficient vehicle passage. Vehicles which have completed their trips during the simulation interval are counted when calculating the average travel time.

\section{Results and Discussion}\label{sec_result}

In this section, we demonstrate the performance of EMVLight and compare it against that of all benchmark methods on four experimentation maps. The results show a clear advantage of EMVLight under the two metrics. 
In addition, we illustrate the difference of underlying route selection by EMVLight and benchmark methods. We further conduct ablation studies to investigate the contribution of different components to EMVLight's performance.

\subsection{Performance Comparison}\label{sec:metrics_comparison}
To evaluate the performance of the proposed EMVLight and all benchmark methods, we conduct SUMO simulation  with five independent runs for each setting. Randomly generated seeds are used in learning-based methods. Means as well as standard deviations of the simulation results are reported for a full numerical assessment. The differences in simulation results for the same setting under independent SUMO runs are coming from configuration noise during generation, such as vehicles' lengths/accelerations/lane-changing eagerness, and, for RL-based methods, random seeds for initialization.
% For the purpose of fairly competing on both metrics, we weigh the EMV passage and traffic signal control tasks the same by default, which is reflected by $\beta = 0.5$ in Eqn.\eqref{eqn:reward2}. Increasing $\beta$ allows EMVLight to shift priority toward EMV passage, and decreasing $\beta$ asks EMVLight to strive for a better congestion management.

We provide implementation details of EMVLight on different experimentation settings in \ref{appendix_a}. Hyper-parameters choices for EMVLight and RL-based benchmarks are provided in \ref{appendix_b}.
\subsubsection{\texorpdfstring{Synthetic $\text{Grid}_{5\times 5}$ results}{Synthetic Grid results}}
% EMV Travel Time in Synthetic Map w/o emergency capacity
\begin{table}[h]
\centering
\fontsize{9.0pt}{10.0pt} \selectfont
\begin{tabular}{@{}ccccc@{}}
\toprule
\multirow{2}{*}{Method}           & \multicolumn{4}{c}{EMV Travel Time $T_{\textrm{EMV}}$ [s]}       \\ \cmidrule(l){2-5} 
                                  & Config 1 & Config 2 & Config 3 & Config 4 \\ \midrule
FT w/o EMV                        &    N/A      &    N/A      &    N/A      &    N/A      \\ \midrule
W + static + FT           &   258.18 $\pm$ 5.32   &  273.32  $\pm$ 9.74       &  256.40 $\pm$ 6.20      &   240.84  $\pm$ 4.43      \\
W + static + MP         &    260.22 $\pm$ 10.87      &    272.40 $\pm$ 10.92       &   265.74 $\pm$ 11.98       &    242.32 $\pm$ 9.48      \\
W + static + CDRL  &    269.42 $\pm$ 7.32      &   282.20 $\pm$ 5.28      &   276.14 $\pm$ 2.58       &   280.32 $\pm$ 4.82       \\
W + static + PL          &   270.68 $\pm$ 9.13       &  279.14 $\pm$ 9.22        &  281.42 $\pm$ 5.62        &   266.10 $\pm$ 8.32       \\
W + static + CL  &    255.72 $\pm$ 4.23      &   272.06 $\pm$ 8.13      &   270.22 $\pm$ 2.81       &  277.12 $\pm$ 6.10       \\
\midrule
W + dynamic + FT          &   229.38 $\pm$ 8.28       &    212.87 $\pm$ 3.17     &   218.46 $\pm$ 4.28       &    220.69 $\pm$ 7s.96      \\
W + dynamic + MP       &   220.48 $\pm$ 9.26      &  208.08 $\pm$ 12.90         &  212.46 $\pm$ 9.82        &   220.98 $\pm$ 10.62       \\
W + dynamic + CDRL &  239.84 $\pm$ 5.24        &   219.15 $\pm$ 8.26       &  211.86 $\pm$ 7.13        &   232.46 $\pm$ 10.16       \\
W + dynamic + PL          &   243.32 $\pm$ 13.86       &   244.82 $\pm$ 10.52       &  250.12 $\pm$ 8.13        &   255.02 $\pm$ 12.76       \\
W + dynamic + CL  &    220.12 $\pm$ 4.19      &  209.12 $\pm$ 4.76      &   224.00 $\pm$ 5.31       &   226.32 $\pm$ 4.13       \\
 \midrule
EMVLight    & \textbf{195.46} $\pm$ 7.48         &  \textbf{190.66} $\pm$ 8.28         &   \textbf{183.12} $\pm$ 6.43        &    \textbf{189.44} $\pm$ 8.32       \\ \bottomrule
\end{tabular}
\caption{EMV travel time in the four configurations of Synthetic $\text{Grid}_{5\times 5}$. Lower value indicates better performance and the lowest values are highlighted in bold.}
\label{tab_synthetic_emv}
\end{table}
% Average Travel Time in Synthetic Map w/o emergency capacity
\begin{table}[h]
\centering
\fontsize{9.0pt}{10.0pt} \selectfont
\begin{tabular}{@{}ccccc@{}}
\toprule
\multirow{2}{*}{Method}           & \multicolumn{4}{c}{Average Travel Time $T_{\textrm{avg}}$ [s]}       \\ \cmidrule(l){2-5} 
                                  & Config 1 & Config 2 & Config 3 & Config 4 \\ \midrule
FT w/o EMV                        &    353.43 $\pm$ 4.65    &    371.13  $\pm$ 4.58     &    314.25  $\pm$ 2.90    &    334.10 $\pm$ 3.73     \\ 
\midrule
W + static + FT           &   380.42 $\pm$ 13.35  &  395.17 $\pm$ 15.37      &  350.16 $\pm$ 13.66    &   363.90 $\pm$ 15.39     \\
W + static + MP         &  355.10  $\pm$ 15.36      &    362.09  $\pm$ 16.15     &   318.76  $\pm$ 14.90   &     330.69  $\pm$ 15.52   \\
W + static + CDRL  &   559.19 $\pm$ 3.60 & 540.81 $\pm$ 12.04  & 568.13 $\pm$ 13.25      &  568.13 $\pm$ 6.67       \\
W + static + PL  &   369.52 $\pm$ 8.72    &   372.32 $\pm$ 16.05     &   339.18  $\pm$ 7.17     &   339.12  $\pm$ 5.33     \\ 

W + static + CL  &   365.64 $\pm$ 14.08    &   380.13 $\pm$ 8.20     &   328.42  $\pm$ 17.52     &   333.74  $\pm$ 5.76     \\ 
\midrule
W + dynamic + FT          &  380.76  $\pm$ 10.70       &   404.81 $\pm$ 18.76      &   345.09  $\pm$ 11.60     &   358.90  $\pm$  15.27   \\
W + dynamic + MP       &   360.38 $\pm$ 10.31    &  365.10 $\pm$ 8.33       &   327.98  $\pm$ 18.90     &  351.62  $\pm$ 3.79      \\
W + dynamic + CDRL          & 565.38  $\pm$ 16.10      &   544.29  $\pm$ 19.23     &  598.73  $\pm$ 11.01      &    572.22  $\pm$ 13.94    \\
W + dynamic + PL &  373.17  $\pm$ 17.98      &   387.25 $\pm$ 13.98      &  349.12 $\pm$ 16.25     &   330.21  $\pm$ 17.23     \\ 
W + dynamic + CL &  359.14  $\pm$ 9.52      &   370.45 $\pm$ 4.02      &  320.64 $\pm$ 4.10     &   335.27  $\pm$ 7.62     \\ 
\midrule
EMVLight    &  \textbf{335.09}  $\pm$ 4.13    &   \textbf{333.28}  $\pm$ 8.81     & \textbf{307.90}   $\pm$ 3.89       &   \textbf{321.02} $\pm$ 5.87     \\ \bottomrule
\end{tabular}
\caption{Average travel time for all vehicles which have completed trips in the four configurations of Synthetic $\text{Grid}_{5\times 5}$. }
\label{tab_synthetic_avg}
\end{table}

Table~\ref{tab_synthetic_emv} and \ref{tab_synthetic_avg} present the experimental results on average EMV travel time and average travel time on Synthetic $\text{Grid}_{5\times 5}$. In terms of EMV travel time $T_{\textrm{EMV}}$, the dynamic routing benchmark performs better than static routing benchmarks. This is expected since dynamic routing considers the time-dependent nature of traffic conditions and update optimal route accordingly. The best learning and non-learning benchmark methods are dynamics routing with CoLight and Max Pressure, respectively. EMVLight further reduces EMV travel time by 16\% on average as compared to dynamic routing benchmarks. This advantage in performance can be attributed to the design of secondary pre-emption agents. This type of agents learns to ``reserve a link" by choosing signal phases that help clear the vehicles in the link to encourage high speed EMV passage (Eqn.~\eqref{eqn:reward}). 

As for average travel time $T_{\textrm{avg}}$, we first notice that the traditional pre-emption technique (W + static + FT) indeed increases the average travel time by around 10\% as compared to a traditional Fix Time strategy without EMV (denoted as ``FT w/o EMV" in Table \ref{tab_synthetic_avg}), thus decreasing the efficiency of vehicle passage. Different traffic signal control strategies have a direct impact on overall efficiency. Fixed Time is designed to handle steady traffic flow. Max Pressure, as a SOTA traditional method, outperforms Fix Time and, surprisingly, nearly outperforms all RL benchmarks in terms of overall efficiency. This shows that pressure is an effective indicator for reducing congestion and this is why we incorporate pressure in our reward design. Coordinate Learner performs the worst probably because its reward is not based on pressure. PressLight doesn't beat Max Pressure because it has a reward design that focuses on smoothing vehicle densities along a major direction, e.g. an arterial. Grid networks with the presence of EMV make PressLight less effective.
CoLight achieves similar performance as Max Pressure and is the best learning benchmark method.
% \hs{CoLight achieves a smaller $T_{\textrm{avg}}$ than other traffic signal control strategies, attributing to a shorter $T_{\textrm{EMV}}$ simultaneously since roads are overall less congested.} 
Our EMVLight improves its pressure-based reward design to encourage smoothing vehicle densities of all directions for each intersection. This enable us to achieve an advantage of 7.5\% over our best benchmarks (Max Pressure).

\subsubsection{Emergency-capacitated Synthetic \texorpdfstring{$\text{Grid}_{5\times 5}$}{Grid} results}

\begin{table}[h]
\centering
\fontsize{9.0pt}{10.0pt} \selectfont
\begin{tabular}{@{}ccccc@{}}
\toprule
\multirow{2}{*}{Method}           & \multicolumn{4}{c}{EMV Travel Time $T_{\textrm{EMV}}$ [s]}       \\ \cmidrule(l){2-5} 
                                  & Config 1 & Config 2 & Config 3 & Config 4 \\ \midrule
FT w/o EMV                        &    N/A      &    N/A      &    N/A      &    N/A      \\ \midrule
W + static + FT           &   254.04 $\pm$ 7.42   &  260.18  $\pm$ 12.03       &  252.12 $\pm$ 11.03      &   232.47  $\pm$ 12.23      \\

W + static + MP         &    233.76 $\pm$ 8.05      &    258.60 $\pm$ 9.06       &   252.74 $\pm$ 13.05      &    233.20 $\pm$ 8.96      \\

W + static + CDRL  &    240.10 $\pm$ 8.65      &   266.28 $\pm$ 8.54      &   258.10 $\pm$ 9.27       &   270.43 $\pm$ 6.18       \\

W + static + PressLight           &   265.28 $\pm$ 7.28       &  269.10 $\pm$ 6.65        &  270.18 $\pm$ 8.83        &   259.20 $\pm$ 7.13       \\

W + static + CoLight  &    250.82 $\pm$ 6.73      &   267.08 $\pm$ 10.21      &   266.12 $\pm$ 4.13       &  270.18 $\pm$ 8.12       \\

\midrule

W + dynamic + FT          &   210.28 $\pm$ 8.90       &    206.18 $\pm$ 7.19     &   210.28 $\pm$ 8.81       &    207.64 $\pm$ 10.02      \\

W + dynamic + MP       &   202.28 $\pm$ 8.54      &  203.20 $\pm$ 9.07         &  206.64 $\pm$ 7.98        &   210.86 $\pm$ 8.59       \\

W + dynamic + CDRL &  218.36 $\pm$ 8.12        &   209.28 $\pm$ 7.19       &  208.180$\pm$ 10.54        &   230.22 $\pm$ 9.22       \\

W + dynamic + PressLight          &   270.08 $\pm$ 10.20       &   238.10 $\pm$ 9.22       &  242.10 $\pm$ 6.98        &   248.24 $\pm$ 10.24       \\

W + dynamic + CoLight  &    216.04 $\pm$ 4.91      &  206.12 $\pm$ 6.27      &   219.26 $\pm$ 6.87      &   223.78 $\pm$ 5.10       \\

 \midrule
EMVLight    & \textbf{150.28} $\pm$ 7.48         &  \textbf{158.20} $\pm$ 6.28         &   \textbf{154.28} $\pm$ 4.19        &    \textbf{159.28} $\pm$ 6.03       \\ \bottomrule
\end{tabular}
\caption{EMV travel time in the four configurations of Synthetic $\text{Grid}_{5\times 5}$ with an emergency capacity co-efficient of 0.25.}
\label{tab_synthetic_ec_emv}
\end{table}

Table~\ref{tab_synthetic_ec_emv} shows $T_{\textrm{EMV}}$ of all the methods implemented on the  emergency-capacitated synthetic \texorpdfstring{$\text{Grid}_{5\times 5}$}{Grid} map. 
By comparing Table~\ref{tab_synthetic_emv} and Table~\ref{tab_synthetic_ec_emv}, we conclude that the emergency-capacitated map exhibits overall shorter $T_{\textrm{EMV}}$ in all configurations. 
% according to the patterns shown in Table~\ref{tab_synthetic_ec_emv}. 
For benchmark methods, the nonzero emergency capacity shorten $T_{\textrm{EMV}}$ by an average of approximately 12 seconds. 
In particular, dynamic routing-based methods benefit more from the additional emergency capacity, resulting in an average reduction of in 16.28 seconds $T_{\textrm{EMV}}$. This is due to the adaptive nature of dynamic navigation. 
By comparing EMVLight and benchmark methods in Table~\ref{tab_synthetic_ec_emv}, we observe that EMVLight reduce $T_{\textrm{EMV}}$ by up to 50 seconds (25\%) as compared to the best benchmark method in all configurations. 
% EMVLight dedicates an significant decrease in $T_{\textrm{EMV}}$ in all configurations within the emergency capacity settings. Emergency vehicles saves up to approximately 25\% traversing through this network. 
The substantial difference in $T_{\textrm{EMV}}$ reduction between benchmark methods and EMVLight is due to high success rate of emergency lane forming under coordination, which is investigated further in Section~\ref{subsec_route_selection}. 
% \dz{[do we need to define emergency yielding? or is it a well known concept in the field? I think at least we should mention the phrase in the caption of Figure 3.]}
% will be explained by routing behavior deviation among methods in \ref{subsec_route_selection}.

\begin{table}[h]
\centering
\fontsize{9.0pt}{10.0pt} \selectfont
\begin{tabular}{@{}ccccc@{}}
\toprule
\multirow{2}{*}{Method}           & \multicolumn{4}{c}{Average Travel Time $T_{\textrm{avg}}$ [s]}       \\ \cmidrule(l){2-5} 
                                  & Config 1 & Config 2 & Config 3 & Config 4 \\ \midrule
FT w/o EMV                        &    353.43 $\pm$ 4.65    &    371.13  $\pm$ 4.58     &    314.25  $\pm$ 2.90    &    334.10 $\pm$ 3.73     \\ 
\midrule

W + static + FT           &   395.28 $\pm$ 5.17  &  410.94 $\pm$ 9.16      &  365.82 $\pm$ 6.14    &   379.64 $\pm$ 6.21     \\

W + static + MP         &  370.52  $\pm$ 6.18      &    375.44  $\pm$ 6.48     &   331.62  $\pm$ 5.92   &     345.13  $\pm$ 8.63 \\

W + static + CDRL  &   575.28 $\pm$ 7.76 & 555.62 $\pm$ 10.04  & 574.91 $\pm$ 19.86      &  585.20 $\pm$ 7.53\\

W + static + PL  &   385.28 $\pm$ 12.09    &   380.83 $\pm$ 10.07     &   360.09  $\pm$ 11.62     &   369.72  $\pm$ 18.02     \\ 

W + static + CL  &   382.17 $\pm$ 6.02    &   380.13 $\pm$ 8.21     &   344.19  $\pm$ 16.02     &   352.07  $\pm$ 4.10     \\ 
\midrule

W + dynamic + FT          &  389.12  $\pm$ 18.21       &   411.98 $\pm$ 15.31      &   353.72  $\pm$ 9.09     &   367.74  $\pm$  16.82   \\

W + dynamic + MP       &   370.28 $\pm$ 12.51    &  362.82 $\pm$ 9.05      &   335.10  $\pm$ 9.16     &  360.02  $\pm$ 17.18     \\

W + dynamic + CDRL          & 575.05  $\pm$ 9.67      &   550.92  $\pm$ 14.06     &  609.26 $\pm$ 11.12      &    578.10  $\pm$ 12.09    \\

W + dynamic + PL &  380.29  $\pm$ 6.10      &   395.28 $\pm$ 5.62      &  359.16 $\pm$ 14.07     &   337.26  $\pm$ 4.96     \\ 

W + dynamic + CL &  366.14  $\pm$ 8.21      &   380.74 $\pm$ 15.84      &  330.44 $\pm$ 17.29     &   343.58  $\pm$ 14.27     \\ 
\midrule

EMVLight    &  \textbf{334.96}  $\pm$ 5.52    &   \textbf{336.18}  $\pm$ 17.09     & \textbf{309.10}   $\pm$ 15.56       &   \textbf{323.76} $\pm$ 17.24     \\ \bottomrule
\end{tabular}
\caption{Average travel time for all vehicles which have completed trips in the four configurations of Synthetic $\text{Grid}_{5\times 5}$ with an emergency capacity coefficient of 0.25. }
\label{tab_synthetic_ec_avg}
\end{table}

As the EMV travels faster and more emergency lanes are formed, the average travel time of non-emergency vehicles increases. Table~\ref{tab_synthetic_ec_avg} shows $T_{\textrm{avg}}$ with emergency capacity added. By comparing Table~\ref{tab_synthetic_avg} and Table~\ref{tab_synthetic_ec_avg}, we observe that with added emergency capacity, the average increase in $T_{\textrm{avg}}$ for non-learning-based and learning-based benchmarks are 12.04 seconds and 7.78 seconds, respectively. Learning-based methods lead to a smaller average increase since agents gradually learn to direct non-EMVs, which are interrupted by EMV passages, to resume their trips as soon as possible. As a result, potential congested queues would not be formed on these segments, effectively reducing the overall $T_{\textrm{avg}}$.

EMVLight, surprisingly, manages to achieve nearly no increase in $T_{\textrm{avg}}$ with added emergency capacity, even though more 
emergency lanes are formed as indicated by smaller $T_{\textrm{EMV}}$. This result shows that EMVLight is able to learn a strong traffic signal coordination strategy while navigating EMVs simultaneously.  The proposed multi-class agent design demonstrates EMVLight's capability of addressing the coupled problems of EMV routing and traffic signal control simultaneously. EMVLight manages to prepare segments for incoming EMVs by reducing the number of vehicles on those segments and restores the impacted traffic in a timely manner after the EMV passage.
% compared with its performance in non-emergency-capacitated counterpart, which again reinforces the traffic signal coordination capability for non-EMVs, especially for those affected by yielding to EMVs.
% \dz{[Again, I don't know how to explain the effect of agents intuitively here. Let's discuss.][coupling - explain good performance on Tavg]}

\subsubsection{\texorpdfstring{$\textrm{Manhattan}_{16 \times 3}$ results}{Manhattan results}} 
% Please add the following required packages to your document preamble:
% \usepackage{multirow}

\begin{table}[h]
\centering
\fontsize{10.0pt}{12.0pt} \selectfont
\begin{tabular}{ccc}
\hline
\multirow{2}{*}{Method} & \multicolumn{2}{c}{$\textrm{Manhattan}_{16 \times 3}$}               \\ \cline{2-3} 
                        & $T_{\textrm{EMV}}$     & $T_{\textrm{avg}}$     \\ \hline
FT w/o EMV              &        N/A              &  1649.64                    \\ \hline
W + static + FT         &    817.37 $\pm$ 17.40                  &      1816.43 $\pm$ 68.96                \\
W + static + MP         &      686.72    $\pm$ 19.23          &    917.52  $\pm$ 52.16                 \\
W + static + CDRL        &    702.62 $\pm$ 24.29          & 1247.67  $\pm$ 83.47  \\
W + static + PL         &      626.88    $\pm$ 24.82            &      992.06 $\pm$ 47.67                \\
W + static + CL         &      545.26 $\pm$ 30.21                &       855.28 $\pm$ 41.29               \\ \hline
W + dynamic + FT         &     820.54 $\pm$ 28.86                  &    1808.25 $\pm$ 68.04                  \\
W + dynamic + MP         &    632.68  $\pm$ 13.29                  &   921.18  $\pm$ 49.29                   \\
W + dynamic + CDRL        &      680.62  $\pm$ 20.17              &   1262.39 $\pm$ 60.09                   \\
W + dynamic + PL         & 521.42 $\pm$ 27.62 & 977.62 $\pm$ 53.45   \\
W + dynamic + CL         &    501.26 $\pm$ 28.71                  &  862.94 $\pm$ 45.19                  \\ \hline
EMVLight                &     \textbf{292.82} $\pm$ 16.23                  &         \textbf{782.13} $\pm$ 39.31             \\ \hline
\end{tabular}
\caption{$T_{\textrm{EMV}}$ and $T_{\textrm{avg}}$ for $\textrm{Manhattan}_{16 \times 3}$. The average travel time without the presence of EMVs (1649.64) is retrieved from data.}
\label{tab_Manhattan_results}
\end{table}

Table~\ref{tab_Manhattan_results} presents EMV travel time and average travel time of all the methods on the $\textrm{Manhattan}_{16 \times 3}$ map. In terms of $T_{\textrm{EMV}}$, dynamic routing benchmarks in general result in faster EMV passasge, as expected. 
% as expected, EMV navigation based on dynamic routing benchmarks travels faster than static routing. 
Compared with benchmark methods, EMVLight produces a considerably low average $T_{\textrm{EMV}}$ of 292.82 seconds, which is 38\% faster than the best benchmark (W+static+CL). 
As for $T_{\textrm{avg}}$, We have similar observation as in the synthetic maps that Max Pressure achieves a similar level of performance on reducing congestion as PressLight and beats CDRL by a solid margin of 25\%. 
CoLight stands out among benchmarks regarding both metrics. Particularly, CoLight shortens $T_{\text{avg}}$ by approximately one minute than Max Pressure strategies on this map.
% Regarding $T_{\textrm{avg}}$, we can also witness that Max Pressure achieves \textit{similar} level of congestion reliving performance with PressLight and \textit{beats} CDRL by a solid margin of 25\%. 

\begin{figure}[h!]
    \centering
    \includegraphics[width=\linewidth]{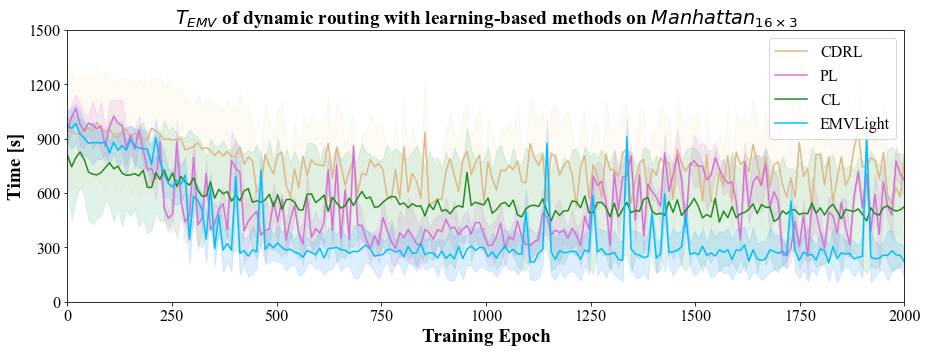}
  \caption{$T_{\textrm{EMV}}$ convergence by learning-based dynamic routing strategies on $\text{Manhattan}_{16\times3}$.}
  \label{fig_emv_learning_curves}
\end{figure}

\begin{figure}[h!]
    \centering
    \includegraphics[width=\linewidth]{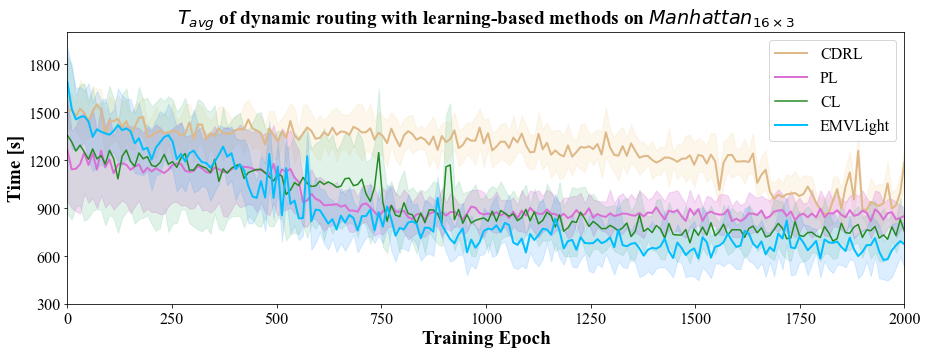}
  \caption{$T_{\textrm{avg}}$ convergence by learning-based dynamic routing strategies on $\text{Manhattan}_{16\times3}$.}
  \label{fig_avg_learning_curves}
\end{figure}

Fig.~\ref{fig_emv_learning_curves} and Fig.~\ref{fig_avg_learning_curves} shows the learning curves of  $T_{\textrm{EMV}}$ and $T_{\textrm{avg}}$, respectively,  in the four RL methods. 
% The learning curves by RL benchmarks for $T_{\textrm{EMV}}$ and $T_{\textrm{avg}}$ are presented in Fig.\ref{fig_emv_learning_curves} and Fig.\ref{fig_avg_learning_curves}, respectively. 
From both figures, we observe that EMVLight has the fastest convergence - in 500 epochs for $T_{\textrm{EMV}}$ and in 1000 epoches for $T_{\textrm{avg}}$ - among all four methods. 
% Both figures illustrate that EMVLight converges to the lowest numerical values within 2000 epochs of training. 
% In addition, EMVLight is also the \textit{first} to converge, experiencing relatively minor fluctuations during learning as an evidence of its robustness. 
In Fig.~\ref{fig_emv_learning_curves}, both PressLight and CDRL struggle with converging to a stable $T_{\textrm{EMV}}$. 
In Fig.~\ref{fig_avg_learning_curves}, CDRL converges very slowly as compared to the other three methods. 
Both figures shows that CDRL behaves the worst since its DQN design hardly scales with an increasing number of intersections. 
These learning curves demonstrate the fast and stable learning of EMVLight. 
% At the same time, other RL benchmarks and EMVLight show learning stability after 750 episodes, striving for a low $T_{\textrm{avg}}$.
% The divergence in training results between EMVLight's $T_{\textrm{EMV}}$ and $T_{\textrm{EMV}}$ from compared benchmarks suggests EMVLight's overall coordination efficiency on simultaneous routing and signal control in this scenario.

\subsubsection{\texorpdfstring{$\textrm{Hangzhou}_{4 \times 4}$ results}{Hangzhou results}}
\begin{table}[h]
\centering
\fontsize{10.0pt}{12.0pt} \selectfont
\begin{tabular}{ccc}
\hline
\multirow{2}{*}{Method} & \multicolumn{2}{c}{$\textrm{Hangzhou}_{4 \times 4}$}               \\ \cline{2-3} 
                        & $T_{\textrm{EMV}}$     & $T_{\textrm{avg}}$     \\ \hline
FT w/o EMV              &        N/A              &   764.08                   \\ \hline
W + static + FT         & 466.19 $\pm$ 10.25 & 779.13 $\pm$ 12.90 \\
W + static + MP         & 377.20 $\pm$ 14.42 & 404.37 $\pm$ 8.12 \\
W + static + CDRL        & 409.56 $\pm$ 12.06 & 749.10 $\pm$ 10.02 \\
W + static + PL         & 380.82 $\pm$ 6.72 & 425.46 $\pm$ 9.74 \\
W + static + CL         & 368.20 $\pm$ 14.66 & 366.14 $\pm$ 8.25 \\ 
\hline
W + dynamic + FT         & 415.63 $\pm$ 9.03  & 783.89 $\pm$ 10.03\\
W + dynamic + MP         & 328.42 $\pm$ 12.28 & 410.25 $\pm$ 6.23 \\
W + dynamic + CDRL        & 401.08 $\pm$ 15.25 & 755.28 $\pm$ 12.82   \\
W + dynamic + PL         & 321.52 $\pm$ 14.58 & 431.27 $\pm$ 8.24\\
W + dynamic + CL         & 319.84 $\pm$ 11.09 & 370.20 $\pm$ 7.13\\ \hline
EMVLight                & \textbf{194.52} $\pm$ 9.65 & \textbf{331.42} $\pm$ 6.18
\\ \hline
\end{tabular}
\caption{$T_{\textrm{EMV}}$ and $T_{\textrm{avg}}$ for $\textrm{Gudang}_{4 \times 4}$. The average travel time without the presence of EMVs (764.08) is retrieved from data.}
\label{tab_Gudang_results}
\end{table}

Table \ref{tab_Gudang_results} presents $T_{\textrm{EMV}}$ and $T_{\textrm{avg}}$ of EMVLight and benchmark methods on $\textrm{Hangzhou}_{4 \times 4}$. EMVLight achieves the lowest $T_{\textrm{EMV}}$ of 194.52 seconds, \textit{beating} the best benchmark (W+dynamic+CL) by 115 seconds (37\%).
As for $T_{\textrm{avg}}$, EMVLight also has excellent performance, exhibiting a 10\% advantage over W+dynamic+CL, and a 20\% advantage over W+dynamic+MP.
Once again, Max Pressure \textit{outperforms} PressLight in terms of $T_{\textrm{avg}}$. This is consistent with our observations with other maps above, suggesting Max Pressure result in great traffic signal coordination strategies to reduce overall congestion.
% is capable of reducing general congestion

Based on results of all introduced experiments, Max Pressure has evinced the consistency to restrict congestion, particularly when restoring stagnant traffic after EMV passages. Consequently, Max Pressure accomplishes the comparable congestion reduction acquirement with, if not better than, RL-based strategies.

% However, lack of coordination for EMV navigation prevents Max Pressure from resulting in a lower $T_{\textrm{EMV}}$ than other methods. 

\begin{figure}[h]
    \centering
    \includegraphics[width=\linewidth]{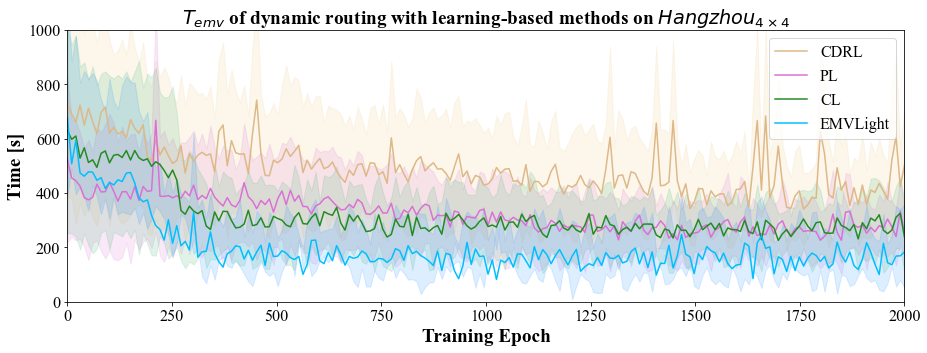}
  \caption{$T_{\textrm{EMV}}$ convergence by learning-based dynamic routing strategies on $\text{Hangzhou}_{4\times4}$.}
  \label{fig_emv_Gudang_learning_curves}
\end{figure}

\begin{figure}[h]
    \centering
    \includegraphics[width=\linewidth]{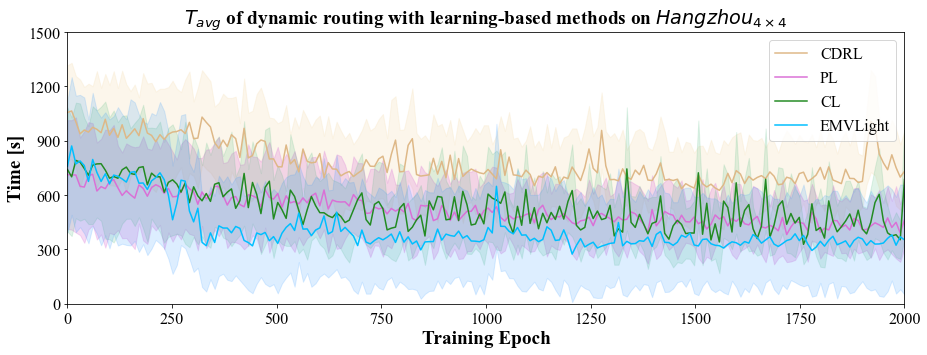}
  \caption{$T_{\textrm{avg}}$ convergence by learning-based dynamic routing strategies on $\text{Hangzhou}_{4\times4}$.}
  \label{fig_avg_Gudang_learning_curves}
\end{figure}

CDRL fails to learn an effective coordination strategy to manage congestion, where its $T_{\textrm{avg}}$ trivially differs from Fix Time strategy's $T_{\textrm{avg}}$.
CoLight surpasses other methods by shortening $T_{\textrm{avg}}$ by at least 40 seconds but it does not illustrate significant improvement in terms of $T_{\textrm{EMV}}$ from alternative benchmarks.
Learning curves in Fig.~\ref{fig_emv_Gudang_learning_curves} and \ref{fig_avg_Gudang_learning_curves} show that EMVLight converges in 500 epochs. 
% EMVLights remains the optimal control method by saving an additional 10\% of $T_{\textrm{avg}}$ from (W+dynamic+CL) and 20\% from (W+dynamic+MP).


\subsection{Routing choices}
\label{subsec_route_selection}
In this section, we investigate the reason behind EMVLight's best performance in $T_{\textrm{EMV}}$ from a routing perspective. In particular, we show how EMVLight is able to leverage emergency capacity to achieve maximum speed passage. 
% The differences of $T_{\textrm{EMV}}$ in Table. \ref{tab_synthetic_emv} results from sequences of decisions on which directions to turn encountering intersections during EMVs' trips. 
% These decisions, given the traffic condition at the moment, determine the travel speed on the selected segment and impact the ultimate trip time for EMVs.
To understand different route choices between benchmark methods and EMVLight,
we analyze EMV routes on emergency-capacitated Synthetic $\textrm{Grid}_{5\times5}$ as well as $\textrm{Hangzhou}_{4\times4}$ to gain insight into the advantage of EMVLight.

\paragraph{Emergency-capacitated Synthetic \texorpdfstring{$\textrm{Grid}_{5\times5}$}{Grid} routes} 

\begin{figure}
\centering
\subfloat[Config 1]{\label{fig_routing_config1} \includegraphics[width=0.20\textwidth]{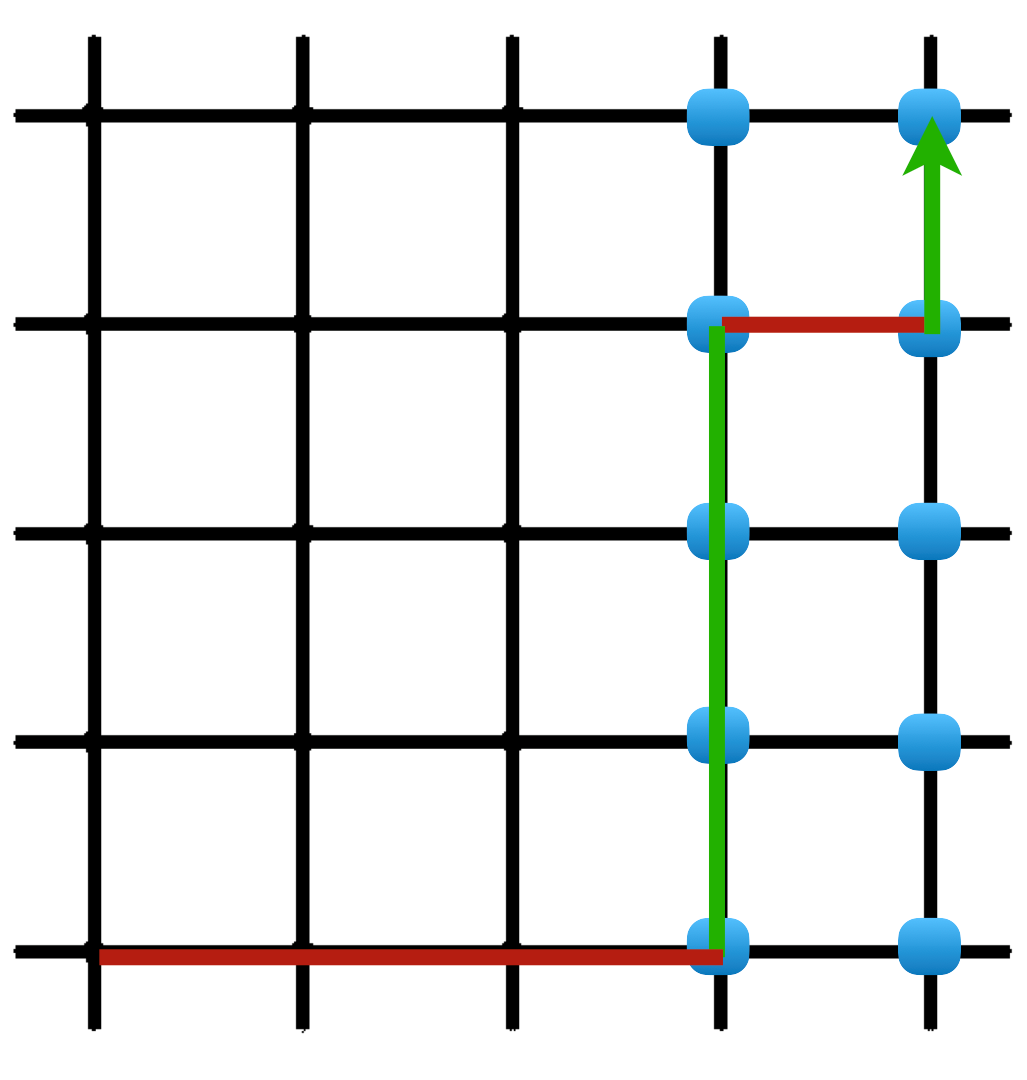}}% The "%" masks the line break.
\hfill
\subfloat[Config 2]{\label{fig_routing_config2} \includegraphics[width=0.20\textwidth]{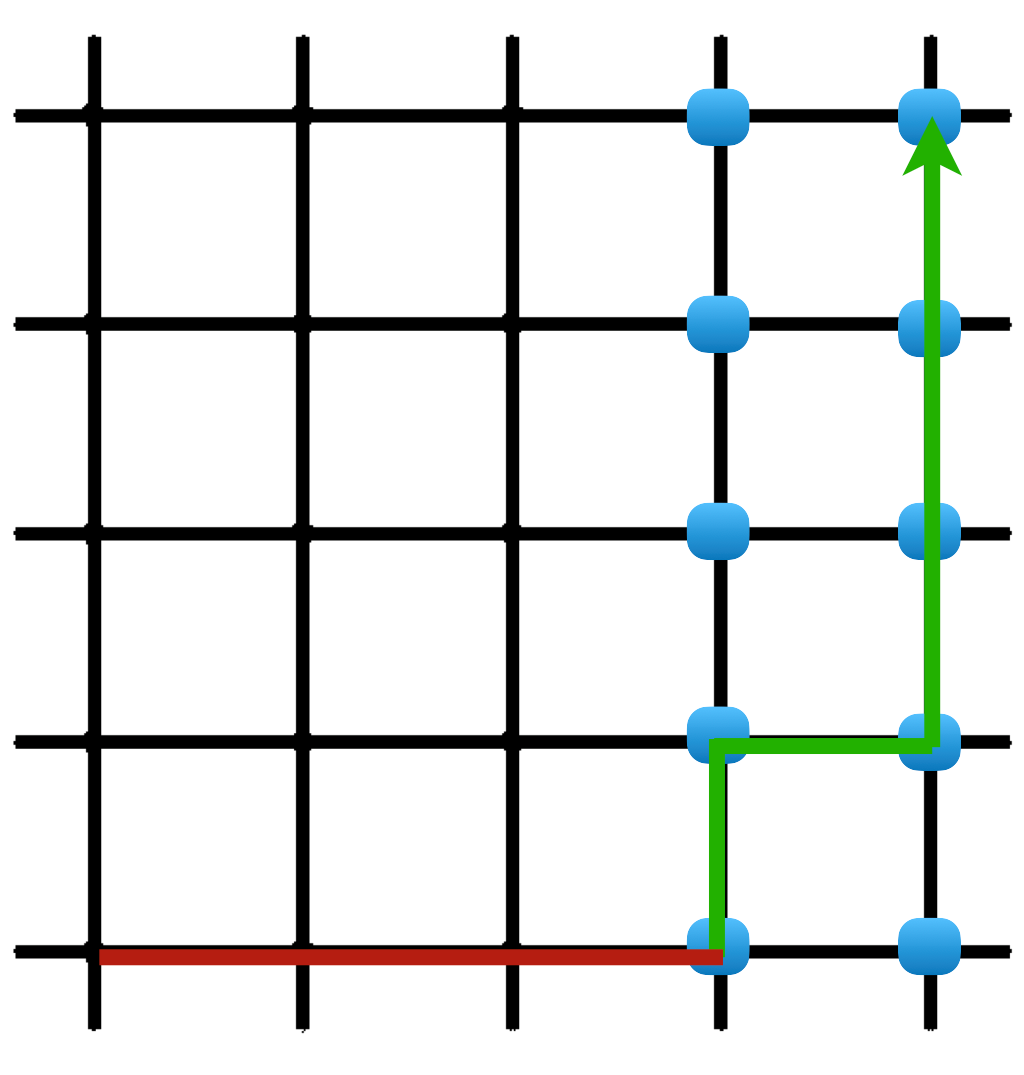}}%
\hfill
\subfloat[Config 3]{\label{fig_routing_config3} \includegraphics[width=0.20\textwidth]{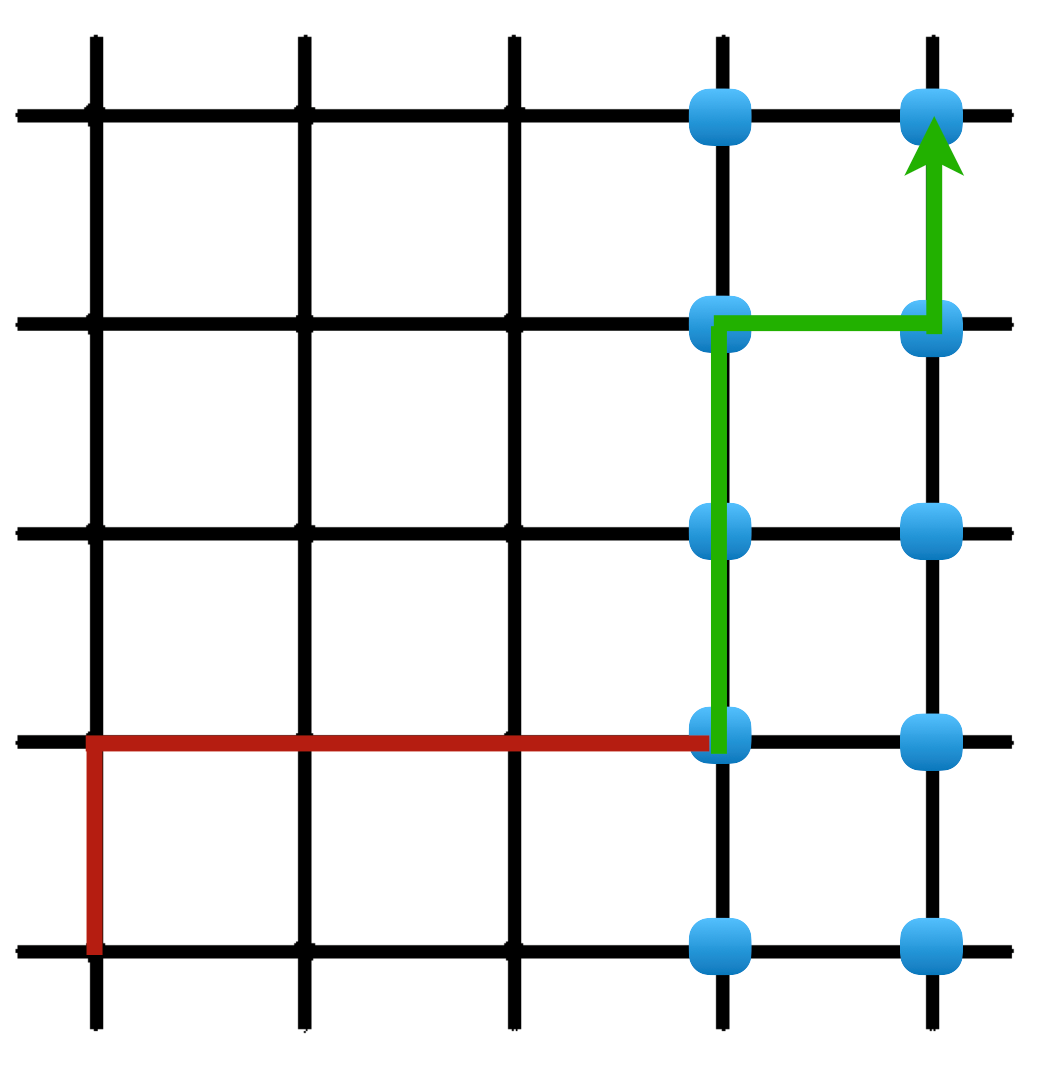}}%
\hfill
\subfloat[config 4]{\label{fig_routing_config4} \includegraphics[width=0.20\textwidth]{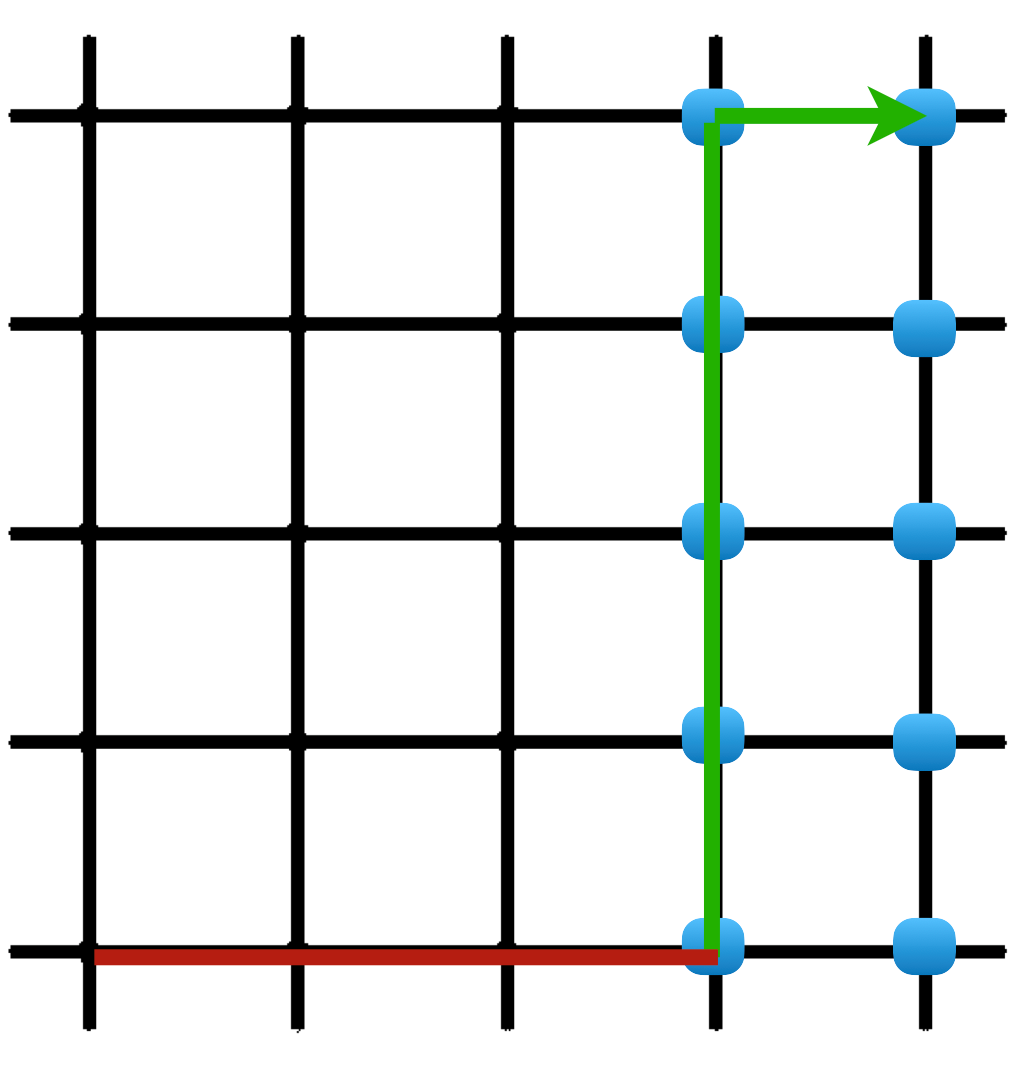}}%
\caption{EMV's routing choice on Emergency-capacitated Synthetic $\text{Grid}_{5\times 5}$ based on EMVLight. Emergency lane established for EMV passage on segments highlighted in green, and not established on segments in red.}
\label{fig_routing_synthetic}
\end{figure}

Routes selected by EMVLight are demonstrated in Fig.~\ref{fig_routing_synthetic} for all four configurations. As this is a regular grid, first we notice that the length of all EMV routes are the Manhattan distance between the origin and destination. This is the shortest length possible to achieve successful EMV dispatch. 
The four routes confirm that EMVLight directs EMVs to enter the region in the east as soon as possible to leverage the extra emergency capacity for full speed passage. This results in 4 full speed links in Config 1 and 3 as well as 5 full speed links in Config 2 and 4.

\begin{figure}[h]
\centering
\begin{subfigure}{.5\textwidth}
  \centering
  \includegraphics[width=.4\linewidth]{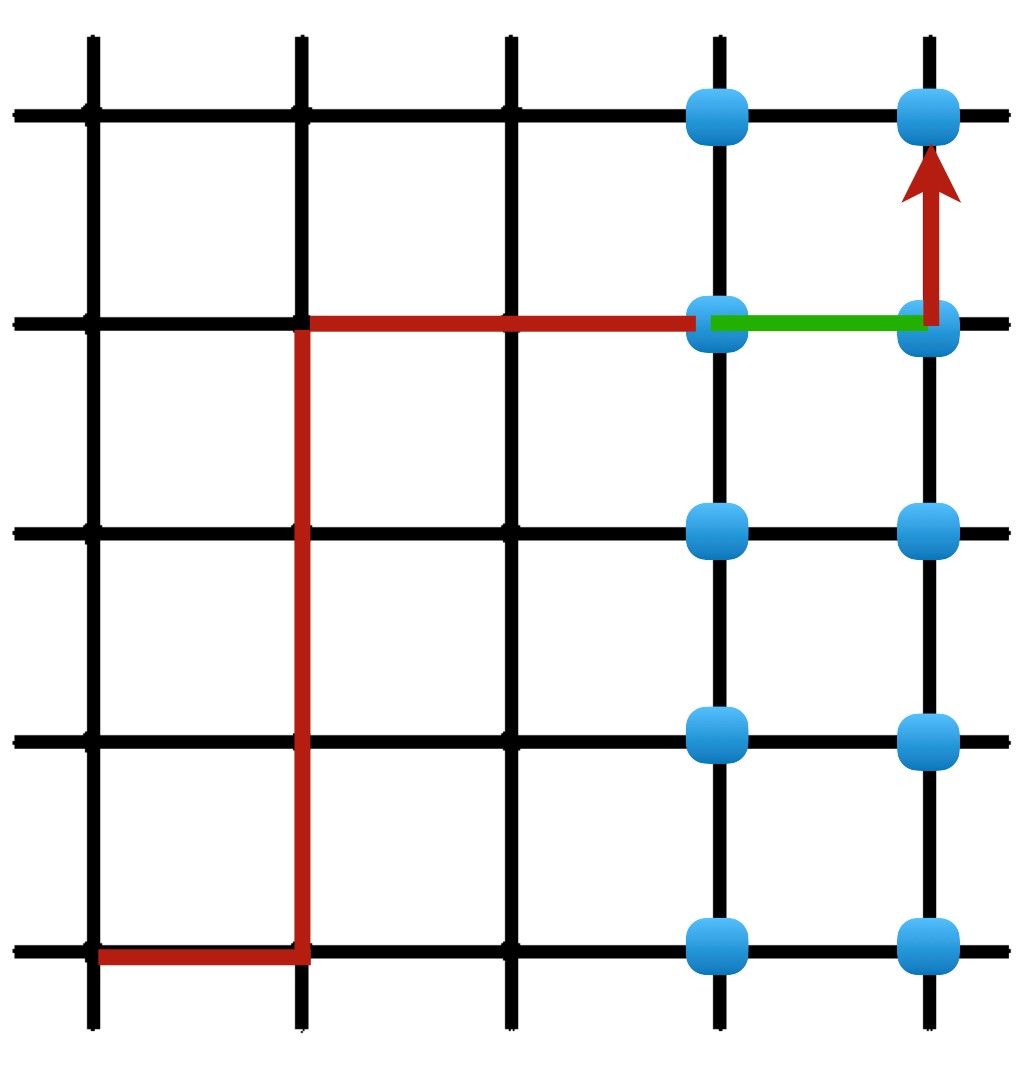}
  \caption{W+static+MP}
  \label{fig_route_choices_static}
\end{subfigure}%
\begin{subfigure}{.5\textwidth}
  \centering
  \includegraphics[width=.4\linewidth]{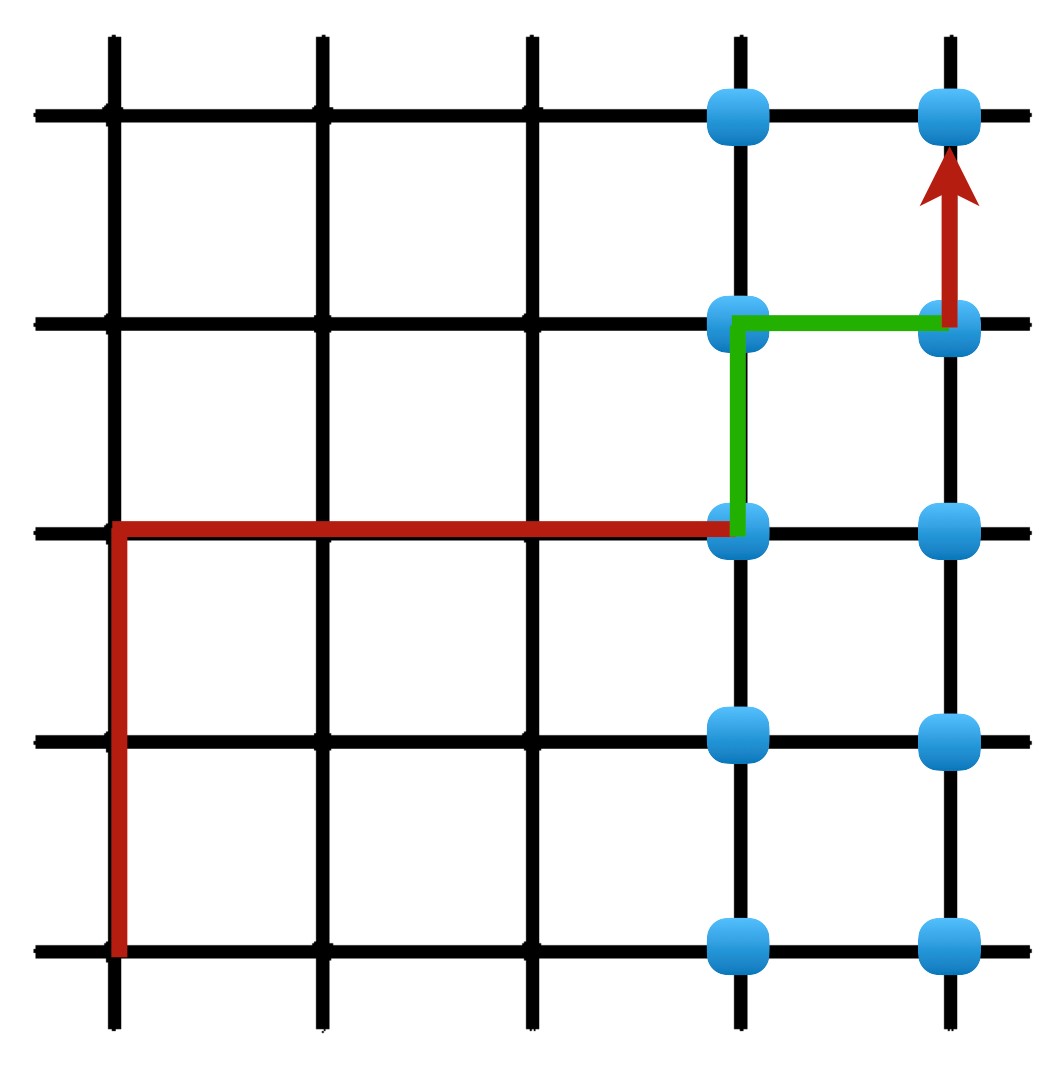}
  \caption{W+dynamic+MP}
  \label{fig_route_choices_dynamics}
\end{subfigure}
\caption{EMV's routing choice on config 1 of Emergency-capacitated Synthetic $\text{Grid}_{5\times 5}$ under MP-based benchmarks.}
\label{fig_route_choices}
\end{figure}

We also present the route choices of W+static+MP and W+dynamic+MP in Config 1, of this map, as shown in Fig.~\ref{fig_route_choices}. 
By comparing these routes with EMVLight (Fig.~\ref{fig_routing_config1}), we can clearly see the benchmark method cannot actively leverage the emergency capacity, generating routes with only 1 full speed segment (Fig.~\ref{fig_route_choices_static}) and 2 full speed segment (Fig.~\ref{fig_route_choices_dynamics}). 
% \hs{Comparing with MP-based strategies (see Fig. \ref{fig_route_choices}), we reveal that neither benchmarks proactively guide EMVs into emergency-capacitated zone. Static routing alternative shows only one full speed segment for EMVs and dynamic routing experiences 2 full speed segments, elaborating the significant difference in $T_{\textrm{EMV}}$}.

% The route choices present that EMVLight, after training, is capable of guiding EMVs to immediately traverse into the emergency-capacitated ``zone'' to take the advantage of potential emergent lanes. 
% Notice that in Fig.\ref{fig_routing_config1} an emergent lane has not been established on an segment with emergency capacity. A plausible explanation is EMVLight faced an unanticipated overflow of non-EMVs which get pushed onto that particular segment. This unexpected overflow may come from the noise of the non-EMVs' routing model or an underestimation of emergency capacity of this particular segment. We consider this as a learning limitation and leave it as a future work. 

\begin{figure}[ht]
\centering
\begin{subfigure}{.30\textwidth}
    \centering
    \includegraphics[width=.95\linewidth]{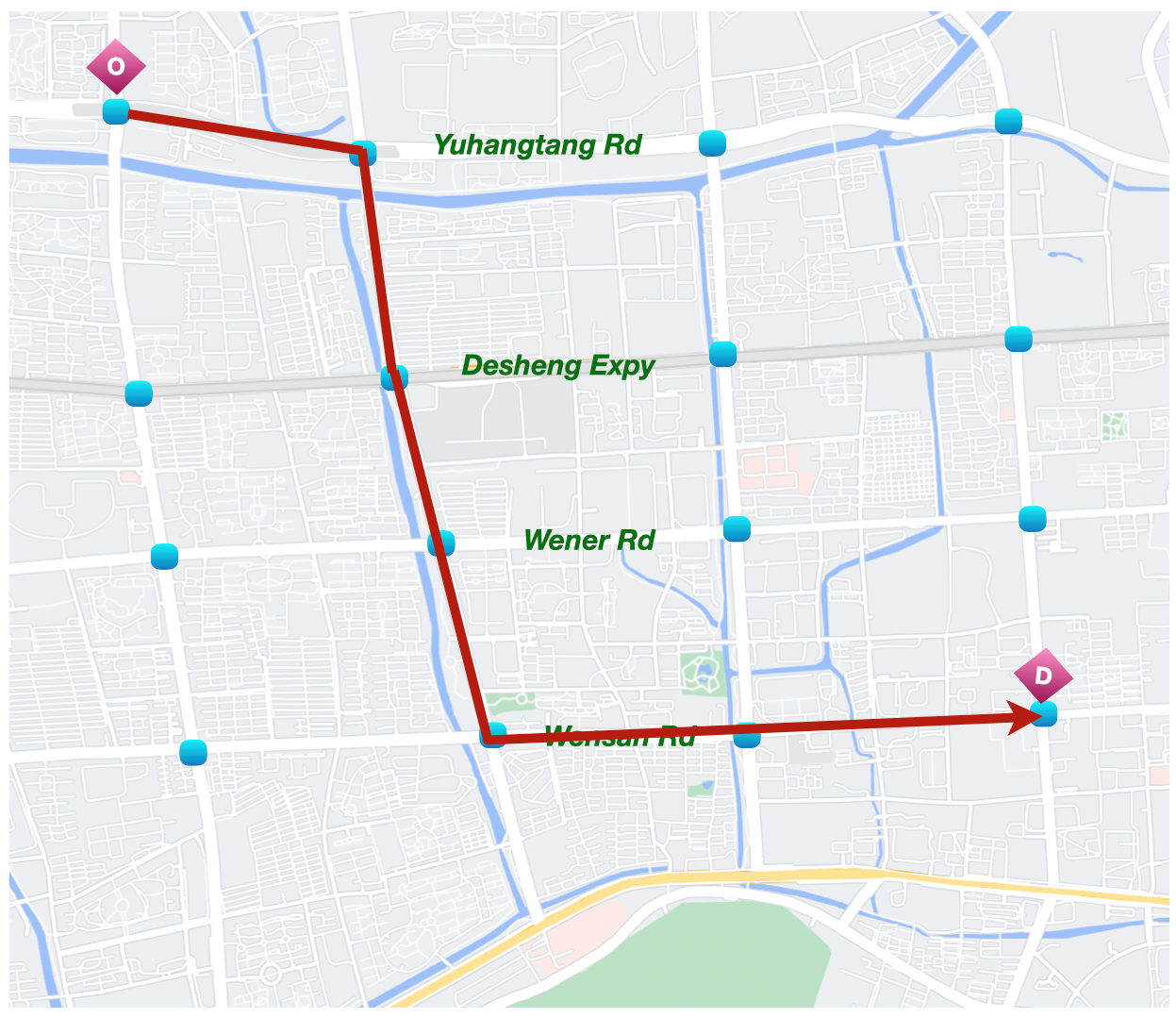}  
    \caption{\bm{$5.5$}km, $277.20 \pm 14.42$s}
    \label{SUBFIGURE LABEL 1}
\end{subfigure}
\begin{subfigure}{.30\textwidth}
    \centering
    \includegraphics[width=.95\linewidth]{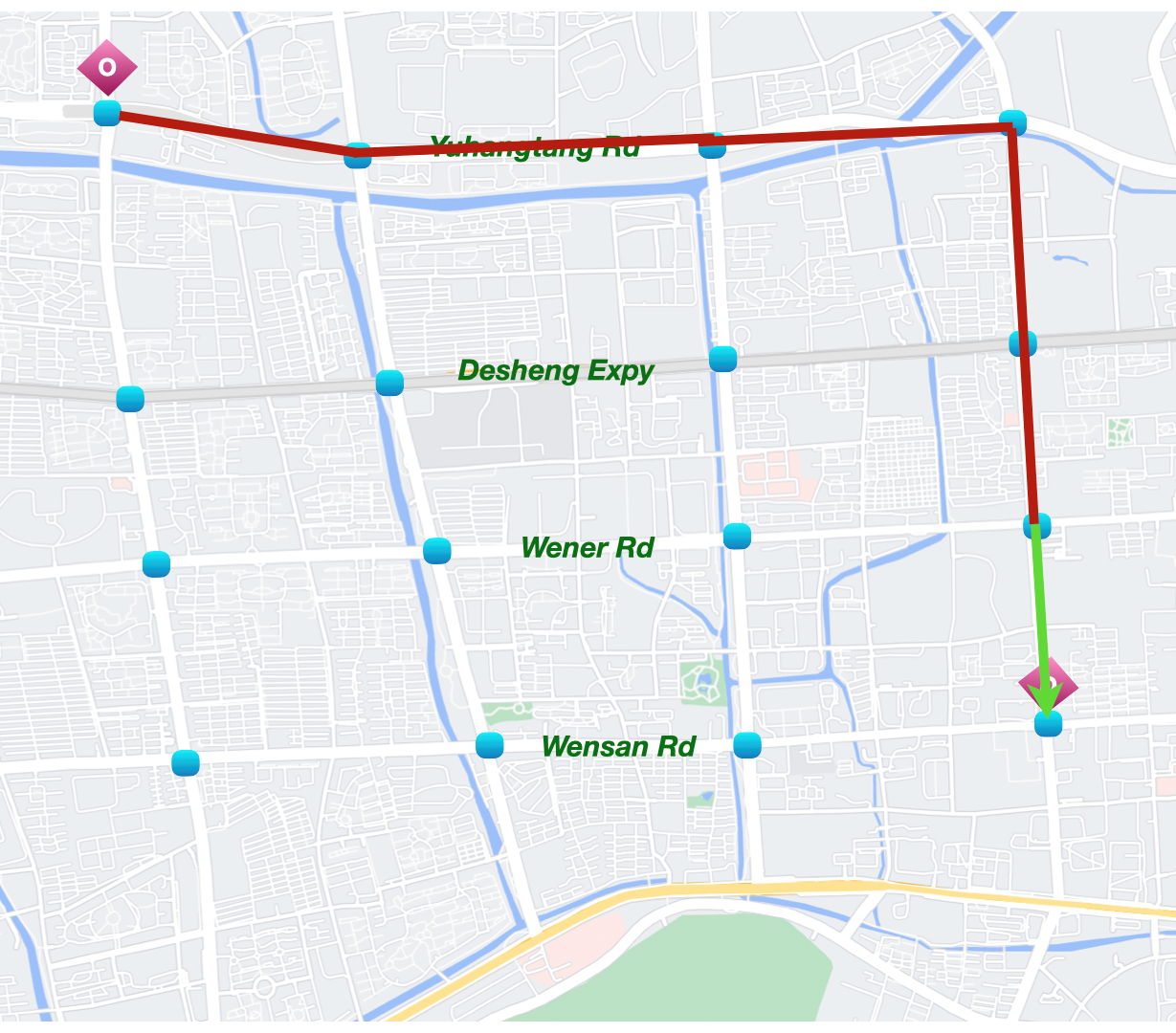}  
    \caption{$5.9$km, $228.42\pm 12.28$s}
    \label{SUBFIGURE LABEL 2}
\end{subfigure}
\begin{subfigure}{.30\textwidth}
    \centering
    \includegraphics[width=.95\linewidth]{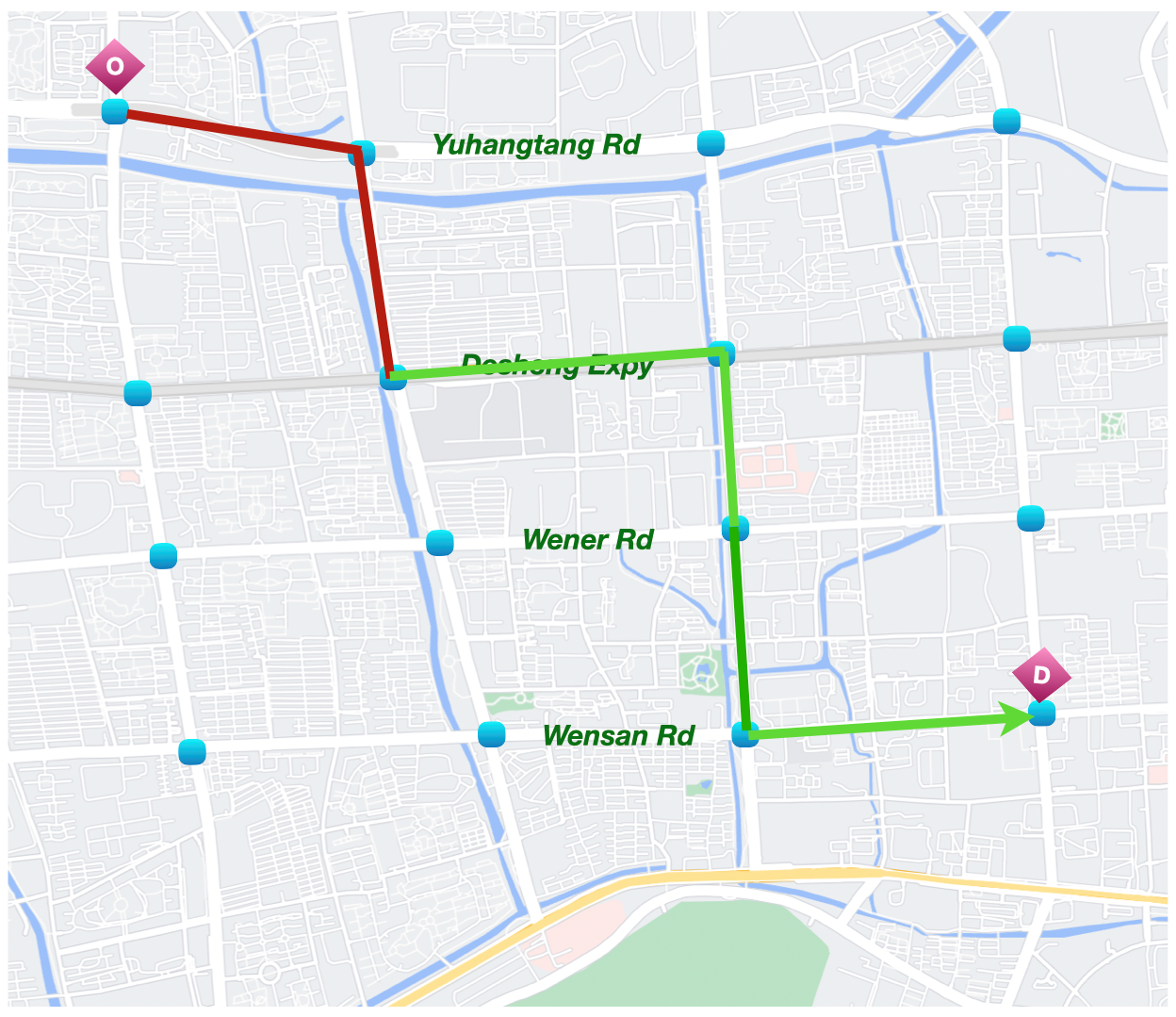}  
    \caption{$6.0$km, $\bm{194.52}\pm9.65$s}
    \label{fig_emvlight}
\end{subfigure}
\caption{The corresponding route selected by (a) W+static+MP, (b) W+dynamic+MP, (c) EMVLight on $\text{Hangzhou}_{4 \times 4}$. Distance and $T_{\textrm{EMV}}$ for the selected path are indicated. The lowest values are highlighted in bold.}
\label{fig_gudang_routings}
\end{figure}

\paragraph{\texorpdfstring{$\textrm{Hangzhou}_{4 \times 4}$}{Hangzhou} routes} 
Since $\textrm{Hangzhou}_{4 \times 4}$ is an irregular grid where different links have different lengths, the routes optimized by different algorithms have different total lengths. This provides another perspective on evaluating routing performance of different models. 
Fig.~\ref{fig_gudang_routings} shows EMV routes given by W+static+MP, W+dynamic+MP and EMVLight in this grid.
% as an illustration of static/dynamic routing difference.
By comparing the total distances of the three routes, we find that the route chosen by EMVLight is the longest among the three models. However, EMVLight achieves the smallest EMV travel time on this route. This is because EMVLight is able to coordinate traffic signals to leverage the emergency capacity to let the EMV travel at its maximum speed on more than half of the route, indicated by the green segments. 

As for static routing (W+static+MP), it chooses a fixed route given the traffic conditions upon dispatching, favoring the shortest-in-distance route.
% It is not hard to notice that EMVLight chooses the longest-in-distance but shortest-in-time path for EMV passage. Static routing method chooses a fixed route given the traffic conditions upon dispatching, favoring the shortest-in-distance route.
The dynamic routing method (W+dynamic+MP) recalculate the time-based shortest path every 30 seconds as EMV travels. Most of the time, however, Max Pressure fails to reduce the number of vehicles on the upcoming links to enable an emergency lane. Fig.~\ref{SUBFIGURE LABEL 2} shows that the emergency lane is only formed in one of the six links.
% while it fails to effectively communicate with upcoming traffic intersections. 
EMVLight is able to further reduce EMV travel time partly because the time-based shortest path is updated in real time. More importantly, EMVLight is able to reduce the number of vehicles in upcoming links so that emergency lanes can be formed. This can be attributed to the design of primary and secondary preemption agents in EMVLight, which will be further examined in Sec.\ref{sec_ablation_reward}.

% \dz{[If it is possible, maybe we can provide evidence to this statement by running the ablation study (replace secondary-preemption agent by normal agents) on this map to confirm that it indeed outputs fewer green segments]}

% EMVLight, however, computes the time-based shortest path in real time and pro-actively communicate with upcoming intersections. 
% These intersections, after learning, manages to ``reserve'' links to establish an emergent lane for incoming EMVs. It is not hard to observe that EMV make the most of emergency capacities and traverse almost freely on 4 segments, out of 6 segments in total, and arrive at the destination within the shortest amount of time. 

\subsection{Ablation Studies}

\subsubsection{Ablation studies on reward}\label{sec_ablation_reward}
We propose three types of agents and design their rewards (Eqn.~\eqref{eqn:reward}) based on our improved pressure definition and heuristics. 
In order to see how our improved pressure definition and proposed agent types influence the results, we propose three ablation studies:
\begin{enumerate}
    \item replacing our pressure definition by that defined in PressLight
    \modi{\item replacing our pressure definition by $
    P_{i} = \sum _{l\in \mathcal{I}_i} w(l)$}
    \item replacing secondary pre-emption agents with normal agents
    \item replacing primary pre-emption agents with normal agents
\end{enumerate}

\begin{table}[h]
\centering
\fontsize{10.0pt}{10.0pt} \selectfont
\begin{tabular}{@{}ccccc|c@{}}
\toprule[1pt]
Ablations                        & Ablation 1 & \modi{Ablation 2} & Ablation 3 & Ablation 4 & EMVLight\\ \midrule
$T_{\text{EMV}}$ [s]         & 205.20 $\pm$ 6.92   &  \modi{237.40 $\pm$ 10.08}  & 311.52   $\pm$ 5.18                   & 384.71 $\pm$ 8.52    & \textbf{194.52} $\pm$ 9.65       \\
$T_{\text{avg}}$ [s]  & 389.14 $\pm$ 8.40 & \modi{382.07 $\pm$ 6.92}  & 442.73 $\pm$ 6.65 & 444.15 $\pm$ 7.02  & \textbf{331.42} $\pm$ 6.18      \\ \bottomrule[1pt]
\end{tabular}
\caption{Ablation studies on pressure-based reward design and agent types. Experiments are conducted on $\textrm{Hangzhou}_{5 \times 5}$. The lowest value are highlighted in bold.}
\label{tab_ablation_reward}
\end{table}

\begin{figure}[h]
\centering
\begin{subfigure}{.5\textwidth}
  \centering
  \includegraphics[width=.6\linewidth]{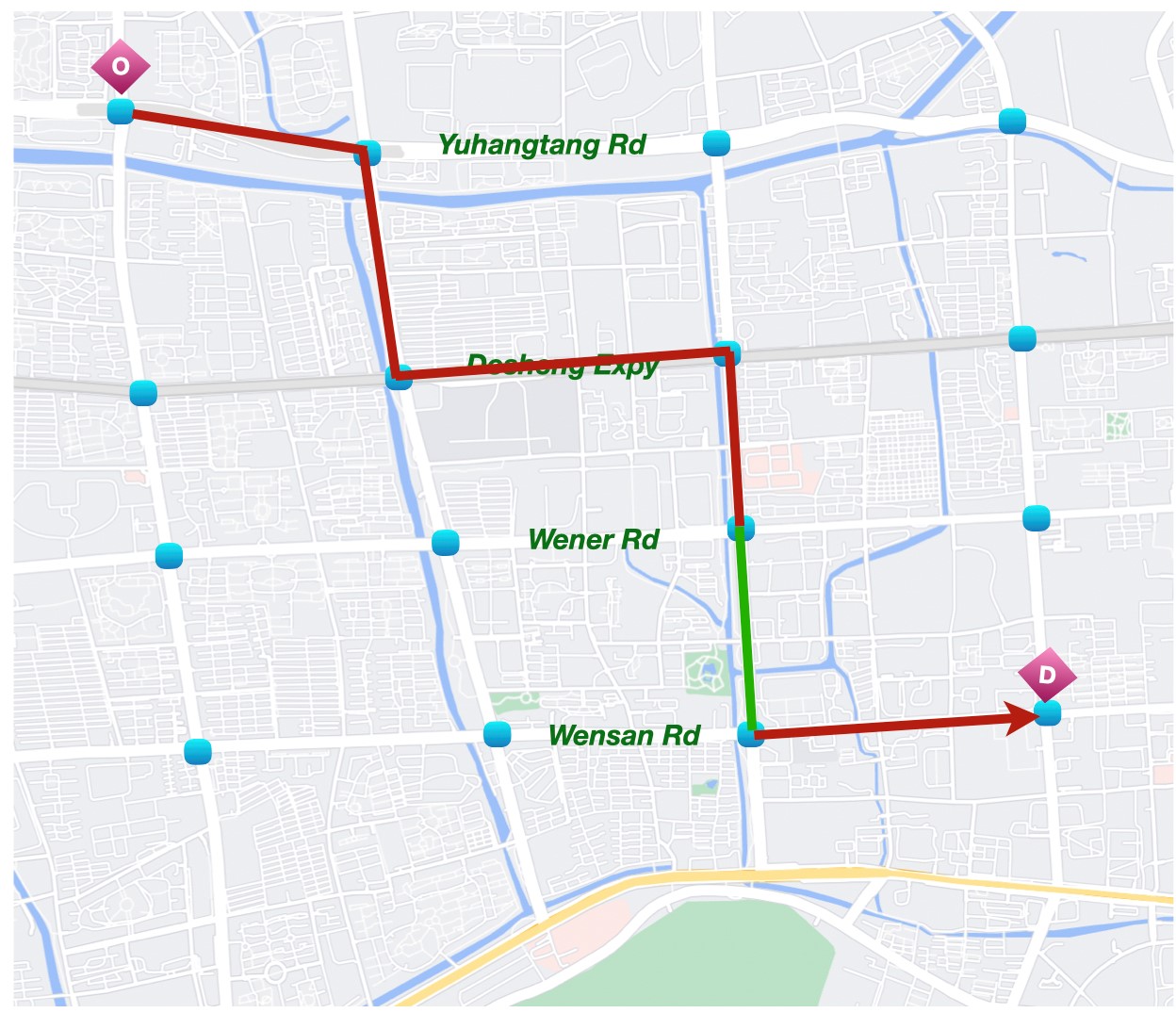}
  \caption{Without secondary agents}
  \label{fig_route_choices_ablation_secondary}
\end{subfigure}%
\begin{subfigure}{.5\textwidth}
  \centering
  \includegraphics[width=.6\linewidth]{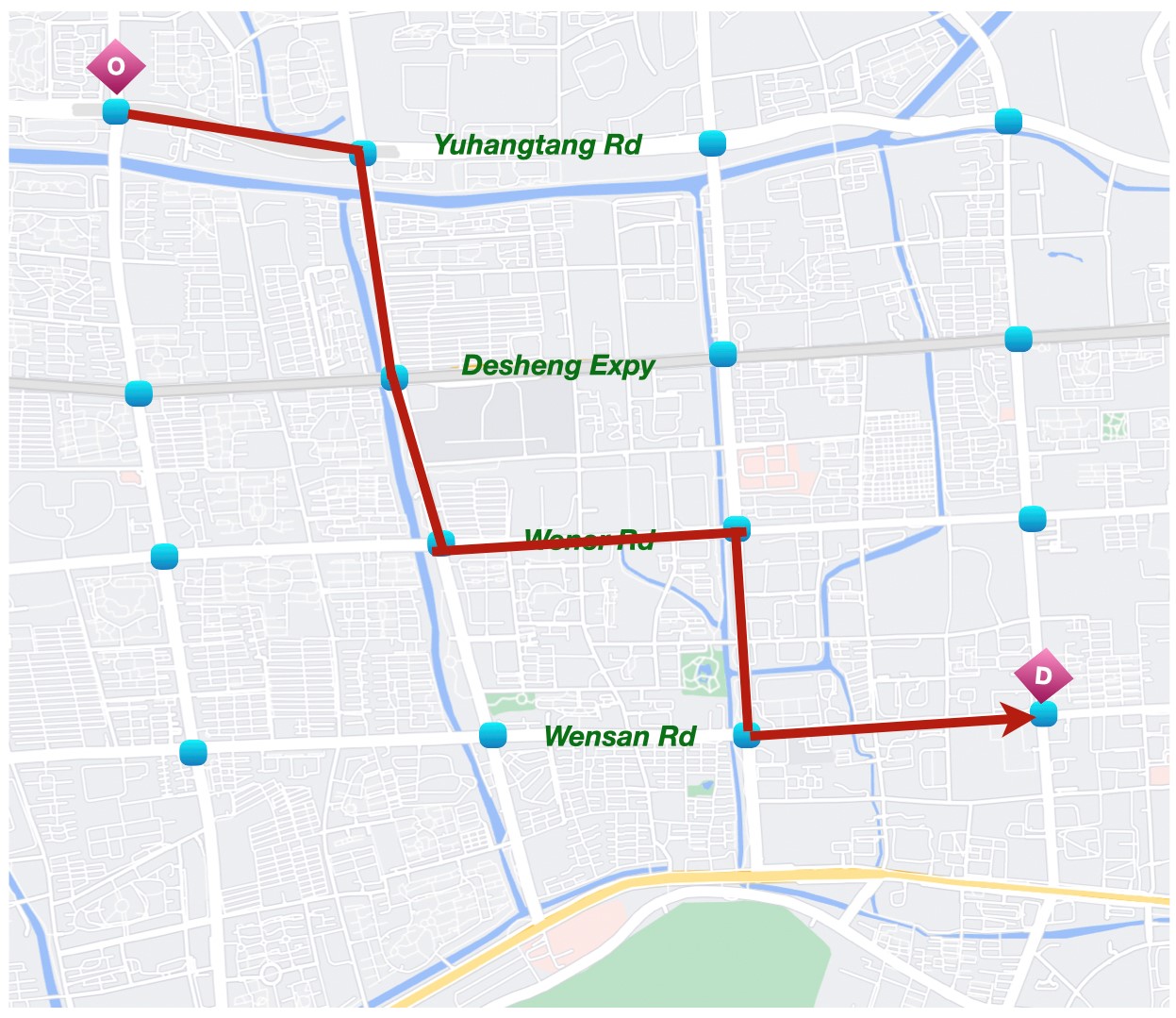}
  \caption{Without primary agents}
  \label{fig_route_choices_ablation_primary}
\end{subfigure}
\caption{EMV's route choice on $\text{Hangzhou}_{4\times 4}$ with replaced primary (a) and secondary (b) agents.}
\label{fig_route_choices_ablation}
\end{figure}

Table \ref{tab_ablation_reward} shows the results of these ablations on the $\text{Hangzhou}_{4\times 4}$ map. We observe that PressLight-style pressure yields a slightly larger $T_{\textrm{EMV}}$ but significantly increases the $T_{\textrm{avg}}$. Without secondary pre-emption agents, $T_{\textrm{EMV}}$ increases by 60\% since almost no ``link reservation" happened. \modi{In Ablation 2, if we do not take the average while keep the sum of absolute value of lane pressures as the intersection pressure, we notice a slightly smaller $T_{\text{avg}}$ but a 15\% increase in $T_{\text{EMV}}$. EMVLight's pressure design outperforms the ones in Ablation 1 and 2 and has been proven as the most suitable pressure design for this particular task.} Moreover, without primary pre-emption agents, $T_{\textrm{EMV}}$ increases considerably, which again proves the importance of pre-emption. We can further confirm the importance of agent designs by inspecting the selected routes in the last two ablation studies. 
Fig. \ref{fig_route_choices_ablation_secondary} and \ref{fig_route_choices_ablation_primary} shows routes selected by EMVLights after replacing secondary agents and primary agents with normal agents, respectively. 
Even though the routes are similar as that in Fig.~\ref{fig_emvlight}, much fewer emergency lanes are successfully formed. This failure of utilizing emergency capacity lead to the significant increase in EMV travel time as shown in Table~\ref{tab_ablation_reward}.
As we can see from the routes, EMVs barely take advantage of emergency yielding during their trips.

% This is because agents now treat EMVs as normal vehicles and do not select the corresponding pre-emption phase, resulting in the EMV travel time slightly smaller than the average travel time.

\subsubsection{Ablation study on policy exchanging}
\begin{figure}[ht]
    \centering
    \includegraphics[width=0.9\linewidth]{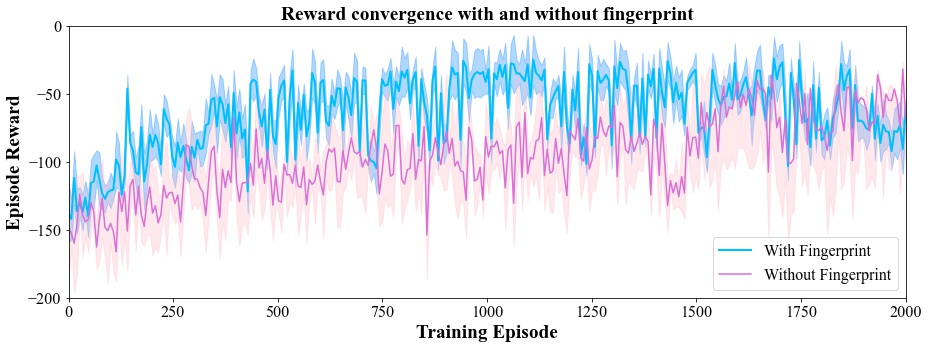}
  \caption{Reward convergence with and without fingerprint. Experiments are conducted on Config 1 synthetic $\text{grid}_{5 \times 5}$.}
  \label{fig_FP_comparison}
\end{figure}

In multi-agent RL, fingerprint has been shown to stabilize training and enable faster convergence.
% is a common technique for faster and more stable training performance. 
In order to see how fingerprint affects training in EMVLight, we remove the fingerprint design, i.e., policy and value networks are changed from $\pi_{\theta_i}(a_i^t|s^t_{\mathcal{V}_i}, \pi^{t-1}_{\mathcal{N}_i})$ and  $V_{\phi_i}(\Tilde{s}^t_{\mathcal{V}_i}, \pi^{t-1}_{\mathcal{N}_i})$ to $\pi_{\theta_i}(a_i^t|s^t_{\mathcal{V}_i})$ and  $V_{\phi_i}(\Tilde{s}^t_{\mathcal{V}_i})$, respectively. 
Fig.~\ref{fig_FP_comparison} shows the influence of fingerprint on training. With fingerprint, the reward converges faster and suffers from less fluctuation, confirming the effectiveness of fingerprints, i.e. policy exchanging.
% We provide a comparison here, see Figure \ref{fig_FP_comparison}, on synthetic $\text{grid}_{5 \times 5}$ with configuration 1 to justify whether or not fingerprint facilitates our training.
% As we can see, enabling policy sharing between agents indeed accelerate agents' learning. Training without fingerprint, although eventually reach the similar reward level, experiences more fluctuations in the learning curve.

% \subsection{Sensitivity test on emergency capacity}
% Comparison between Synthetic $\textrm{Grid}_{5\times5}$ and Emergency-capacitated Synthetic $\textrm{Grid}_{5\times5}$ reveals
\subsection{Sensitivity Analysis on \texorpdfstring{$\beta$}{beta}}
\modi{
The value of $\beta$'s in Eqn.~\eqref{eqn:reward2} represents the weights assigned to the two tasks. When $\beta = 1$, the secondary pre-emption agents will act like normal agents and only focus on congestion alleviation. When $\beta = 0$, the secondary pre-emption agents concentrate on emptying non-EMVs and reserve the whole segment for the approaching EMV. In order to illustrate the tradeoff, we conduct experimental runs on the emergency-capacitated synthetic $\text{Grid}_{5\times 5}$ with configuration 1 as the sensitivity analysis on $\beta$, see Table~\ref{tab:beta}. We perform 5 independent runs for each $\beta$'s value. 

\begin{table}[ht]
\fontsize{9.0pt}{10.0pt} \selectfont
\centering
\begin{tabular}{@{}cccccc@{}}
\toprule
           & 0 & 0.25 & 0.5 & 0.75 & 1 \\ \midrule
$T_{\text{EMV}}$ & $119.10 \pm 5.14$  & $134.40 \pm 5.83$     & $158.20 \pm 6.28$    & $172.60 \pm 5.71$   &  $185.20 \pm 4.19$ \\
$T_{\text{avg}}$ & $340.23 \pm 4.75$  &  $338.17 \pm 7.25$     & $334.92 \pm 5.52$    &  $331.81 \pm 9.28 $ & $327.10 \pm 8.18$  \\ \bottomrule
\end{tabular}
\caption{EMVLight's performance on Emergency-capacitated Synthetic $\text{Grid}_{5\times 5}$ with Configuration 1 given different values of $\beta$.}
\label{tab:beta}
\end{table}

As we can see from Table \ref{tab:beta}, $T_{\textrm{EMV}}$ obtains the minimum value when $\beta$ is 0 since the EMV's path has been cleared and reserved in advance. Comparing to $T_{\text{EMV}}$ when $\beta = 1$, we witness a $35\%$ reduction in $T_{\text{EMV}}$ when secondary agents effortlessly reserve the segments. Likewise, $T_{\text{avg}}$ hits the smallest value when secondary agents shift their responsibility onto congestion reduction entirely. Under this scenario, pre-emption only happens on the segment which EMV is currently traveling on and the EMV will travel with the slowest speed since more vehicles appear ahead of its road. Although every non-EMV saves about 10 seconds for the whole trip in this case, an ambulance might delay for more than one minute and a precious life might be lost.
We believe it would be the best for the traffic system administrator to decide $\beta$ given the emergency of the case.
}
\section{Summary}\label{sec_conclusion}
In this paper, we introduced a decentralized reinforcement learning framework, EMVLight, to facilitate the efficient passage of EMVs and reduce traffic congestion at the same time.
Leveraging the multi-agent A2C framework, agents incorporate dynamic routing and cooperatively control traffic signals to reduce EMV travel time and average travel time of non-EMVs.
Our work considers the realistic settings of emergency capacitated road segments and traffic patterns before, during and after EMV passages. Extending Dijkstra's algorithm and embedding into the multi-class RL agent design, EMVLight fundamentally addresses the coupling challenge of EMV's dynamic routing and traffic signal control, filling the research gap on this particular task.
% Adopting the multi-agent A2C framework for the agents' communication, we manages to dynamically updating the shortest path for EMVs as well as reduces average travel time for non-EMVs.
Evaluated on both synthetic and real-world map, EMVLight shortens the EMV travel time and average travel time by up to $42.6\%$ and $23.5\%$ respectively, comparing with existing methods from traditional and learning-based traffic signal control for EMV-related managements. Both quantitative and qualitative assessments on EMVLight, including its EMV navigation as well as pre-clear and restore traffic conditions under emergency state, have concluded that EMVLight is a promising control scheme for such task.

Non-trivial future directions for this study including, but not limited to, the followings. First, the interactions among vehicles, especially under emergency, are extremely complicated. As an effort to close the gap between simulation and reality, we are motivated to extend current ETA estimation model to capture more realistic traffic patterns with EMVs so that agents are more responsive in the field tests. Second, we are looking forward to navigating multiple EMVs simultaneously in the same traffic network. The reward design for pre-emption agents (imagine two or more EMVs appears within one intersection) is worth a technical and ethical discussion. Last but not least, as one of the RL applications in the field of transportation, our method has required a massive number of updated iterations, even in the simulated environments, to achieve the desirable result. How to learn efficiently so that our trail-and-error attempts would not bring catastrophic impacts on real traffic is a critical question for all ITS RL applications.

\chapter{Dynamic Queue-jump Lane for Emergency Vehicles: a Multi-agent Proximal Policy Optimization Approach}\label{chap:mappo-dqjl}
\section{Introduction}
\label{sec:introduction}
The increasing population and urbanization have made it exceedingly difficult to operate urban emergency services efficiently. Over the past decade, response times for emergency vehicles (EMVs), such as ambulances, fire trucks, and police vehicles, have worsened due to rising traffic congestion. Historical data from New York City, USA~\cite{NY2019}, shows that the number of annual emergency incidents increased from 1,114,693 in 2004 to 1,352,766 in 2014, with average response times increasing from 7 minutes 53 seconds to 9 minutes 23 seconds~\cite{Emergency2014}. This represents a 20\% increase in response times over a decade. Furthermore, in fiscal year 2023, the city's emergency response times worsened by an additional 82 seconds—16.1\% higher than in 2019~\cite{Calder_2023}. These delays are especially critical in life-threatening situations, such as cardiac arrests, where every minute without defibrillation reduces survival chances by 7\% to 10\%, with survival rates dropping sharply after 8 minutes~\cite{Heart2013}.

Given the urgent need to reduce EMV response times, technological solutions must be explored to mitigate the impact of traffic congestion. One promising direction is the use of connected vehicle technologies, specifically vehicle-to-everything (V2X) communication, which allows for real-time information sharing between EMVs, non-EMVs, and traffic management systems. These systems enable EMVs to receive optimal routing information based on current traffic conditions, and they facilitate coordination between non-EMVs to clear paths for approaching EMVs.

This study focuses on a critical yet specific challenge within the focused context of intra-link movement of EMVs: the establishment of dynamic queue jump lanes (DQJL) to expedite EMV passage through congested road segments.

\begin{definition}[Dynamic Queue Jump Lane]\label{def:dqjl}
A DQJL is a dynamically formed lane created while an EMV is approaching or en route, clearing downstream space to facilitate its passage. 
\end{definition}

A queue jump lane (QJL) is typically used in bus operations to allow buses to bypass long queues before intersections~\cite{Zhou2005Performance,Cesme2015Queue}. 
In the context of EMV movement, we introduced DQJL as in Definition~\ref{def:dqjl}. The formation of these lanes is facilitated by V2X technology, which enables connected and autonomous vehicles (CAVs) to coordinate their maneuvers based on real-time instructions. However, the challenge is not only to ensure rapid EMV passage but also to minimize the disruption to surrounding traffic, particularly the non-connected human-driven vehicles (HDVs), which are not directly controllable.

The task of DQJL formation is highly intricate, as it requires coordinating CAVs to influence the behavior of HDVs indirectly, a task fraught with uncertainty due to the unpredictability of human drivers. The key objective is to minimize the time it takes for the EMV to traverse the congested segment while ensuring reduced maneuvers from non-EMVs, particularly HDVs, motivating all maneuvers to be efficient. Striking a balance between these objectives—optimizing EMV passage time while reducing the complexity of maneuvers for non-EMVs—requires sophisticated coordination strategies.

To address these challenges, we propose modeling the DQJL formation as a partially observable Markov decision process (POMDP), which allows for the incorporation of both controllable CAVs and uncontrollable HDVs. The POMDP framework enables agents (CAVs) to make real-time decisions based on partial observations of the traffic environment. We adopt a multi-agent actor-critic deep reinforcement learning approach to derive optimal coordination strategies, enabling CAVs to dynamically clear lanes for EMVs while minimizing the impact on other vehicles. This approach accounts for partial connectivity, ensuring flexibility with varying levels of CAV penetration rates in mixed traffic environments.

The contributions of this work are threefold. First, we introduce the concept of DQJL as a novel traffic management strategy, providing a scalable framework for its dynamic formation in mixed traffic environments. This framework addresses the inherent complexities of heterogeneous traffic systems by integrating both CAVs and HDVs. Second, we design a multi-agent reinforcement learning algorithm tailored to optimize DQJL formation. This algorithm leverages centralized training with decentralized execution to effectively coordinate CAVs while accounting for the stochastic and reactive behavior of HDVs, ensuring robust performance under varying traffic conditions. Finally, through extensive SUMO-based simulations, we validate the proposed system’s effectiveness. Our results demonstrate substantial improvements over baseline scenarios, achieving significant reductions in EMV passage time (up to 39.8\%) and overall lane-changing maneuvers (up to 55.7\%), particularly under scenarios with increasing CAV penetration rates.

The remainder of this chapter is organized as follows. In Section~\ref{sec:literature_review}, we review relevant studies on queue-jump lanes and multi-agent reinforcement learning (MARL) frameworks in connected vehicle applications. Section~\ref{sec:modeling} formulates the DQJL problem as a partially observable Markov decision process (POMDP), differentiating between CAVs and HDVs in mixed traffic scenarios. In Section~\ref{sec:learning}, we present our MARL-based approach for solving the DQJL problem, detailing the network architecture and learning algorithms. Section~\ref{sec:experimental_setup} outlines the experimental setup. In Section~\ref{sec:discussion}, we present experimental results and key insights and a conclusion is drawn in Section~\ref{sec:conclusion}.

\section{Literature Review}
\label{sec:literature_review}
In this section, we discuss literature available on queue jump lane in Subsec.~\ref{sec:dqjl_literature}, and existing multi-agent deep reinforcement learning techniques on similar connected vehicle applications in Subsec. \ref{sec:marl_literature}.

\subsection{Queue Jump Lane for EMVs}
\label{sec:dqjl_literature}
Deep learning has recently demonstrated significant promise across various applications \cite{you2019ct,you2018structurally,lyu2018super,lyu2019super,you2019low,you2020unsupervised,guha2020deep}.
Queue jump lanes (QJLs) have never been used for EMV deployment and represent a novel operational strategy that leverages CAV technologies for EMV deployment. Although QJLs are relatively new, the literature already demonstrates their positive effects in reducing travel time variability, particularly when combined with transit signal priority (TSP). However, these studies are primarily based on moving-bottleneck models for buses \cite{Zhou2005Performance,Cesme2015Queue,cheng2017body,cheng2016hybrid}. We adapt this bus operation strategy for EMV deployment in our setting, given that EMVs typically travel faster than non-EMVs and can “preempt” non-EMV traffic due to their priority status.

At the same time, traditional siren technologies often fail to provide adequate warning time for other vehicles, and there is frequently a lack of clarity regarding which route should be avoided \cite{chen2021adaptive,you2021mrd,liu2021aligning}. This confusion, particularly under highly congested conditions, leads to increased delays, as mentioned earlier, and contributes to accident rates that are 4 to 17 times higher \cite{Buchenscheit2009AVE,cheng2016identification}, as well as increased collision severity \cite{Yasmin2012Effects,cheng2016random,li2019novel}, both of which can further delay EMV response times. A study by Savolainen et al.~\cite{Savolainen2010Effects}, utilizing data from Detroit, Michigan, confirmed the sensitivity of driver behaviors to various ITS communication methods for conveying EMV route information. 
Kohneh et al.~\cite{Kohneh2024twoways} adopts a hybrid non-dominated sorting genetic algorithm II-particle swarm optimization (NSGAII-PSO) approach to solve the bi-objective optimization problem which minimizes EMV travel time while maximizing safety.

Wu et al.~\cite{WU2020preclearing} proposed a lane pre-clearing strategy for normal road segments, employing cooperative driving with CAVs to maintain EMV speed while minimizing traffic disruption. Their approach formulates the problem as a bi-level optimization and employs an EMV sorting algorithm, effectively clearing lanes and identifying a linear relationship between optimal solutions and road density to improve EMV routing decisions. Hannoun et al.~\cite{Hannoun2019Facilitating} utilized mixed-integer linear programming (MILP) to develop a static path clearance control strategy for approaching EMVs. A follow-up work~\cite{hannoun2021sequential}, relying on V2X, extended the problem to larger distances by dividing roads into segments, allowing for a divide-and-conquer approach to optimization. These approaches simplify the road network into cells, approximating vehicle positions and velocities, limiting its precision and applicability to complex real-world traffic scenarios. Moreover, the computational complexity of MILP-based approaches makes them less suited for real-time applications in dynamic traffic settings. 

\begin{figure}[htbp]
    \centering
    \begin{subfigure}{0.7\textwidth}
        \centering
        \includegraphics[width=\linewidth]{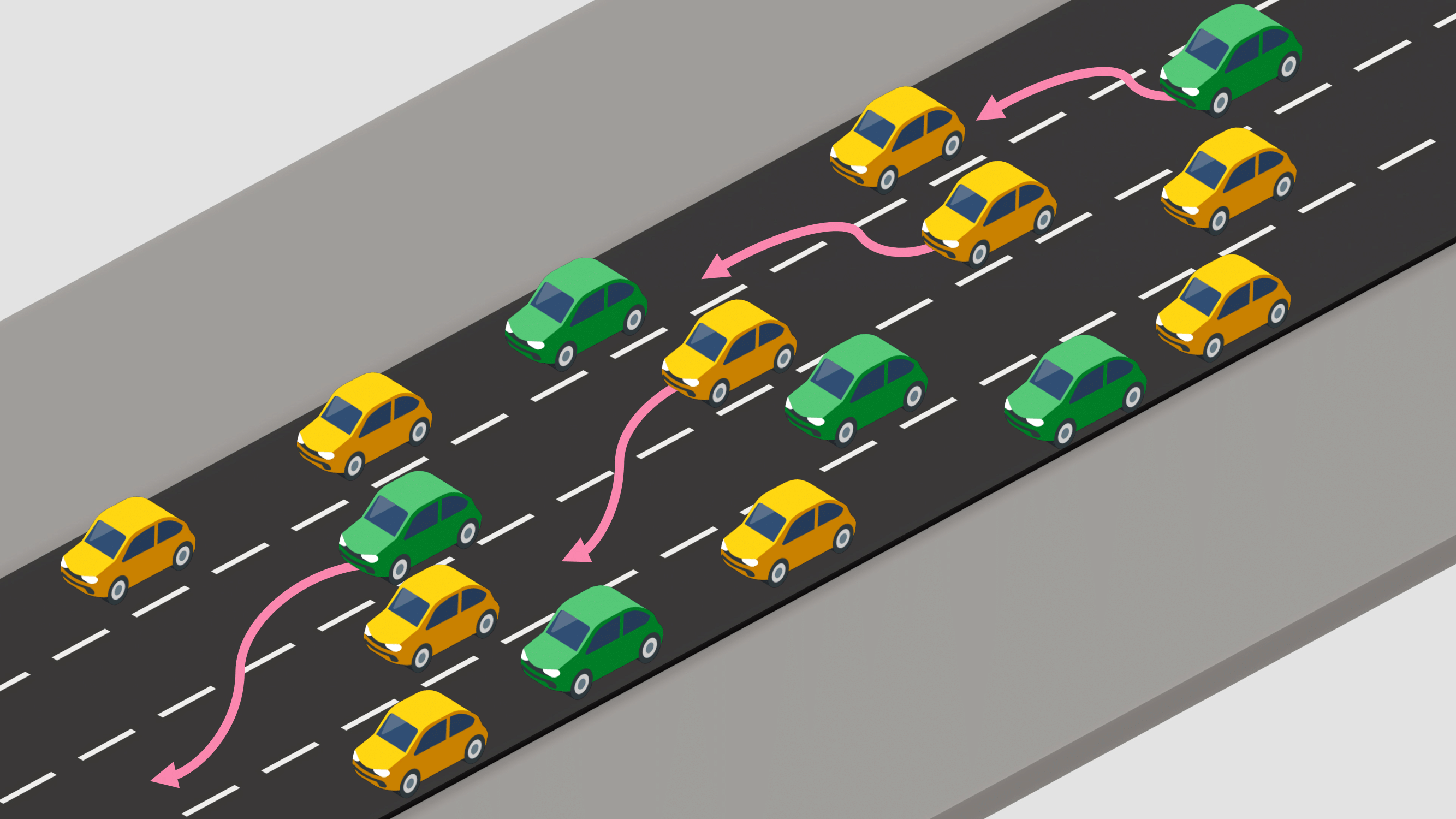}
        \caption{Actions planning.}
    \end{subfigure}

    \vspace{0.2cm} % Adds some vertical space between the figures

    \begin{subfigure}{0.7\textwidth}
        \centering
        \includegraphics[width=\linewidth]{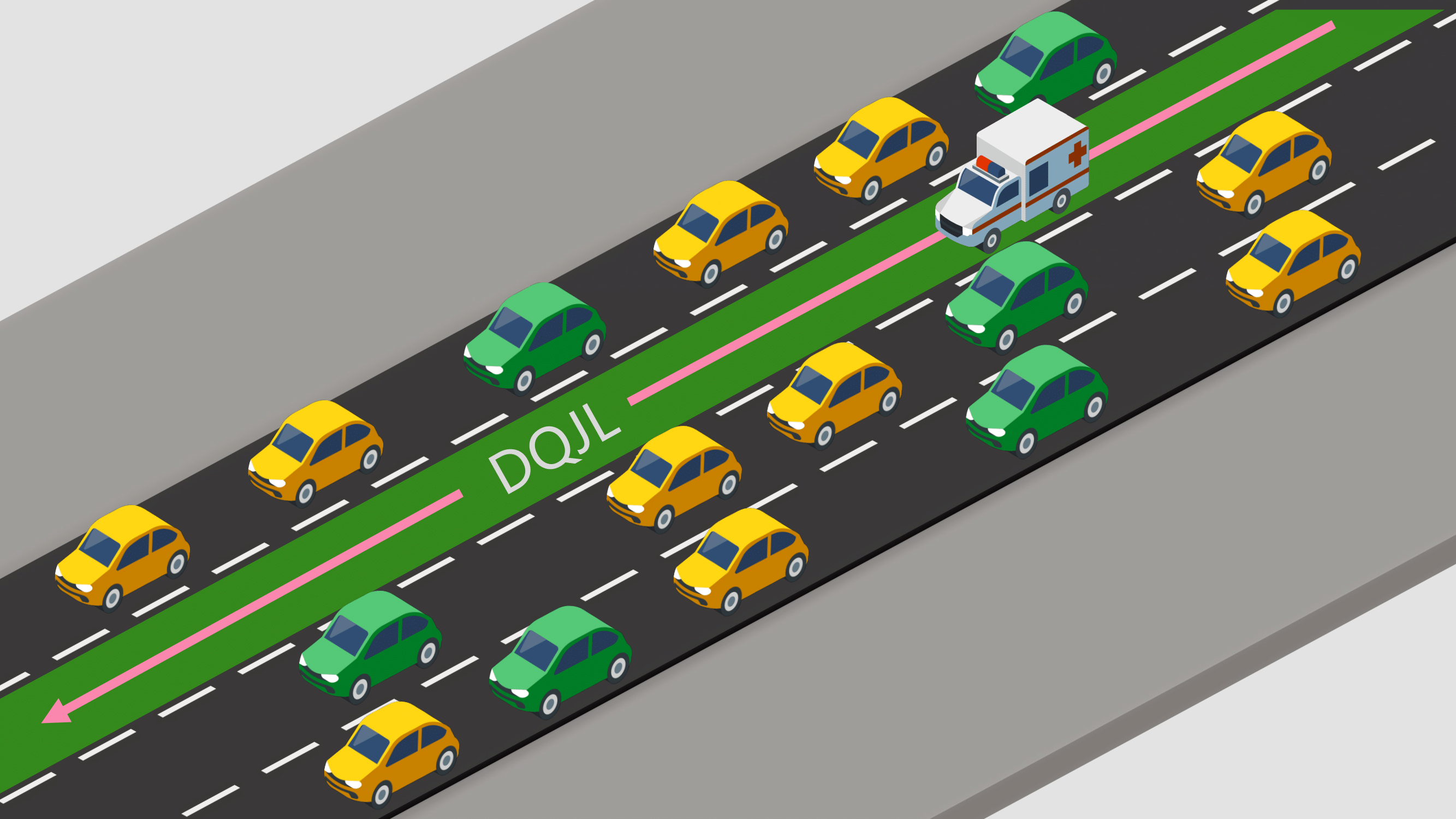}
        \caption{DQJL established.}
    \end{subfigure}
    \caption{Illustration of a DQJL formation: Green vehicles are CAVs and yellow ones are HDVs. HDVs yields and CAVs perform coordinated lane changes for a DQJL.}
    \label{fig:dqjl}
\end{figure}

QJLs have not yet been studied as a dynamic control strategy, underscoring the need to account for uncertainties in realistic, non-deterministic traffic conditions, such as yielding to an approaching EMV \cite{liu2021auto,you2022megan,liu2022retrieve}. To address this gap, we introduce the concept of dynamic queue jump lanes (DQJLs), as illustrated in Fig.~\ref{fig:dqjl}. During the DQJL clearing process for an approaching EMV, non-EMVs are continuously monitored and instructed, enabling the rapid and safe establishment of DQJLs. Additionally, we explore the application of DQJLs in partially connected scenarios, where only a subset of vehicles are equipped with communication devices, allowing for partial communication between vehicles and infrastructure.

\begin{figure}[h]
    \centering
    \begin{subfigure}{0.7\textwidth}
        \centering
        \includegraphics[width=\linewidth]{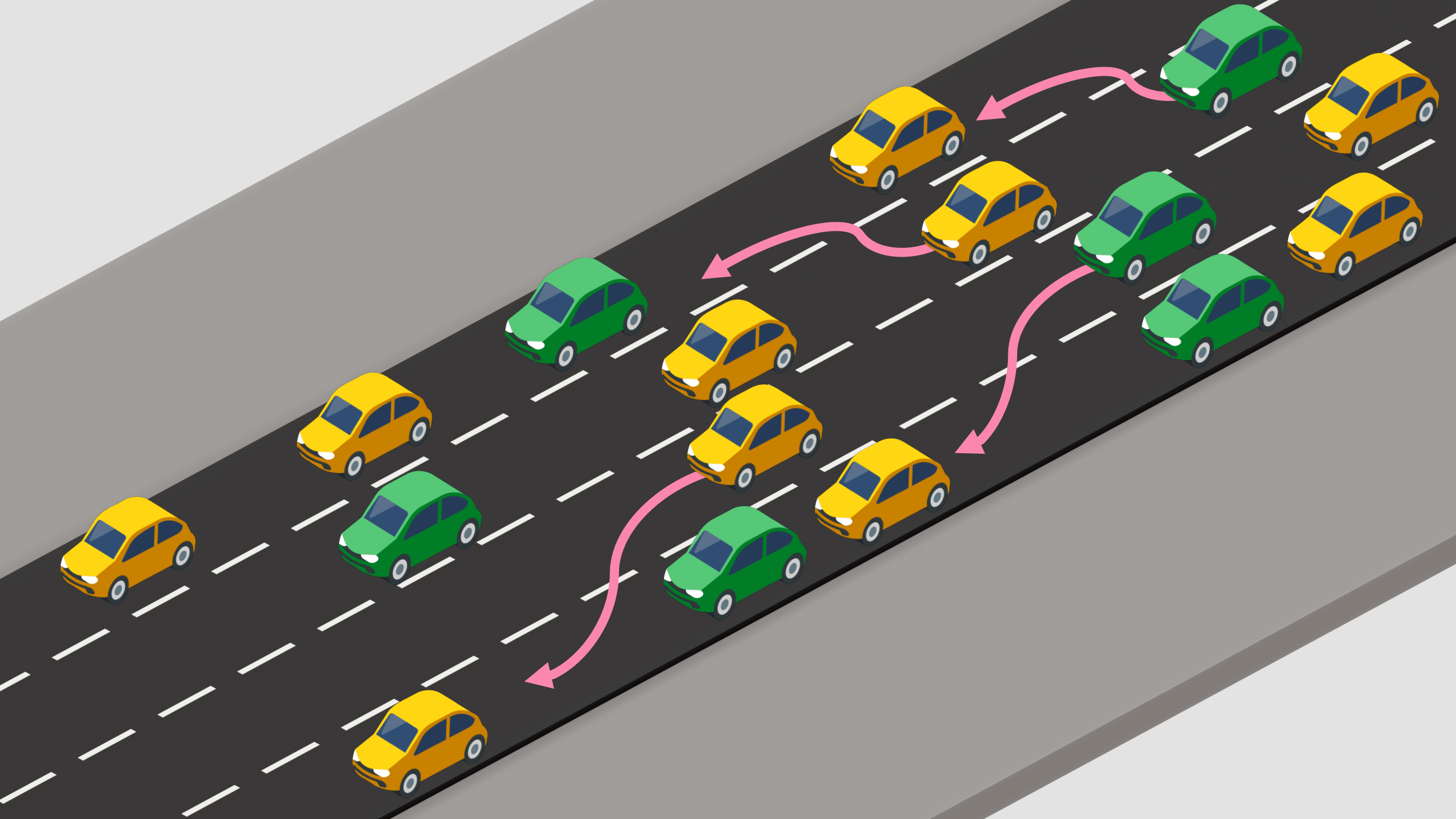}
        \caption{Actions planning.}
    \end{subfigure}

    \vspace{0.2cm} % Adds some vertical space between the figures

    \begin{subfigure}{0.7\textwidth}
        \centering
        \includegraphics[width=\linewidth]{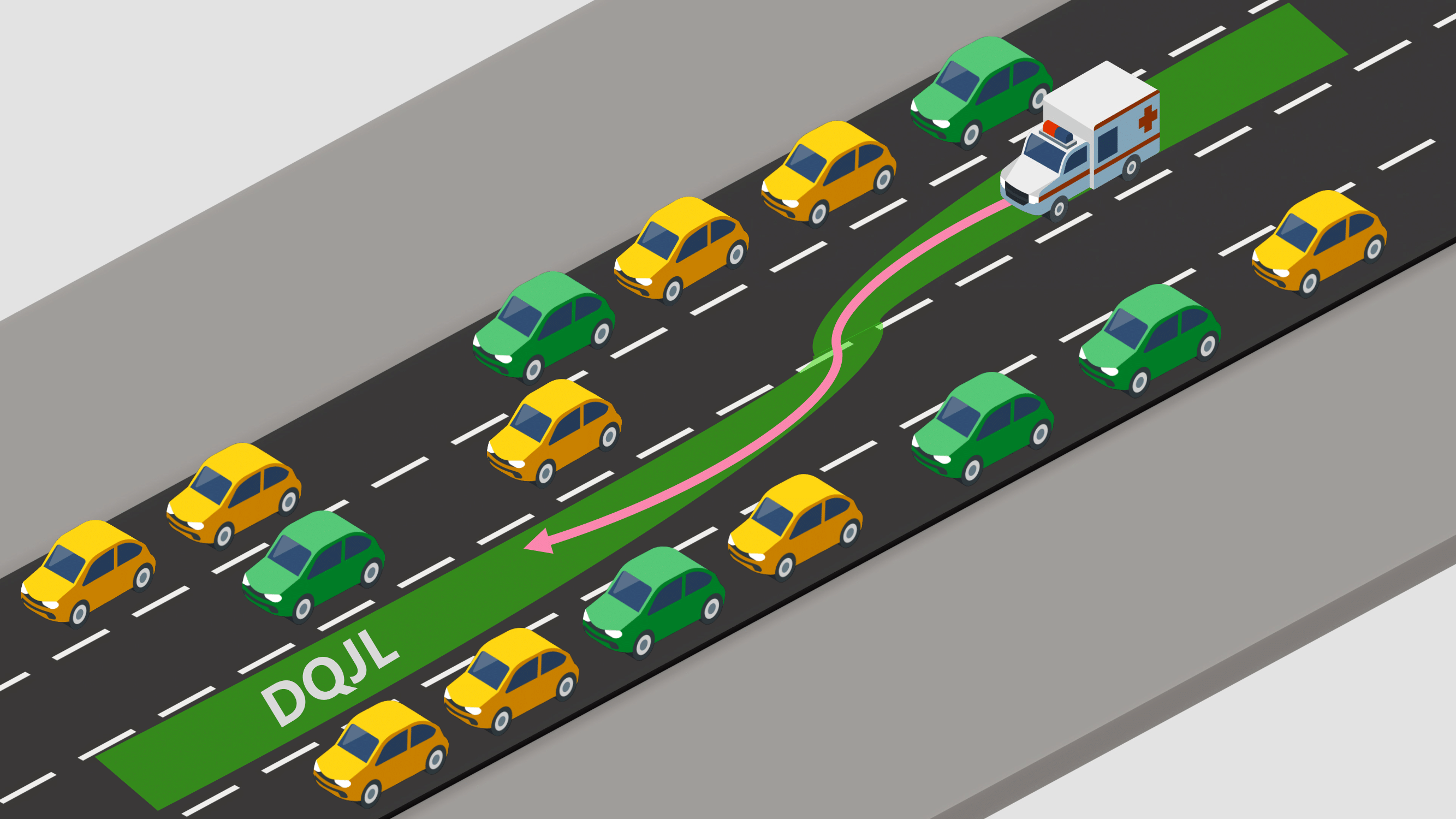}
        \caption{DQJL established.}
    \end{subfigure}
    \caption{An alternative DQJL formation, where the EMV will change lane once.}
    \label{fig:lane_change_dqjl}
\end{figure}

A related task, referred to as corridor clearance, is investigated by Suo et al.~\cite{suo2024model,liu2020rethinking}, who propose a proximal policy optimization (PPO) method to enhance clearance between intersections. In their approach, the corridor is formed by designating a single CAV as a "splitting point", which blocks following traffic to allow the EMV to switch into a less congested lane. However, their study adopts a single-agent perspective, focusing solely on the interaction between a CAV and an EMV, thereby limiting the collaborative potential of multiple CAVs in the system. Additionally, the simulation was conducted on a simple two-lane road segment, failing to capture the more complex interactions between CAVs and background traffic in multi-lane environments.
This "splitting point" design, however, suggests that DQJLs do not always have to be a straight line. Instead, they can consist of different lanes that require minimal lane-change maneuvers by the EMV. While this might slightly increase the EMV's passage time, it minimizes disruption to the surrounding traffic, see Fig.~\ref{fig:lane_change_dqjl}.

\subsection{Multi-agent Reinforcement Learning for CAVs}\label{sec:marl_literature}
The proliferation of V2X communication technologies has significantly increased the presence of CAVs, driving extensive research into multi-agent reinforcement learning (MARL) frameworks for vehicular applications \cite{zhou2023survey,henao2023multiclass,liu2023medical,cai2023unsupervised}. In these systems, individual vehicles are modeled as autonomous agents collaborating toward shared objectives. This paradigm offers several advantages, including decentralized decision-making, scalability, adaptability to dynamic traffic conditions, and the capacity for collaborative learning among vehicles.

MARL in CAV scenarios presents a promising approach to addressing complex transportation challenges, particularly where traditional methods have shown limitations. Recent studies have applied MARL to various aspects of traffic management and vehicle coordination, highlighting its versatility and effectiveness. For instance, in the realm of congestion mitigation, Ha et al.~\cite{ha2020leveraging} developed a MARL-based control model using graph convolutional networks (GCN) and deterministic policy gradient (DDPG) techniques for highway scenarios, outperforming traditional speed harmonization methods even with low CAV penetration rates. Similarly, Nakka et al.~\cite{Nakka2022Merging} applied a decentralized multi-agent deterministic policy gradient (MADDPG) approach to highway merging, successfully eliminating stop-and-go traffic.

Intersection management has also been a key focus for MARL applications \cite{you2023rethinking,liu2023benchmarking,yan2024after,han2024hybrid,ma2024segment}. Guo et al.~\cite{guo2022coordination} employed an enhanced monotonic value function factorisation (QMIX) algorithm to coordinate CAVs at unsignalized intersections in mixed-autonomy traffic, while Antonio et al.~\cite{antonio2022multi} introduced an advanced autonomous intersection management (AIM) system using MARL and Curriculum through Self-Play, significantly surpassing traditional traffic light systems in performance.

In vehicle control, Zhou et al.~\cite{zhou2022multi} proposed an MA2C method for lane-changing decisions in mixed traffic environments, utilizing a multi-objective reward function. Shi et al.~\cite{SHI2021Connected} presented a cooperative longitudinal control strategy for CAVs in mixed traffic, improving both string stability and efficiency. Additionally, Chen et al.~\cite{Chen2023OnRamp} addressed mixed-traffic highway on-ramp merging with a MARL framework, enabling autonomous vehicles to collaborate and adapt to human-driven vehicles (HDVs).

Beyond traffic flow management, MARL has demonstrated potential in other CAV applications \cite{liang2024rescuing,cao2024multi,you2024calibrating}. Zhang et al.~\cite{ZHANG2022parking} developed a MARL framework for online parking assignment in mixed CV and non-connected vehicle (NCV) environments, achieving substantial improvements over conventional methods. In the domain of robotic driving simulation, Palanisamy et al.~\cite{Palanisamy2020Driving} introduced MACAD-Gym, a multi-agent simulation platform for CAVs, which supports research into deep reinforcement learning (DRL)-based algorithms in complex, multi-agent environments beyond geo-fenced operational design domains.

Parada et al.~\cite{parada2023safe} have proposed a MAPPO learning framework for autonomous vehicle maneuvering in the presence of EMVs. However, this study assumes a fully autonomous traffic setting and focuses primarily on collision risk avoidance. While effective for collision minimization, this design de-prioritizes the reduction of EMV passage time, which is a critical factor in emergency response. Such an approach may also not be feasible for near-term deployment in mixed autonomy environments.

These diverse applications of MARL in CAV contexts underscore its potential to revolutionize traffic management and vehicle coordination \cite{you2024mine,sun2024medical,huang2024cross}. For those interested in a more comprehensive review of MARL frameworks for CAV applications in mixed traffic environments, Yadav et al.~\cite{yadav2023comprehensive} and Dinneweth et al.~\cite{Dinneweth2022Survey} provide valuable insights into the current state of the art and emerging trends in this rapidly advancing field.

Despite the extensive research on MARL in CAV applications, there is currently no work that employs MARL and focuses on pre-clearing lanes for EMVs in heterogeneous traffic settings. This gap presents an exciting opportunity for this research to explore how MARL can be leveraged to optimize the formation and management of DQJLs, potentially resulting in more efficient and adaptive traffic management strategies in urban environments.

\section{Dynamic Queue Jump Lanes in Mixed Traffic}
In this section, we describe the modeling of DQJL formation as a discrete-time partially observed Markov Decision Process (POMDP) in Subsec.~\ref{subsec:system_modeling}. Additionally, in Subsec.~\ref{subsec:agent_design}, we introduce two distinct agent designs for CAVs and HDVs, enabling the study of system performance under varying levels of connectivity penetration. This approach allows for a comprehensive evaluation of DQJL formation dynamics in mixed traffic environments.
\label{sec:modeling}

\subsection{System Modeling}
\label{subsec:system_modeling}
We formulate the establishment of DQJLs in heterogeneous traffic as a collaborative multi-agent system. Each non-EMV is an autonomous agent, with the agent set $\mathcal{N}$ comprising two subsets:

\begin{equation}
\mathcal{N} = \mathcal{N}_{\text{CAV}} \cup \mathcal{N}_{\text{HDV}}
\end{equation}

DQJLs can be formed either before or when EMVs enter a road segment. To investigate interactions between EMVs and non-EMVs, we assume an EMV begins traversing a congested road segment at time $t_0$, where the EMV itself is not considered an active agent but passively awaits the clearing of a path. In this scenario, CAVs leverage V2X communication capabilities, while HDVs rely solely on local sensory information. All agents collaborate to create a clear path for the EMV.

This mixed autonomy environment is modeled as a Partially Observable Markov Decision Process (POMDP), represented by the tuple $(S, A, T, R, \Omega, O, \gamma)$. In this formulation, $S$ denotes the global state space that represents the overall traffic environment, including the positions, speeds, and states of CAVs, HDVs, and the EMV. Each agent takes actions from the action space $A$, with choices such as lane changes or speed adjustments aimed at optimizing DQJL formation. The transition function $T: S \times A \rightarrow S'$ defines how the state of the environment evolves probabilistically in response to agents' actions.

The reward function $R_i$, specific to each agent $i \in \mathcal{N}$, balances global system-wide objectives with individual goals. Observations are partial, where $\Omega$ represents the observation space for each agent, and the observation function $O$ maps the global state to the local observation available to each agent. Notably, the observation space and observation function differ between CAVs and HDVs, reflecting their distinct capabilities, which will be further discussed in Subsec~\ref{subsec:agent_design}. The process is further governed by a discount factor $\gamma$, which moderates the trade-off between immediate and future rewards.

\subsection{Agent Design}
\label{subsec:agent_design} 
This study proposes two realistic and straightforward designs for CAVs and HDVs, tailored to their respective communication, perception capabilities, and maneuverability. 

\subsubsection{Connected and Automated Vehicles (CAVs)}
The POMDP for the \(i\)-th CAV agent is defined as:

\begin{equation}
\text{POMDP}_{\text{CAV}}^i = \left( S^i, A^i, T^i, R^i, \Omega^i, O^i, \gamma \right)
\end{equation}

\paragraph{State Space}
The state space for the \(i\)-th CAV is given by:

\begin{equation}
S^i = \{x, y, v, d_\text{EMV}, l_\text{EMV}, d_\text{clear}\}
\end{equation}

In this formulation, \(x\) denotes the longitudinal position of the CAV, while \(y\) refers to the lateral position, represented as an integer indicating the lane index. The variable \(v\) describes the velocity of the CAV. The term \(d_\text{EMV}\) represents the longitudinal distance to the approaching EMV, and \(l_\text{EMV}\) denotes the lane difference with respect to the EMV. This lane difference is expressed as an integer, where positive or negative values indicate the relative lane position of the CAV compared to the EMV, allowing for multiple lane differences (e.g., \(-2, -1, 0, 1\)). Lastly, \(d_\text{clear}\) stands for the clear distance ahead, providing essential feedback for the agent to learn lane-changing strategies and evaluate the completeness of DQJL formation. Based on this design, the responsibility for effective DQJL formation predominantly rests with CAVs.

\paragraph{Action Space}
The action space for the \(i\)-th CAV is decomposed into longitudinal and lateral actions:

\begin{equation}
A^i = A_{\text{long}} \times A_{\text{lat}}
\end{equation}

The longitudinal action space \( A_{\text{long}} \) consists of discrete control actions for speed adjustment, allowing a CAV to accelerate, decelerate, or maintain its current speed. Specifically, we discretize the action space as follows:

\[
A_{\text{long}} = \{-2.5 \text{ m/s}^2,\, 0 \text{ m/s}^2,\, +1.5 \text{ m/s}^2\}.
\]

Here, a negative value corresponds to deceleration, zero indicates maintaining the current speed, and a positive value corresponds to acceleration. These choices reflect representative urban driving behaviors, enabling the CAV to respond flexibly to varying traffic conditions, including the distance to the EMV (\(d_{\text{EMV}}\)) and the clear distance ahead (\(d_{\text{clear}}\)).

The lateral action space, \(A_{\text{lat}}\), determines whether the CAV performs a lane change. It includes actions to indicate the direction of lane changes, where \(A_{\text{lat}} = 1\) denotes a lane change to the right, \(A_{\text{lat}} = -1\) indicates a lane change to the left, and \(A_{\text{lat}} = 0\) means no lane change. These lateral actions are based on the CAV’s current lane position \(y\), the lane difference relative to the EMV (\(l_{\text{EMV}}\)), and the state of neighboring vehicles \(\mathbf{N}_{\text{local}}\), allowing the agent to perform lane changes that contribute to efficient DQJL formation.

\paragraph{Transition Function}
The transition function \(T^i\) represents the probabilistic model of how the environment changes in response to the CAV’s actions. Given the current state \(S^i\) and action \(A^i\), \(T^i\) models the transition to the next state. This includes the changes in the CAV's position, speed, and the dynamic states of neighboring vehicles based on both local actions and external factors like traffic flow and EMV behavior. Mathematically, this can be expressed as:
\[
T^i_{\text{CAV}}(S^i_t, A^i_t) \sim P(S^i_{t+1} \mid S^i_t, A^i_t),
\]

\paragraph{Observation Space and Function}
The observation space \(\Omega^i\) consists of information that the \(i\)-th CAV can perceive from the environment. The observation function \(O^i\) maps the global state \(S^i\) to the partial observation \(\Omega^i\) available to the CAV. Mathematically, this can be expressed as:

\[
\Omega^i = O^i(S^i) \quad \text{where} \quad O^i: S^i \rightarrow \Omega^i
\]

The observation space \(\Omega^i\) is a concatenation of the state of the ego vehicle and the states of all vehicles within a certain radius. If a neighboring vehicle is a CAV, full state information is communicated, including position, velocity, and lane index. For HDVs, the CAV can only perceive partial information, limited to the position \((x, y)\) and velocity \(v\). Therefore, \(\Omega^i\) is defined as:

\[
\Omega^i = \{ x, y, v, \mathbf{N}_{\text{CAV}}, \mathbf{N}_{\text{HDV}} \},
\]

where \(\mathbf{N}_{\text{CAV}}\) denotes the full state information of neighboring CAVs, including their position \((x, y)\), velocity \(v\), and \(\mathbf{N}_{\text{HDV}}\) represents the partial state information of neighboring HDVs, limited to their position \((x, y)\) and velocity \(v\).

This design aligns with the advanced communication and perception capabilities of CAVs, positioning them as the proactive agents in traffic control. By leveraging their ability to exchange comprehensive state information with other CAVs and perceive partial data from HDVs, CAVs can make informed, real-time decisions that optimize traffic flow and ensure smoother interactions with CAVs and HDVs. 
\subsubsection{Human-Driven Vehicles (HDVs)}
The POMDP for the \(i\)-th HDV agent is defined as:

\begin{equation}
\text{POMDP}^i = \left( S^i, A^i, T^i, R^i, \Omega^i, O^i, \gamma \right)
\end{equation}

\paragraph{State Space}
The state space for the \(i\)-th HDV is given by:

\begin{equation}
S^i = \{x, y, v, b_{\text{alert}}\}
\end{equation}

In this formulation, \(x\) denotes the longitudinal position of the HDV, while \(y\) refers to the lateral position, represented as an integer indicating the lane index. The variable \(v\) represents the velocity of the HDV. The boolean variable \(b_{\text{alert}}\) indicates whether the HDV has been alerted to the EMV’s presence, either through visual or auditory cues such as sirens or traffic signals.

\paragraph{Action Space}
The action space for the \(i\)-th HDV is simplified to account for basic human driving maneuvers:

\begin{equation}
A^{i} = A_{\text{yield}} \times A_{\text{lat}}
\end{equation}

The longitudinal action space, \(A_{\text{yield}}\), is binary, indicating whether the HDV decides to yield (1) or not (0) in response to an EMV alert. The lateral action space, \(A_{\text{lat}}\), governs lane-changing behavior. Similar to CAVs, \(A_{\text{lat}} = 1\) represents a lane change to the right, \(A_{\text{lat}} = -1\) signifies a lane change to the left, and \(A_{\text{lat}} = 0\) indicates no lane change. Under normal driving conditions, HDVs do not yield. However, when hearing sirens or observing visual cues from an approaching EMV, the HDVs' actions are locked into \(A_{\text{yield}} = 1\), prompting them to find the next available free space to change lanes, pull over, or stop. These simplified actions reflect the limited complexity of human drivers' decision-making when responding to EMVs. It is anticipated that CAVs, being more proactive agents, can learn the yielding patterns of HDVs and adjust accordingly.

\paragraph{Transition Function}
The transition function \(T^i\) for the \(i\)-th HDV models how the environment evolves based on the HDV’s actions. Given the current state \(S^i\) and action \(A^i\), \(T^i\) describes the transition to the next state, accounting for changes in the HDV’s position, speed, and interactions with other vehicles. Since HDVs rely on limited sensory information, external factors such as CAVs' maneuver and EMV behavior have significant influence on these transitions. Mathematically, this can be expressed as:
\[
T^i_{\text{HDV}}(S^i_t, A^i_t) \sim P(S^i_{t+1} \mid S^i_t, A^i_t),
\]

\paragraph{Observation Space and Function}
HDVs operate with a simplified observation space, limited to the ego vehicle's state and nearby vehicle positions and velocities. Lacking communication capabilities, HDVs rely solely on local perceptions from onboard sensors, resulting in restricted and naive decision-making. This limitation underscores the critical role of CAVs in compensating for HDVs' reduced situational awareness and reactivity.

Despite their mostly passive behavior, HDVs are still modeled as agents for several reasons. First, their actions, such as deciding to yield or not, directly influence the efficiency of DQJL formation and the passage time of the EMV. Modeling HDVs as agents allows the framework to incorporate the stochastic and reactive nature of human drivers, which is critical for robust policy learning by CAVs. Second, this design enables the simulation of realistic interactions between CAVs and HDVs, allowing CAVs to anticipate HDV behavior and adapt accordingly. Finally, HDVs contribute to the learning process by generating reward signals (see Subsubsec.~\ref{subsubsec:reward_design}), ensuring that the policies learned by CAVs account for both cooperative and uncooperative HDV behaviors. By treating HDVs as agents, the framework effectively captures the dynamics of mixed traffic environments and ensures scalability under varying levels of CAV penetration.

\subsubsection{Reward Design}\label{subsubsec:reward_design}

The reward structure is designed to encourage both global cooperation and individual efficiency. For each agent \(i\), the reward function combines a shared global reward and an agent-specific local reward:

\begin{equation}\label{eqn:global_reward}
R_{\text{global}} = -\Biggl(
\underbrace{\beta_1 \cdot t_{\text{EMV}}}_{\textcolor{BrickRed}{\text{Time Penalty}}}
+ \underbrace{\beta_2 \cdot n_{\text{EMV}}^{\text{lane change}}}_{\textcolor{RoyalBlue}{\text{Lane Change Penalty}}} 
+ \underbrace{\beta_3 \cdot \left(1 - \frac{d_{\text{clear}}}{L}\right)}_{\textcolor{ForestGreen}{\text{DQJL Completeness Penalty}}}
\Biggr)
\end{equation}

\begin{equation}\label{eqn:return}
R^i = \alpha R_{\text{global}} - (1 - \alpha) \cdot R_{\text{local}}^i
\end{equation}

The global reward \(R_{\text{global}}\) is shared among all agents and reflects system-wide objectives: minimizing EMV passage time, reducing EMV lane changes, and ensuring complete DQJL formation. This shared reward promotes coordination between agents toward common goals.

The local reward \(R_{\text{local}}^i\) varies by agent type:

1. **For CAVs**:
   \[
R_{\text{local}}^{\text{CAV}} = 
\underbrace{-\eta_1 \cdot n_{\text{CAV}}^{\text{lane change}}}_{\textcolor{RoyalBlue}{\text{Local: CAV Lane Change Penalty}}}
\;+\;
\underbrace{-\eta_2 \cdot |a_{\text{long}}|}_{\textcolor{Sepia}{\text{Local: CAV Acceleration Penalty}}}
\]
   The local reward for CAVs penalizes excessive lane changes and sudden acceleration or deceleration, encouraging smoother driving behavior.

2. **For HDVs**:
\[
R_{\text{local}}^{\text{HDV}} = 
\underbrace{- \eta_3 \cdot n_{\text{HDV}}^{\text{lane change}}}_{\textcolor{RoyalBlue}{\text{Local: HDV Lane Change Penalty}}}
\]
   For HDVs (passive agents), the local reward simply tracks lane-changing penalties.

The parameter \(\alpha \in [0,1]\) balances the trade-off between global coordination and local efficiency. A higher \(\alpha\) emphasizes system-wide performance, while a lower \(\alpha\) prioritizes individual agent behavior.

\section{Multi-Agent Learning for DQJLs}
\label{sec:learning}
To address the challenge of coordinating multiple agents in forming DQJLs, we employ the Multi-Agent Proximal Policy Optimization (MAPPO) algorithm. This section details MAPPO's adaptation for the DQJL task, starting with preliminaries in Subsec.~\ref{subsec:preliminaries}, an explanation of MAPPO's alignment with DQJL in Subsec.~\ref{subsec:MAPPO_DQJL}, the MAPPO-DQJL algorithm in Algorithm~\ref{algo:mappo_dqjl}, and network architectures in Subsec.~\ref{subsec:network_architecture}.

\subsection{Preliminaries}
\label{subsec:preliminaries}
MAPPO is an on-policy reinforcement learning algorithm designed to optimize individual agent policies while accounting for the collective behavior of all agents. Schulman et al.~\cite{schulman2017proximal} introduced Proximal Policy Optimization (PPO), which has since become a widely adopted algorithm for model-free reinforcement learning due to its balance of stability and performance. Yu et al.~\cite{yu2022surprising} highlight the exceptional performance of multi-agent reinforcement learning in collaborative settings, further emphasizing its potential for diverse applications.

In the context of DQJL formation, this approach enables agents to quickly form DQJLs by learning cooperative strategies. MAPPO operates under centralized training and decentralized execution, making it ideal for this scenario. During centralized training, agents share information and are trained with a global perspective of the environment. This allows the system to leverage full-state information, ensuring that the policies learned reflect the global objective of minimizing EMV passage time while maintaining traffic flow. 

In contrast, decentralized execution allows agents to make decisions based on local observations. This ensures scalability and real-world applicability, as agents act independently, responding to their immediate environment in real-time. This autonomy is crucial in mixed autonomy settings where communication might be limited, allowing agents to function effectively despite partial observability. The decentralized execution phase ensures robust behavior in dynamic traffic conditions, while the centralized training phase guarantees that these local actions contribute to the broader goal of DQJL formation.

A key feature of MAPPO is its centralized value function estimation, where a centralized critic evaluates each agent’s actions based on the global state of the system. This helps agents understand how their actions contribute to the collective objective. Policy updates using the PPO objective ensure a balance between exploration and exploitation, leading to stable and efficient learning. The reward structure encourages both individual and collective coordination, optimizing DQJL formation and minimizing disruptions to surrounding traffic, thus ensuring the safe and swift passage of EMVs.

\subsection{MAPPO for DQJL Formation}
\label{subsec:MAPPO_DQJL}
In our approach to DQJL formation, we adapt MAPPO with parameter sharing for CAVs, designing policy and value networks specifically tailored to the dynamic, interactive nature of traffic systems. All CAV agents utilize a common policy network \(\pi_{\theta_{\text{CAV}}}(a_i \mid o_i)\), where \(\theta_{\text{CAV}}\) denotes the shared policy parameters, ensuring that these agents learn a unified, generalizable strategy based on their local observations. HDVs are modeled as passive agents with fixed policy networks \(\pi_{\theta_i}(a_i \mid o_i)\), enabling a unified agent-based representation of the traffic system while maintaining realistic human driving patterns. This design choice not only simplifies state representation but also provides the flexibility to incorporate different HDV behavioral models within the same framework. Our experimental results demonstrate that treating HDVs as agents rather than environmental elements leads to superior performance in mixed-autonomy scenarios.

To foster coordination across all vehicles, we employ a centralized value network \(V_{\phi}(s)\) that evaluates the global state \(s\), which includes information from both active (CAVs) and passive (HDVs) agents. This unified value function, parameterized by \(\phi\), enables consistent evaluation of vehicle interactions regardless of autonomy level, providing a comprehensive assessment of the traffic situation. The centralized evaluation mechanism guides CAVs toward cooperative behaviors while realistically accounting for HDV responses, ultimately supporting efficient DQJL formation.

The reward function \(R_i\) for each CAV reconciles individual-level goals—such as safety and maneuver efficiency—with the broader aim of expediting EMV passage. By balancing local incentives with global objectives, this hierarchical reward structure motivates agents to coordinate their actions in a manner that promotes improved overall traffic flow and reduced disruptions.

To ensure stable learning and robust policy improvement, we employ the clipped PPO objective. This training paradigm maintains a careful equilibrium between exploration and exploitation, controlling volatility in performance as the CAVs refine their shared policy. By combining parameter-sharing among CAVs, a unified agent-based representation for both CAVs and HDVs, and a carefully crafted reward structure, our MAPPO-based framework facilitates the rapid, secure, and cooperative implementation of DQJLs.

\subsubsection{MAPPO-DQJL Algorithm}
We present the pseudocode for our MAPPO-DQJL algorithm in Algorithm~\ref{algo:mappo_dqjl}, followed by a detailed walk-through of its key components.
\begin{algorithm}[htbp]
\caption{MAPPO-DQJL Training with Parameter Sharing}
\label{algo:mappo_dqjl}
\SetKwInOut{Input}{Input}
\SetKwInOut{Output}{Output}

\Input{
$T$: maximum time step of an episode \\
$N_{\text{bs}}$: batch size \\
$\eta_{\theta}$: learning rate for policy networks \\
$\eta_{\phi}$: learning rate for value networks
}

\Output{
$\phi$: learned parameters for the shared value network \\
$\theta_{\text{CAV}}$: learned parameters for the shared CAV policy network
}

\BlankLine
Initialize $\phi, \theta_{\text{CAV}}, k \gets 0, B \gets \varnothing$ \;
Initialize simulation environment (SUMO), $t \gets 0$, get $\{s_t^i\}_{i \in \mathcal{N}}$ \;

\Repeat{\textnormal{Convergence}}{
    \tcc{Generate trajectories}
    \ForEach{CAV $i \in \mathcal{N}_{\text{CAV}}$ \textnormal{(in parallel)}}{
        Sample $a_t^i$ from $\pi_{\theta_{\text{CAV}}}$ \;
        Execute fixed policy for HDVs $j \in \mathcal{N}_{\text{HDV}}$ \;
        Receive $r_t^i$ and $s_{t+1}^i$ \;
        $B \gets B \cup \{(s_t^i, a_t^i, s_{t+1}^i, r_t^i)\}$ \;
    }
    
    $t \gets t + 1, k \gets k + 1$ \;

    \If{$t == T$}{
        Reset simulation, $t \gets 0$, get $\{s_0^i\}_{i \in \mathcal{N}}$ \;
    }

    \If{$k == N_{\text{bs}}$}{
        \tcc{Update using shared parameters for CAVs}
        Compute advantages $\{\hat{A}_t^i\}_{i \in \mathcal{N}_{\text{CAV}}}$ using GAE \;
        Compute returns $\{\hat{R}_t^i\}_{i \in \mathcal{N}_{\text{CAV}}}$ \;
        
        Aggregate gradients over CAVs in $\mathcal{N}_{\text{CAV}}$ \;
        $\phi \gets \phi - \eta_\phi \nabla_{\phi} L_V(\phi)$ \;
        $\theta_{\text{CAV}} \gets \theta_{\text{CAV}} + \eta_\theta \nabla_{\theta_{\text{CAV}}} L^{\text{CLIP}}(\theta_{\text{CAV}})$ \;
        
        $k \gets 0$, $B \gets \varnothing$ \;
    }
}
\end{algorithm}

\paragraph{Centralized Training with Decentralized Execution}
The algorithm adopts the centralized training and decentralized execution (CTDE) paradigm, which is well-suited for the DQJL task. During decentralized execution, each CAV $i$ observes its local state $o_{i,t}$, including neighboring vehicles, and independently selects actions $a_{i,t}$ based on its shared policy $\pi_{\theta_{\text{CAV}}}$. This decentralized approach enables real-time autonomy, scalability, and robust responses to dynamic traffic conditions without requiring communication between agents during deployment.

During centralized training, given our reward design that incorporates both shared global objectives and agent-specific local rewards, a critic evaluates the global state $s_t$ through a value function:

\begin{equation}
V_\phi(s_t) = \mathbb{E}_{\pi} \left[ \sum_{k=0}^\infty \gamma^k \left(R_{\text{global}}(s_{t+k}, \mathbf{a}_{t+k}) + \sum_{i \in \mathcal{N}_{\text{non-EMV}}} R^i_{\text{local}}(s^i_{t+k}, a^i_{t+k})\right) \mid s_t \right]
\end{equation}

where $R_{\text{global}}$ represents the shared system-wide reward focusing on EMV-related objectives, and $R^i_{\text{local}}$ captures the local efficiency of each CAV $i$.

This centralized value function provides a comprehensive assessment that:
1. Tracks global DQJL formation progress through $R_{\text{global}}$
2. Aggregates individual CAV contributions through $\sum_{i \in \mathcal{N}_{\text{CAV}}} R^i_{\text{local}}$
3. Accounts for HDV interactions implicitly through state transitions

The value network $V_{\phi}$ is updated by minimizing the squared error between this estimated value and the observed returns:
\begin{equation}
    L_V(\phi) = \mathbb{E}_{\mathcal{B}} \left[ \left( V_{\phi}(s_t) - \hat{R}_t \right)^2 \right]
\end{equation}

where $\hat{R}_t$ represents the actual discounted sum of rewards observed during rollouts. This CTDE framework enables agents to learn coordinated behaviors that optimize both system-wide efficiency and individual performance while maintaining scalability in real-world deployments.

\paragraph{Generalized Advantage Estimation (GAE)}
In the MAPPO framework, the update of policy networks relies on accurate estimation of the advantage function $\hat{A}_t$, which measures how much better or worse an action $a_{i,t}$ is relative to the current policy. In our context, this helps the CAVs learn how their actions (e.g., lane changes, acceleration) contribute to efficient DQJL formation. To compute the advantage, we use Generalized Advantage Estimation (GAE), which reduces variance in the policy gradient estimates while maintaining manageable bias.

The advantage function at time $t$ is given by:
\begin{equation}\label{eqn:gae}
\hat{A}_t = \sum_{k=0}^{\infty} (\gamma \lambda)^k \delta_{t+k}
\end{equation}

where $\delta_t = r_t + \gamma V_{\phi}(s_{t+1}) - V_{\phi}(s_t)$ represents the temporal difference (TD) error, and $r_t$ is the reward received at time $t$. The parameters $\gamma$ and $\lambda$ control the trade-off between emphasizing immediate and future rewards, which in this context ensures that agents focus not only on short-term lane clearing but also long-term efficiency in EMV passage.

\paragraph{Policy Network Update and Parameter Sharing for CAVs}
Our framework employs PPO with a clipped objective to refine the policy parameters shared among all CAV agents, while HDVs maintain fixed policies that reflect realistic driving behaviors. By consolidating the learning process into a single, unified policy $\pi_{\theta_{\text{CAV}}}$ for all CAVs, we enable these vehicles to collectively learn and improve their decision-making behaviors from one another's experiences. This parameter sharing is particularly advantageous in DQJL formation, where CAVs must jointly coordinate their maneuvers while accounting for HDV responses to ensure effective lane-clearing and prompt EMV passage.

The PPO objective with the CLIP function is given by:
\begin{equation}\label{eqn:ppo_objectives}
    L^{\text{CLIP}}(\theta) = \mathbb{E}_{\mathcal{B}} \Bigl[ \min \bigl( r_t(\theta) \hat{A}_t,\; \text{clip}\bigl( r_t(\theta), 1 - \epsilon, 1 + \epsilon \bigr) \hat{A}_t \bigr) \Bigr],
\end{equation}
where
\[
r_t(\theta) = \frac{\pi_{\theta}(a_t \mid o_t)}{\pi_{\theta_{\text{old}}}(a_t \mid o_t)}
\]
represents the probability ratio between the updated and previous shared policies. By constraining \(r_t(\theta)\) to \([1-\epsilon, 1+\epsilon]\), the policy avoids large, destabilizing jumps. This controlled adaptation is critical in a DQJL context, where sudden, drastic policy shifts might lead to erratic maneuvers and reduced cooperation among CAVs.

The policy update is performed via gradient ascent:
\[
\theta \gets \theta + \alpha_{\pi} \nabla_{\theta} L^{\text{CLIP}}(\theta),
\]
with \(\alpha_{\pi}\) as the learning rate. Over successive updates, all CAVs collectively converge toward more reliable and efficient DQJL formation strategies, as knowledge gained from one scenario disseminates through the shared parameters to benefit the entire CAV fleet.

% \paragraph{Value Network Update}
% The value network $V_{\phi}$ is updated by minimizing the squared error between the estimated value $V_{\phi}(s_t)$ and the observed return $\hat{R}_t$. This centralized value function evaluates the global state incorporating both active CAVs and passive HDVs, where HDVs execute predetermined fixed policies. This unified approach enables comprehensive assessment of the traffic state, accounting for both learnable CAV actions and predictable HDV behaviors in the context of DQJL formation.

% The value network objective is defined as:
% \begin{equation}\label{eqn:clip}
%     L_V(\phi) = \mathbb{E}_{\mathcal{B}} \left[ \left( V_{\phi}(s_t) - \hat{R}_t \right)^2 \right],
% \end{equation}
% where $\hat{R}_t = \sum_{k=0}^{T} \gamma^k r_{t+k}$ denotes the discounted cumulative reward. The shared value network leverages experiences from CAVs while accounting for fixed HDV behaviors, guiding the optimization of CAV policies toward efficient DQJL formation.
\subsection{Transformer-based Network Architecture}
\label{subsec:network_architecture}
This study implements a transformer-based architecture for both policy and value networks in our MAPPO-DQJL algorithm. The transformer architecture, introduced by Vaswani et al.~\cite{vaswani2017attention}, is selected for its proven capability in modeling complex dependencies through self-attention mechanisms. In the context of DQJL formation, where vehicles must coordinate across varying distances and time horizons, the transformer's ability to capture long-range interactions and process information in parallel makes it particularly suitable for real-time decision-making in dense traffic conditions~\cite{parisotto2020stabilizing,chen2021decision}.

\subsubsection{Policy Network}
Under our parameter-sharing framework, all CAV agents utilize a single shared policy network \(\pi_{\theta}\). The network processes each CAV's local observation \(o_i\) through the following sequence:
\begin{enumerate}
    \item Embedding layer transforms observations into a higher-dimensional feature space
    \item Positional encodings incorporate spatial and temporal information
    \item \(N\) transformer encoder layers process the embedded input using self-attention
    \item Policy head (fully connected layer) computes action distribution parameters
\end{enumerate}

Formally, for any CAV agent \(i\):
\[
\pi_{\theta}(a_i \mid o_i) = \text{PolicyHead}\Bigl( \text{TransformerEncoder}\bigl( \text{PositionalEncoding}( \text{Embedding}(o_i) ) \bigr) \Bigr),
\]
where \(\pi_{\theta}(a_i \mid o_i)\) represents the probability distribution over actions \(a_i\), conditioned on \(o_i\). The parameter sharing enables efficient knowledge transfer across the CAV fleet. See Subsections~\ref{subsec:appendix_policy_network} for implementation details.

\subsubsection{Value Network}
The centralized value network \(V_{\phi}\) evaluates the global state \(s\) through a similar transformer-based architecture:
\[
V_{\phi}(s) = \text{ValueHead}\bigl( \text{TransformerEncoder}( \text{Embedding}(s) ) \bigr).
\]
This shared value network provides a unified assessment of how joint actions impact the global traffic state, guiding CAVs toward optimal DQJL formation strategies. See Subsections~\ref{subsec:appendix_value_network} for implementation details.

\subsubsection{Transformer Encoder Layer}
Each transformer encoder layer maintains the standard architecture: multi-head attention followed by a position-wise feed-forward network, with layer normalization and residual connections after each sub-layer. Detailed implementation specifications and parameter estimations are provided in Subsections~\ref{subsec:appendix_tf_encoder_layer} and~\ref{subsec:number_of_parameters}, respectively.

\section{Experimental Setup}
\label{sec:experimental_setup}
In this section, we delineate the methodology used to evaluate the performance of MAPPO-DQJL. Subsection~\ref{subsec:objectives} specifies the central research objectives that guide our experimental investigations. Subsection~\ref{subsec:metrics} details the evaluation metrics employed to gauge the framework’s effectiveness. 
Subsection~\ref{subsec:simulation} describes the simulation environment utilized in these experiments. Finally, Subsection~\ref{subsec:benchmarks} introduces the benchmarks that serve as comparative baselines for assessing MAPPO-DQJL. 

\subsection{Experimental Objectives}
\label{subsec:objectives}

The experiments are designed to address the following key research questions:

\begin{itemize}
    \item \textbf{Effectiveness of MAPPO-DQJL:} To what extent does our MARL-based approach for DQJL formation outperform established benchmarks with respect to EMV passage time and holistic traffic flow optimization?
    \item \textbf{Impact of Penetration Rates:} How does varying CAV penetration rates influence the performance of the proposed framework, particularly regarding DQJL coordination and operational efficiency?
    \item \textbf{Impact of Traffic Density:} Similarly, how does varying traffic density influence the performance of the proposed framework?
\end{itemize}

These focal points collectively shape the principal objectives of our study. By rigorously examining these questions, we aim to elucidate the complex interplay between DQJL formation strategies, CAV adoption levels, and performance trade-offs, ultimately informing near-term policy decisions and strategic deployment of intelligent traffic management solutions.

\subsection{Evaluation Metrics}
\label{subsec:metrics}

Although a DQJL is typically formed prior to EMV ingress into a given road segment, the underlying agent behaviors remain consistent throughout the EMV’s traversal. Consequently, the temporal duration required to establish the DQJL is effectively captured by the EMV’s passage time. We employ the following evaluation metrics:

\begin{itemize}
    \item \textbf{EMV Passage Time ($t_{\text{EMV}}$):} This metric represents the duration required for the EMV to traverse the designated road segment once the DQJL is in place, emphasizing the primary objective of expediting emergency response.

    \item \textbf{Number of Lane-changing for EMVs ($N_{\text{EMV}}^{\text{LC}}$) and Non-EMVs ($N_{\text{non-EMV}}^{\text{LC}}$):} We also measure the lane-changing efforts of both EMVs and non-EMVs, including CAVs and HDVs. Slight increments in EMV passage time may substantially diminish the maneuvering burden imposed on non-EMVs. Our analysis focuses on carefully balancing these trade-offs to minimize overall disruption while maintaining efficient EMV transit.
\end{itemize}
\subsection{Simulation Environment}
\label{subsec:simulation}
We employ Simulation of Urban Mobility (SUMO) as our simulation environment. SUMO's ability to realistically model fine-grained interactions in mixed traffic scenarios, coupled with its native support for V2X communication and precise vehicle dynamics, provides a high-fidelity platform for examining DQJL formation and its implications on traffic flow. The simulation is conducted on a representative roadway environment, reflecting a typical urban street layout.

To ensure simulation reliability, we conducted a comprehensive calibration and validation process using empirical traffic flow data from the NYC Mobility Report~\cite{nycdot2019mobility}. The calibration focused on key parameters including car-following behavior, lane-changing dynamics, and intersection delays, leveraging Manhattan's peak hour metrics such as average vehicle speeds and traffic volumes. Our model validation against these real-world observations achieved a goodness-of-fit of over 85\% for key metrics including average travel times, queue lengths, and throughput. These calibrated parameters were then applied consistently across all experimental scenarios to maintain reliability in our comparative analyses. For vehicle behavior modeling, we utilized SUMO's built-in car-following and lane-changing models (detailed in Subsec.~\ref{subsec:appendix_car_following} and Subsec.~\ref{subsec:appendix_lane_changing}), adapting them to implement fixed policies for HDVs. Rather than treating HDVs as purely rule-based environmental elements, we represented them as passive agents, ensuring consistent representation of all vehicle types within our MAPPO framework while preserving SUMO's realistic vehicle dynamics.

To ensure a focused investigation of the interplay between CAVs and HDVs, we assume homogeneous configurations for each class of vehicle. Specifically, all CAVs share a uniform set of learning parameters, while all HDVs share a common fixed policy derived from realistic driving behaviors.

\paragraph{Road Configurations} We adopt three distinct road configurations in SUMO: Two-Lane, Three-Lane, and Four-Lane roadways, each spanning a length of 1500 meters. Traffic flow is varied across different simulation settings, ensuring a realistic depiction of dynamic traffic conditions. The EMV on duty enters the roadway already populated with vehicles on any lane randomly, simulating real-world congestion scenarios. Vehicles continuously enter and leave the roadway as the EMV traverses, and the observation window concludes once the EMV has exited the roadway.

\subsection{Benchmarks}
\label{subsec:benchmarks}
Although DQJL is a novel concept specifically designed to facilitate EMV passage through congested urban roads, there are other frameworks aimed at improving EMV intra-link movements in similar traffic conditions. Two control frameworks that address tasks closely related to DQJL in heterogeneous traffic environments are as follows:

\begin{enumerate}
    \item \textbf{No Control}: This scenario represents a baseline condition where no specialized control strategy is employed. Non-EMVs adhere strictly to their default car-following and lane-changing behaviors in SUMO, reacting only to immediate traffic conditions without any global coordination. Although vehicles may occasionally yield when the EMV is in close proximity, these responses are purely local and uncoordinated, often resulting in suboptimal clearance times for the EMV.

    \item \textbf{PPO-based Corridor Clearance (PPO-CC)}: Suo et al.~\cite{suo2024model} introduced a single-agent PPO-based control framework where a CAV acts as a "splitting point" agent to assist EMVs in clearing corridors. The CAV dynamically adjusts its speed and positioning based on local traffic conditions, creating a corridor for the EMV to pass through with minimal disruption to other vehicles. 

    \item \textbf{MAPPO-DQJL-CAV}: A reduced version of our proposed framework that considers only CAVs as agents, treating HDVs as part of the environment. The agent design for CAVs remains consistent with MAPPO-DQJL. This benchmark evaluates whether the inclusion of HDVs as agents significantly impacts performance.
\end{enumerate}

\section{Results and Discussion}
\label{sec:discussion}
In this section, we present the results from the simulated experiments. The effects of penetration rate, traffic density, and the adoption of contra-flow lanes on MAPPO-DQJL performance are detailed in Subsec.~\ref{subsec:sensitivity}. Additionally, the contributions of different components within MAPPO-DQJL are analyzed in Subsec.~\ref{subsec:ablations}. Environment configurations, MDP parameters, and training hyper-parameters selected to produce the following results are listed in Table~\ref{tab:parameters_summary}.

\subsection{Sensitivity Studies}\label{subsec:sensitivity}
Many factors influence the effectiveness of candidate lane-clearing schemes. Among these, two primary determinants stand out: the penetration rate of CAVs and the prevailing traffic density. Variations in CAV penetration rates influence the degree of cooperative behaviors and the system’s capacity to execute sophisticated maneuvers, while shifts in traffic density alter the underlying operational conditions, potentially affecting both safety and efficiency outcomes. 

To ensure a fair comparison of performance among the candidate schemes, all other parameters remain consistent across different experimental conditions. Traffic flow is fixed as \textbf{15} veh/lane/min for penetration rate study and penetration rate is set as \textbf{50\%} for traffic density study. All statistics reported in the tables are based on a total of 20 runs per scenario.

\subsubsection{Penetration Rate}
\paragraph{Two-Lane Roadway} When the traffic density is set to 15 veh/lane/min, the EMV passage time, the number of EMV lane-change maneuvers, and the number of non-EMVs’ lane-change maneuvers are presented Table~\ref{tab:penetration_study_two_lane}.

\begin{table}[htbp]
\centering
{\fontsize{9}{11}\selectfont
\begin{tabular}{lccccc}
\toprule
\textbf{Method} & \textbf{0\%} & \textbf{25\%} & \textbf{50\%} & \textbf{75\%} & \textbf{100\%} \\
\midrule
\textbf{\textit{$t_{\text{EMV}} (s)$}} & & & & & \\ 
\multirow{1}{*}{No Control} & \multicolumn{5}{c}{191.2 $\pm$ 9.7} \\[6pt]
PPO-CC~\cite{suo2024model} 
& 190.3 $\pm$ 9.7 
& 183.5 $\pm$ 10.3
& 171.5 $\pm$ 10.5 
& 152.2 $\pm$ 8.9 
& 145.6 $\pm$ 7.3 \\[6pt]
MAPPO-DQJL-CAV 
& 189.8 $\pm$ 9.9 
& 175.0 $\pm$ 10.0
& 162.3 $\pm$ 10.1 
& 145.7 $\pm$ 7.9 
& 143.0 $\pm$ 6.5 \\[6pt]
MAPPO-DQJL 
& \textbf{189.6} $\pm$ 9.5 
& \textbf{170.8} $\pm$ 10.8
& \textbf{150.9} $\pm$ 11.5 
& \textbf{138.4} $\pm$ 8.8 
& \textbf{133.8} $\pm$ 7.6 \\
\midrule
\textbf{\textit{\boldmath$N_{\text{EMV}}^{\text{LC}}$}} & & & & & \\ 
\multirow{1}{*}{No Control} & \multicolumn{5}{c}{4.0 $\pm$ 0.3} \\[6pt]
PPO-CC~\cite{suo2024model} 
& \textbf{1.0} $\pm$ 0.0 & \textbf{1.0} $\pm$ 0.0 & 1.0 $\pm$ 0.0 & 1.0 $\pm$ 0.0 & 1.0 $\pm$ 0.0 \\[6pt]
MAPPO-DQJL-CAV 
& 3.9 $\pm$ 0.4 & 2.8 $\pm$ 0.4 & 1.6 $\pm$ 0.3 & 1.3 $\pm$ 0.2 & 1.1 $\pm$ 0.2 \\[6pt]
MAPPO-DQJL 
& 4.0 $\pm$ 0.3 & 1.8 $\pm$ 0.3 & \textbf{0.9} $\pm$ 0.2 & \textbf{0.7} $\pm$ 0.1 & \textbf{0.6} $\pm$ 0.1 \\
\midrule
\textbf{\textit{\boldmath$N_{\text{non-EMV}}^{\text{LC}}$}} & & & & & \\ 
\multirow{1}{*}{No Control} & \multicolumn{5}{c}{20.3 $\pm$ 2.7} \\[6pt]
PPO-CC~\cite{suo2024model} 
& \textbf{19.3} $\pm$ 2.0 & 17.0 $\pm$ 1.8 & 14.7 $\pm$ 1.5 & 12.1 $\pm$ 1.2 & 9.5 $\pm$ 1.0 \\[6pt]
MAPPO-DQJL-CAV 
& 19.5 $\pm$ 2.0 & 18.0 $\pm$ 1.8 & 16.0 $\pm$ 1.6 & 13.0 $\pm$ 1.3 & 11.0 $\pm$ 1.1 \\[6pt]
MAPPO-DQJL 
& 20.5 $\pm$ 1.8 & \textbf{16.9} $\pm$ 1.7 & \textbf{14.0} $\pm$ 1.4 & \textbf{10.7} $\pm$ 1.2 & \textbf{8.3} $\pm$ 0.9 \\
\bottomrule
\end{tabular}
}
\caption{Performance for all methods on evaluation metrics in the Two-Lane Roadway across different penetration rates. Bold values indicate the best performance.}
\label{tab:penetration_study_two_lane}
\end{table}
The results presented in Table~\ref{tab:penetration_study_two_lane} demonstrate that the proposed MAPPO-DQJL framework outperforms all other methods across all non-zero penetration rates with respective to all metrics. Although MAPPO-DQJL-CAV also leverages multi-agent coordination, its reliance on treating HDVs as part of the environment rather than as learning agents leads to noticeably longer EMV passage times compared to MAPPO-DQJL. This suggests that fully integrating HDVs into the learning process is crucial for achieving more efficient cooperation and faster EMV clearance.

As the CAV penetration rate increases, the passage time for EMVs under MAPPO-DQJL ultimately reduces by approximately 30\% relative to the No Control scenario, underscoring the substantial benefits gained from widespread CAV adoption. When comparing MAPPO-DQJL to PPO-CC, the advantage is less pronounced—on the order of 8\% at intermediate penetration levels—yet still meaningful. These findings highlight the importance of fully integrated, multi-agent cooperation among CAVs and HDVs to attain near-optimal traffic flow conditions and expedite EMV passage.

A notable observation is that the PPO-CC approach consistently reports a mean of $1.0 \pm 0.0$ lane changes. This behavior arises from its design: PPO-CC designates a CAV as a stationary “splitting point,” ensuring that the EMV must execute exactly one lane-changing maneuver to travel through the cleared corridor. In contrast, as the penetration rate reaches the 50\% mark, MAPPO-DQJL achieves an average lane-change count below 1.0, indicating that for than half times, the EMV can traverse through the corridor without changing lane. This improvement becomes even more pronounced at higher penetration rates, where MAPPO-DQJL consistently outperforms all other methods, demonstrating its efficiency in coordinating maneuvers and reducing unnecessary maneuvers.

On the other hand, the MAPPO-DQJL-CAV variant, which treats HDVs as part of the environment rather than as active learning agents, cannot achieve this level of efficiency. Consequently, the EMV under MAPPO-DQJL-CAV often performs more than one lane-change on average, reflecting a non-ideal scenario where less integrated coordination with HDVs leads to suboptimal maneuvering behaviors.

The results also indicate that the proposed MAPPO-DQJL method outperforms the other approaches across all non-zero CAV penetration rates. In particular, MAPPO-DQJL-CAV, which does not treat HDVs as learning agents, suffers from a lack of coordination between CAVs and HDVs. This oversight allows PPO-CC, despite its simpler structure, to achieve fewer lane-change maneuvers in certain intermediate scenarios than MAPPO-DQJL-CAV. Nonetheless, at 100\% CAV penetration, MAPPO-DQJL not only reduces non-EMV lane-change maneuvers by roughly 60\% compared to the No Control baseline, but also maintains about a 10\% margin of improvement over PPO-CC. These findings underscore the value of fully integrated cooperation among CAVs and HDVs, enabling more substantial efficiency gains at higher penetration levels.

\begin{figure}[hbtp]
    \centering
    \begin{subfigure}[b]{0.48\linewidth}
        \centering
        \includegraphics[width=\linewidth]{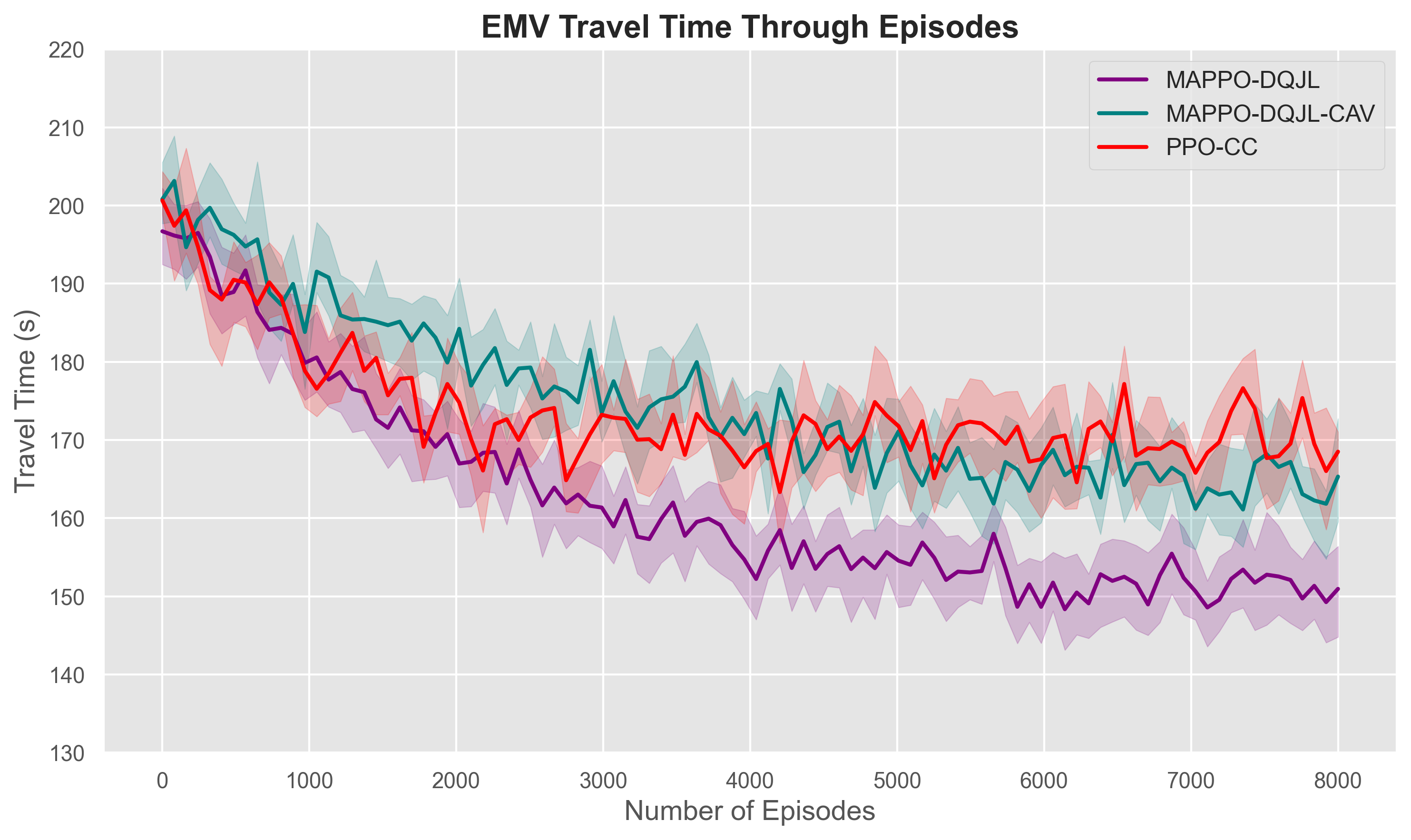}
        \caption{EMV travel time.}
        \label{fig:emv_travel_time}
    \end{subfigure}
    \hfill
    \begin{subfigure}[b]{0.48\linewidth}
        \centering
        \includegraphics[width=\linewidth]{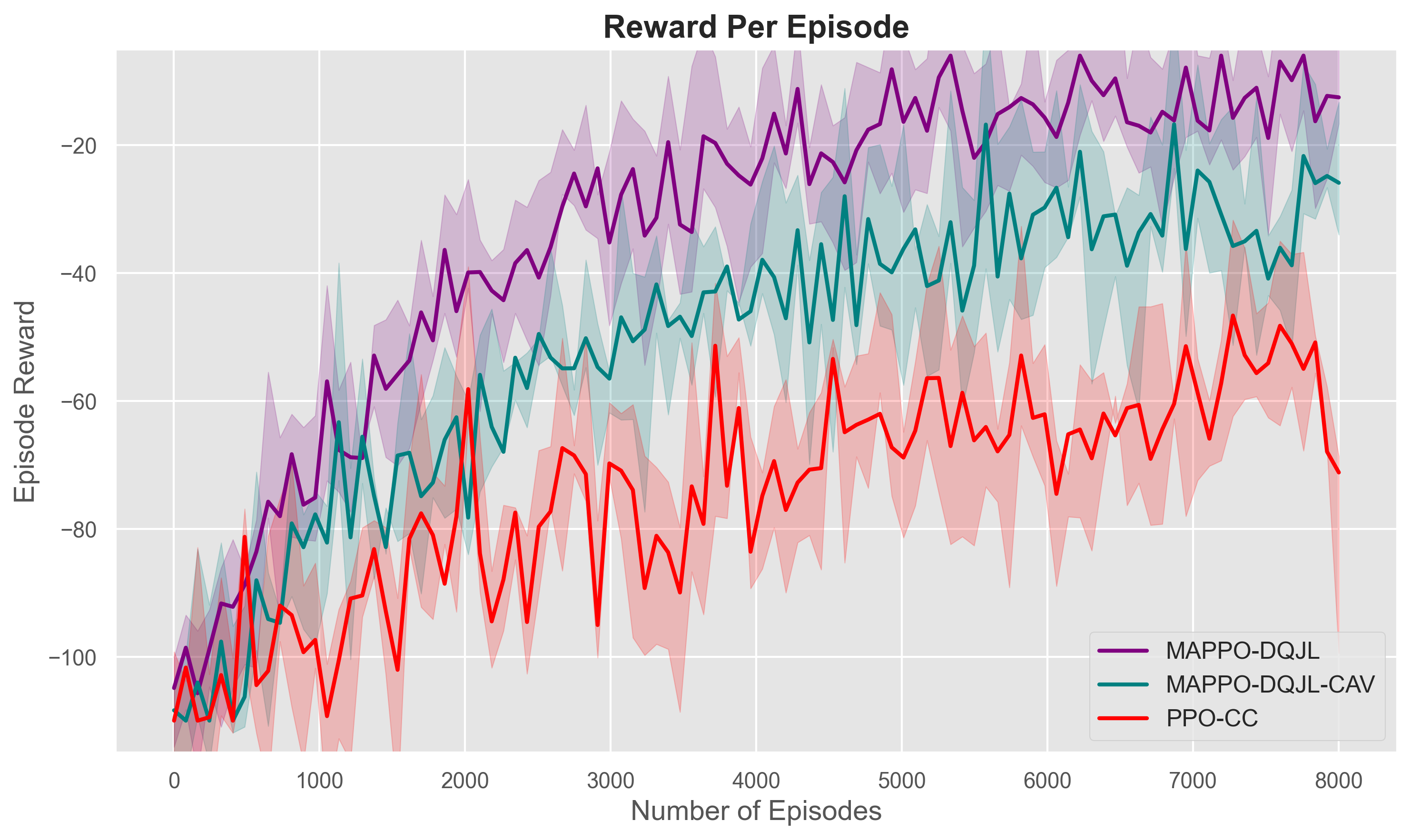}
        \caption{Reward.}
        \label{fig:reward_per_episode}
    \end{subfigure}
    \caption{Comparison of EMV travel time (a) and reward (b) per episode achieved by learning-based methods.}
    \label{fig:training}
\end{figure}

Fig.~\ref{fig:training} illustrates the training patterns of MAPPO-DQJL, MAPPO-DQJL-CAV, and PPO-CC across episodes for the Two-Lane Roadway with 15 veh/lane/min under 50\% penetration rate. As shown in Fig.~\ref{fig:emv_travel_time}, PPO-CC converges around 3000 episodes, but $t_{\text{EMV}}$ fluctuates around 170s at convergence. MAPPO-DQJL converges near 7500 episodes, achieving the lowest $t_{\text{EMV}}$ overall. MAPPO-DQJL-CAV demonstrates a similar learning pattern but stabilizes at a higher $t_{\text{EMV}}$ of around 162s. Fig.~\ref{fig:reward_per_episode} reflects this trend, also highlighting greater fluctuations in MAPPO-DQJL-CAV compared to MAPPO-DQJL.

\paragraph{Three-Lane Roadway} 
Table~\ref{tab:penetration_rate_three_lane} shows the performance achieved by all methods on the Three-Lane Roadway. The outcomes mirror the trends observed in the two-lane scenario, with MAPPO-DQJL consistently yielding superior performance as CAV penetration increases.
\begin{table}[htbp]
\centering
{\fontsize{9}{11}\selectfont
\begin{tabular}{lccccc}
\toprule
\textbf{Method} & \textbf{0\%} & \textbf{25\%} & \textbf{50\%} & \textbf{75\%} & \textbf{100\%} \\
\midrule
\textbf{\textit{$t_{\text{EMV}} (s)$}} & & & & & \\ 
\multirow{1}{*}{No Control} & \multicolumn{5}{c}{180.0 $\pm$ 9.2} \\[6pt]
PPO-CC~\cite{suo2024model} 
& 179.5 $\pm$ 9.5 & 165.3 $\pm$ 10.1 & 150.4 $\pm$ 10.3 & 138.7 $\pm$ 9.1 & 130.1 $\pm$ 8.5 \\[6pt]
MAPPO-DQJL-CAV 
& 178.8 $\pm$ 9.7 & 159.6 $\pm$ 10.5 & 141.8 $\pm$ 10.6 & 134.2 $\pm$ 9.0 & 127.5 $\pm$ 8.0 \\[6pt]
MAPPO-DQJL 
& \textbf{178.4} $\pm$ 9.3 & \textbf{156.4} $\pm$ 10.8 & \textbf{135.9} $\pm$ 10.9 & \textbf{127.8} $\pm$ 9.2 & \textbf{125.7} $\pm$ 8.7 \\
\midrule
\textbf{\textit{\boldmath$N_{\text{EMV}}^{\text{LC}}$}} & & & & & \\ 
\multirow{1}{*}{No Control} & \multicolumn{5}{c}{3.2 $\pm$ 0.4} \\[6pt]
PPO-CC~\cite{suo2024model} 
& \textbf{1.0} $\pm$ 0.0 & \textbf{1.0} $\pm$ 0.0 & 1.0 $\pm$ 0.0 & 1.0 $\pm$ 0.0 & 1.0 $\pm$ 0.0 \\[6pt]
MAPPO-DQJL-CAV 
& 2.9 $\pm$ 0.4 & 2.1 $\pm$ 0.3 & 1.5 $\pm$ 0.2 & 1.2 $\pm$ 0.2 & 1.1 $\pm$ 0.2 \\[6pt]
MAPPO-DQJL 
& 3.1 $\pm$ 0.3 & 1.7 $\pm$ 0.3 & \textbf{0.9} $\pm$ 0.2 & \textbf{0.5} $\pm$ 0.1 & \textbf{0.4} $\pm$ 0.1 \\
\midrule
\textbf{\textit{\boldmath$N_{\text{non-EMV}}^{\text{LC}}$}} & & & & & \\ 
\multirow{1}{*}{No Control} & \multicolumn{5}{c}{27.3 $\pm$ 2.1} \\[6pt]
PPO-CC~\cite{suo2024model} 
& \textbf{26.8} $\pm$ 2.2 & 21.6 $\pm$ 1.8 & 17.2 $\pm$ 1.4 & 14.6 $\pm$ 1.2 & 12.3 $\pm$ 1.0 \\[6pt]
MAPPO-DQJL-CAV 
& 27.4 $\pm$ 2.2 & 24.5 $\pm$ 1.9 & 19.3 $\pm$ 1.5 & 14.0 $\pm$ 1.2 & 11.5 $\pm$ 1.0 \\[6pt]
MAPPO-DQJL 
& 26.9 $\pm$ 1.9 & \textbf{22.0} $\pm$ 1.6 & \textbf{16.5} $\pm$ 1.4 & \textbf{12.3} $\pm$ 1.1 & \textbf{10.2} $\pm$ 0.9 \\
\bottomrule
\end{tabular}
}
\caption{Performance metrics on the Three-Lane Roadway across different penetration rates. Bold values indicate the best performance.}
\label{tab:penetration_rate_three_lane}
\end{table}
As shown in Table~\ref{tab:penetration_rate_three_lane}, MAPPO-DQJL yields lower EMV passage times than all other methods at every non-zero penetration level. Similar to the two-lane scenario, MAPPO-DQJL-CAV remains at a disadvantage due to its limited integration of HDVs. However, the introduction of a third lane provides additional flexibility for non-EMVs to yield, thereby creating more free space and further reducing EMV passage times relative to the two-lane case.

Similarly, the presence of an additional lane provides the EMV with greater lateral flexibility, reducing the number of lane changes required during its traversal. This increased spatial availability allows the EMV to complete its passage with fewer maneuvers, highlighting the benefits of enhanced clearance. However, this improvement comes at the cost of increased lane changes by non-EMVs, as they actively yield space for the EMV. This trend is particularly evident in the evaluation of $N_{\text{non-EMV}}^{\text{LC}}$, where the total number of non-EMV maneuvers increases to accommodate the EMV’s passage. 

Nevertheless, it is important to note that the rise in $N_{\text{non-EMV}}^{\text{LC}}$ is partially attributable to the increased number of non-EMVs present on the roadway. When normalized by the number of vehicles, $N_{\text{non-EMV}}^{\text{LC}}$ per vehicle is lower compared to other methods. This observation underscores the effectiveness of MAPPO-DQJL in minimizing additional disturbances.

\paragraph{Four-Lane Roadway} With the same traffic flow configurations as above, we present the performance metrics in Table~\ref{tab:penetration_study_four_lane}.
\begin{table}[htbp]
\centering
{\fontsize{9}{11}\selectfont
\begin{tabular}{lccccc}
\toprule
\textbf{Method} & \textbf{0\%} & \textbf{25\%} & \textbf{50\%} & \textbf{75\%} & \textbf{100\%} \\
\midrule
\textbf{\textit{$t_{\text{EMV}} (s)$}} & & & & & \\ 
\multirow{1}{*}{No Control} & \multicolumn{5}{c}{184.1 $\pm$ 9.9} \\[6pt]
PPO-CC~\cite{suo2024model}
& 179.0 $\pm$ 9.2 & 165.0 $\pm$ 9.8 & 150.5 $\pm$ 10.0 & 128.5 $\pm$ 8.8 & 120.0 $\pm$ 8.2 \\[6pt]
MAPPO-DQJL-CAV
& 178.8 $\pm$ 9.5 & 150.2 $\pm$ 10.2 & 130.8 $\pm$ 10.3 & 122.0 $\pm$ 9.0 & 115.0 $\pm$ 8.5 \\[6pt]
MAPPO-DQJL
& \textbf{178.5} $\pm$ 9.3 & \textbf{147.5} $\pm$ 10.5 & \textbf{125.0} $\pm$ 10.7 & \textbf{118.5} $\pm$ 8.8 & \textbf{113.0} $\pm$ 8.0 \\
\midrule
\textbf{\textit{\boldmath$N_{\text{EMV}}^{\text{LC}}$}} & & & & & \\ 
\multirow{1}{*}{No Control} & \multicolumn{5}{c}{2.5 $\pm$ 0.3} \\[6pt]
PPO-CC~\cite{suo2024model} 
& \textbf{1.0} $\pm$ 0.0 & \textbf{1.0} $\pm$ 0.0 & 1.0 $\pm$ 0.0 & 1.0 $\pm$ 0.0 & 1.0 $\pm$ 0.0 \\[6pt]
MAPPO-DQJL-CAV
& 2.2 $\pm$ 0.3 & 1.6 $\pm$ 0.3 & 1.1 $\pm$ 0.2 & 0.9 $\pm$ 0.2 & 0.7 $\pm$ 0.1 \\[6pt]
MAPPO-DQJL
& 2.4 $\pm$ 0.3 & 1.4 $\pm$ 0.3 & \textbf{0.9} $\pm$ 0.2 & \textbf{0.6} $\pm$ 0.1 & \textbf{0.4} $\pm$ 0.1 \\
\midrule
\textbf{\textit{\boldmath$N_{\text{non-EMV}}^{\text{LC}}$}} & & & & & \\ 
\multirow{1}{*}{No Control} & \multicolumn{5}{c}{34.1 $\pm$ 2.7} \\[6pt]
PPO-CC~\cite{suo2024model}
& 33.7 $\pm$ 2.6 & 27.4 $\pm$ 2.3 & 22.3 $\pm$ 2.0 & 18.5 $\pm$ 1.8 & 15.6 $\pm$ 1.3 \\[6pt]
MAPPO-DQJL-CAV
& 34.3 $\pm$ 2.5 & 28.6 $\pm$ 2.4 & 21.7 $\pm$ 1.9 & 16.9 $\pm$ 1.5 & 15.8 $\pm$ 1.3 \\[6pt]
MAPPO-DQJL
& \textbf{33.5} $\pm$ 2.8 & \textbf{26.6} $\pm$ 2.2 & \textbf{19.6} $\pm$ 1.8 & \textbf{16.2} $\pm$ 1.3 & \textbf{15.1} $\pm$ 1.0 \\
\bottomrule
\end{tabular}
}
\caption{Performance metrics on the Four-Lane Roadway across different penetration rates. Bold values indicate the best performance.}
\label{tab:penetration_study_four_lane}
\end{table}
As indicated in Table~\ref{tab:penetration_study_four_lane}, relative to the three-lane scenario, the EMV passage time under MAPPO-DQJL is further reduced, and the EMV requires even fewer lane changes at high CAV penetration. In this configuration, the extra lane further diminishes the EMV’s need to shift lanes. At 100\% penetration, MAPPO-DQJL reduces EMV lane changes to approximately 0.4 on average, implying that under most circumstances, the EMV can traverse the segment with virtually no lane-changing. This outcome suggests that the proposed method has effectively reached a peak level of coordination and efficiency in facilitating EMV passage.

Meanwhile, although we observe a slight increase in the total number of lane-change maneuvers for all non-EMVs as their numbers grow, MAPPO-DQJL still ensures that the average number of lane-change maneuvers per non-EMV remains the smallest. In other words, while non-EMVs collectively adjust their positions more frequently to accommodate the EMV, the overall system benefits from greater lane-change efficiency. This pattern reinforces that, from a system-wide perspective, MAPPO-DQJL optimizes the trade-off between EMV clearance and non-EMV maneuvering complexity. In this setting, we observe a 39.8\% reduction in EMV passage time and 55.7\% decrease in non-EMV lane-changing maneuvers with the adopt of MAPPO-DQJL.

The findings clearly indicate that as the penetration rate of CAVs increases, both EMV passage time and the number of maneuvers for EMV consistently decrease, as anticipated. This reduction occurs because, in the worst-case scenario, CAV behavior approximates that of HDVs, thereby providing at least the baseline level of performance. Moreover, the diminishing marginal improvement in reward with increasing penetration rates suggests that, beyond a certain threshold of CAV presence in the mixed traffic, the system approaches near-optimal conditions for DQJL formation. Achieving this critical mass of CAVs is therefore key to unlocking substantial gains in traffic efficiency. These results align with the findings of related studies investigating various traffic management tasks in a mixed environment \cite{wu2017flow,Houshmand2019Penetration,ding2020penetration,argote2015connected,du2017coordination,khondaker2015variable}, further emphasizing the importance of CAV penetration rates in shaping overall traffic dynamics.

\subsubsection{Traffic Density}
\paragraph{Two-Lane Roadway}
\begin{table}[htbp]
\centering
\begin{tabular}{lccc}
\toprule
\textbf{Method} & \textbf{5 veh/lane/min} & \textbf{15 veh/lane/min} & \textbf{25 veh/lane/min} \\
\midrule
\textbf{\textit{$t_{\text{EMV}} (s)$}} & & & \\ 
No Control & 120.3 $\pm$ 9.1 & 171.2 $\pm$ 10.7 & 212.4 $\pm$ 13.2 \\
PPO-CC~\cite{suo2024model} & 118.7 $\pm$ 9.5 & 169.8 $\pm$ 11.2 & 202.6 $\pm$ 12.5 \\
MAPPO-DQJL-CAV & 117.4 $\pm$ 9.3 & 159.7 $\pm$ 10.3 & 191.5 $\pm$ 11.4 \\
MAPPO-DQJL & \textbf{114.9} $\pm$ 8.7 & \textbf{149.9} $\pm$ 11.8 & \textbf{188.2} $\pm$ 10.9 \\
\midrule
\textbf{\textit{\boldmath$N_{\text{EMV}}^{\text{LC}}$}} & & & \\ 
No Control & 2.2 $\pm$ 0.5 & 4.0 $\pm$ 0.3 & 3.5 $\pm$ 0.6 \\
PPO-CC~\cite{suo2024model} & 1.0 $\pm$ 0.0 & 1.0 $\pm$ 0.0 & 1.0 $\pm$ 0.0 \\
MAPPO-DQJL-CAV & 1.3 $\pm$ 0.4 & 1.6 $\pm$ 0.3 & 2.5 $\pm$ 0.5 \\
MAPPO-DQJL & \textbf{0.5} $\pm$ 0.1 & \textbf{0.8} $\pm$ 0.2 & \textbf{1.2} $\pm$ 0.3 \\
\midrule
\textbf{\textit{\boldmath$N_{\text{non-EMV}}^{\text{LC}}$}} & & & \\ 
No Control & 8.8 $\pm$ 1.2 & 15.0 $\pm$ 1.7 & 21.0 $\pm$ 2.2 \\
PPO-CC~\cite{suo2024model} & 9.2 $\pm$ 1.1 & 12.8 $\pm$ 1.5 & 18.5 $\pm$ 2.1 \\
MAPPO-DQJL-CAV & 9.7 $\pm$ 1.2 & 16.3 $\pm$ 1.7 & 20.9 $\pm$ 2.3 \\
MAPPO-DQJL & \textbf{7.9} $\pm$ 1.2 & \textbf{10.9} $\pm$ 1.4 & \textbf{16.2} $\pm$ 1.4 \\
\bottomrule
\end{tabular}
\caption{Performance metrics on the Two-Lane Roadway across varying traffic densities. Bold values indicate the best performance.}
\label{tab:density_two_lane}
\end{table}
As indicated in Table~\ref{tab:density_two_lane}, under moderate traffic conditions (e.g., 15 veh/lane/min), MAPPO-DQJL demonstrates relatively greater time savings compared to fully congested scenarios (25 veh/lane/min). This finding suggests that while MAPPO-DQJL’s coordinated strategy excels at reducing EMV passage time and maneuver complexity at moderate densities, its relative benefits diminish as the system approaches full saturation. In other words, while MAPPO-DQJL significantly improves performance under intermediate traffic loads, the advantage it offers naturally tapers off once congestion reaches near-maximum levels.

Under low-density conditions, the EMV performs approximately 0.5 lane-change maneuvers on average, demonstrating the near-elimination of unnecessary lane changes. Even as density increases to 25 veh/lane/min, MAPPO-DQJL maintains fewer EMV maneuvers than any other method, underscoring its capacity to ensure smoother EMV passage. At moderate congestion levels, MAPPO-DQJL’s EMV lane-change frequency is only about 20\% of that observed with No Control, and even under heavily congested conditions (25 veh/lane/min), it remains approximately one-third. These observations align with earlier findings, suggesting that as spatial flexibility declines, the coordination potential of MAPPO-DQJL diminishes, yet still surpasses that of alternative approaches.
Meanwhile, in low-density conditions (5 veh/lane/min), non-EMVs make relatively few lane changes overall, and MAPPO-DQJL still achieves the lowest number. At moderate density (15 veh/lane/min), the values are in line with previous analyses, reaffirming MAPPO-DQJL’s advantage. Under heavy congestion (25 veh/lane/min), non-EMVs must undertake more lane changes to accommodate the EMV, but MAPPO-DQJL continues to yield fewer maneuvers than the other methods, demonstrating its enduring benefits even as the system approaches saturation.

\paragraph{Three-Lane Roadway}
\begin{table}[htbp]
\centering
\begin{tabular}{lccc}
\toprule
\textbf{Method} & \textbf{5 veh/lane/min} & \textbf{15 veh/lane/min} & \textbf{25 veh/lane/min} \\
\midrule
\textbf{\textit{$t_{\text{EMV}} (s)$}} & & & \\ 
No Control        & 94.7 $\pm$ 7.4 & 156.7 $\pm$ 10.9 & 216.2 $\pm$ 12.3 \\
PPO-CC~\cite{suo2024model} & 92.9 $\pm$ 7.2 & 150.1 $\pm$ 10.4 & 206.4 $\pm$ 11.9 \\
MAPPO-DQJL-CAV    & 91.3 $\pm$ 7.6 & 141.6 $\pm$ 10.5 & 199.1 $\pm$ 11.3 \\
MAPPO-DQJL        & \textbf{89.4} $\pm$ 6.9 & \textbf{135.7} $\pm$ 10.8 & \textbf{196.3} $\pm$ 10.0 \\
\midrule
\textbf{\textit{\boldmath$N_{\text{EMV}}^{\text{LC}}$}} & & & \\ 
No Control        & 1.9 $\pm$ 0.2 & 1.3 $\pm$ 0.2 & 2.9 $\pm$ 0.5 \\
PPO-CC~\cite{suo2024model} & 1.0 $\pm$ 0.0 & 1.0 $\pm$ 0.0 & 1.0 $\pm$ 0.0 \\
MAPPO-DQJL-CAV    & 1.4 $\pm$ 0.3 & 1.6 $\pm$ 0.2 & 2.2 $\pm$ 0.3 \\
MAPPO-DQJL        & \textbf{0.8} $\pm$ 0.1 & \textbf{0.9} $\pm$ 0.2 & \textbf{1.4} $\pm$ 0.2 \\
\midrule
\textbf{\textit{\boldmath$N_{\text{non-EMV}}^{\text{LC}}$}} & & & \\ 
No Control        & 13.7 $\pm$ 1.4 & 21.9 $\pm$ 1.8 & 31.3 $\pm$ 2.3 \\
PPO-CC~\cite{suo2024model} & 13.2 $\pm$ 1.3 & 17.2 $\pm$ 1.4 & 25.4 $\pm$ 2.0 \\
MAPPO-DQJL-CAV    & 14.8 $\pm$ 1.5 & 19.2 $\pm$ 1.6 & 27.4 $\pm$ 2.1 \\
MAPPO-DQJL        & \textbf{12.9} $\pm$ 1.4 & \textbf{16.7} $\pm$ 1.3 & \textbf{23.1} $\pm$ 2.2 \\
\bottomrule
\end{tabular}
\caption{Performance metrics on the Three-Lane Roadway across varying traffic densities. Bold values indicate the best performance.}
\label{tab:density_three_lane}
\end{table}

The performance metrics across varying traffic densities on the Three-Lane Roadway, as summarized in Table~\ref{tab:density_three_lane}, highlight the consistent advantages of MAPPO-DQJL. For \textit{\boldmath$t_{\text{EMV}}$}, MAPPO-DQJL demonstrates shorter durations across all densities. At a moderate density of 15 veh/lane/min, these results align closely with prior benchmarks, confirming its robustness. Even as traffic density increases to 25 veh/lane/min, MAPPO-DQJL maintains a relative advantage over competing methods, underscoring its effectiveness in high-density scenarios.

The trends in EMV lane-change maneuvers further emphasize the benefits of MAPPO-DQJL. As traffic density rises, the additional lanes of the roadway contribute to keeping EMV maneuvers low. At 15 veh/lane/min, MAPPO-DQJL's performance is consistent with earlier analyses, while at higher densities, it continues to effectively limit the EMV's lane changes compared to other methods.

Similarly, non-EMV lane-change maneuvers exhibit the anticipated trend of increasing with traffic density. At 15 veh/lane/min, MAPPO-DQJL’s results align with previously established patterns, demonstrating fewer lane changes than competing methods. Even at 25 veh/lane/min, where maneuver frequencies are heightened, MAPPO-DQJL preserves a distinct advantage, reflecting its capacity to handle high-density traffic scenarios more efficiently.

Additionally, it is noteworthy that MAPPO-DQJL-CAV induces more non-EMV lane changes than both PPO-CC and MAPPO-DQJL. This outcome arises from the fact that MAPPO-DQJL-CAV, by treating HDVs purely as environmental elements rather than learning agents, fails to incorporate the nuanced, diverse behaviors of HDVs into its decision-making process. Consequently, the control framework lacks the sophistication to minimize disruptions effectively, leading to a higher frequency of non-EMV maneuvers.

\paragraph{Four-Lane Roadway}
\begin{table}[htbp]
\centering
\begin{tabular}{lccc}
\toprule
& 5 veh/lane/min & 15 veh/lane/min & 25 veh/lane/min \\
\midrule
\multicolumn{4}{l}{\textbf{\textit{$t_{\text{EMV}} (s)$}}} \\[4pt]
No Control        & 84.3 $\pm$ 6.9  & 147.2 $\pm$ 10.4 & 206.7 $\pm$ 11.7 \\
PPO-CC~\cite{suo2024model} & 82.4 $\pm$ 6.7  & 140.3 $\pm$ 10.1 & 196.6 $\pm$ 10.9 \\
MAPPO-DQJL-CAV    & 81.6 $\pm$ 6.8  & 131.4 $\pm$ 10.2 & 186.2 $\pm$ 10.6 \\
MAPPO-DQJL        & \textbf{78.7} $\pm$ 5.9 & \textbf{125.4} $\pm$ 10.6 & \textbf{183.1} $\pm$ 9.9 \\
\midrule
\textbf{\textit{\boldmath$N_{\text{EMV}}^{\text{LC}}$}} & & & \\ 
No Control        & 1.6 $\pm$ 0.3 & 1.3 $\pm$ 0.2 & 2.3 $\pm$ 0.4 \\
PPO-CC~\cite{suo2024model} & 1.0 $\pm$ 0.0 & 1.0 $\pm$ 0.0 & 1.0 $\pm$ 0.0 \\
MAPPO-DQJL-CAV    & 1.1 $\pm$ 0.2 & 1.2 $\pm$ 0.2 & 1.7 $\pm$ 0.3 \\
MAPPO-DQJL        & \textbf{0.5} $\pm$ 0.1 & \textbf{0.6} $\pm$ 0.1 & \textbf{1.1} $\pm$ 0.2 \\
\midrule
\textbf{\textit{\boldmath$N_{\text{non-EMV}}^{\text{LC}}$}} & & & \\ 
No Control        & 17.4 $\pm$ 1.7 & 26.3 $\pm$ 2.1 & 35.4 $\pm$ 2.4 \\
PPO-CC~\cite{suo2024model} & 16.6 $\pm$ 1.6 & 19.9 $\pm$ 1.9 & 28.4 $\pm$ 2.2 \\
MAPPO-DQJL-CAV    & 17.9 $\pm$ 1.5 & 21.3 $\pm$ 1.8 & 29.3 $\pm$ 2.3 \\
MAPPO-DQJL        & \textbf{15.2} $\pm$ 1.3 & \textbf{16.4} $\pm$ 1.5 & \textbf{24.7} $\pm$ 2.1 \\
\bottomrule
\end{tabular}
\caption{Performance metrics on the Four-Lane Roadway across varying traffic densities. Bold entries indicate best performance.}
\label{tab:density_four_lane}
\end{table}
As indicated in Table~\ref{tab:density_four_lane}, the introduction of an additional lane in the four-lane scenario further reduces EMV passage times at low and moderate densities, aligning with previously established trends. Although EMV passage times rise noticeably at 25 veh/lane/min, MAPPO-DQJL still achieves the shortest durations. A similar pattern is evident when examining EMV lane changes: under moderate densities, the EMV can maintain even fewer lane changes, and while maneuver counts increase slightly at high density, MAPPO-DQJL continues to enable the fewest EMV lane changes overall. In parallel, non-EMVs must also perform additional lane changes as density escalates. While MAPPO-DQJL moderates this increase more effectively than other approaches, the underlying trend of heightened complexity at greater densities persists, mirroring the patterns observed in both two-lane and three-lane configurations.

Experimental results for all three scenarios have cleary indicate that, under low-density conditions, the performance gap between MAPPO-DQJL and other candidate schemes remains modest, as even conventional approaches can efficiently handle sparse traffic. As density increases and approaches moderate congestion levels, MAPPO-DQJL’s advantage becomes most pronounced, significantly outpacing other schemes in reducing EMV passage times. However, once the roadway becomes fully congested, this benefit diminishes. In scenarios of near-total saturation, no scheme can substantially expedite EMV passage, resulting in a vanishing performance gap. This finding underlines the importance of intermediate traffic densities, where MAPPO-DQJL’s strategic coordination yields the greatest relative gains over competing approaches. Studies \cite{stern2018dissipation, wang2015caccdensity, feng2021densityimpact}, exploring mixed traffic tasks similarly identify “sweet spots” in density. These insights emphasize that operational conditions, not just automation levels, are crucial for realizing the full benefits of advanced traffic management interventions.

Across the entire density sensitivity study, MAPPO-DQJL-CAV consistently underperformed compared to MAPPO-DQJL. This disparity arises because MAPPO-DQJL-CAV treats HDVs purely as part of the environment, rather than as learning agents. Without incorporating HDVs into the coordination process, the control framework cannot anticipate or adapt to the distinct and often stochastic behaviors of each HDV. Consequently, as traffic densities increase, the system struggles to maintain efficient flow, leading to more frequent lane changes and prolonged EMV passage times. In other words, omitting HDVs from the learning process significantly impairs the scheme’s overall coordination capabilities, particularly under congested conditions. 

This observation reinforces the importance of designing a learning framework that accounts for both CAVs and HDVs as active participants, thereby optimizing system performance across a wide range of traffic scenarios.
\FloatBarrier
\subsection{Ablation Studies}
\label{subsec:ablations}
A range of ablation studies is conducted to clarify the influence of key components in the MAPPO-DQJL framework. By selectively altering certain design choices, we can isolate their effects and better understand the fundamental drivers of performance in DQJL formation.

\subsubsection{Calibration of \texorpdfstring{$\alpha$}{alpha}}\label{subsec:reward}
Reward shaping plays a pivotal role in guiding agent behaviors. By adjusting the balance between global and local rewards, we can observe how different configurations influence EMV passage efficiency and overall traffic stability. All experiments conducted in Subsec.~\ref{subsec:sensitivity} adopt $\alpha = 0.75$. In this ablation study, we systematically vary the balancing coefficient \(\alpha\) in Eq.~\eqref{eqn:return} on the Three-Lane Roadway under 15 veh/lane/min traffic density and 50\% CAV penetration. Each unique value of \(\alpha\) is associated with a separate training phase for the agents, followed by 20 simulation runs to evaluate performance. The results are presented in Table~\ref{tab:alpha_ablations}.
\begin{table}[htbp]
\centering
\begin{tabular}{lccccc}
\toprule
 & \(\alpha = 0\) & \(\alpha = 0.25\) & \(\alpha = 0.5\) & \(\alpha = 0.75\) & \(\alpha = 1.0\) \\
\midrule
$t_{\text{EMV}}$ (s) & 200.3 $\pm$ 11.8 & 170.7 $\pm$ 11.9 & 150.2 $\pm$ 11.3 & 135.9 $\pm$ 10.9 & 125.4 $\pm$ 10.1 \\[6pt]
$N_{\text{EMV}}^{\text{LC}}$ & 4.1 $\pm$ 0.4 & 2.9 $\pm$ 0.3 & 1.7 $\pm$ 0.3 & 0.9 $\pm$ 0.2 & 0.7 $\pm$ 0.1 \\[6pt]
$N_{\text{non-EMV}}^{\text{LC}}$ & 10.2 $\pm$ 1.1 & 13.7 $\pm$ 1.6 & 16.3 $\pm$ 1.7 & 19.7 $\pm$ 1.3 & 23.4 $\pm$ 1.9 \\
\bottomrule
\end{tabular}
\caption{Effects of different \(\alpha\) values on EMV passage time and lane-change maneuvers on the Three-Lane Roadway with 15 veh/lane/min and 50\% CAV penetration rate.}
\label{tab:alpha_ablations}
\end{table}

As shown in Table~\ref{tab:alpha_ablations}, increasing \(\alpha\) leads to lower \(t_{\text{EMV}}\) and \(N_{\text{EMV}}^{\text{LC}}\), as the agents shift their focus toward optimizing global EMV-related objectives. This adjustment in reward composition enhances the EMV’s passage efficiency and reduces its lane-change frequency. However, this improvement comes with a trade-off. As \(N_{\text{EMV}}^{\text{LC}}\) decreases, \(N_{\text{non-EMV}}^{\text{LC}}\) tends to increase. The process of dynamically clearing lanes for the EMV necessitates multiple layers of yielding maneuvers, compelling non-EMVs to adjust their positions more frequently. In other words, while prioritizing the EMV’s global objectives streamlines its trajectory, the surrounding traffic must accommodate these changes, resulting in a higher number of lane-change maneuvers by non-EMVs. This provides valuable insight into how EMV operations can cause disruption to overall traffic flow, emphasizing the need for carefully balanced strategies when managing EMV passage under various traffic conditions.
\subsubsection{Parameter Sharing}\label{subsec:parameters_sharing}
Parameter sharing among CAVs facilitates knowledge transfer and promotes consistent strategies across agents. To evaluate the impact of this design choice, we conducted an ablation study comparing our shared policy approach against a variant where each CAV maintains an independent policy network while HDVs retain their fixed policies. This investigation helps understand whether independent learning could potentially lead to better coordination strategies for DQJL formation.

Due to the computational demands of training multiple independent networks, we constrained the experiment to 15 CAVs and 15 HDVs, with HDVs maintaining their fixed behavioral policies throughout. The simulation environment was configured as a Three-Lane Roadway spanning 500 meters, allowing for focused analysis of how parameter sharing affects CAV coordination capability.

\begin{figure}[htbp]
    \centering
    \includegraphics[width=0.7\textwidth]{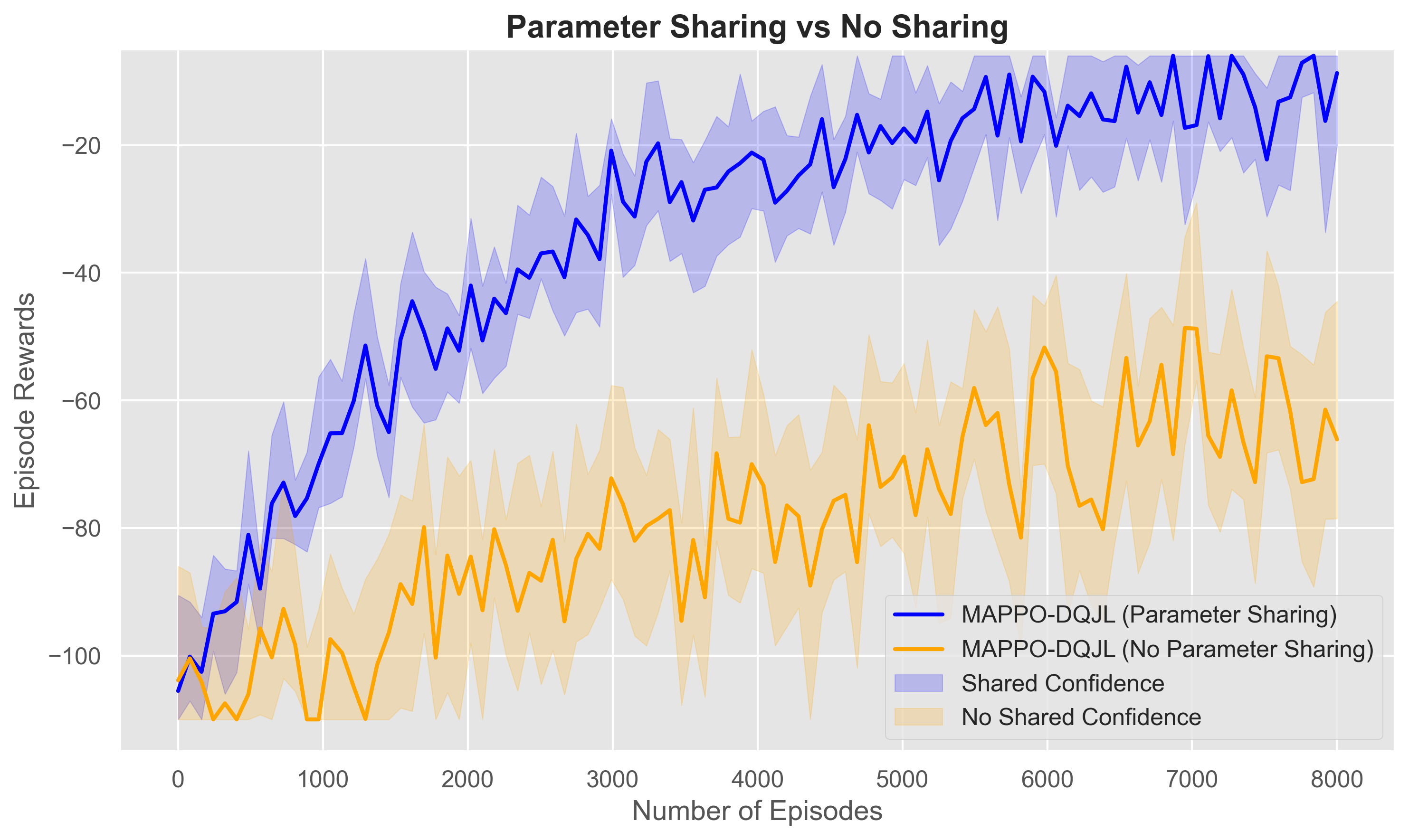}
    \caption{Reward convergence with and without parameter sharing for CAVs.}
    \label{fig:parameter_sharing}
\end{figure}
See Fig.~\ref{fig:parameter_sharing} for the learning patterns of CAV agents with and without shared parameters. Without a shared policy network, CAV agents are treated as independent learners, optimizing their policies in isolation. This limits generalization and knowledge transfer, making it harder to converge on optimal strategies for establishing DQJLs.

In contrast, treating CAV agents as homogeneous with a shared policy network offers key advantages. Parameter sharing facilitates knowledge transfer, enabling consistent strategies across agents, especially in stochastic environments. It also reduces learning complexity, accelerating convergence, and inherently promotes cooperation by aligning agent objectives. These benefits highlight the efficacy of MAPPO-DQJL’s shared policy design.

\section{Summary}
\label{sec:conclusion}

This study introduces DQJLs as a practical solution for enabling EMVs to navigate congested urban traffic efficiently. Using the MAPPO framework, the proposed system coordinates CAVs and HDVs to improve EMV passage times while minimizing disruptions to overall traffic flow.

Results from SUMO simulations validate the framework’s ability to reduce EMV passage times with minimal lane changes, for both EMVs and non-EMVs. By treating HDVs as dynamic agents rather than passive obstacles, the system accounts for their stochastic behaviors, enhancing traffic interaction modeling and coordination. The framework scales effectively to multi-lane scenarios, achieving smoother traffic flow and fewer EMV lane changes. Higher CAV penetration rates amplify these benefits, leading to shorter EMV passage times and fewer maneuvers for all vehicles, emphasizing the importance of adopting V2X technologies.

The CTDE approach, combined with a transformer-based architecture, enhances scalability and robustness. This design models complex inter-agent dependencies, ensuring reliable performance in high-density, dynamic traffic conditions. The results demonstrate the framework’s capacity to manage real-world traffic complexities while maintaining system efficiency.

DQJLs have clear potential for real-world ITS applications, particularly in congested urban settings. By reducing EMV response times, the proposed approach improves outcomes for ambulances, fire trucks, and law enforcement, safeguarding lives and property. As V2X technologies continue to advance, this framework offers a path toward more cooperative and efficient roadways, transforming EMV assistance into a streamlined, society-wide effort to enhance urban safety and resilience. Future work will focus on incorporating more realistic features, expanding scalability to address higher-dimensional uncertainties, and bridging the sim-to-real gap for practical deployment.

\chapter{Intersection-Aware Assessment of EMS Accessibility in NYC: A Data-Driven Approach}\label{chap:ems_accessibility}

\section{Introduction}

\label{sec:introduction_access}
Emergency response times in urban areas, particularly in densely populated metropolitan regions such as New York City (NYC), are profoundly influenced by traffic congestion. Research has established that reducing the travel time of emergency vehicles (EMVs)—a critical component of overall emergency response time (ERT)—can significantly improve outcomes for urgent incidents, including emergency medical services (EMS)~\cite{blackwell2002response, SCHUSTER2024104017}, fires~\cite{jaldell2017important,challands2010relationships}, and police activities~\cite{blanes2018effect}. However, as New York City continues to undergo rapid urbanization, the growing congestion on its roadways has exacerbated this challenge. Recent reports indicate that the city's emergency response time for life-threatening situations has increased by 29\% over the past decade, reflecting a concerning trend of slower responses in high-priority cases~\cite{EMS1_NYC_Response_Times}. According to the Mayor’s Management Report, response times in New York City have risen by 82 seconds—16.1\% higher than the 8 minute and 28 second average in fiscal year 2019—surpassing even the elevated times seen during the onset of the COVID-19 pandemic in fiscal 2020, despite a 52\% increase in civilian fire deaths in fiscal 2023~\cite{NYPost_Response_Times_2023}.

Various approaches have been introduced to mitigate delays in EMV travel time, such as traffic signal preemption and route optimization. However, these methods are typically implemented in isolation and fail to address the dynamic and evolving nature of urban traffic congestion. For instance, signal preemption strategies prioritize EMVs by altering traffic signals, but often at the cost of significant disruption to non-EMV traffic. These inefficiencies propagate through the network, creating broader delays and unintended consequences for overall traffic management. Moreover, these traditional methods are static and lack the adaptability needed to respond to real-time fluctuations in traffic conditions and congestion patterns. As a result, their long-term effectiveness is limited, particularly in a complex urban environment like New York City, where traffic patterns are unpredictable and can change rapidly. Addressing these challenges necessitates more integrated and adaptive systems that holistically coordinate traffic management, minimizing disruption to non-EMV traffic while ensuring swift emergency response. 

A recent study on emergency response accessibility in NYC identified significant disparities in service coverage, particularly in underserved areas such as Staten Island and parts of Queens~\cite{chung2024access}. These regions, referred to as "emergency service deserts," experience prolonged response times due to inadequate proximity to fire stations and EMS. The study further highlights a correlation between population density and emergency service accessibility, with less dense areas facing longer delays. While these findings emphasize the need for innovative traffic management solutions to enhance emergency response across all boroughs, the proposed accessibility model in~\cite{chung2024access} has been shown to underestimate EMV travel times within the city. An improved EMS accessibility model is therefore essential to accurately identify and address vulnerable areas in NYC.

Building upon these findings, this study aims to develop an EMS accessibility model that incorporates the potential adoption of large-scale traffic signal control (TSC) systems to facilitate EMV passage in congested networks. Using this EMS accessibility model, we can simulate travel times from the nearest medical facility to any location in the city, enabling the identification of vulnerable regions in terms of EMS coverage. Additionally, this model allows for the evaluation of how a TSC scheme, such as \textit{EMVLight}~\cite{su2023emvlight}, could support these vulnerable areas. By integrating advanced modeling capabilities with insights from the accessibility study, this study is able to demonstrate the potential to significantly enhance emergency response times in NYC’s most vulnerable and congested regions.

This study makes the following contributions:
\begin{enumerate}
    \item It integrates the New York City road network, intersection data, and population distribution into a unified analytical framework.
    \item It proposes an EMS accessibility model that accounts for the impact of intersection density along emergency routes.
    \item It identifies vulnerable regions using the proposed model, investigates the EMS accessibility and illustrates the potential improvements in accessibility achievable through the incorporation of \textit{EMVLight}.
\end{enumerate}

The structure of this study is as follows. Section~\ref{sec:literature_review_access} reviews existing literature on traffic signal controls for EMVs, Intersection delay and EMS accessibility. The datasets employed in this study are introduced in Section~\ref{sec:methodology}, which also presents the proposed intersection-aware EMS accessibility model. Section~\ref{sec:results} details the travel time analysis and the vulnerable regions based on the proposed method. Demographic evaluations are conducted on the result in Section~\ref{sec:discussion_access} and an adoption of \textit{EMVLight} is carried out in Subsec.~\ref{subsec:improvement}, highlighting the areas of New York City that stand to benefit the most. Finally, conclusions are summarized in Section~\ref{sec:conclusion_access}.

\section{Related Works}
\label{sec:literature_review_access}
This section reviews key literature supporting our study. Subsection~\ref{tsc_for_emvs} examines advancements in traffic signal control (TSC) for EMVs, Subsection~\ref{subsec:intersection_delay} focuses on intersection delays as a critical factor in EMV travel times and Subsection~\ref{subsec:ems_accessibility} explores EMS accessibility and its disparities. Together, these insights provide the foundation for our proposed methodology.

\subsection{EMS accessibility}\label{subsec:ems_accessibility}
The accessibility of EMS is a critical component of public health and safety, particularly in urban areas where rapid response times are essential for saving lives \cite{yang2020nuset,ma2021undistillable,ma2021good,you2020towards,you2020contextualized,xu2021semantic}. Studies have developed a variety of methodologies to evaluate and improve EMS accessibility. Novak and Sullivan~\cite{novak2014link} proposed a link-focused approach for assessing access to emergency services, emphasizing the importance of connectivity within urban road networks. Similarly, Xia et al.~\cite{xia2019measuring} utilized big GPS data to measure spatiotemporal accessibility, highlighting the dynamic nature of EMS access influenced by traffic and temporal factors.

Geographic Information Systems (GIS) have played a pivotal role in analyzing and addressing spatial disparities in EMS accessibility \cite{you2022sbilsan,ma2022sparse,feng2022kergnns}. Tansley et al.~\cite{tansley2015spatial} employed GIS to assess spatial access to EMS in low- and middle-income countries, revealing significant geographic inequities. Hashtarkhani et al.~\cite{hashtarkhani2020age} introduced an age-integrated GIS-based methodology to evaluate EMS accessibility, demonstrating the potential to tailor EMS services to different demographic groups. In Dhaka, Bangladesh, Ahmed et al.~\cite{ahmed2019impact} examined the impact of traffic variability on EMS access, highlighting the challenges posed by urban congestion in providing equitable emergency healthcare.

Several studies have addressed the influence of environmental and infrastructural factors on EMS accessibility \cite{chen2021self,you2021knowledge,ma2022stingy}. Green et al.~\cite{green2017city} explored city-scale accessibility during flood events, underscoring the importance of resilient infrastructure to maintain EMS operations during natural disasters. Similarly, Yin et al.~\cite{yin2017evaluating} evaluated the cascading impacts of sea-level rise and coastal flooding on EMS access in Lower Manhattan, New York City, revealing vulnerabilities in emergency response under climate-induced hazards. Albano et al.~\cite{albano2014gis} developed a GIS-based model to estimate the operability of emergency response structures during floods, providing critical insights for urban planning in disaster-prone areas.

Addressing urban-rural disparities has also been a key focus in EMS accessibility research \cite{you2022class,you2022end,you2022simcvd}. Luo et al.~\cite{luo2022locating} examined strategies for locating EMS facilities to reduce urban-rural inequalities, while Carr et al.~\cite{carr2009access} highlighted access challenges in rural areas within the United States. Holguin et al.~\cite{holguin2018access} proposed urban planning methodologies aimed at generating equity in EMS access, emphasizing the need for targeted interventions in underserved regions. The spatiotemporal aspect of EMS accessibility has been further explored in studies like those by Hu et al.~\cite{hu2020impact}, who analyzed the impact of traffic on spatial accessibility variations in inner-city Shanghai. Shi et al.~\cite{shi2022spatial} conducted a spatial accessibility assessment of urban tourist attractions in Shanghai, demonstrating the need for comprehensive urban planning to address emergency response challenges in high-density areas.

Chung et al.~\cite{chung2024access} present a geospatial analysis of emergency service accessibility in New York City, identifying significant disparities in coverage, particularly in Staten Island and parts of Queens. The study introduces an accessibility index that combines spatial proximity, road network connectivity, and response time data to highlight regions with low emergency service availability. Additionally, the authors examine the impact of traffic congestion on EMS response times, emphasizing challenges in densely populated areas.

In summary, EMS accessibility is influenced by a combination of spatial, temporal, and demographic factors \cite{han2023medgen3d,cai2023cross,zeng2023fast,you2023implicit}. The integration of advanced GIS techniques, real-time traffic data, and tailored urban planning approaches is critical for addressing disparities and ensuring equitable access to emergency services across diverse geographic and socio-economic contexts \cite{ma2023pre,you2023actionplus}.
\subsection{Intersections delay}\label{subsec:intersection_delay}
The influence of intersection density and topology on travel time has been a subject of extensive research. Higher intersection densities often lead to increased delays due to frequent stops and interruptions in traffic flow. Ewing and Cervero's meta-analysis~\cite{EwingCervero2010} established that although higher intersection densities are associated with reduced vehicle miles traveled (VMT), they often result in longer travel times due to frequent interruptions. Al-Dabbagh et al.~\cite{al2019impact} further emphasize the role of intersection topology, demonstrating how intersection design impacts traffic congestion in urban cities. The \textit{Highway Capacity Manual} (HCM)~\cite{HCM2010} offers methodologies for assessing urban street performance, highlighting the influence of signalized intersections on travel time \cite{liu2022retrieval}. 

Specific studies have also examined how intersections affect travel times across different transportation modes \cite{liu2022graph,chen2022exploring,you2022momentum}. Feng et al.~\cite{feng2015quantifying} quantified the joint impact of stop locations, signalized intersections, and traffic conditions on bus travel times, highlighting the compounded delays experienced by public transport. Strauss and Miranda-Moreno~\cite{strauss2017speed} explored cyclist travel patterns, demonstrating how delays at intersections significantly influence travel times in the Montreal network. Furthermore, Park et al.~\cite{park2016measuring} utilized Bluetooth-based data to measure intersection performance, illustrating how advanced data collection methods can provide detailed insights into intersection-related delays. Ivanov et al.~\cite{AssessingTrafficCapacity2022} also highlight the importance of intersection geometry and control mechanisms in mitigating delays in high-density areas. These findings collectively underscore the critical role of intersection design and performance in influencing travel time and highlight the need for targeted urban planning interventions \cite{han2023diffeomorphic,yan2023representation,chen2023bridge}. 

When it comes to EMV travel time, reducing delays at intersections is particularly critical, as every second lost can directly impact emergency outcomes \cite{dong2022multi,you2022incremental,dong2022flow}. Therefore, the ability to dynamically manage and optimize intersection performance is essential for ensuring timely emergency response, forming the focus of this study \cite{sun2023hybrid,liu2023llmrec,cao2023multi}.

\subsection{TSC for EMVs}\label{tsc_for_emvs}
The optimization of TSC for EMS is paramount in reducing response times and enhancing public safety \cite{lam2023large,you2023bootstrapping,yang2023multimodal}. Traditional TSC methods, such as signal preemption, temporarily adjust traffic signals to grant EMVs priority passage through intersections~\cite{obeck1991traffic}. While effective in facilitating EMV movement, these approaches often disrupt the flow of non-EMV traffic, leading to broader network inefficiencies~\cite{hashim2013traffic}. Advancements in intelligent transportation systems (ITS) have introduced more adaptive and dynamic TSC strategies. Bhate et al.~\cite{bhate2018iot} proposed an IoT-based intelligent traffic signal system that leverages real-time data to prioritize EMVs, demonstrating significant reductions in response times. Similarly, Noori et al.~\cite{noori2016connected} developed a connected vehicle-based TSC strategy, utilizing vehicle-to-infrastructure communication to enhance EMV preemption efficiency.
Predictive control models have also been explored to anticipate EMV arrivals and adjust signal timings accordingly. Qin and Khan~\cite{qin2012control} investigated control strategies for traffic signal timing transitions during EMV preemption, highlighting methods to minimize disruptions to regular traffic flow. Additionally, Shanmughasundaram et al.~\cite{shanmughasundaram2018li} introduced a Li-Fi-based automatic traffic signal control system, offering a novel approach to EMV prioritization through high-speed data transmission.

The integration of multi-agent reinforcement learning (MARL) frameworks has further advanced TSC for EMVs \cite{li2023nonconvex}. Su et al.~\cite{su2023emvlight} presented EMVLight, a decentralized MARL framework that optimizes both EMV routing and traffic signal coordination, effectively reducing response times while maintaining overall traffic efficiency. This approach underscores the potential of machine learning techniques in addressing the complexities of urban traffic management.

Comprehensive reviews by Bin Wan Hussin et al.~\cite{bin2019review} and Yu et al.~\cite{yu2022state} provide extensive analyses of existing TSC techniques for EMVs, encompassing traditional methods and emerging technologies. These reviews offer valuable insights into the evolution of TSC strategies and underscore the necessity for continued innovation in this domain.
\section{Methodology}
\label{sec:methodology}
In this section, we introduce the datasets used for this study in Subsec.~\ref{subsec:datasets} and the EMS accessibility model in \hs{Subsec.~\ref{subsec:ems_model}.}
\subsection{Datasets}\label{subsec:datasets}
In this subsection, we introduce the datasets used for the proposed methodology.
\subsubsection{Emergency medical services sites}
\begin{figure}[h!]
    \centering
    \includegraphics[width=\textwidth]{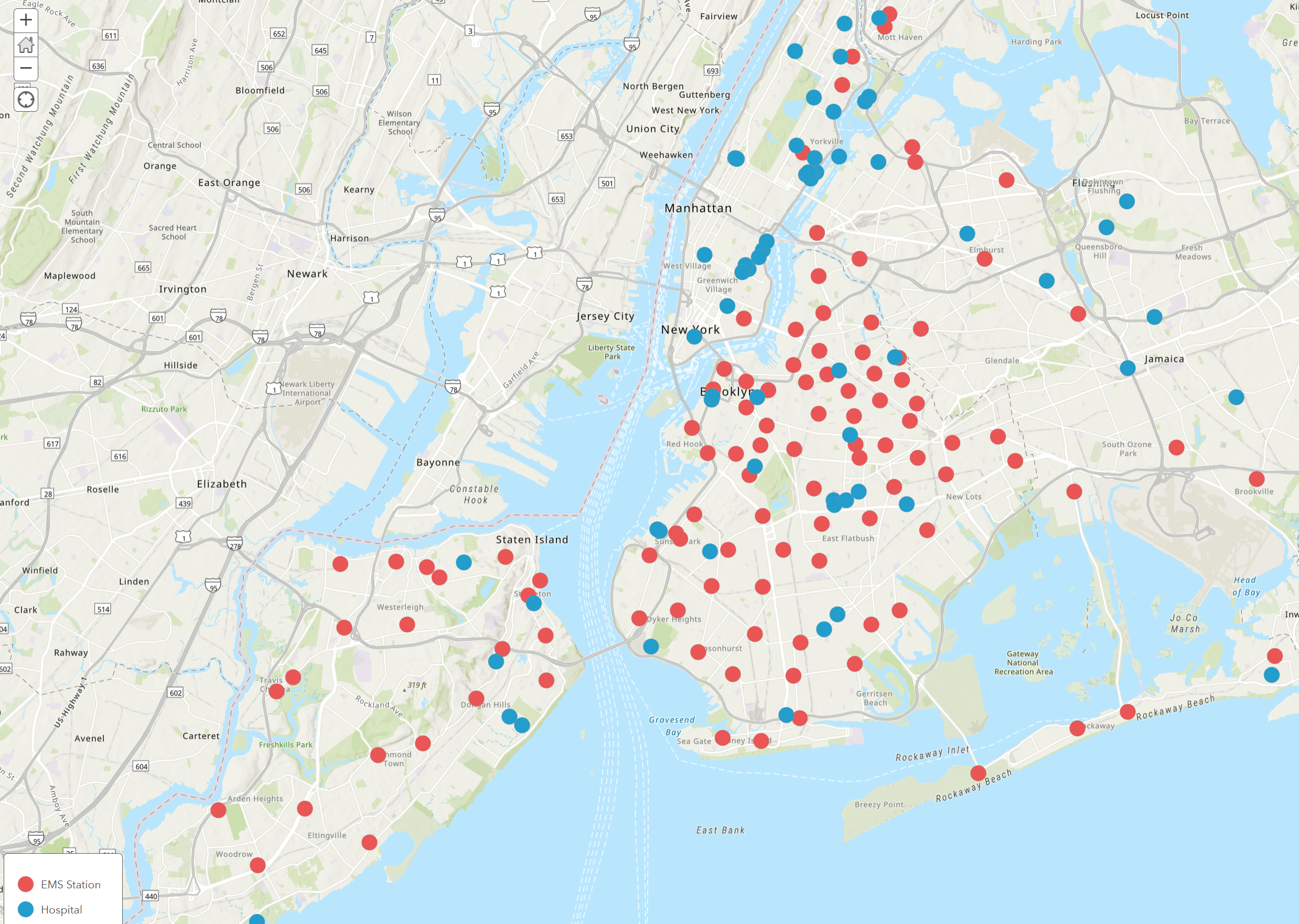} 
    \caption{EMS Stations (Blue) and Hospitals (Red) in NYC.}
    \label{fig:hospitals_and_ems_stations}
\end{figure}

The locations of emergency service sites are sourced from the Homeland Infrastructure Foundation-Level Data (HIFLD) Open Data repository \cite{yan2023localized,li2023marganvac}. Given that this study concentrates on optimizing the transit of ambulances and other Emergency Medical Services (EMS) vehicles, we specifically extract the precise locations of hospitals~\cite{HIFLD_hospital} and EMS stations~\cite{HIFLD_EMS_Stations} to ensure targeted analysis and effective modeling of emergency response routes. Note that the EMS stations dataset also includes certain fire stations, as these facilities provide EMS services as well. The numbers of sites by each borough is provided in Tab.~\ref{table:hospitals_and_ems_stations} and their locations are represented in Fig.~\ref{fig:hospitals_and_ems_stations}.
\begin{table}[ht]
\centering
\begin{tabular}{@{}lccc@{}}
\toprule
\textbf{Borough} & \textbf{EMS Stations} & \textbf{Hospitals} & \textbf{Total} \\ \midrule
Bronx            & 19                    & 12                 & 31             \\
Brooklyn         & 76                    & 15                 & 91             \\
Manhattan        & 28                    & 23                 & 51             \\
Queens           & 15                    & 10                 & 25             \\
Staten Island    & 23                    & 6                 & 29             \\ \midrule
\textbf{Total NYC} & \textbf{161}        & \textbf{66}       & \textbf{227}   \\ \bottomrule
\end{tabular}
\caption{Number of EMS Stations and Hospitals by boroughs in NYC}
\label{table:hospitals_and_ems_stations}
\end{table}

\subsubsection{Road network}
\begin{figure}[h!]
    \centering
    \includegraphics[width=\textwidth]{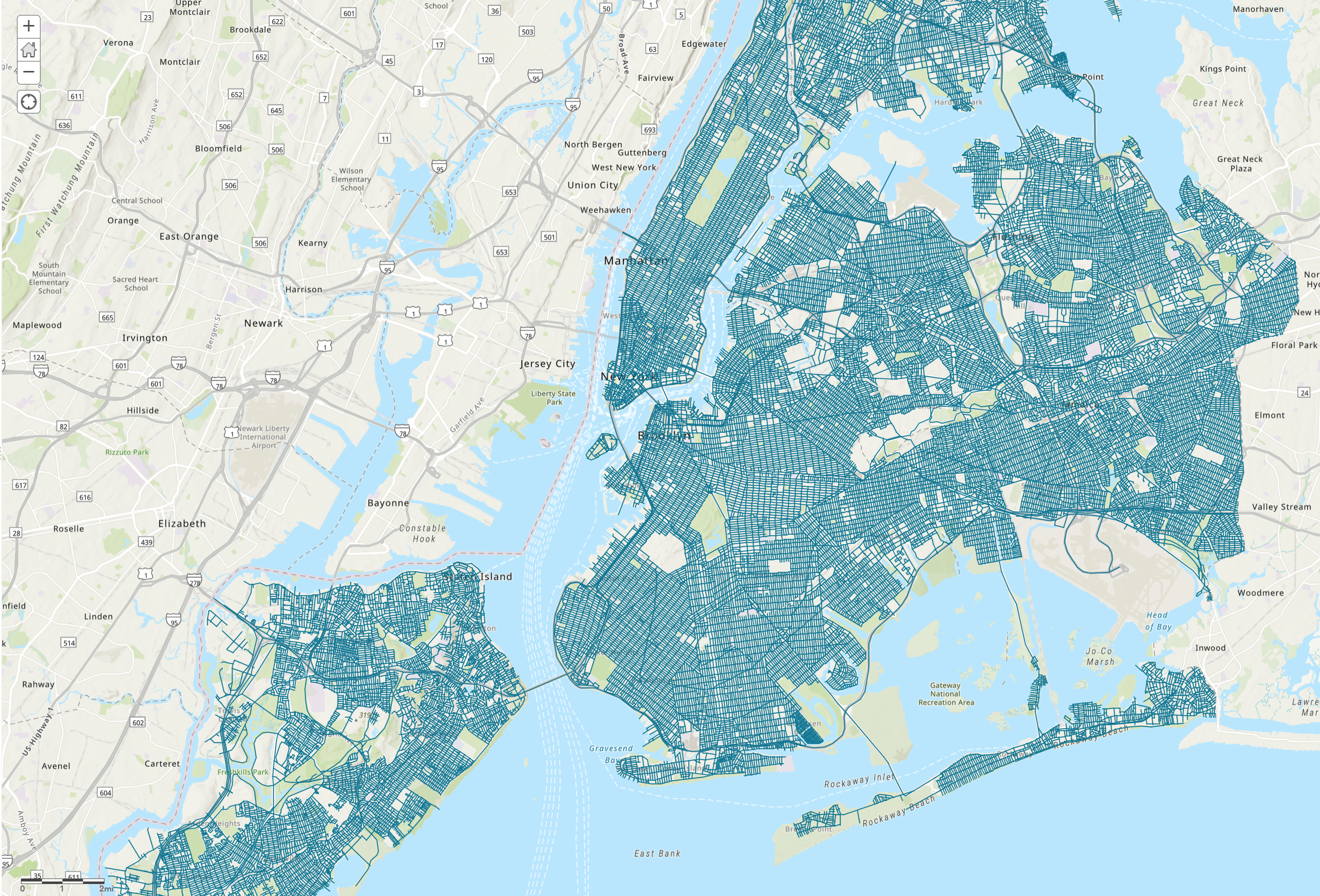} 
    \caption{All NYC streets following NYC Street Centerline.}
    \label{fig:nyc_street_centerline}
\end{figure}

OpenStreetMap (OSM) is employed to visualize the streets and intersections of New York City. The road network is represented as a symmetric multigraph, capturing the intricate connections and pathways characteristic of the urban transportation infrastructure. To ensure precision and clarity in the representation, the NYC Street Centerline (CSCL) dataset~\cite{nyc_cscl} is utilized.  An overview of the network is illustrated in Fig.~\ref{fig:nyc_street_centerline}. This representation serves as the foundation for analyzing the accessibility of EMS site locations, with travel times along the network serving as a key metric for reachability and service efficiency. By employing this approach, the study offers a realistic depiction of urban mobility and its implications for emergency response performance.

\subsubsection{Signalized intersections}
\begin{figure}[h!]
    \centering
    \includegraphics[width=\textwidth]{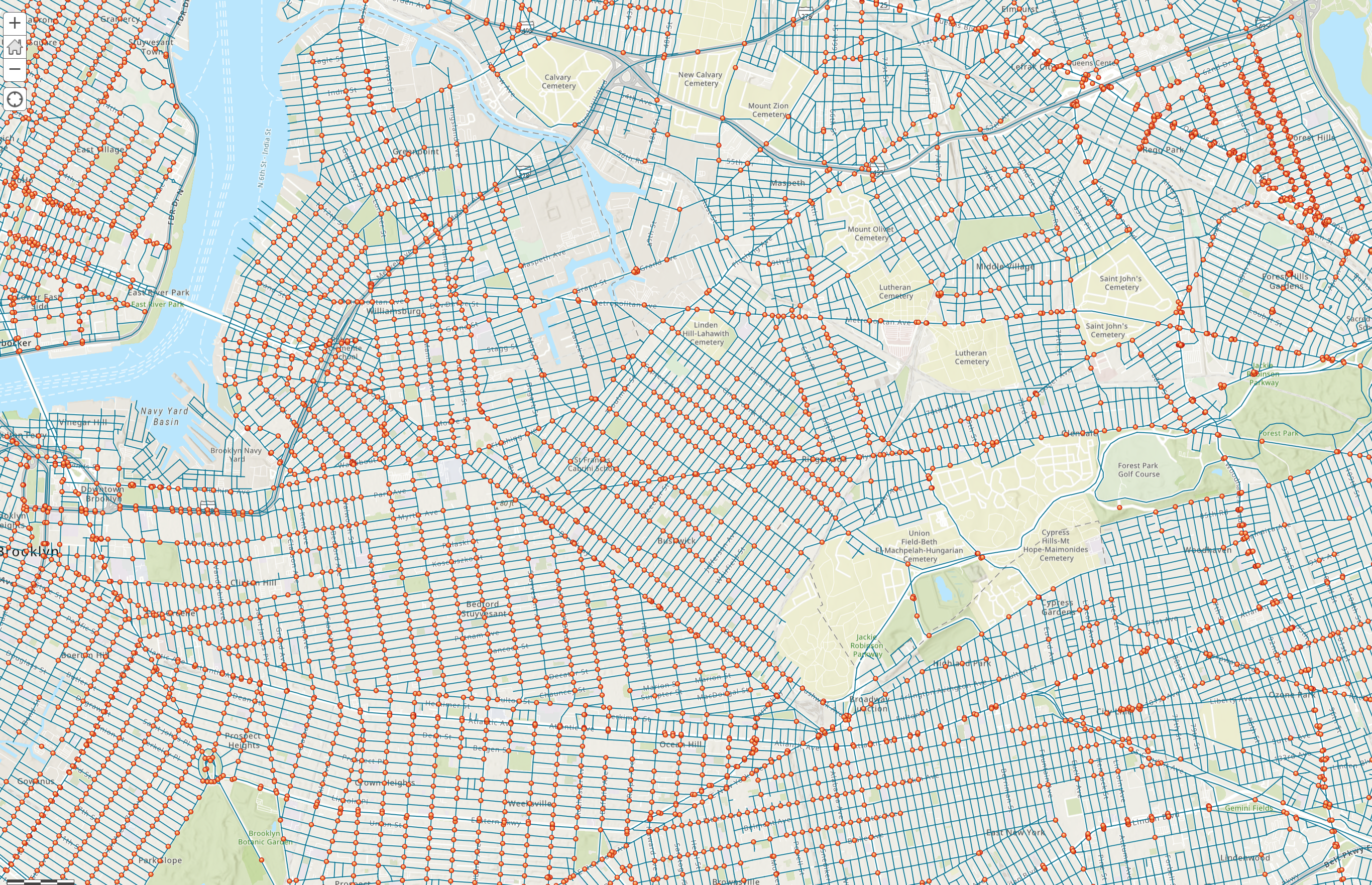}
    \caption{Signalized intersections layout in Brooklyn and Queens}
\label{fig:signalized_intersections}
\end{figure}

Currently, there is no publicly available open-source dataset that comprehensively identifies all signalized intersections in New York City. However, we have transformed the NYC Street Centerline (CSCL) shapefile to generate an approximate map of signalized intersection locations. Following pre-processing of the CSCL data, we filtered the dataset to include only road types classified as streets or highways and restricted traffic directions to multi-way configurations. Intersection points were derived by extracting line endpoints from the filtered road geometries. To address precision errors and reduce overlap caused by minor positional discrepancies, the coordinates of these endpoints were rounded. Duplicate intersection points were subsequently removed, resulting in a refined and clean dataset of potential signalized intersection locations. Refer to Fig.~\ref{fig:signalized_intersections} for an illustrative visualization of the signalized intersection map in New York City, where blue lines represent streets or highways as part of the network in Subsec.~\ref{subsubsec:graph} and orange dots represent traffic lights.

Given that each signalized intersection can function as an agent within a large-scale TSC system such as \textit{EMVLight}, we regard this map as a realistic near-future representation of the network for facilitating EMV passage in NYC.\footnote{\hs{This intersections map is available at: \url{https://zenodo.org/records/14280428}}}

\subsubsection{NYC census data}
\begin{figure}[h!]
    \centering
    \includegraphics[width=\textwidth]{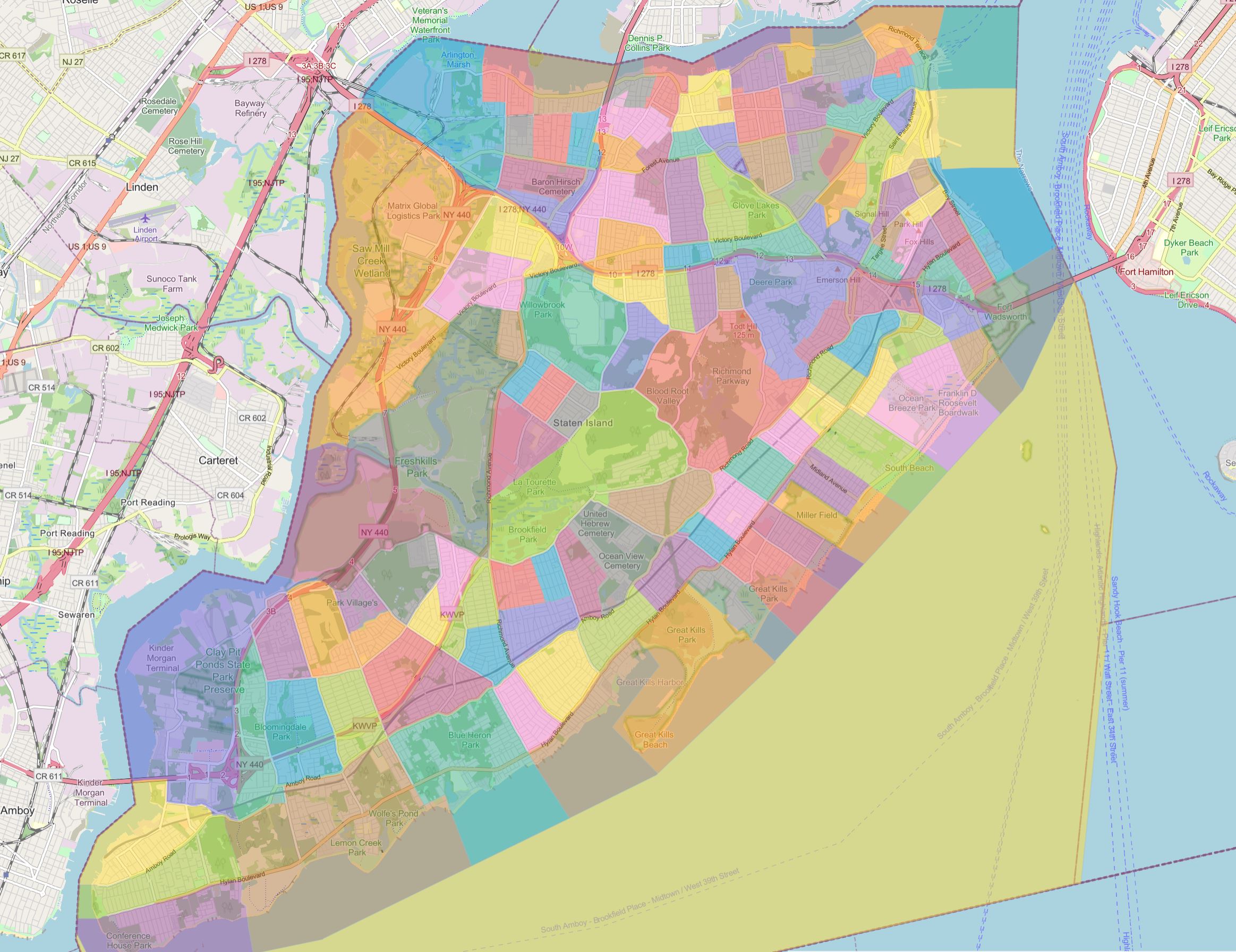} 
    \caption{Census tracts in Staten Island. Each polygon represents a census tract.}
    \label{fig:staten_island_census_tracts}
\end{figure}

\begin{figure}[h!]
    \centering
    \includegraphics[width=\textwidth]{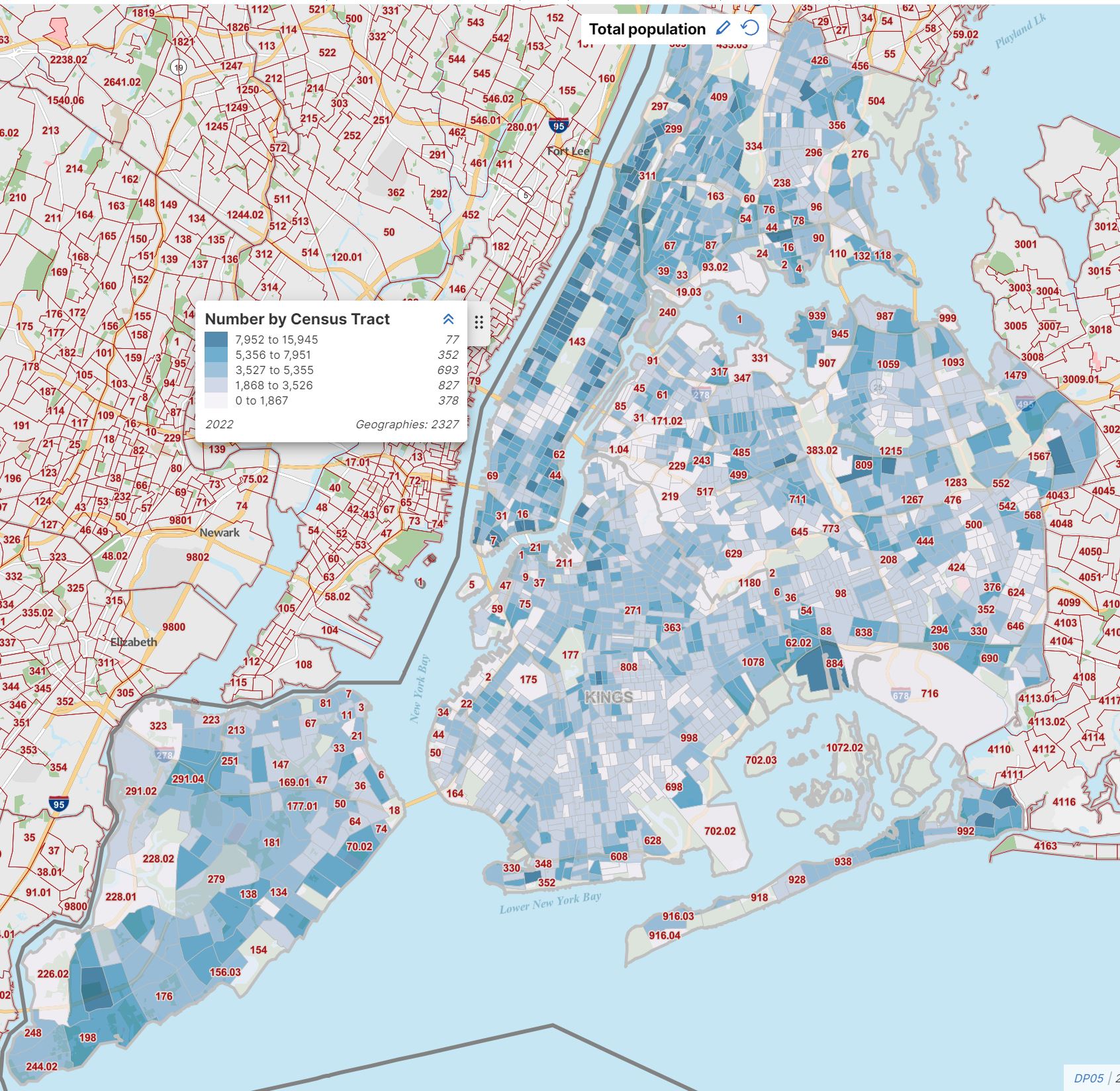} 
    \caption{Population distribution of NYC census tracts.}
    \label{fig:total_population}
\end{figure}

Census data plays a pivotal role in conducting analyses of accessibility and vulnerability. While \hs{Chung ~\cite{chung2024access}} employs a five-year average for such assessments, this study utilizes 2023 population data at the census tract level provided by the U.S. Census Bureau~\cite{bureau2023census}, thereby enhancing temporal precision. The spatial configurations of census tracts are sourced from the U.S. Census Bureau's dataset~\cite{us_census_bureau_census_tracts_2023}, with a representative example for Staten Island illustrated in Fig.~\ref{fig:staten_island_census_tracts}. 
The demographic data are DP05 from American Community Survey (ACS) Demographic and Housing Estimate~\cite{acs2021demographic} 2022 5-year estimate data profiles. In subsequent sections of this study, we will undertake a detailed examination of the spatial distribution of various demographic characteristics across New York City. This analysis aims to provide deeper insights into issues surrounding EMS accessibility. To provide an overview of the census data, a population density map illustrating the spatial distribution of the population is presented in Fig.~\ref{fig:total_population}.

\subsection{Intersection-aware EMS Accessibility Model}\label{subsec:ems_model}
\hs{Accessibility is commonly defined as the ease of reaching desired destinations, with measures classified into infrastructure-based, location-based, person-based, and utility-based categories~\cite{geurs2004accessibility}. Among these, location-based measures, particularly travel time within a predefined threshold, are widely used for their simplicity and relevance to EMS planning. Travel time thresholds, such as the NFPA 1710 standard of \(\tau = 4\) minutes~\cite{NFPA1710}, serve as benchmarks for evaluating whether populations can access critical EMS facilities.

This study adopts a travel time-based measure of accessibility, incorporating intersection density to account for delays in densely connected urban areas. Building on the methodology proposed by Chung et al.~\cite{chung2024access}, the following subsections outline the model’s graph representation, travel time assignment, and population weighting framework, providing a refined evaluation of EMS accessibility in NYC.}
\subsubsection{Graph representation}\label{subsubsec:graph}
The road network is represented as a graph \( G = (V, E) \), where \( V \) is the set of nodes, each corresponding to either a road intersection or endpoint, and \( E \) denotes the set of edges, each representing a distinct road segment connecting two nodes. Attributes associated with each edge, such as road length and speed limit, facilitate the initial computation of travel time across the network under idealized conditions.
\subsubsection{Travel time assignment}
To capture the influence of urban intersection density on travel times, an intersection density metric \( I(v) \) is defined for each node \( v \in V \). This metric is calculated as the number of intersections within a fixed radius \( r \) around the node, normalized by the area of the radius, such that:
\begin{equation}
    I(v) = \frac{\text{Number of intersections within radius } r}{\pi r^2}.
\end{equation}
A higher \( I(v) \) value implies a denser concentration of intersections, which is expected to contribute to travel delays, particularly affecting emergency response times in congested urban zones. Based on Leadership in Energy and Environmental Design (LEED)~\cite{LEED_IntersectionDensity}, we adopt \( r = 800 \, \mathrm{m} \) for intersection density evaluations.

To compute travel times in the road network, we follow the graph representation introduced in Subsec.~\ref{subsubsec:graph}. Each edge \( e \in E \) is assigned a baseline travel time \( T(e) \), calculated as the edge, i.e., road segment, length divided by the speed limit. The default speed limit on most New York City streets is \( 25 \, \mathrm{mph} \), implemented under the Vision Zero initiative~\cite{nyc_speed_limit} to enhance traffic safety. On highways and parkways, the limit is generally higher, ranging from \( 50 \, \mathrm{mph} \) to \( 55 \, \mathrm{mph} \) depending on signage. The baseline travel time matrix \( T \) is then defined as:
\begin{equation}
    T_{ij} = \text{Shortest path travel time from node } v_i \text{ to node } v_j,
\end{equation}
where \( T_{ij} \) represents the cumulative travel time across all edges along the shortest path between nodes \( v_i \) and \( v_j \), under traffic-free conditions (no delays). 

To account for delays caused by intersection density, we adjust the travel time for each edge \( e \) using a node-based intersection delay design. The delay for an edge \( e \) connecting nodes \( v_m \) and \( v_n \) is modeled as:
\begin{equation}
    D_\text{intersection}(e) = \alpha \cdot \frac{I(v_m) + I(v_n)}{2},
\end{equation}
where \( I(v_m) \) and \( I(v_n) \) are the intersection densities of the nodes at the two ends of the edge \( e \), and \( \alpha \) is a calibration factor reflecting the impact of intersection density on delays, in units of [minutes $\cdot$ square meters per intersection]. This design ensures that travel times accurately capture the influence of densely connected nodes at the start and end of each road segment, aligning with the principle that congestion delays are more pronounced in highly interconnected regions.

The adjusted travel time for each edge is then given by:
\begin{equation}
    T'(e) = T(e) + D_\text{intersection}(e),
\end{equation}
and the corresponding adjusted travel time matrix \( T' \) is:
\begin{equation}
    T'_{ij} = \text{Shortest path travel time from node } v_i \text{ to node } v_j \text{ incorporating intersection delays.}
\end{equation}

A node \( v_i \) is considered accessible to/from an EMS site (EMS station or hospital) \( s_j \) if:
\begin{equation}
    T'_{ij} \leq \tau,
\end{equation}
where \( \tau \) is a benchmark travel time, typically set as \( 4 \, \mathrm{minutes} \), aligned with the National Fire Protection Association standards~\cite{NFPA1710}.

By integrating both road characteristics and node-based intersection delays, this framework offers a more realistic evaluation of travel times, enabling the identification of vulnerable regions where response times exceed the benchmark \( \tau \). While the model operates under the assumption that the EMV can travel unimpeded between nodes—an idealization that does not fully align with real-world conditions—it effectively captures the concept of accessibility and accounts for intersection delays along the route, providing valuable insights regarding EMS accessibility.

\subsubsection{Population assignment}\label{subsubsec:pop_assignment}
To accurately gauge the number of residents affected by EMS accessibility, population assignment is also incorporated extending the idea proposed by \cite{chung2024access}. Each node \( v \in V \) is assigned a population density by intersecting census tracts with node-based Voronoi polygons\cite{burrough20158}. Census tracts provide demographic data, including the population count \( P(c) \) and livable area \( L(c) \). 

For simplicity, the livable area of each census tract in our study is considered synonymous with its total area, which can be acquired directly from \cite{bureau2023census}.
The effective population density \( \rho^*(c) \) is defined as:
\begin{equation}
    \rho^*(c) = \frac{P(c)}{L(c)}.
\end{equation}
See Table~\ref{tab:population_density_by_borough} for total area, population and population density by each borough.
\begin{table}[ht]
    \centering
    \renewcommand{\arraystretch}{1.2} % Adjust row spacing
    \setlength{\tabcolsep}{10pt} % Adjust column spacing
    \begin{tabular}{lccc}
        \toprule
        \textbf{Borough} & \textbf{$L(c)$ [mi$^2$]} & \textbf{$P(c)$ [M]} & \textbf{$\rho^*(c)$ [k ppl/mi$^2$]} \\
        \midrule
        Bronx          & 42.2  & 1.42 & 33.65 \\
        Brooklyn       & 69.4  & 2.57 & 37.04 \\
        Manhattan      & 22.7  & 1.63 & 71.81 \\
        Queens         & 108.7 & 2.27 & 20.88 \\
        Staten Island  & 57.5  & 0.47 & 8.17  \\
        \midrule
        \textbf{Total NYC} & 300.5 & 8.36 & 27.82 \\
        \bottomrule
    \end{tabular}
    \caption{Total area, population and population density for boroughs and NYC.}
    \label{tab:population_density_by_borough}
\end{table}

The Voronoi region \( R(v) \) associated with node \( v \) defines the area influenced by that node within the road network. The population assigned to each node \( v \), \( P(v) \), is then computed by proportionally weighting the overlap of \( R(v) \) with each census tract \( c \in C \), as follows:
\begin{equation}
    P(v) = \sum_{c \in C} \left( |R(v) \cap c| \cdot \rho^*(c) \right),
\end{equation}
where \( |R(v) \cap c| \) represents the area of overlap between the Voronoi region \( R(v) \) and census tract \( c \). This assignment ensures that each node in the network has a population weighting, allowing for the assessment of accessibility impacts across densely and sparsely populated areas.

By integrating both intersection-adjusted travel times and population density metrics, the modified EMS model provides a more granular perspective on emergency accessibility in urban environments, particularly highlighting regions where dense intersections may exacerbate response delays.

\subsubsection{Vulnerability determination}
Vulnerability to insufficient EMS coverage is assessed by comparing the shortest adjusted travel time \( T'_{i,\text{EMS}} \) from each node \( v_i \) to the nearest EMS facility against a benchmark threshold \( \tau \), following National Fire Protection Association (NFPA) guidelines\cite{NFPA1710}. Nodes with \( T'_{i,\text{EMS}} > \tau \) are deemed inaccessible.

Regions are classified as “vulnerable” if they contain clusters of inaccessible nodes. To quantify the impact, population weights are assigned to each inaccessible node, allowing for an aggregate measure of residents underserved by EMS. This population-weighted vulnerability assessment highlights priority areas for potential TSC enhancement.

\section{Results}\label{sec:results}
In this section, we present travel times using the proposed EMS accessibility model outlined in Subsec.\ref{subsec:travel_time}. We then compare the simulated EMV travel time with the ground-truth travel time in Subsec.~\ref{subsec:comparison_w_real_world}. Vulnerable regions with respect to EMS accessibility are identified in Subsec.~\ref{subsec:vulnerable_regions}. 
\subsection{Travel Time}\label{subsec:travel_time}

Table~\ref{tab:travel_time_summary} summarizes the travel time statistics for EMS stations, hospitals, and a combined category considering EMS stations and hospitals as one type of facility providing comprehensive medical services based on the simulation of the proposed EMS accessibility model. The statistics are presented across different \(\alpha\) values, where \(\alpha\) is the intersection delay factor introduced in Subsec.~\ref{subsec:ems_model}, which represents the additional delay induced by intersection density. \(\alpha\) is expressed in unit of seconds $\cdot$ square meters for better readability.

Travel times are reported in minutes and include key percentiles (25\%, 50\%, 75\%, 97.5\%, and 100\%) alongside the average travel time for each category. The 97.5\% percentile represents the practically longest travel time while excluding a small number of extreme outliers in the complete simulated travel time distribution. 

Travel time to EMS stations are typically shorter due to the higher density and wider distribution of EMS stations compared to hospitals. Hospitals, although fewer in number, extend the range of travel times because of their greater average distance in urban areas. Consequently, the overall travel time to the nearest emergency medical facility is slightly shorter than that to the nearest EMS station, as the inclusion of hospitals increases the number of accessible options.
\begin{table}[ht]
    \centering
    \renewcommand{\arraystretch}{1.2} 
    \setlength{\tabcolsep}{5pt} 
    \begin{tabular}{lccccccc}
        \toprule
        \textbf{Category} & \(\boldsymbol{\alpha (\text{s}\cdot\text{m}^2)}\) & \textbf{25\%} & \textbf{50\%} & \textbf{75\%} & \textbf{97.5\%} & \textbf{100\%} & \textbf{Average} \\
        \midrule
        EMS Stations & 0   & 2.15 & 3.10 & 4.18 & 6.60 & 7.95 & 4.80 \\
                     & 5   & 2.28 & 3.23 & 4.35 & 6.85 & 8.12 & 4.97 \\
                     & 10  & 2.50 & 3.48 & 4.72 & 7.28 & 8.42 & 5.19 \\
                     & 15  & 2.95 & 3.98 & 5.12 & 7.98 & 8.93 & 5.68 \\
        \midrule
        Hospital     & 0   & 3.45 & 5.03 & 6.82 & 9.35 & 10.90 & 6.70 \\
                     & 5   & 3.68 & 5.11 & 6.93 & 9.46 & 11.15 & 6.95 \\
                     & 10  & 3.85 & 5.29 & 7.12 & 9.71 & 11.41 & 7.22 \\
                     & 15  & 4.10 & 5.81 & 7.60 & 10.32 & 11.92 & 7.50 \\
        \midrule
        Overall      & 0   & 2.07 & 2.98 & 3.92 & 6.03 & 7.22 & 3.92 \\
             & 5   & 2.18 & 3.13 & 4.08 & 6.12 & 7.34 & 4.09 \\
             & 10  & 2.35 & 3.42 & 4.47 & 6.48 & 7.52 & 4.19 \\
             & 15  & 2.73 & 3.86 & 4.99 & 7.28 & 8.38 & 4.41 \\
        \bottomrule
    \end{tabular}
    \caption{Summary of travel time [mins] for EMS stations, hospitals, and combined (Overall) given different \(\alpha\) values.}
    \label{tab:travel_time_summary}
\end{table}

\begin{figure}[h]
    \centering
    \includegraphics[width=\textwidth]{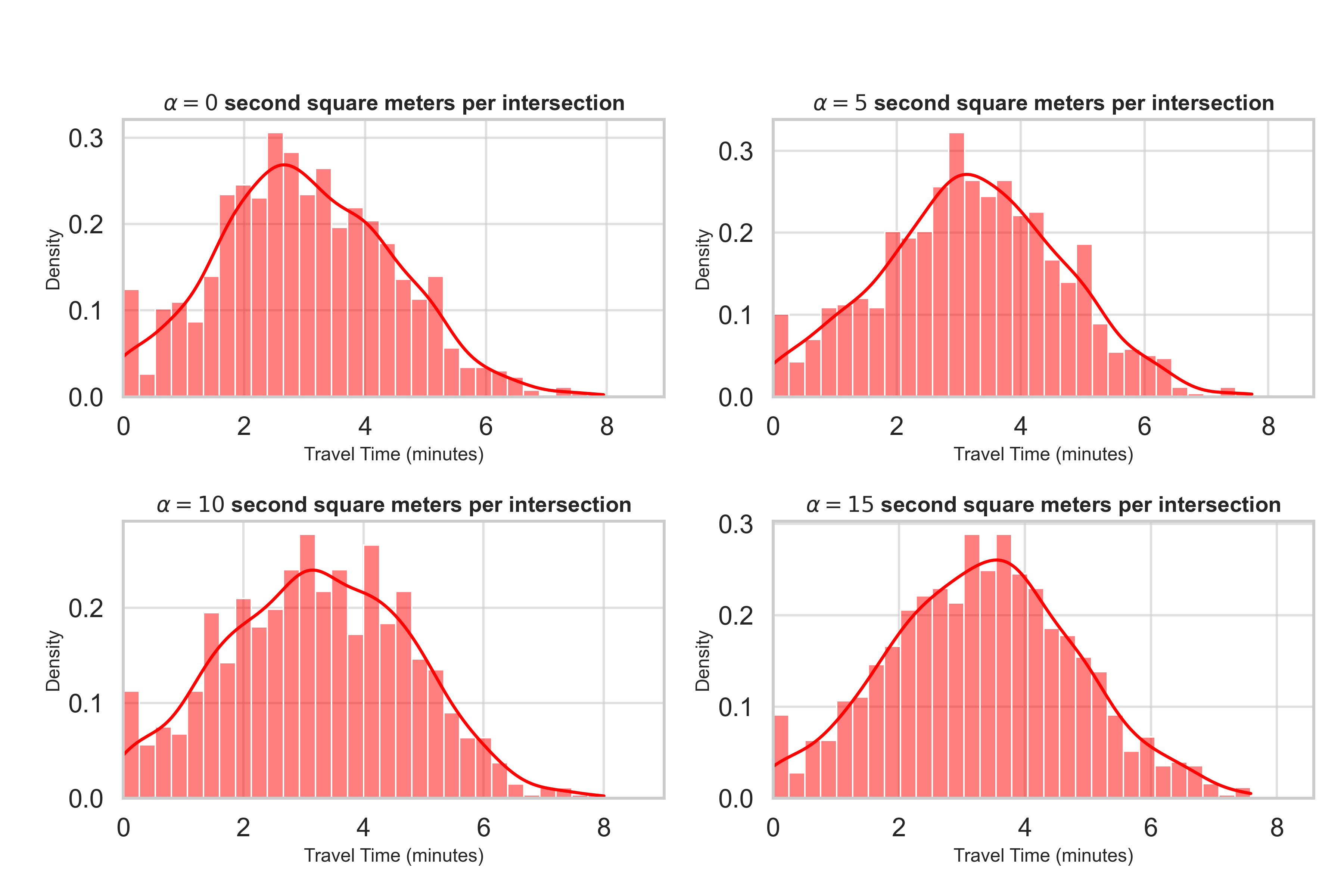}
    \caption{Travel time distributions for EMS stations across different \(\alpha\) values.}
    \label{fig:ems_travel_time}
\end{figure}

\begin{figure}[h]
    \centering
    \includegraphics[width=\textwidth]{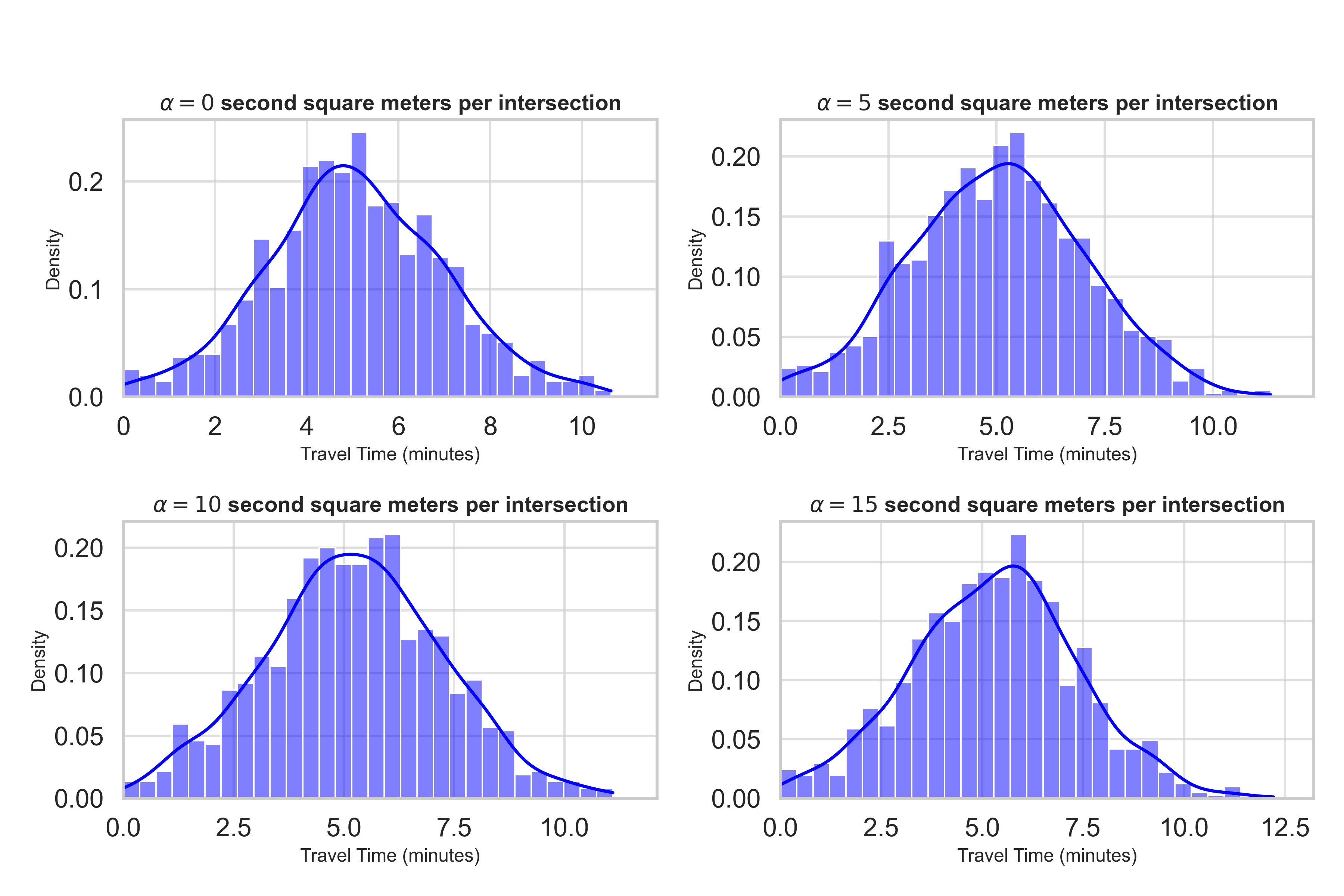}
    \caption{Travel time distributions for Hospital across different \(\alpha\) values.}
    \label{fig:hospital_travel_time}
\end{figure}

Figures~\ref{fig:ems_travel_time} and~\ref{fig:hospital_travel_time} illustrate the travel time distributions for EMS stations and hospital, respectively, under different \(\alpha\) values. Each figure consists of subplots representing travel time distributions for \(\alpha = 0, 5, 10,\) and \(15\) $\text{s}\cdot\text{m}^2$. The distributions show an increasing pattern in travel times as \(\alpha\) grows, reflecting the impact of intersection-induced delays. 

\subsection{Calibration of \texorpdfstring{\boldmath$\alpha$}{alpha}}\label{subsec:comparison_w_real_world}
To validate the correctness of the proposed model’s predicted travel time, which accounts for delays due to intersections, we summarize the EMV travel times recorded in the 911 NYC End-to-End dataset~\cite{NYC911Data}. These travel times are derived from the time difference between agency dispatch and agency arrival and span the period from the week of November 18, 2023, to October 21, 2024.
\begin{figure}[h!]
    \centering
    \includegraphics[width=\textwidth]{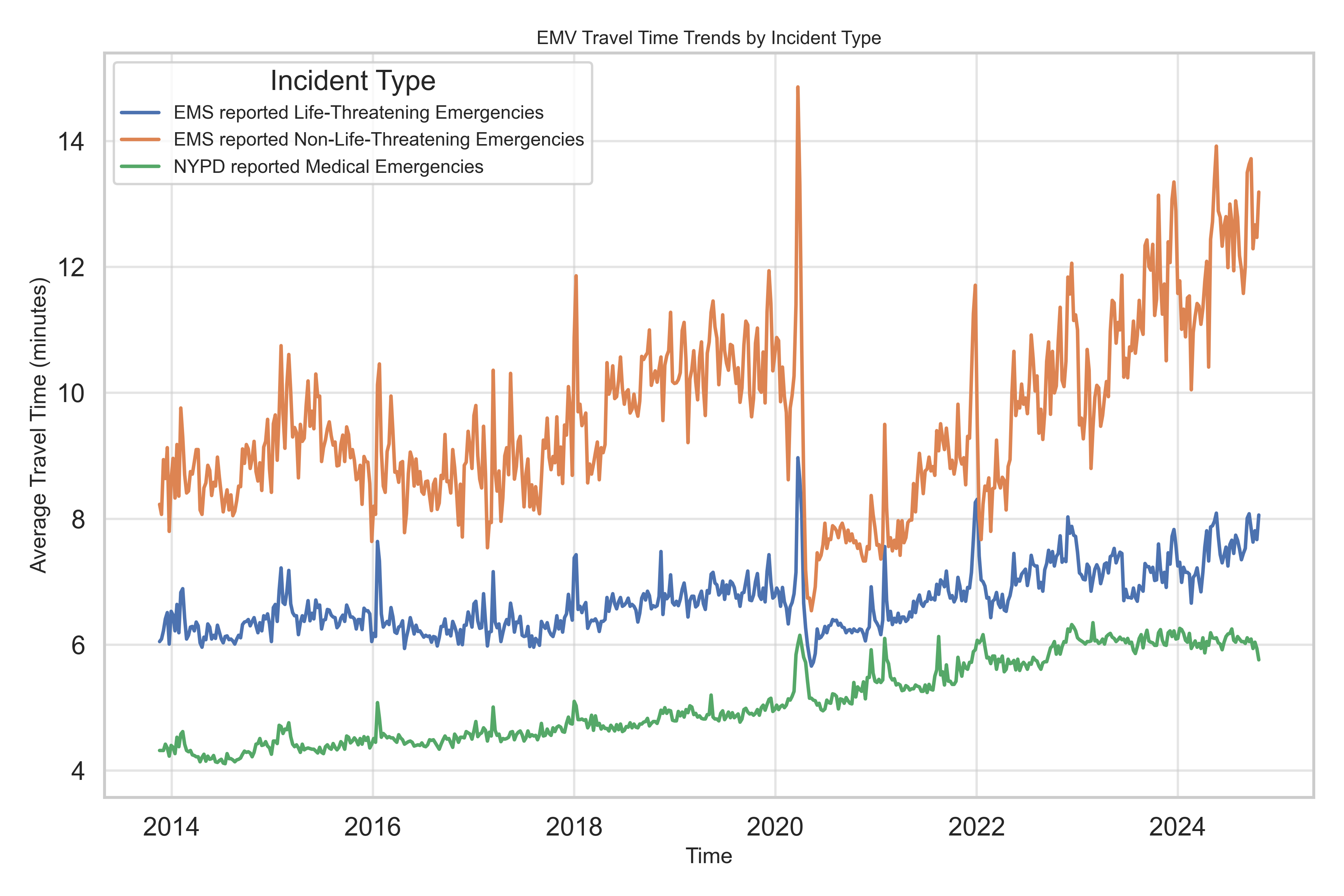}
    \caption{EMV travel time trend based on 911 NYC End-to-End.}
    \label{fig:emv_travel_time_trend}
\end{figure}
For reference, Fig.~\ref{fig:emv_travel_time_trend} illustrates the trend of EMV travel time over the past decade, revealing a continuous increase in EMV travel time for NYPD-reported medical emergencies throughout the study period. Notably, EMV travel time for both life-threatening and non-life-threatening emergencies experienced an abrupt decline during the outbreak of Covid-19 in the city due to shelter-in-place, followed by a gradual rise in the post-Covid-19 period.
\begin{table}
    \centering
    \renewcommand{\arraystretch}{1.2}
    \setlength{\tabcolsep}{6pt} 
    \begin{tabular}{lcc}
        \toprule
        \textbf{} & \textbf{$T_{actual}$ (minutes)} & \textbf{$T_{actual}/T_{simulated}$}\\
        \midrule
        25\% & 6.39 & 2.34\\
        50\% & 7.62 & 1.92\\
        75\% & 9.40 & 1.89\\
        97.5\% & 12.34 & 1.69\\
        100\% & 14.86 & 1.77\\
        Average & 7.96 & 1.80\\
        \bottomrule
    \end{tabular}
    \caption{Statistics for actual travel time ($T_{actual}$) based on 911 End-to-End reports. The ratio $T_{actual}/T_{simulated}$ compares actual travel time to simulated travel time for overall medical service when $\alpha = 15\,\text{s}\cdot\text{m}^2$.}
    \label{tab:real_statistics_travel_time}
\end{table}

\hs{
If we calculate the statistics of the actual EMV travel time $T_{actual}$ for all medical incidents and compare it with the simulated travel time $T_{simulated}$ when $\alpha = 15\,\text{s}\cdot\text{m}^2$, as shown in Table~\ref{tab:real_statistics_travel_time}, we observe that the simulated travel times in Table~\ref{tab:travel_time_summary} tend to underestimate the actual travel times. This observation aligns with the findings of \cite{chung2024access}, as both studies assume that EMVs always travel at the speed limit between intersections. Furthermore, we note that for shorter trips, the ratio $T_{actual}/T_{simulated}$ tends to be larger compared to longer trips. This can be attributed to intersection delay factors, which play a proportionally greater role in shorter trips, whereas the simulation-to-reality gap narrows for longer trips due to the influence of extended road segments.

While it may be tempting to scale the intersection delay factor $\alpha$ by the average underestimation ratio ($1.8\times 15 = 27\,\text{s}\cdot\text{m}^2$), this approach oversimplifies the underlying dynamics of EMV travel times. Firstly, attributing the majority of travel time to intersection delay overlooks the fact that urban travel times are influenced by numerous interacting factors, such as traffic congestion, vehicle acceleration and deceleration, driver behavior, and road geometry~\cite{zhang2007intersection}. These factors are not uniformly distributed across all trips and vary based on location, time of day, and road conditions. Assigning a higher $\alpha$ uniformly would disproportionately amplify the role of intersections and fail to capture these complexities. For instance, segments with fewer intersections but heavy congestion would still exhibit significant delays, which a simple scaling approach cannot account for.

Secondly, the model assumes that intersection delay is additive rather than multiplicative, meaning the travel time for a road segment is modeled as the sum of its free-flow travel time and an intersection delay component. Scaling $\alpha$ directly would imply a shift in this underlying assumption, potentially leading to inconsistent results. Increasing $\alpha$ would exaggerate delays at intersections without considering the diminishing marginal impact of additional intersections, especially in areas where traffic congestion or coordinated signals mitigate delay. This would likely result in unrealistic travel time predictions in regions with high intersection densities but well-managed traffic flow.

Finally, experiments conducted on the $\text{Synthetic Grid}_{5\times5}$ map in Section~\ref{subsec:improvement} demonstrate that $\alpha = 15\,\text{s}\cdot\text{m}^2$ aligns closely with observed delays in scenarios lacking pre-emption. This consistency underscores that $\alpha = 15\,\text{s}\cdot\text{m}^2$ provides a reasonable baseline for evaluating EMS accessibility without overestimating intersection effects. While the model underestimates absolute travel times due to idealized assumptions, $\alpha = 15\,\text{s}\cdot\text{m}^2$ strikes a balance, effectively capturing relative differences in EMS accessibility and identifying vulnerable regions.
}

\subsection{Vulnerable regions}\label{subsec:vulnerable_regions}
\begin{figure}[h]
    \centering
    \includegraphics[width=\textwidth]{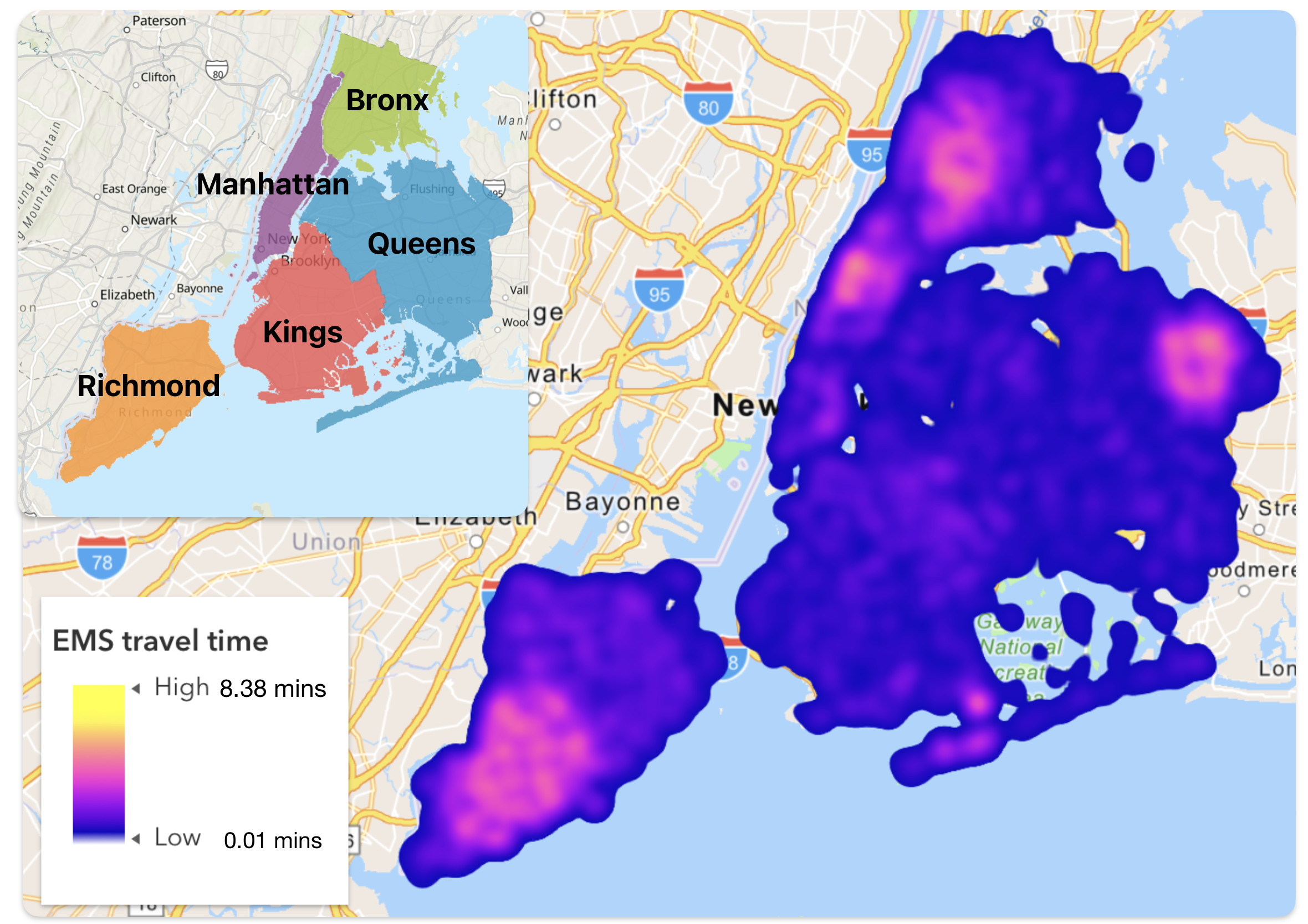} 
    \caption{EMS travel time heatmap, with heated regions highlight\hs{ing} the vulnerable regions.}
    \label{fig:vulnerable_regions}
\end{figure}

Based on the simulated travel time in Subsec.~\ref{subsec:travel_time} for the overall EMS accessibility with $\alpha = 15\text{s}\cdot\text{m}^2$, we here highlight the EMS accessibility vulnerable regions in NYC, see Fig.~\ref{fig:vulnerable_regions}.

We identified four regions within New York City that are particularly vulnerable in terms of EMS accessibility, all exhibiting average EMS travel times exceeding the benchmark of $\tau = 4 \, \text{minutes}$. First, the southern and western parts of Staten Island are characterized by sparse population distribution and a suburban layout with a limited number of medical facilities. This geographic configuration leads to prolonged response times and reduced accessibility for residents. Second, the outer ring of Queens faces a dual challenge of low medical facility density and relatively high intersection density. These geographically expansive areas experience significant gaps in EMS coverage, particularly within densely populated residential neighborhoods. Third, the Upper West Side of Manhattan, a densely populated residential neighborhood, is marked by a dense network of intersections and frequent traffic congestion. These conditions pose substantial barriers to efficient EMS response, further highlighting the region's accessibility challenges. Fourth, large areas of the Bronx, which lack sufficient medical facilities, also suffer from inadequate service coverage. Additionally, minor vulnerable regions with average EMS travel times around $\tau = 4 \, \text{minutes}$ are recognized, such as the Rockaway Peninsula. This area, being geographically distant from the city center, faces challenges in accessing general medical services.
\section{Discussion}
\label{sec:discussion_access}
In this section, we first conduct a sensitivity study on the value of $\tau$ in Subsec.~\ref{subsec:tau}. We examine accessibility' with respect to demographic factors in Subsec.~\ref{subsec:demographic_analysis}, and then further investigate \textit{EMVLight}'s potential to facilitate these EMS accessibility vulnerable area in Subsec.~\ref{subsec:improvement}.

\subsection{Feasibility of \texorpdfstring{$\tau$}{tau}}\label{subsec:tau}
\begin{figure}[ht]
    \centering
    \includegraphics[width=0.75\textwidth]{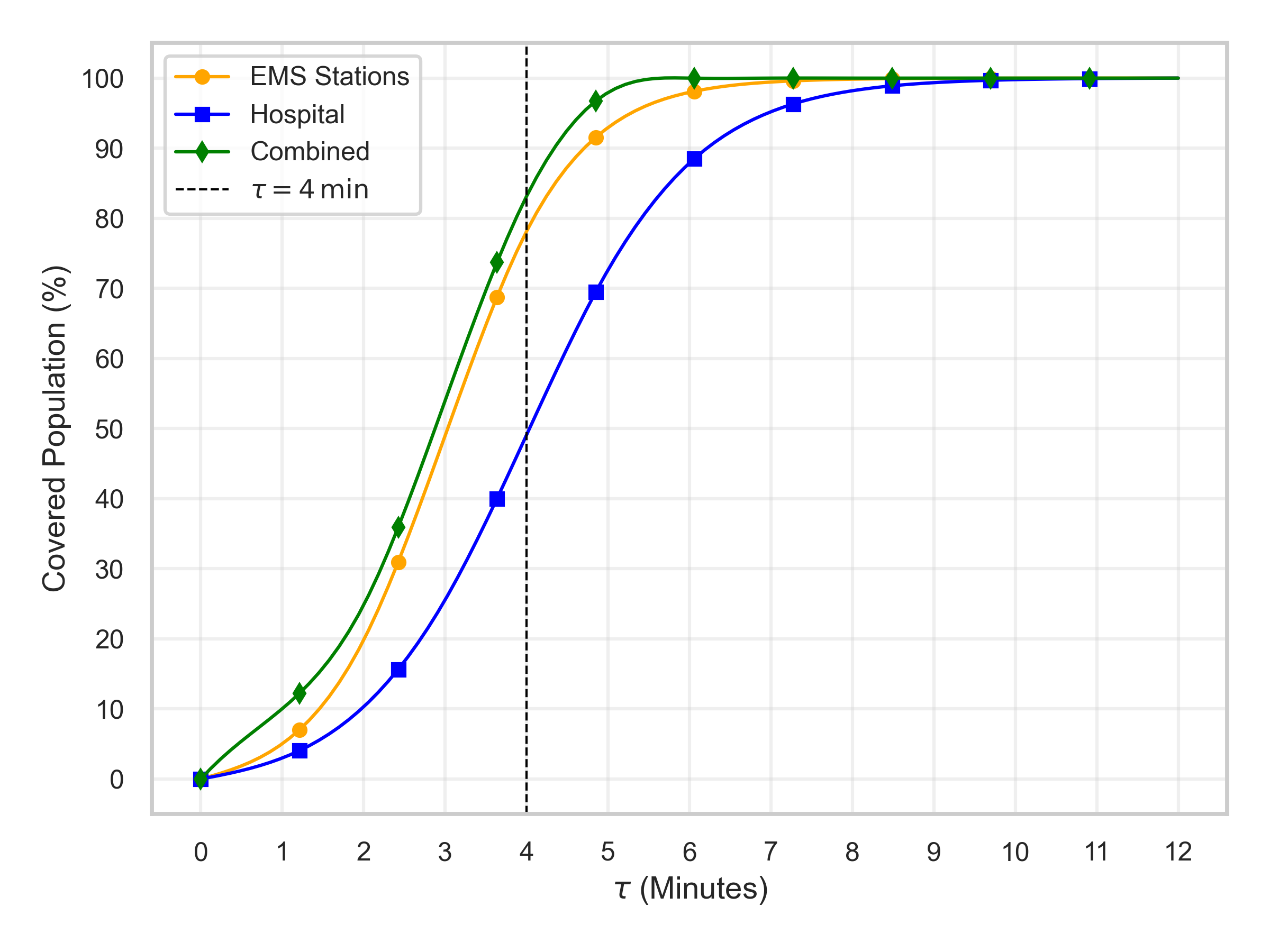} 
    \caption{Population covered by EMS accessibility as a function of $\tau$.}
    \label{fig:covered_func}
\end{figure}

Even though $\tau = 4 \, \text{minutes}$ is suggested by NFPA 1710 standards~\cite{NFPA1710} and widely adopted, its feasibility in NYC remains uncertain. Using the population assignment framework proposed in Subsubsec.~\ref{subsubsec:pop_assignment}, we evaluate the percentage of the population covered by EMS accessibility as a function of benchmark travel time $\tau$. Fig.~\ref{fig:covered_func} illustrates the accessibility of emergency services (EMS stations, hospitals, and combined) as a function of $\tau$. The results indicate a rapid increase in population coverage for smaller $\tau$, followed by a plateau as $\tau$ increases. At $\tau = 4 \, \text{minutes}$, EMS stations cover approximately 80\% of the population, aligning with NFPA 1710 standards, while hospital coverage remains lower, at around 50\%.

This disparity underscores the critical role of EMS stations in ensuring rapid response and highlights the need for infrastructure improvements to enhance hospital accessibility. Beyond $\tau = 4 \, \text{minutes}$, additional gains in coverage diminish, making $\tau = 4 \, \text{minutes}$ a practical benchmark for EMS station accessibility. When EMS stations and hospitals are considered jointly, 83\% of the population can be reached within 4 minutes. However, hospitals, with greater capacity to handle critical emergencies, remain indispensable for specific medical situations. \hs{The current allocation of hospitals suggests that a benchmark travel time of $\tau = 5.5 \, \text{minutes}$ is more realistic for hospital accessibility in NYC, which exceeds the NFPA 1710 requirement of 4 minutes. This finding highlights the urgent need for additional hospital locations to ensure equitable and timely access to critical medical care.}

\subsection{Demographic analysis}\label{subsec:demographic_analysis}
To explore correlations and extend vulnerability analysis to socio-economic dimensions, we compare the identified vulnerable regions with demographic distributions, such as age and median income, derived from the U.S. Census~\cite{bureau2023census}. Figures are available in Appendix~\ref{appendix_access}.
\subsubsection{Age}
The demographic distribution of NYC reveals significant variations in age structure across census tracts. Fig.~\ref{fig:65+_distribution} highlights that the 65+ population is concentrated in Staten Island and southern Brooklyn, where several census tracts have over 1,700 elderly residents. In contrast, the Bronx and parts of northern Manhattan have notably lower concentrations, with many tracts reporting fewer than 338 individuals aged 65+. Fig.~\ref{fig:median_age_distribution}, depicting median age, complements this observation, showing that Staten Island and southern Brooklyn exhibit the highest median ages (above 50 years), indicative of older populations. Conversely, the Bronx and northern Brooklyn demonstrate significantly younger median ages, often below 30 years. This spatial dichotomy underscores that demographic aging in Staten Island and southern Brooklyn renders these areas more medically vulnerable, as they are likely to demand greater age-specific medical services.

Overlaying the identified EMS-vulnerable regions with demographic data, particularly the distribution of the population aged 65 and over, reveals several critical insights. The Upper West Side of Manhattan stands out with a notably high concentration of elderly residents, a demographic that is especially reliant on timely emergency medical services. This significant proportion of older adults, combined with the area's existing EMS accessibility challenges, amplifies the region’s vulnerability. Similarly, the peripheral areas of Queens exhibit comparable risks, where limited medical infrastructure coincides with a growing elderly population. Staten Island further highlights this intersection of demographic and geographic challenges, particularly in its southern and western census tracts, which feature a high ratio of residents aged 65 and older. As Staten Island is broadly considered an EMS accessibility desert, the substantial presence of seniors in these areas exacerbates the already critical gaps in emergency medical response. These findings underscore the urgent need for targeted interventions to improve EMS accessibility in regions with a high concentration of vulnerable elderly populations.

\subsubsection{Median Income}
We present maps highlighting pronounced socioeconomic disparities across NYC, with Fig.~\ref{fig:median_income_distribution} depicting the median income per census tract and Fig.~\ref{fig:poverty_percentage_distribution} illustrating the poverty population ratio per census tract. High-income areas, primarily located in Manhattan (e.g., Upper East Side, Midtown), exhibit median incomes exceeding \$99,680, which align with low poverty ratios (below 10.8\%). Conversely, the Bronx stands out as the most economically vulnerable borough, characterized by median incomes below \$34,450 and poverty rates exceeding 34.1\%, followed by economically distressed neighborhoods in Brooklyn, such as East New York and Brownsville. This inverse correlation between income and poverty highlights the potential larger medical needs in these underserved areas, particularly in the Bronx and select Brooklyn neighborhoods.

Overlaying the identified EMS-vulnerable regions with median income distribution data reveals critical socioeconomic disparities impacting EMS accessibility. Notably, the peripheral areas of Queens exhibit higher poverty rates compared to surrounding neighborhoods, intensifying the challenges posed by limited medical infrastructure and relatively high intersection densities. In Staten Island, particularly the southern and western regions, residents experience relatively low income levels, compounding the area's designation as an EMS desert and further hindering timely emergency responses. Bronx registers the lowest median household income among New York City's boroughs and also exhibits the most severe EMS accessibility challenge.

\subsection{Improvement with EMVLight}\label{subsec:improvement}
\textit{EMVLight}~\cite{su2023emvlight} is a state-of-the-art large-scale multi-agent reinforcement learning (MARL) framework designed to optimize real-time TSC for facilitating the efficient passage of EMVs in congested road networks. By dynamically assigning the roles of pre-emption agents to signalized intersections as EMVs traverse the network, \textit{EMVLight} enables real-time shortest path navigation and coordinated signal control, ensuring seamless EMV mobility through highly congested environments.

Experimental evaluations on synthetic maps (e.g., Synthetic Grid\(_{5 \times 5}\)) 
and real-world maps (e.g., Manhattan, \(16 \times 3\); Hangzhou, \(4 \times 4\)) 
demonstrated that \textit{EMVLight} can reduce EMV travel time (\(T_{\text{EMV}}\)) 
by 60\% to 80\%, depending on whether the network is emergency-capacitated, 
compared to benchmarks without pre-emption. These results highlight the 
framework’s ability to substantially enhance EMV traversal efficiency across 
varying network configurations and densities.

To assess \textit{EMVLight}'s ability to reduce $T_{\text{EMV}}$ within the perspective of the EMS accessibility model proposed in Subsec.~\ref{subsec:ems_model}, we can study the amount of time the EMV takes passing through intersections against the benchmark to estimate the potential intersection delay reduction. 

\begin{figure}[h]
    \centering
    \includegraphics[width=\textwidth]{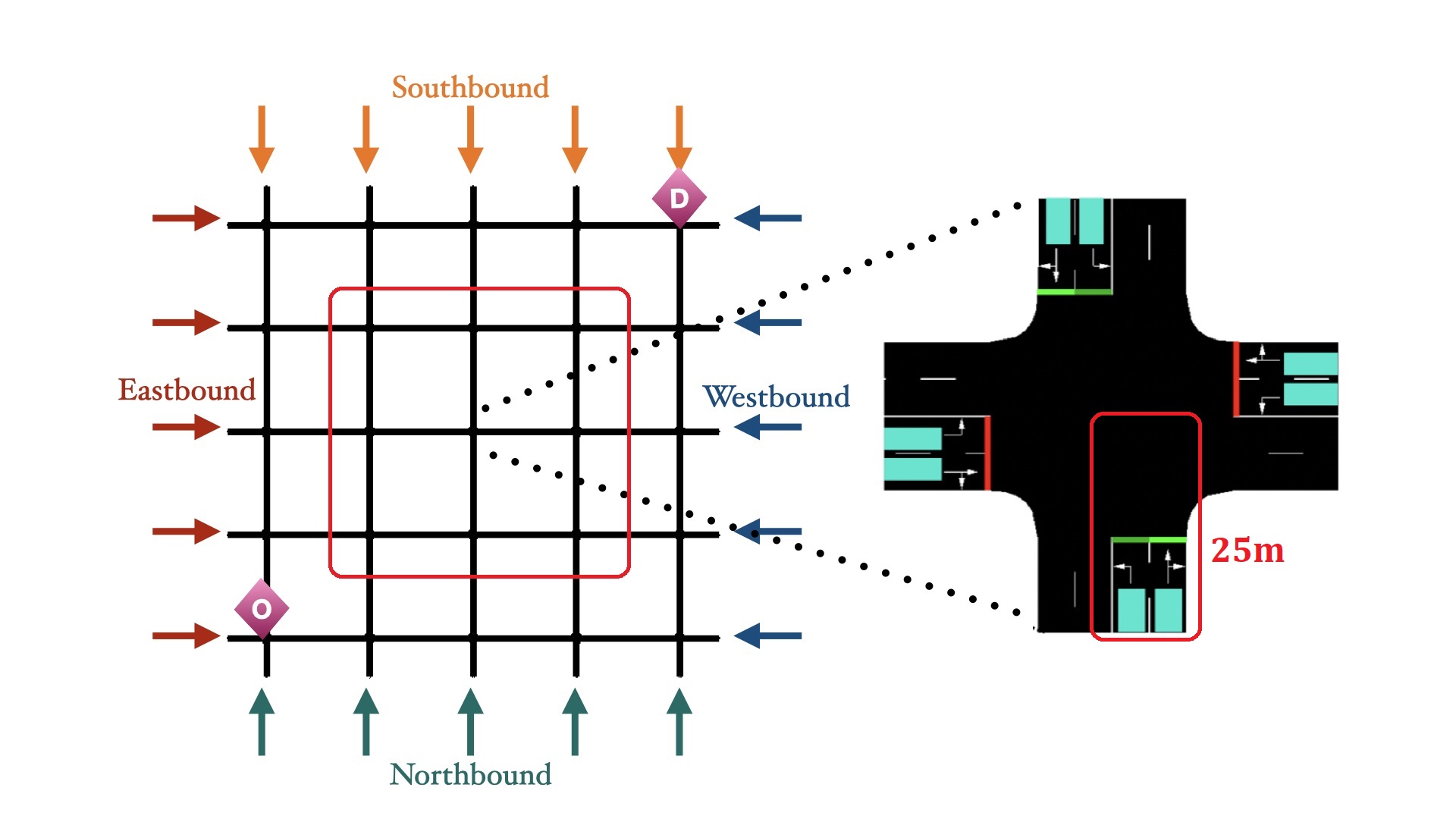} 
    \caption{$\text{Synthetic Grid}_{5\times5}$ and the inner 9 intersections are observed for intersection delay.}
    \label{fig:synthetic_grid}
\end{figure}

Let us revisit the $\text{Synthetic Grid}_{5 \times 5}$ map, shown in Fig.~\ref{fig:synthetic_grid}, where intersections are interconnected through bi-directional links, each consisting of two lanes. For this experiment, we assume that all links have zero emergency capacity. The traffic configuration selected for this study involves northbound and southbound flows transitioning to eastbound and westbound directions, with a non-peak flow of 200 vehicles per lane per hour and a peak flow of 240 vehicles per lane per hour. The origin (O) and destination (D) of the EMV are explicitly labeled on the map to highlight the designated travel route. The traffic flow for this synthetic map spans a duration of 1200 seconds, during which varying congestion levels are simulated. To ensure the network is sufficiently congested, the EMV is dispatched at \( t = 600 \, \text{s} \), a point when traffic conditions have reached compactness.

We record the time taken by the EMV as it traverses any of the inner nine intersections, with the measurement boundaries defined as 25 meters from the center of each intersection in the incoming direction to 25 meters in the outgoing direction. This precise delineation ensures consistency in capturing the traversal times across intersections. The benchmarks for comparison are 1.\textbf{Walabi}~\cite{bieker2019modelling}, which is a rule-based control scheme aiming Green Wave for EMVs in SUMO environment, and 2. no pre-emption for EMVs. 

\begin{figure}[h]
    \centering
    \includegraphics[width=0.9\textwidth]{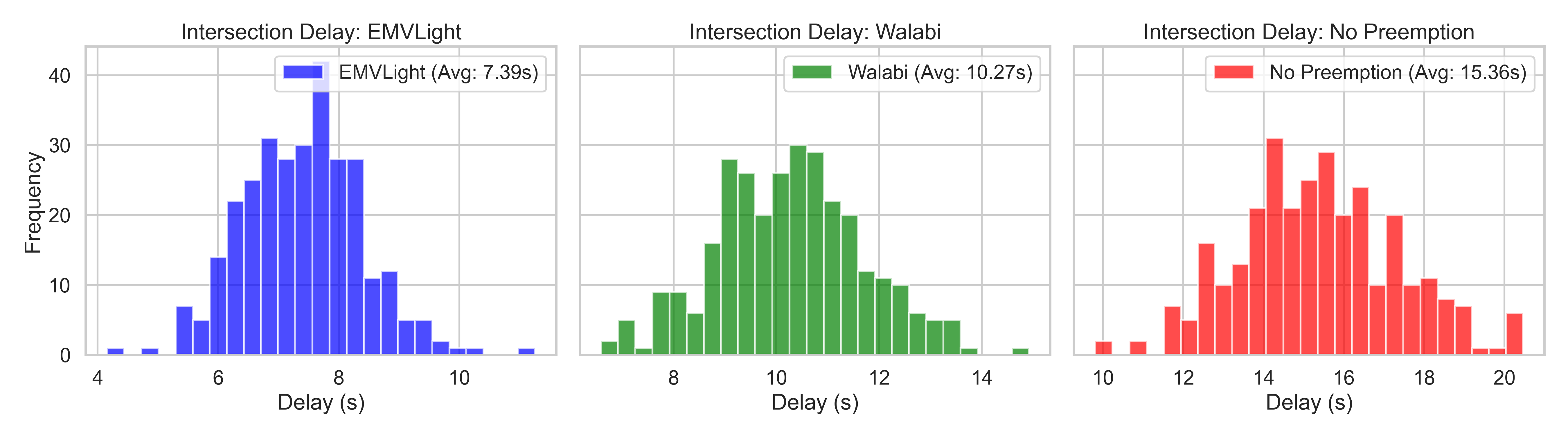} 
    \caption{Amount of time to pass intersections for EMVs.}
    \label{fig:intersection_delay}
\end{figure}

After running the simulations 120 times for each scenario, we aggregate and present the time required for the EMV to traverse the observed intersections in Fig.~\ref{fig:intersection_delay}. The results clearly demonstrate that \textit{EMVLight} reduces intersection delays by approximately 30\% compared to the Walabi and achieves a time savings of over 50\% relative to no pre-emption scenarios. 

\begin{figure}[h]
    \centering
    \includegraphics[width=.75\textwidth]{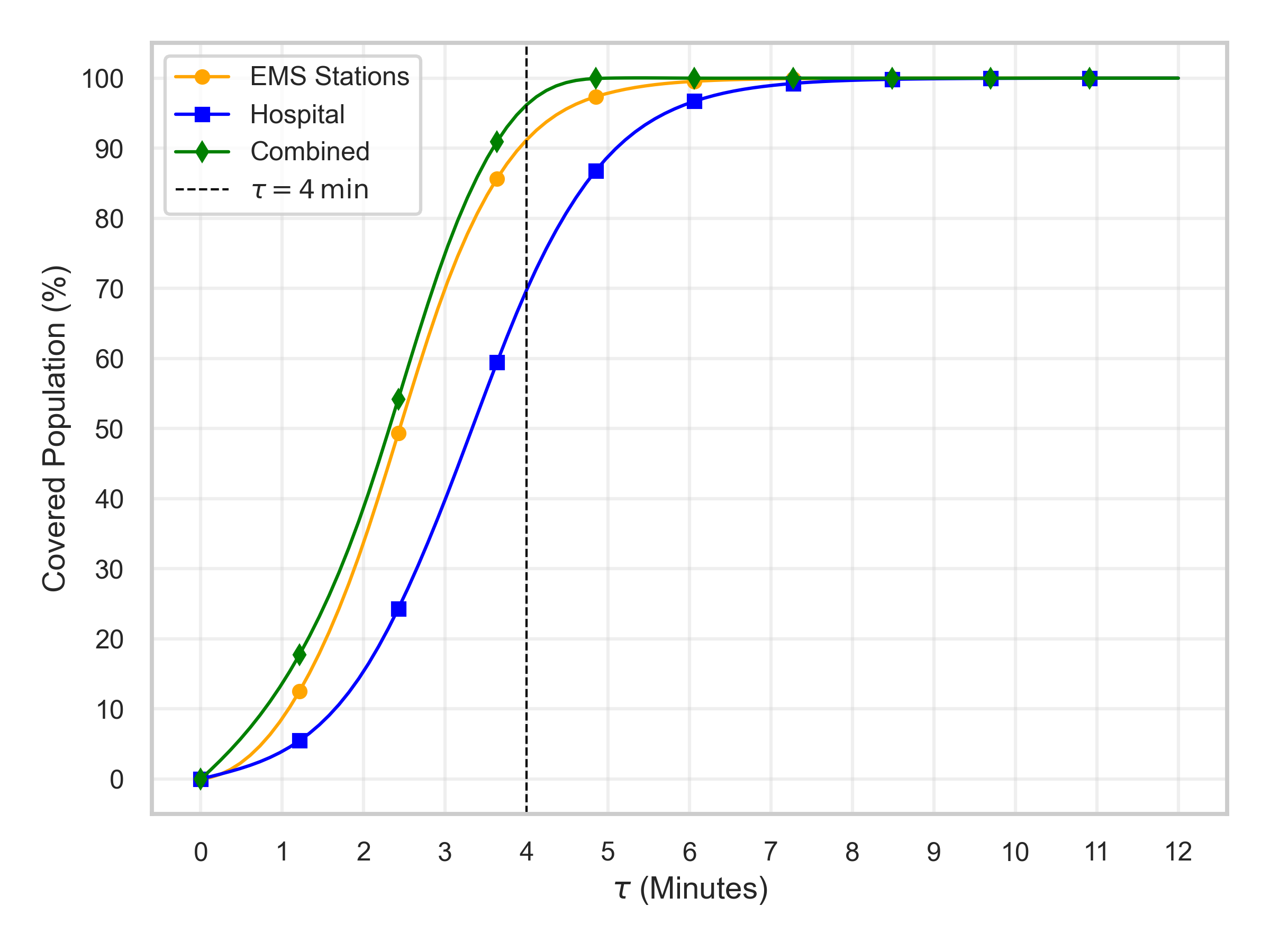} 
    \caption{Population coverage function against $\tau$ with $\alpha = 7.5 \text{s}\cdot\text{m}^2 $}
    \label{fig:emvlight_pop_covered}
\end{figure}

The results indicate that incorporating \textit{EMVLight} across the entire city effectively reduces the intersection delay factor, \(\alpha\), by half. By adopting \(\alpha = 7.5 \, \text{s} \cdot \text{m}^2\), the updated population coverage function, shown in Fig.~\ref{fig:emvlight_pop_covered}, reveals significant improvements in EMS accessibility. Approximately 70\% of the population gains access to hospitals, and 95\% of NYC residents can be served within \(\tau = 4 \, \text{minutes}\). These results highlight the transformative potential of \textit{EMVLight} in enhancing emergency service coverage across the city.

\section{Summary}\label{sec:conclusion_access}
This study presents an intersection-aware EMS accessibility model to comprehensively evaluate medical emergency service coverage in NYC. By incorporating road network characteristics, intersection delays, and demographic data, the model identifies critical EMS-vulnerable regions, including Staten Island, the peripheral regions of Queens, and parts of Manhattan. These findings emphasize the challenges faced by underserved areas, particularly those with limited medical infrastructure, high intersection densities, or significant demographic vulnerabilities. The analysis further demonstrates the capability of \textit{EMVLight} to significantly improve EMS performance by halving intersection delays, increasing hospital accessibility to 70\%, and ensuring 95\% of NYC residents are served within the benchmark travel time of \(\tau = 4 \, \text{minutes}\).

Despite its contributions, the study is subject to certain limitations. The absence of real-time traffic condition data in the travel time assignment likely underestimates congestion-induced delays, and the assumption of free-flow travel between intersections simplifies the complexity of urban traffic dynamics. These constraints highlight the need for future work to incorporate dynamic traffic data and congestion modeling to refine the accessibility framework. Additionally, the static nature of the EMS station locations in the model does not account for adaptive dispatch strategies, which could further enhance emergency response capabilities in underserved regions.

Nonetheless, this study provides a robust foundation for understanding EMS accessibility and exploring the role of advanced TSC systems like \textit{EMVLight} in improving emergency vehicle navigation. The intersection-aware framework and the integration of demographic and network data offer valuable insights into the spatial and temporal disparities in EMS access. Future studies should focus on validating the model using real-world traffic data, exploring the scalability of TSC systems across larger urban networks, and assessing the cost-effectiveness of implementing such systems. By addressing these aspects, the findings of this study can inform policy decisions and urban planning strategies aimed at reducing response times and ensuring equitable access to emergency medical services in cities like NYC and beyond.

\chapter{Conclusions}\label{chap:conclusion}
This chapter synthesizes the primary findings outlined in Section~\ref{sec:findings} and delves into the limitations and prospective future directions detailed in Section~\ref{sec:limitations}. Finally, Section~\ref{sec:closing_remarks} offers reflective closing remarks on the contributions and implications of the research.
\section{Research Findings}\label{sec:findings}
\subsection{\textit{EMVLight}}
\textit{EMVLight}, introduced in Chapter~\ref{chap:emvlight}, demonstrated the feasibility and efficacy of integrating decentralized MARL to simultaneously address EMV routing and traffic signal pre-emption. It effectively tackles two key network-level challenges: (1) coupling dynamic EMV routing with traffic signal control, and (2) reducing EMV travel time while simultaneously mitigating overall congestion. Experimental results highlight that \textit{EMVLight} achieves a remarkable 42.6\% reduction in EMV travel time and a 23.5\% decrease in average travel time for all vehicles, significantly outperforming both conventional and RL-based traffic control methods in synthetic and real-world scenarios.

Many components contribute to the success of \textit{EMVLight}. The incorporation of emergency lane formation highlights the model's capacity to coordinate between primary pre-emption agents, secondary pre-emption agents, and normal agents to pre-clear links ahead of EMVs effectively. Experimental observations reveal that \textit{EMVLight} often navigates EMVs along longer routes that result in shorter overall travel times. The inclusion of a spatial discount factor, combined with the centralized training and decentralized execution scheme, facilitates optimal trajectory planning across the network during EMV traversal. Additionally, the definition of pressure was modified to align with the agent design, ensuring feasibility in achieving the framework's goals to reduce disturbance and overall traffic congestion. Another critical component is the use of fingerprints, which, as demonstrated in ablation studies, significantly stabilizes training and improves agent coordination. Furthermore, \textit{EMVLight} explores the trade-off between minimizing EMV travel time and mitigating overall congestion, delegating this decision to traffic managers based on operational priorities.

\subsection{MAPPO-DQJL}
Chapter~\ref{chap:mappo-dqjl} introduced MAPPO-DQJL, a novel intra-link coordination strategy leveraging the presence of CAVs. This framework dynamically forms lanes to prioritize EMV passage, utilizing V2X communication to coordinate CAVs and indirectly influence HDVs. Modeling the DQJL formation as a POMDP, the framework employs a transformer-based network design and the MAPPO algorithm to train agents to deliver the optimal DQJL establishment strategy for expedited EMV passage with reduced unnecessary maneuvers from non-EMVs.

Extensive experiments on a variety of simulated multi-lane roadway configurations demonstrated the framework's capacity to clear routes for EMVs while reducing overall traffic disturbances. EMVs changed lanes fewer than once on average across all experimental settings with the proposed methodology. By treating HDVs as agents rather than part of the environment, MAPPO-DQJL significantly enhances performance by enabling CAVs to learn and adapt to HDV behaviors, yielding optimal coordination strategies. The framework achieved up to a 39.8\% reduction in EMV passage time and a 55.7\% reduction in lane-change maneuvers for non-EMVs, underscoring its scalability and robustness in mixed traffic scenarios.

Notably, the framework's performance improved with higher CAV penetration rates, achieving near-optimal coordination when penetration reached 75\% or higher. In multi-lane configurations, MAPPO-DQJL demonstrated that EMVs could navigate segments with fewer than one lane change on average, minimizing disruptions to non-EMVs while expediting emergency responses. Furthermore, it was observed that the benefits of CAV coordination exhibit diminishing returns when penetration rates or traffic density reaches a critical mass, aligning with findings from other CAV applications in tasks similar to DQJL formation.

\subsection{NYC EMS Accessibility Study}
Chapter~\ref{chap:ems_accessibility} presented an intersection-aware EMS accessibility model for NYC, integrating road network characteristics, intersection density, and demographic data. The model accounts for delays occurring at intersections and provides valuable insights into travel times assigned to and from EMS facilities. Additionally, it explores the percentage of NYC residents covered within varying benchmark service times.

The study identified vulnerable areas in terms of EMS response times and further analyzed the demographic characteristics of these regions. By incorporating \textit{EMVLight} into the study, the model demonstrated its potential to reduce intersection delays and improve hospital accessibility.

Results indicated that implementing \textit{EMVLight} could enable over 95\% of NYC residents to access hospitals within benchmark travel times, offering actionable insights for urban planners and policymakers to enhance EMS equity and efficiency. Moreover, the study revealed that areas with high intersection densities continue to face delays due to congestion, underscoring the necessity for integrated signal optimization strategies. By analyzing borough-specific response times and population density distributions, the findings emphasize the critical importance of targeted interventions, such as increasing hospital density and optimizing resource allocation in underserved areas.
\section{Limitations and Future Directions}\label{sec:limitations}

\subsection{\textit{EMVLight}}
While \textit{EMVLight} effectively integrates routing and pre-emption, several limitations remain. The framework relies on accurate real-time traffic data, which may not always be available or reliable in real-world deployments. Future research could explore integrating predictive traffic models and robust communication systems to address data uncertainty. Additionally, the scalability of \textit{EMVLight} in larger urban networks with more complex traffic patterns warrants further investigation. 

A critical consideration for future work lies in the statistical robustness of our experimental results. While our current findings demonstrate improvements in travel times, the statistical significance of these improvements needs more rigorous validation. Future studies should determine the optimal number of experimental runs through power analysis, considering factors such as effect size and desired confidence levels. This would provide more statistically sound conclusions about the framework's performance improvements, especially regarding travel time reductions which might appear marginal in absolute terms but could have significant cumulative impacts on emergency response efficiency.

Non-trivial future directions for this study include, but are not limited to, the following. First, as an effort to bridge the gap between simulation and reality, we are motivated to develop more sophisticated traffic pattern models that better capture the complex dynamics of mixed traffic with EMVs. Second, we aim to address the challenge of navigating multiple EMVs simultaneously in the same traffic network, which raises both technical and ethical considerations in reward design for pre-emption agents. Third, a fundamental challenge in applying reinforcement learning to transportation systems lies in the extensive trial-and-error nature of the learning process. Future research should focus on developing safe learning frameworks that minimize negative impacts during the learning phase.

\subsection{MAPPO-DQJL}
Despite its success in coordinating DQJLs, MAPPO-DQJL faces challenges in handling extreme traffic densities and low CAV penetration rates. The reliance on centralized training may also limit scalability in real-time applications. Future work should focus on adaptive learning techniques to reduce computational overhead and enhance real-time adaptability.

A key area for enhancement lies in the representation of human-driven vehicles within the framework. Current models may oversimplify the stochastic nature of human driving behaviors and the complex decision-making processes of human drivers. Future work should explore incorporating more sophisticated driver behavior models that account for varying levels of aggression, risk tolerance, and decision-making patterns. Additionally, the game-theoretic nature of interactions between selfish human drivers, particularly during emergency scenarios, warrants deeper investigation. This could involve developing mechanism design approaches to incentivize cooperative behavior while acknowledging the inherently competitive nature of individual routing choices.

Experimental validation in real-world mixed autonomy environments would provide critical insights for deployment. Additionally, expanding the framework to incorporate multi-lane dynamic lane assignment strategies could further enhance its effectiveness, particularly in environments with asymmetric traffic flows or high vehicle heterogeneity.

Integrating \textit{\textit{EMVLight}} with MAPPO-DQJL presents a promising avenue for future research, as it combines the strengths of network-wide traffic signal coordination with dynamic intra-link lane-clearing strategies. By leveraging \textit{\textit{EMVLight}}'s decentralized MARL framework for real-time routing and signal optimization alongside MAPPO-DQJL's capability to manage mixed traffic at the lane level, a unified system could holistically address both macro-level and micro-level challenges of EMV passage. This integration could enable seamless coordination between intersections and road segments, dynamically adapting to traffic patterns while ensuring unobstructed EMV movement. Furthermore, the combined framework would be particularly effective in scenarios with varying CAV penetration rates, where coordination across intersections and individual lanes becomes critical. To achieve this, future work could focus on designing reward mechanisms that align global network efficiency with local lane-level optimization, ensuring that the dual objectives of reducing EMV travel time and minimizing disruptions to non-EMVs are met cohesively. Experimental validation in simulation environments and real-world mixed autonomy networks would be instrumental in demonstrating the practical feasibility and benefits of this integrated approach.

\subsection{NYC EMS Accessibility Study}
Chapter~\ref{chap:ems_accessibility} highlighted critical disparities but relied on static data and simplified assumptions about traffic dynamics and intersection delays. Future research could incorporate dynamic traffic simulations and real-time data feeds to refine accessibility models. Expanding the study to include other urban centers with varying demographics and infrastructure would provide a broader understanding of EMS accessibility challenges. Additionally, integrating equity-focused policies, such as targeted investments in underserved areas, could address systemic inequities in EMS coverage.

Another potential direction is to explore the impact of emerging technologies, such as autonomous ambulances or drone-based EMS support, on accessibility metrics. Coupling these innovations with intersection-aware frameworks could revolutionize emergency response strategies, particularly in geographically constrained regions.

\section{Closing Remarks}\label{sec:closing_remarks}
This dissertation presents a comprehensive approach to enhancing EMV passage and EMS accessibility in urban environments through innovative methodologies. The integration of MARL-based frameworks, such as \textit{EMVLight} and MAPPO-DQJL, alongside detailed accessibility studies, demonstrates significant potential to improve emergency response efficiency and equity. By addressing the limitations and exploring the proposed future directions, this research lays a robust foundation for advancing emergency traffic management and urban resilience. Furthermore, the findings underscore the transformative potential of integrating cutting-edge technologies with data-driven strategies to create safer, more equitable, and efficient urban transportation systems.

%%%%% Appendices start %%%%%%%%%%%%%%%%
%% Comment out the following line if your thesis has no appendix

%% Note: If your thesis has more than one appendix, NYU requires a "list of
%% appendices" page before the body of the thesis. I don't provide the tools
%% to create that here, so you're on your own for that one... Sorry.
%\input{app2}

%%%% Input bibliography file %%%%%%%%%%%%%%%
% \input{thesis}
\bibliographystyle{unsrt}
\bibliography{_main} % add intelligently
\appendix
\chapter{Appendix for Chapter 1}
In this appendix, we present the implementation details of the proposed EMVLight framework in Section~\ref{appendix_a}, the hyper-parameters selected for all RL-based methods in Section~\ref{appendix_b}, and the implementation of the emergency lane in SUMO in Section~\ref{appendix_c}.

\section{Implementation Details}\label{appendix_a}
% The architectures used for experiments are provided below.
\textbf{MDP time step.} Although MDP step length can be arbitrarily small enough for optimality, traffic signal phases should be maintained a minimum amount of time so that vehicles and pedestrians can safely cross the intersections. To avoid rapid switching between the phases, we set our MDP time step length to be 5 seconds.

% \BDedt{Also, due to safety concerns, once a signal phase has been initiated, it remains unchanged for certain minimum amount of time, e.g. 5 seconds. Therefore, to avoid rapid switching between the phases, we set our MDP time step length to be 5 seconds.}

\textbf{Implementation details for Non-emergency-capacitated/Emergency capacitated Synthetic \texorpdfstring{$\text{Grid}_{5 \times 5}$}{Grid}}
\begin{itemize}
    \item dimension of $s^t_{\mathcal{V}_i}$: $5 \times (8+8+4+2)=110$
    \item dimension of $\Tilde{s}^t_{\mathcal{V}_i}$: $5 \times (8+8+4+2)=110$
    \item dimension of $\pi^{t-1}_{\mathcal{N}_i}$: $4 \times 8 = 32$
    % \item Input for both policy and value networks:
    % observation as a tensor of $5 \times 22$ dimensions and neighbor policies as a tensor of $4 \times 8$ dimensions. The value network multiplies the neighbor policies with spatial discounted factor. 1 action dimension.
    \item Policy network $\pi_{\theta_i}(a_i^t|s^t_{\mathcal{V}_i}, \pi^{t-1}_{\mathcal{N}_i})$: 
    \texttt{concat[}$110 \xrightarrow[]{\textrm{FC}} 128$ReLu, $32 \xrightarrow[]{\textrm{FC}} 64$ReLu\texttt{]} $ \xrightarrow[]{} 64$LSTM $ \xrightarrow[]{\textrm{FC}}8$Softmax
    \item Value network $V_{\phi_i}(\Tilde{s}^t_{\mathcal{V}_i}, \pi^{t-1}_{\mathcal{N}_i})$: \texttt{concat[}$110 \xrightarrow[]{\textrm{FC}} 128$ReLu, $32 \xrightarrow[]{\textrm{FC}} 64$ReLu\texttt{]} $ \xrightarrow[]{} 64$LSTM $ \xrightarrow[]{\textrm{FC}}1$Linear
    \item Each link is $200m$. The free flow speed of the EMV is $12m/s$ and the free flow speed for non-EMVs is $6m/s$.
    \item Temporal discount factor $\gamma$ is $0.99$ and spatial discount factor $\alpha$ is $0.90$.
    \item Initial learning rates $\eta_\phi$ and $\eta_\theta$ are both 1e-3 and they decay linearly. Adam optimizer is used.
    \item MDP step length $\Delta t = 5s$ and for secondary pre-emption reward weight $\beta$ is $0.5$.
    \item Regularization coefficient is $0.01$.
\end{itemize}
\textbf{Implementation details for \texorpdfstring{$\text{Manhattan}_{16 \times 3}$}{Manhattan}}
The implementation is similar to the synthetic network implementation, with the following differences:
\begin{itemize}
    \item Initial learning rates $\eta_\phi$ and $\eta_\theta$ are both 5e-4.
    \item Since the avenues and streets are both one-directional, the number of actions of each agent are adjusted accordingly. 
    \item Avenues and streets length are based on real Manhattan block size with each block of $80m \times 274m$.
\end{itemize}

\textbf{Implementation details for \texorpdfstring{$\text{Hangzhou}_{4 \times 4}$}{Manhattan}}
The implementation is similar to the synthetic network implementation, with the following differences:
\begin{itemize}
    \item Initial learning rates $\eta_\phi$ and $\eta_\theta$ are both 5e-4.
    \item Streets length are based on real map of Hangzhou Gudang district.
\end{itemize}

\section{Hyperparameters}\label{appendix_b}
We provide the the choice of hyper-parameters for RL-based methods in Table.\ref{tab_RL_hyperparameters}.
\begin{table}[ht]
\centering
\fontsize{9.0pt}{10.0pt} \selectfont
\begin{tabular}{@{}ccccc@{}}
\toprule
Hyper-parameters & CDRL   & PL   & CL   & EMVLight   \\ \midrule
temporal discount  &   \multicolumn{4}{c}{0.99} \\
batch size      &  32     &  128    & 128     &    128        \\
buffer size     & 1e5      &  1e4    &  1e4    &   1e4         \\
sample size     &  2048     & 1000     & 1000     &  1000          \\
$\eta$ and decay  & 0.5\&0.975 & 0.8\&0.95     &  0.8\&0.95    & 0.8\&0.95           \\ \midrule
optimizer       &  Adam     & RMSprop     &  RMSprop    &   Adam         \\
Learning rate   &  0.00025     &  1e-3    & 1e-3     &   1e-3         \\ \midrule
\# Conv layers   &   1    &  -     &  3    &    -        \\
\# MLP layers    &   1    &   4   &  3    &      1      \\
\# MLP units     &   (168,168)    & (32,32)     &  (32,32)    &   (192,1)         \\
MLP Activation  & \multicolumn{4}{c}{ReLU}         \\
Initialization  & \multicolumn{4}{c}{Random Normal} \\
\midrule
step length $\Delta T$      &   \multicolumn{4}{c}{ 5 seconds}      \\
\bottomrule
\end{tabular}
\caption{Hyper-parameters selected for RL-based methods.}
\label{tab_RL_hyperparameters}
\end{table}

\section{Emergency Lane in SUMO}\label{appendix_c}
\begin{figure}[ht]
\includegraphics[width=0.8\linewidth]{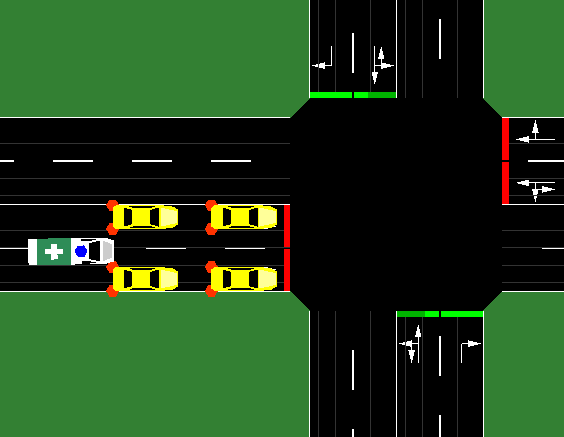}
\centering
\caption{A blue light device approaching an intersection with the sublane model activated. }
\label{fig_emergency_lane_SUMO}
\end{figure}

With increased lateral resolution after enabling the sublane model, SUMO is able to simulates the virtual lane, i.e. emergency lane, formation for emergency traffic. Vehicles can drive between lanes under this mode. At the same time, we define the EMV as a \emph{blue light device} to incorporate with the sublane model. Thus, we can observe that non-EMVs on the left would pull over towards the left side, i.e. latAlignment="left", and non-EMVs on the right would pull over towards the right side, i.e.,  latAlignment="right". When the EMV is traveling on the emergency lane, these vehicles do not perform lane changes and they do not move onto the emergency lane. After the EMV leaves this segments, these vehicles resume normal driving and move back with their previous lateral alignment. These two SUMO built-in modules together is able to achieve our proposed emergency lane. See Fig.~\ref{fig_emergency_lane_SUMO} below from \cite{bieker2018analysis}.

\chapter{Appendix for Chapter 2}
In this appendix, we provide a comprehensive description of the network architectures for CAVs in Section~\ref{sec:network_design_for_CAVs}, along with the SUMO configurations and MAPPO-DQJL parameters detailed in Section~\ref{sec:sumo_configurations}.

\section{Network Details for CAVs}\label{sec:network_design_for_CAVs}
\subsection{Policy Network}\label{subsec:appendix_policy_network}
\[
\pi_{\theta}(a_i \mid o_i) = \text{PolicyHead}\bigl(\text{TransformerEncoder}(\text{PositionalEncoding}(\text{Embedding}(o_i)))\bigr).
\]
\noindent\textbf{Architecture Overview (CAVs):}
\begin{itemize}
    \item \textbf{Input:} Observation \(o_i\)
    \item \textbf{Embedding Layer:} Dimension \( d_{\text{model}} = 256 \)
    \item \textbf{Positional Encoding:} Spatial and temporal context
    \item \textbf{Transformer Encoder:} 
    \begin{itemize}
        \item Number of Layers: \( N = 6 \)
        \item Attention Heads: \( h = 8 \)
        \item Hidden Dimension: \( d_{\text{hidden}} = 528 \)
        \item Feed-Forward Dimension: \( d_{\text{ff}} = 1024 \)
        \item Activation: ReLU, Dropout Rate: 0.1
    \end{itemize}
    \item \textbf{Policy Head:} Fully connected layer producing action probabilities
\end{itemize}
  
\subsection{Centralized Value Network}\label{subsec:appendix_value_network}
\[
V_{\phi}(s) = \text{ValueHead}(\text{TransformerEncoder}(\text{Embedding}(s))).
\]
\noindent\textbf{Architecture Overview (CAVs):}
\begin{itemize}
    \item \textbf{Input:} Global state $s$
    \item \textbf{Embedding Layer:} Dimension $d_{\text{model}} = 256$
    \item \textbf{Transformer Encoder:}
    \begin{itemize}
        \item Number of Layers: $M = 6$
        \item Attention Heads: $h = 8$
        \item Hidden Dimension: $d_{\text{hidden}} = 512$
        \item Feed-Forward Dimension: $d_{\text{ff}} = 2048$
        \item Activation: ReLU, Dropout Rate: $0.1$
    \end{itemize}
    \item \textbf{Value Head:} Fully connected layer producing a scalar value
\end{itemize}

\subsection{Transformer Encoder Layer}\label{subsec:appendix_tf_encoder_layer}
Each transformer encoder layer operates over representations of dimension \( d_{\text{model}} = 256 \) (as adopted in the policy network design for CAVs) and comprises:
\begin{itemize}
    \item \textbf{Multi-Head Attention:} 
    \begin{itemize}
        \item Number of Heads: \( h = 8 \)
        \item Key/Value Dimension per Head: \( d_k = d_v = \frac{d_{\text{model}}}{h} = 32 \)
    \end{itemize}
    The output of the multi-head attention is projected back to \( d_{\text{model}} = 256 \).
    \item \textbf{Feed-Forward Network:} 
    Two fully connected layers with \( d_{\text{ff}} = 4096 \) and ReLU activation.
    \item \textbf{Residual Connections:} 
    Applied after both the multi-head attention and feed-forward sub-layers, improving gradient flow and stability.
    \item \textbf{Layer Normalization:}
    Applied to each sub-layer’s output (after residual addition) to stabilize training dynamics.
\end{itemize}

\subsection{Number of Parameters}\label{subsec:number_of_parameters}
Most parameters arise from the Transformer Encoder layers, each MHA and FFN sub-layers. For both the policy and value networks, MHA (with $8$ heads and $d_{\text{model}}=256$) requires four linear transformations (queries, keys, values, and output), each $256\times256$, yielding about $4 \times 256 \times 256 \approx 262{,}000$ parameters per layer. In the policy network, the FFN has $d_{\text{ff}}=1024$, contributing $2 \times 256 \times 1024 \approx 524{,}000$ parameters per layer. Thus, each encoder layer adds roughly $786{,}000$ parameters; for $6$ layers, this is about $4.7$ million parameters, plus a smaller amount for embeddings and the policy head.

In the value network, with $d_{\text{ff}}=2048$, each layer’s FFN has about $2 \times 256 \times 2048 \approx 1{,}048{,}000$ parameters. Combined with MHA, each layer adds approximately $1.31$ million parameters; over $6$ layers, around $7.9$ million parameters in total, again plus modest contributions from embeddings and the value head.

\section{SUMO Configurations}\label{sec:sumo_configurations}
\subsection{Car-Following Model}\label{subsec:appendix_car_following}
In SUMO, the car-following behavior of vehicles is modeled using the Intelligent Driver Model (IDM), which governs how a vehicle adjusts its speed based on the proximity and dynamics of the vehicle ahead. The acceleration \(a(t)\) of a vehicle is defined as:
\[
a(t) = a_{\text{max}} \left( 1 - \left( \frac{v(t)}{v_0} \right)^4 - \left( \frac{s^*(v(t), \Delta v)}{s(t)} \right)^2 \right),
\]
where \(a_{\text{max}}\) is the maximum achievable acceleration, \(v(t)\) is the current speed, and \(v_0\) denotes the desired (free-flow) speed. The variable \(s(t)\) represents the current gap between the vehicle and the one immediately in front, while \(\Delta v\) is the speed difference between these two vehicles.

The desired gap \(s^*(v(t), \Delta v)\) is given by:
\[
s^*(v(t), \Delta v) = s_0 + v(t)T + \frac{v(t) \Delta v}{2\sqrt{a_{\text{max}} b}},
\]
where \(s_0\) is the minimum safety gap, \(T\) the desired time headway, and \(b\) the comfortable deceleration. By maintaining a safe following distance and adjusting speed in response to the leading vehicle, IDM ensures realistic and safe traffic flow dynamics.

In the context of DQJL formation, IDM serves as a baseline safety constraint. Both HDVs and CAVs adhere to IDM-based accelerations and distance requirements, guaranteeing that their interactions remain safe and well-coordinated even under dynamic and congested traffic conditions.
\subsection{Lane-Changing Model}\label{subsec:appendix_lane_changing}
The lane-changing behavior in SUMO employs a rule-based framework that assesses both safety and incentive conditions prior to executing a maneuver. Safety is paramount, ensuring that a vehicle can merge into the target lane without inducing hazardous situations. The safety condition, accounting for deceleration capabilities and reaction times, is expressed as:
\[
v(t) + bT \leq v_{\text{leader}}(t) \quad \text{and} \quad v_{\text{follower}}(t) + bT \leq v(t),
\]
where \( v_{\text{leader}}(t) \) and \( v_{\text{follower}}(t) \) are the speeds of the leading and following vehicles in the target lane, respectively. The parameter \( b \) denotes the comfortable deceleration, while \( T \) represents the driver’s reaction time or headway. These inequalities guarantee that both the merging vehicle and those in the target lane can safely adjust their speeds if necessary.

In addition to satisfying the safety condition, the lane change must offer a tangible incentive. A vehicle will opt to change lanes only if the anticipated speed in the target lane is substantially higher than in its current lane. This incentive criterion ensures that maneuvers are not only safe, but also yield an improvement in travel efficiency, thereby supporting the formation of DQJLs without compromising overall traffic stability.

\subsection{Parameters, Coefficients and Training Hyper-parameters for MAPPO-DQJL}
\begin{table}[htbp]
\centering
\scriptsize
\begin{tabular}{lccc}
\toprule
\textbf{Parameter} & \textbf{Description} & \textbf{Value} & \textbf{Units}\\
\midrule
\multicolumn{4}{l}{\textit{Car-Following (IDM) Parameters}} \\[6pt]
$s_0$        & Minimum safety gap                         & 2.5        & m \\
$T$          & Desired time headway                       & 1.2        & s \\
$v_0$        & Desired (free-flow) speed                  & 20.0       & m/s \\
$a_{\text{max}}$ & Maximum acceleration                     & 1.5        & m/s$^2$ \\
$b$          & Comfortable deceleration                   & 2.5        & m/s$^2$ \\[6pt]

\multicolumn{4}{l}{\textit{Lane-Changing Model Parameters (Same $b$ and $T$ as IDM)}} \\[6pt]
\midrule
\multicolumn{4}{l}{\textit{Global Reward Coefficients}} \\[6pt]
$\alpha$     & Global-vs-Local reward balance             & 0.75        & -- \\
$\beta_1$    & EMV time penalty weight                    & 0.02       & 1/s \\
$\beta_2$    & EMV lane-change penalty weight             & 1.0        & 1/(\#lane changes) \\
$\beta_3$    & DQJL completeness reward weight            & 1.0        & 1/m \\[6pt]
\midrule
\multicolumn{4}{l}{\textit{Local Reward Coefficients (CAVs)}} \\[6pt]
$\eta_1$     & CAV lane-change penalty weight             & 1.0        & 1/(\#lane changes) \\
$\eta_2$     & CAV acceleration penalty weight            & 0.2        & 1/(m/s$^2$) \\[6pt]
\midrule
\multicolumn{4}{l}{\textit{Local Reward Coefficients (HDVs)}} \\[6pt]
$\eta_3$     & HDV yield reward weight                    & 1.0        & 1/(yield action) \\[6pt]
\midrule
\multicolumn{4}{l}{\textit{MDP and Discount Factor}} \\[6pt]
$\gamma$     & Future reward weighting                    & 0.99       & -- \\
\textbf{MDP Timestep} & Simulation step per action        & 0.5        & s \\[6pt]
\midrule
\multicolumn{4}{l}{\textit{Training Hyper-parameters}} \\[6pt]
Learning Rate & Policy and Value Function Optimizer       & $1 \times 10^{-4}$ & -- \\
Batch Size    & Number of samples per update              & 64         & -- \\
Horizon       & Steps per rollout before update           & 1024       & steps \\
PPO Epochs    & Number of PPO update epochs per batch     & 10         & -- \\
Clip Range    & PPO clip range for policy updates         & 0.2        & -- \\
GAE $\lambda$ & Advantage estimation parameter            & 0.95       & -- \\
\bottomrule
\end{tabular}
\caption{Parameter summary for MAPPO-DQJL simulations}
\label{tab:parameters_summary}
\end{table}

\chapter{Appendix for Chapter 3}
\section{Demographic maps}\label{appendix_access}
\begin{figure}[ht]
    \centering
    \includegraphics[width=\textwidth]{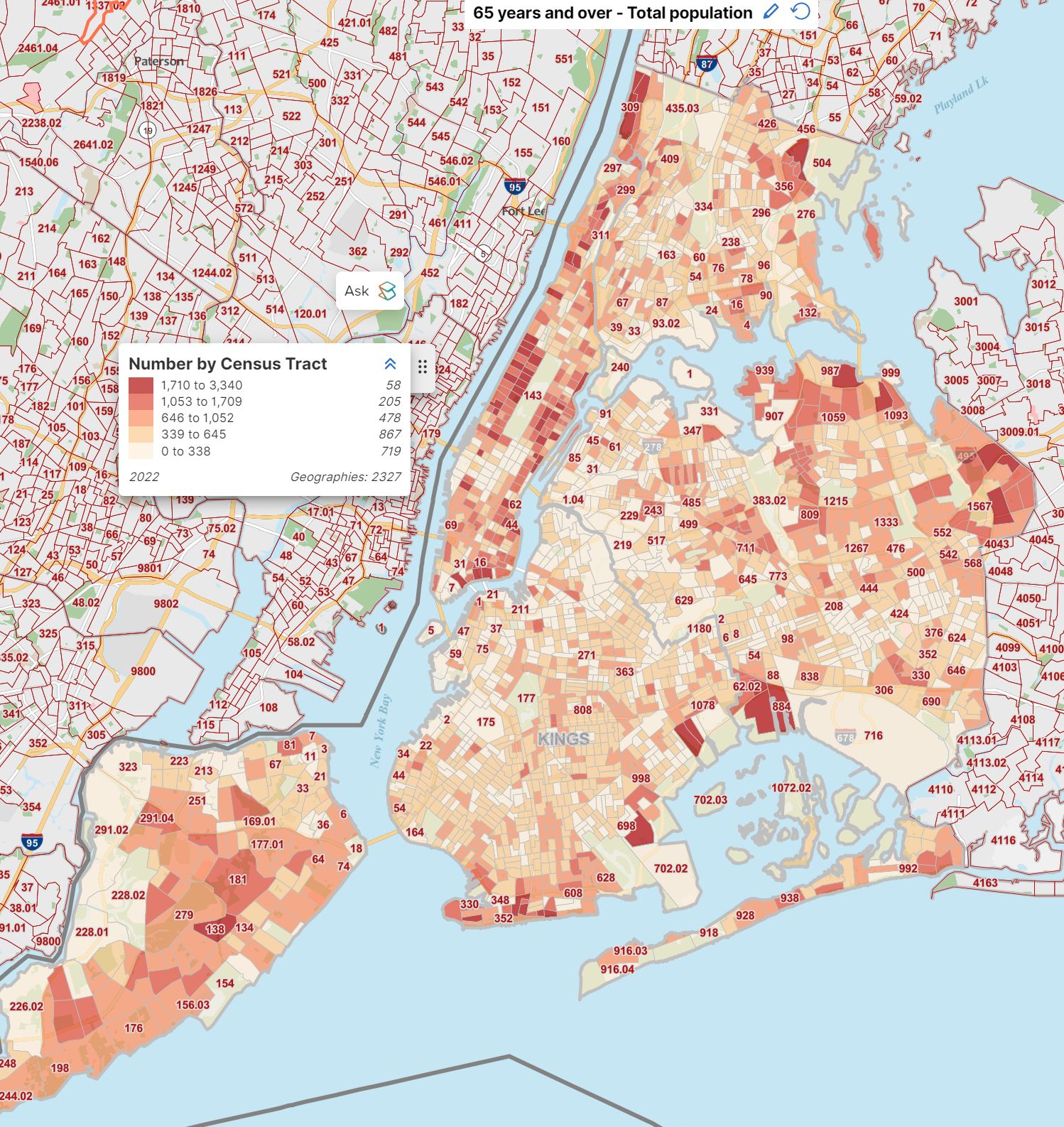} 
    \caption{Population of 65+ years old distribution of NYC census tracts}
    \label{fig:65+_distribution}
\end{figure}

\begin{figure}[ht]
    \centering
    \includegraphics[width=\textwidth]{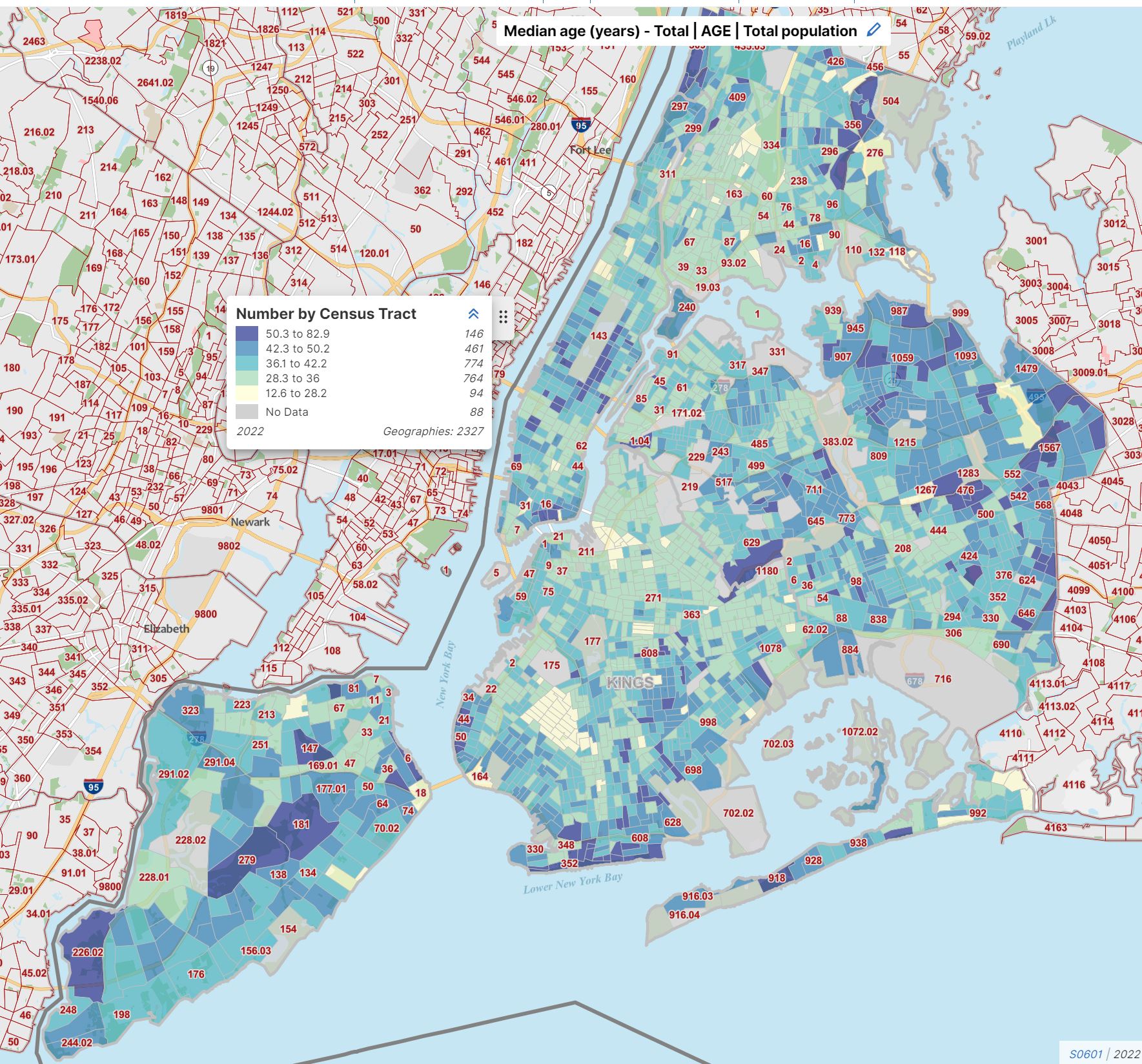} 
    \caption{Median age distribution of NYC census tracts.}
    \label{fig:median_age_distribution}
\end{figure}

\begin{figure}[ht]
    \centering
    \includegraphics[width=\textwidth]{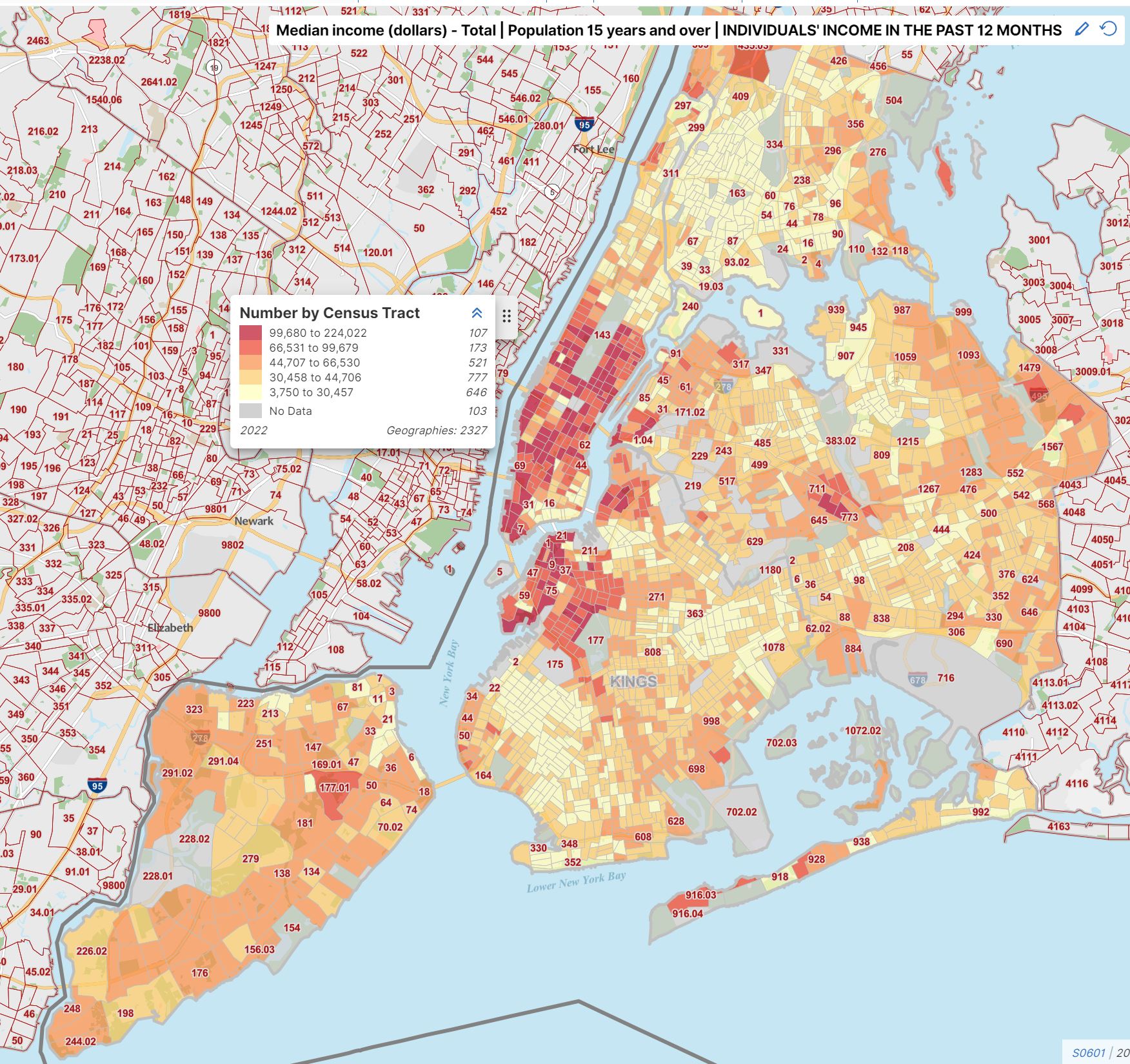} 
    \caption{Median income distribution of NYC census tracts}
    \label{fig:median_income_distribution}
\end{figure}

\begin{figure}[ht]
    \centering
    \includegraphics[width=\textwidth]{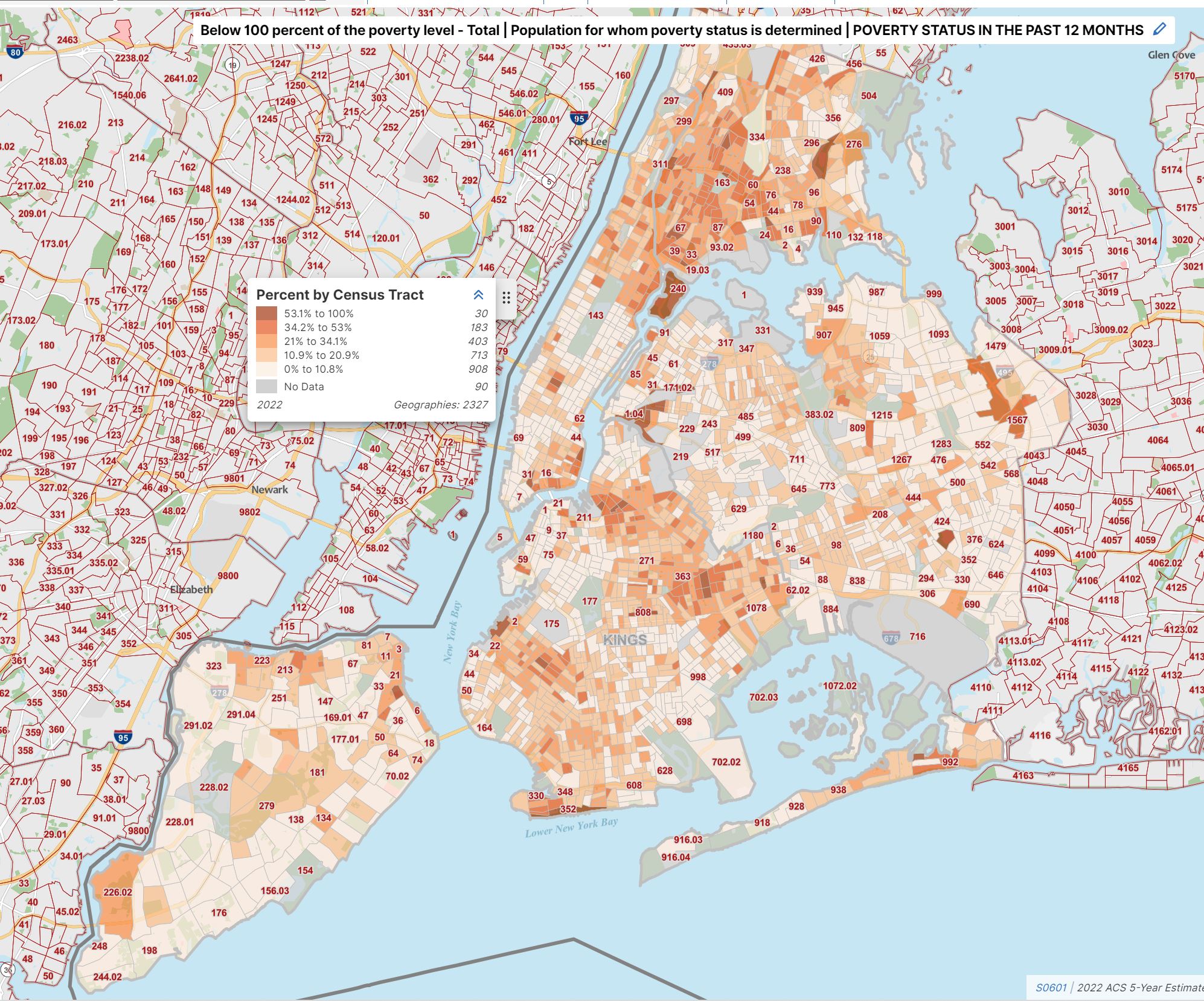} 
    \caption{Poverty percentage distribution of NYC census tracts}
\label{fig:poverty_percentage_distribution}
\end{figure}

In this section, we present detailed population distribution maps based on data from the 2023 US Census~\cite{bureau2023census}. Fig.~\ref{fig:65+_distribution} illustrates the distribution of residents aged 65 years or older across NYC, while Fig.~\ref{fig:median_age_distribution} shows the median age for all census tracts. Fig.~\ref{fig:median_income_distribution} depicts the median income across NYC census tracts, and Fig.~\ref{fig:poverty_percentage_distribution} provides the poverty ratio distribution among residents in all census tracts.


\end{document}